\documentclass[12pt]{article}
\usepackage[letterpaper,left=1in,bottom=1in,top=1in,right=1in]{geometry}

\usepackage[utf8]{inputenc}
\usepackage[T1]{fontenc}
\usepackage[english]{babel}

\usepackage{microtype}
\usepackage{csquotes}

\usepackage{setspace}
\usepackage[en-GB]{datetime2}
\DTMlangsetup[en-GB]{showdayofmonth=false}

\usepackage{titlesec}

\titleformat{\part}[frame]
{\normalfont}
{\filright \footnotesize \enspace PART \thepart \enspace}
{8pt}
{\vspace{20pt} \huge \bfseries \filcenter}

\titlespacing*{\part}{0pt}{-50pt}{10pt}

\usepackage{titletoc}

\usepackage[shortlabels]{enumitem}

\usepackage{graphicx}
\usepackage{tikz}
\usepackage{caption}
\usepackage{subcaption}
\usepackage{float}
\usepackage{booktabs}

\usepackage{xcolor}

\definecolor{linkred}{RGB}{125,25,50}

\usepackage{amsmath}
\allowdisplaybreaks[4]
\usepackage{amssymb}
\usepackage{mathtools}
\usepackage{amsthm}
\usepackage{etoolbox}

\usepackage{amsfonts}
\usepackage{mathrsfs}
\usepackage{dsfont}
\usepackage{bm}

\usepackage{relsize}
\usepackage{nicefrac}
\usepackage{centernot}

\usepackage{accents}

\usepackage{algorithm}

\usepackage{algpseudocode}

\theoremstyle{plain}
\newtheorem{theorem}{Theorem}
\newtheorem{lemma}[theorem]{Lemma}
\newtheorem{proposition}[theorem]{Proposition}
\newtheorem{corollary}[theorem]{Corollary}
\newtheorem{claim}[theorem]{Claim}

\newtheorem{atheorem}{Theorem}

\theoremstyle{definition}
\newtheorem{definition}[theorem]{Definition}

\newtheorem{example}[theorem]{Example}

\theoremstyle{remark}
\newtheorem{remark}[theorem]{Remark}

\AtBeginEnvironment{proof}{\setcounter{localclaim}{0}}



\usepackage{tcolorbox}

\tcbset{
    rounded corners,
    colback = white,
    before skip = 0.25cm,
    after skip = 0.5cm,
    boxrule = 2pt,
    arc = 10pt
}

\newtcolorbox{algbox}{float=!t}

\usepackage[framemethod=tikz]{mdframed}

\newmdenv[
  linewidth=0.6pt,
  linecolor=black!65,
  roundcorner=2pt,
  skipabove=12pt,
  skipbelow=12pt,
  innerleftmargin=9pt,
  innerrightmargin=9pt,
  innertopmargin=7pt,
  innerbottommargin=8pt
]{theoremframe}

\usepackage[numbers]{natbib}
\usepackage{hyperref}
\hypersetup{
    colorlinks=true,
    linkcolor=linkred,
    citecolor=linkred,
    urlcolor=linkred,
    linktocpage=true,
    breaklinks=true,
    pdftitle={PAC Learning with Multiple Correct Answers}
}

\usepackage{xurl}

\usepackage[capitalize,noabbrev]{cleveref}

\newenvironment{keywords}{\vspace{-\baselineskip} \leftskip 30pt \rightskip 30pt \noindent \ignorespaces \small {\bfseries Keywords:}}{\par}

\def\X{{\mathcal X}}
\def\Y{{\mathcal Y}}
\def\H{{\mathcal H}}
\def\D{{\mathcal D}}
\def\E{{\mathbb E}}
\newcommand{\N}{\mathbb{N}}
\newcommand{\cA}{\mathcal{A}}
\newcommand{\cB}{\mathcal{B}}
\newcommand{\cE}{\mathcal{E}}
\newcommand{\cF}{\mathcal{F}}
\newcommand{\cG}{\mathcal{G}}
\newcommand{\cI}{\mathcal{I}}
\newcommand{\cK}{\mathcal{K}}
\newcommand{\cM}{\mathcal{M}}
\newcommand{\cP}{\mathcal{P}}
\newcommand{\cQ}{\mathcal{Q}}
\newcommand{\cU}{\mathcal{U}}
\newcommand{\cV}{\mathcal{V}}
\newcommand{\eps}{\varepsilon}
\newcommand{\ind}{\mathds{1}}

\DeclareMathOperator{\supp}{supp}
\DeclareMathOperator{\Var}{Var}

\DeclareMathOperator{\Unif}{Unif}
\DeclareMathOperator{\Bin}{Bin}
\DeclareMathOperator{\Dec}{Dec}
\DeclareMathOperator{\MsgDec}{MsgDec}

\DeclareMathOperator{\miss}{miss}

\newcommand{\dN}{d^{\cap}_{\mathrm{N}}}
\newcommand{\dDS}{d^{\cap}_{\mathrm{DS}}}
\newcommand{\dpair}{d_{\mathrm{pair}}}
\newcommand{\dpack}{d^{\cap}_{\mathrm{pack},\gamma}}
\newcommand{\dpackr}[1]{d^{\cap}_{\mathrm{pack},#1}}
\newcommand{\mfull}{m^{\mathrm{full}}}
\newcommand{\mfullag}{m^{\mathrm{full,ag}}}
\newcommand{\mband}{m^{\mathrm{band}}}
\newcommand{\mbandag}{m^{\mathrm{band,ag}}}
\newcommand{\mone}{m^{\mathrm{one}}}
\newcommand{\Lset}{L^{\mathrm{set}}}
\newcommand{\RiskMin}{\mathfrak{L}}

\usepackage{color-edits}
\addauthor[Alireza]{az}{teal}
\addauthor[Farnam]{fm}{cyan}
\addauthor[Kun]{kw}{red}
\addauthor[Amirreza]{as}{green}

\title{PAC Learning with Multiple Correct Answers}

\author{
    Steve Hanneke\thanks{Department of Computer Sciences, Purdue University, USA. Email: \texttt{steve.hanneke@gmail.com}} \and
    Farnam Mansouri\thanks{University of Waterloo and Vector Institute, Canada. Email: \texttt{f5mansou@uwaterloo.ca}} \and Alireza F. Pour\thanks{University of Waterloo and Vector Institute, Canada. Email: \texttt{alireza.fathollahpour@uwaterloo.ca}} \and
    Amirreza Shaeiri\thanks{Department of Computer Sciences, Purdue University, USA. Email: \texttt{amirreza.shaeiri@gmail.com}} \and
    Kun Wang\thanks{Department of Computer Sciences, Purdue University, USA. Email: \texttt{wangkun8512@gmail.com}}
}

\date{\today}

\begin{document}

\hypersetup{pageanchor=false}
\maketitle

\thispagestyle{empty}

\begin{abstract}

\footnotesize

In multiclass PAC learning, a prediction is correct only if it equals the label of the example, and learnability is characterized by the Natarajan dimension for finite label spaces and by the DS dimension in general. Many tasks, however, admit several correct answers: a theorem has many valid proofs, and any one of them will do. We study PAC learning when each example comes with a set of at most $\ell$ correct labels and a prediction is correct if it belongs to this set. Training may reveal the \emph{full set}, only whether a proposed label is correct (\emph{bandit feedback}), or one correct label chosen by a demonstrator (\emph{demonstrations}).

Our main finding is that demonstrations can be much harder than full sets, and even than bandit feedback, both in which classes can be learned and in how many examples are needed. For finite label spaces, a \emph{generalized Natarajan dimension} characterizes learnability from full sets and from bandit feedback, in both the realizable and the agnostic setting.
From demonstrations, however, a class is realizably learnable if and only if the \emph{intersections of at most $\ell+1$ of its hypotheses} are learnable from full sets. Already with three labels, this condition can fail for a class learnable from a single full-set example or even a single bandit answer. When the condition holds, we show that $\widetilde O(\eps^{-\ell})$ demonstrations suffice, and the exponent $\ell$ cannot be improved in general, even for classes learnable from $\widetilde O(1/\eps)$ full sets or bandit answers.

For countable label spaces and $\ell \geq 2$, finiteness of the DS dimension, extended to set-valued hypotheses, is neither necessary nor sufficient for learning from full sets. Instead, our \emph{packed dimension}, which asks for many disjoint groups of at most $\ell+1$ hypotheses with no common label, characterizes full-set learnability, and demonstrations still reduce to intersections. Agnostic learning from demonstrations, already for finite label spaces, is not determined by the dimensions of the intersection class that characterize realizable learning. Under a mild nondegeneracy assumption, we characterize it by the \emph{DS dimension} of an induced multiclass class, together with a local compatibility condition.

Finally, we show that the uniform bound $\ell$ on the number of correct labels is essential. Without this bound, it is impossible to characterize learnability by any dimension determined by finite subclasses.
\end{abstract}

\clearpage


\clearpage
\hypersetup{pageanchor=true}

\setcounter{page}{1}
\setcounter{tocdepth}{2}\tableofcontents

\clearpage

\section{Introduction}\label{sec:intro}

A central question in statistical learning theory is which classes can be learned from finitely many examples \cite{vapnik1974theory}. The most prominent framework that makes this question precise is the \emph{probably approximately correct (PAC)} model of \cite{valiant1984theory}. When every example has exactly one correct answer, learnability in this model is now well understood: binary classes are learnable exactly when their VC dimension is finite \cite{vapnik1971uniform,blumer1989learnability}, and a line of work beginning with \cite{natarajan1989learning} culminated in a characterization of multiclass learnability by the DS dimension \cite{daniely2014optimal,brukhim2022characterization}. It is natural to ask what remains of this picture when an example may have several correct answers.

Many learning tasks have this form. A theorem may have many valid proofs, and a prover succeeds if it produces any one of them. Similarly, a specification can be met by many different programs, and a chess position may have several winning moves. In such tasks, a target assigns to every input $x$ a set $h^\star(x)$ of correct labels, and a prediction is correct if it belongs to this set. We ask which classes can be learned under different forms of feedback, and in particular when each training example reveals only one of its correct labels.

Training data of this kind, called \emph{demonstrations}, are what practice usually provides. A proof written by a mathematician, or a program written by a programmer, certifies one correct answer and says nothing about the others. Supervised fine-tuning trains on exactly such data, one correct output per prompt, a setting recently studied formally by \cite{joshi2026learning}. On the other hand, the existing theory of autoregressive learning \cite{malach2024auto,joshi2025theory,hanneke2026sample} measures correctness against a single target output. As \cite{hanneke2026sample} points out, this misses tasks such as proving a theorem, where every correct output should count. Producing one of many valid outputs is also the theme of language generation in the limit \cite{kleinberg2024language}.

This paper develops a PAC theory for tasks with several correct outputs. The learner may see the whole set of correct labels, one correct label, or only whether a label it proposes is correct, as when a checker verifies a proposed proof. Three features significantly distinguish this theory from the multiclass one. First, learning from demonstrations is governed not by the class itself but by the intersections of at most $\ell+1$ of its hypotheses, where $\ell$ bounds the number of correct labels. Second, for infinite label spaces, the natural extension of the DS dimension no longer characterizes learnability, and we introduce a new dimension that does. Third, realizable and agnostic learnability, which coincide in the multiclass theory \cite{david2016supervised,pmlr-v178-hopkins22a,hanneke2025representation}, come apart for demonstrations, and for nondegenerate classes, we characterize agnostic learnability by the DS dimension of an induced multiclass class, together with local compatibility.

\subsection{Multi-Label PAC Learning}\label{sec:model}
Let $\X$ be an instance space and $\Y$ a countable label space.\footnote{For general instance spaces, we impose the regularity assumptions of Appendix~\ref{app:measurability}.} A \emph{multi-label hypothesis} $h$ is defined on a set $\supp(h)\subseteq\X$ and assigns to every $x\in\supp(h)$ a nonempty set $h(x)\subseteq\Y$ of \emph{correct labels}, of size at most a fixed integer $\ell$. The learner outputs a single label (unlike in multi-label classification \cite{zhang2014review}), and it succeeds if this label belongs to the correct set.

As for partial concept classes \cite{alon2021theory}, $h$ may be undefined outside $\supp(h)$, and the data are then drawn from $\supp(h)$. Partial hypotheses arise even from total classes, because the intersection of two hypotheses is undefined where their correct-label sets are disjoint. A class $\H$ is a set of multi-label hypotheses. The learner outputs a \emph{predictor} $g:\X\to\Y$, whose \emph{error} with respect to a target $h^\star\in\H$ and a distribution $P$ on $\supp(h^\star)$ is
\[
L_{P,h^\star}(g)=\Pr_{x\sim P}\bigl[g(x)\notin h^\star(x)\bigr].
\]
During training, the learner receives independent inputs $x_1,\ldots,x_m\sim P$ together with one of three kinds of feedback:
\begin{itemize}
\item[--] \emph{full-set feedback}: the sets $h^\star(x_1),\ldots,h^\star(x_m)$;
\item[--] \emph{bandit feedback}: for each $x_i$, the learner proposes a label $u_i$, possibly chosen adaptively, and observes whether $u_i\in h^\star(x_i)$;
\item[--] \emph{one-correct-label feedback}, or \emph{demonstrations}: for each $x_i$, a label $y_i\in h^\star(x_i)$ drawn independently from an unknown conditional distribution given $x_i$, the \emph{demonstrator}.
\end{itemize}
The class $\H$ is \emph{PAC learnable} under a feedback model if, for every $\eps,\delta\in(0,1)$, there are a learner and a finite sample size $m$ such that, for every target, every distribution and, under one-correct-label feedback, every demonstrator, the learner returns from $m$ examples a predictor with error at most $\eps$ with probability at least $1-\delta$. We write $\mfull_\H(\eps,\delta)$, $\mband_\H(\eps,\delta)$ and $\mone_\H(\eps,\delta)$ for the least such $m$ under the three feedback models. When $\ell=1$, this is multiclass PAC learning.

In the \emph{agnostic} setting, examples $(x,A)$ are drawn from an arbitrary distribution $\D$ on inputs and label sets, and a prediction is correct if it belongs to the sampled set $A$. A hypothesis predicts a whole set, and, as with the zero-one loss in the multiclass case, we count it as wrong when this set differs from the sampled one. Thus the learner competes with the benchmark
\[
\Lset_\D(\H)=\inf_{h\in\H}\Pr_{(x,A)\sim\D}[h(x)\neq A].
\]
This benchmark is natural: for each $h\in\H$, a predictor that outputs a label of $h(x)$ has error at most $\Pr[h(x)\neq A]$. For $\ell=1$, it is the usual multiclass benchmark. Under each feedback model, agnostic learning asks for error at most $\Lset_\D(\H)+\eps$, where demonstrations are now labels drawn from the sampled sets, and we write $\mfullag_\H(\eps,\delta)$ for the sample complexity under full-set feedback. The main question of this work is the following.
\begin{center}
\emph{Which classes are PAC learnable when an input may have several correct labels,\\ and how does the answer depend on the feedback?}
\end{center}

\subsection{Main Results}\label{sec:results}
The answer depends on the size of the label space, the available feedback, and whether the number of correct labels is uniformly bounded. Table~\ref{tab:summary} summarizes our results, and the rest of this section states them.

\begin{table}[t]
\centering
\small
\renewcommand{\arraystretch}{1.3}
\begin{tabular}{@{}lll@{}}
\toprule
Setting & Learnable if and only if & Sample complexity\\
\midrule
finite $\Y$, full set & $\dN(\H)<\infty$ & $\widetilde O_\ell\bigl((d+\log\frac1\delta)/\eps\bigr)$\\
finite $\Y$, bandit & $\dN(\H)<\infty$ & $\widetilde O_\ell\bigl(k\,d\,\eps^{-1}\log\frac1\delta\bigr)$\\
finite $\Y$, one correct label & $\dN(\cI_{\ell+1}(\H))<\infty$ & $\widetilde O_\ell\bigl(d\,\eps^{-\ell}\log\frac1\delta\bigr)$\\
countable $\Y$, full set & $\dpack(\H)<\infty$ & $\widetilde O_{\ell,\gamma}\bigl((p+\log\frac1\delta)/\eps\bigr)$\\
countable $\Y$, one correct label & $\dpack(\cI_{\ell+1}(\H))<\infty$ & $\widetilde O_{\ell,\gamma}\bigl(p\,\eps^{-\ell}\log\frac1\delta\bigr)$\\
agnostic, one correct label & \multicolumn{2}{l}{$\dDS(\widetilde\H)<\infty$ and local compatibility (nondegenerate $\H$)}\\
no uniform bound $\ell$ & \multicolumn{2}{l}{no dimension with finite witnesses}\\
\bottomrule
\end{tabular}
\medskip
\caption{Summary of results. Here $d$ and $p$ denote the dimensions in the second column, $k=|\Y|$, $\cI_{\ell+1}(\H)$ is the class of intersections of at most $\ell+1$ hypotheses of $\H$ (defined before Theorem~\ref{thm:B}), $\widetilde\H$ is the collapsed class (defined before Theorem~\ref{thm:F}), and $\widetilde O$ hides logarithmic factors. The bounds are for the realizable setting; the bandit bound is from Theorem~\ref{thm:finite-bandit}. Under full-set feedback, and under bandit feedback with finite label spaces, realizable and agnostic learnability coincide.}
\label{tab:summary}
\end{table}

\subsubsection*{Finite Label Spaces}
Our first result extends the classical characterization of multiclass learnability by \cite{natarajan1989learning} and \cite{ben1995characterizations} to several correct labels, under both full-set and bandit feedback. We use generalizations of the Natarajan and DS dimensions to set-valued hypotheses. The generalized Natarajan dimension requires, at each input, a family of correct-label sets with empty intersection, such that every combination of choices across inputs is realized by the class. Formally, distinct inputs $x_1,\ldots,x_n$ are \emph{product-shattered} by $\H$ if there are finite families $\cA_1,\ldots,\cA_n$ of label sets such that the sets in each $\cA_i$ have no common label and, for every choice of $A_1\in\cA_1,\ldots,A_n\in\cA_n$, some $h\in\H$ satisfies $h(x_i)=A_i$ for all $i$. The \emph{generalized Natarajan dimension} $\dN(\H)$ is the largest size of a product-shattered sequence. When $\ell=1$, a family of singletons with no common label must contain two different labels, so $\dN$ recovers the Natarajan dimension.

Two examples illustrate the definition. Unions of at most $r$ intervals on the line, where both labels $0$ and $1$ are correct at the endpoints of the intervals, form a class with $\dN=2r$, matching the VC dimension of unions of $r$ intervals (Example~\ref{ex:intervals}). In nearest-site classification on the line, a hypothesis is given by $k$ sites $s_1<\cdots<s_k$, and the correct labels at an input are the indices of its nearest sites, so two labels are correct at each midpoint between adjacent sites; this class has $\dN=k-1$ (Example~\ref{ex:sites}).

The \emph{joint DS dimension} $\dDS(\H)$ uses a weaker, local condition. A \emph{pattern}
on $x_1,\ldots,x_n$ is a tuple
$(h(x_1),\ldots,h(x_n))$ for some $h\in\H$ defined on
these inputs. A \emph{DS witness} is a finite nonempty set $V$ of patterns with the following property: for each $v\in V$ and each $x_i$, the patterns of $V$ that agree with $v$ on all other inputs have no common label at $x_i$. In other words, revealing the correct-label sets at all inputs but one never leaves a label that is correct at the hidden input for all consistent patterns. The dimension $\dDS(\H)$ is the largest $n$ for which there exist distinct $x_1,\ldots,x_n$ that admit a DS witness. When $\ell=1$, it reduces to the usual DS dimension. Section~\ref{sec:prelim-dimensions} gives the precise definitions of these and the other dimensions below.

\begin{atheorem}[Full-set and bandit feedback]\label{thm:A}
Let $\Y$ be finite, and let $\H$ be a class whose correct-label sets have size at most $\ell$.
The following are equivalent.
\begin{itemize}
\item[--] $\dN(\H)<\infty$.
\item[--] $\dDS(\H)<\infty$.
\item[--] $\H$ is PAC learnable from full sets, in the realizable or, equivalently, in the agnostic setting.
\item[--] $\H$ is PAC learnable with bandit feedback, in the realizable or, equivalently, in the agnostic setting.
\end{itemize}
Moreover, if $d=\dN(\H)<\infty$ and $k=|\Y|$, then
\[
\mfull_\H(\eps,\delta)=O_\ell\Bigl(\frac{1}{\eps}\Bigl(d\log\frac{(d+1)(k+1)}{\eps}+\log\frac{2}{\delta}\Bigr)\Bigr).
\]
\end{atheorem}

For $d\ge2$ and fixed failure probability, the dependence on $d/\eps$ is optimal up to logarithmic factors (Proposition~\ref{prop:nat-lower}). Under bandit feedback our rates carry an extra factor $k$, since a uniform query tests a given label only with probability $1/k$. The qualitative part of Theorem~\ref{thm:A} for realizable full-set learning can also be derived from \cite{bressan2025fine} (Appendix~\ref{app:cost-sensitive}); our direct construction adds a uniform bound with only logarithmic dependence on $k$ for fixed $\ell$.

The common starting point of our algorithms is a
transductive rule that knows a pool of $n$ inputs
in advance and makes at most
$O_\ell(d\log(nk))$ mistakes. Its mistake bound
yields a realizable full-set learner and a finite
cover which is used for agnostic and bandit learning.
Section~\ref{sec:overview} explains the counting
argument behind the rule, and the proofs are given
in Sections~\ref{sec:finite-full}
and~\ref{sec:finite-bandit}
(Theorems~\ref{thm:finite-full}
and~\ref{thm:finite-bandit}).

The next result shows that learning from a single correct label is governed by a different class. Let $\cI_{\ell+1}(\H)$ be the class of all intersections $h_1\cap\cdots\cap h_r$ of $r\leq\ell+1$ hypotheses of $\H$, restricted to any set of inputs on which the intersection is nonempty. If $\H$ is learnable from demonstrations, then $\cI_{\ell+1}(\H)$ is learnable from full sets. Indeed, a label in the intersection is a valid demonstration for every $h_i$, so the demonstration learner, fed with such labels, has small error for every $h_i$ simultaneously and hence, by a union bound, for their intersection. The converse is the main technical result of this paper.

\begin{atheorem}[One correct label]\label{thm:B}
Let $\Y$ be finite with $k=|\Y|$, and let $\H$ be a class whose correct-label sets have size at most $\ell$.
The following are equivalent.
\begin{itemize}
\item[--] $\H$ is PAC learnable from one correct label.
\item[--] $\cI_{\ell+1}(\H)$ is PAC learnable from full sets.
\item[--] $\dN(\cI_{\ell+1}(\H))<\infty$.
\item[--] $\dDS(\cI_{\ell+1}(\H))<\infty$.
\end{itemize}
Moreover, if $d=\dN(\cI_{\ell+1}(\H))<\infty$, then
\[
\mone_\H(\eps,\delta)=O_\ell\Bigl(\frac{d+1}{\eps^{\ell}}\log^2\Bigl(\frac{(d+2)(k+1)}{\eps}\Bigr)\log\frac{2}{\delta}\Bigr).
\]
\end{atheorem}

The intersection class cannot be replaced by $\H$ itself, even for a class learnable from a single full-set example. In Example~\ref{ex:three-labels}, every hypothesis but one accepts the label $1$ everywhere, together with a label in $\{2,3\}$ that depends on the input, and the remaining hypothesis accepts exactly $2$ and $3$ everywhere. The dimensions $\dN$ and $\dDS$ of this class equal one, and a single full-set example, or a single bandit answer, tells the two kinds of targets apart, after which a constant prediction is always correct. A demonstrator who never shows the label $1$, however, reveals nothing about the kind of the target, and the class is not learnable from one correct label.

The number $\ell+1$ comes from a simple fact: if sets of size at most $\ell$ have no common label, then at most $\ell+1$ of them already have no common label (Lemma~\ref{lem:small}). Deeper intersections therefore add nothing for learnability: if $\cI_{\ell+1}(\H)$ is learnable from full sets, then so is $\cI_R(\H)$ for every fixed $R$ (Corollary~\ref{cor:all-depths}). The dependence
on $\eps^{-\ell}$ is also necessary in the worst case,
as Theorem~\ref{thm:C} shows. The proof of
Theorem~\ref{thm:B} is in Sections~\ref{sec:reduction}
and~\ref{sec:finite-one}
(Theorem~\ref{thm:finite-one}).

\begin{atheorem}[The exponent $\ell$ is necessary]\label{thm:C}
For every $\ell\geq2$ and $d\geq1$, there is a class $\H$ with label space $[\ell]$ such that $d\leq\dN(\cI_{\ell+1}(\H))\leq d+(\ell+1)^\ell$ and
\[
\mone_\H(\eps,1/8)>\frac{d}{4(32(\ell+1)\eps)^{\ell}}
\qquad\text{for every } 0<\eps\leq1/256.
\]
\end{atheorem}

Since $\H\subseteq\cI_{\ell+1}(\H)$, this class has $\dN(\H)\leq d+(\ell+1)^\ell$, so by Theorem~\ref{thm:A} it is learnable from full sets with $\widetilde O_\ell((d+\log(1/\delta))/\eps)$ examples. Thus one-correct-label feedback can cost a factor of order $\eps^{-(\ell-1)}$ compared with full-set feedback. This is a statement about the worst class of a given dimension: some classes with the same dimension are learnable from demonstrations with $O((d+\log(1/\delta))/\eps)$ examples (Remark~\ref{rem:worst-only}). The proof is in Section~\ref{sec:worst-case} (Theorem~\ref{thm:worst-case}).

\subsubsection*{Countable Label Spaces}
When $\Y$ is infinite, finiteness of the generalized Natarajan dimension no longer suffices, already for $\ell=1$ \cite{brukhim2022characterization}. One might expect the joint DS dimension to take over, as in the multiclass case. We show that it does not once $\ell\geq2$, and we introduce a dimension that characterizes learnability. The difficulty is that, with sets of size two or more, an empty intersection may force very few errors: in a DS witness, if all but one of the candidates that remain at the hidden input contain the label $1$, predicting $1$ is wrong for only one of them. Our dimension therefore asks for many disjoint small conflicts at every input, each of which forces an error whatever the learner predicts.

Fix $0<\gamma<1/(\ell+1)$. Consider a finite nonempty list of patterns on distinct inputs $x_1,\ldots,x_n$, possibly with repetitions. A \emph{conflict block} at $x_i$ is a set of at most $\ell+1$ entries of the list that agree on all inputs other than $x_i$ and have no common label at $x_i$, so every label predicted at $x_i$ is wrong for at least one entry of the block. The \emph{packed dimension} $\dpack(\H)$ is the largest $n$ for which some list has, at every $x_i$, pairwise disjoint conflict blocks that together contain at least $\gamma$ times as many entries as the list.\footnote{Whether $\dpack(\H)$ is finite does not depend on the choice of $\gamma$ in this range (Remark~\ref{rem:fixed-gamma}).} Such a list forces errors on at least a $\gamma/(\ell+1)$ fraction of its entries at every input.

\begin{atheorem}[Full-set feedback]\label{thm:D}
Let $\Y$ be countable, and let $\H$ be a class whose correct-label sets have size at most $\ell$.
The following are equivalent.
\begin{itemize}
\item[--] $\dpack(\H)<\infty$.
\item[--] $\H$ is PAC learnable from full sets.
\item[--] $\H$ is agnostically PAC learnable from full sets.
\end{itemize}
Moreover, if $p=\dpack(\H)<\infty$ and $B=p+1+\log(2/\delta)$, then
\[
\begin{aligned}
\mfull_\H(\eps,\delta)&=O_{\ell,\gamma}\Bigl(\frac{B}{\eps}\log^2\frac{B}{\eps}\Bigr),\\
\mfullag_\H(\eps,\delta)&=O_{\ell,\gamma}\Bigl(\frac{B}{\eps^2}\log^2\frac{B}{\eps}\Bigr).
\end{aligned}
\]
In contrast, finiteness of the joint DS dimension is neither necessary nor sufficient: for every $\ell\geq2$, there are two total classes with at most two correct labels per input such that one has $\dDS(\H)=1$ and is not learnable from full sets, and the other has $\dDS(\H)=\infty$ and is learnable from full sets.
\end{atheorem}

For the upper bounds, the packed dimension yields a weak learner, which we boost into a sample compression scheme. The proof is in Section~\ref{sec:countable-full} (Theorem~\ref{thm:countable-full}), and the two separating classes are in Appendix~\ref{app:separations} (Theorems~\ref{thm:rank-example} and~\ref{thm:large-ds-learnable}).

For learning from demonstrations, the reduction to the intersection class holds for countable label spaces as well. Together with the preceding full-set characterization,
this gives a characterization in terms of the packed
dimension of $\cI_{\ell+1}(\H)$. The next theorem also gives sufficient conditions in terms of $\dDS$ and of the \emph{disjoint-neighbor dimension} $\dpair$, which strengthens the DS condition: for each pattern $v$ of the witness and each $x_i$, some pattern of the witness that agrees with $v$ on all other inputs must have a correct-label set at $x_i$ that is disjoint from that of $v$.

\begin{atheorem}[One correct label]\label{thm:E}
Let $\Y$ be countable, and let $\H$ be a class whose correct-label sets have size at most $\ell$.
The following are equivalent.
\begin{itemize}
\item[--] $\H$ is PAC learnable from one correct label.
\item[--] $\cI_{\ell+1}(\H)$ is PAC learnable from full sets.
\item[--] $\dpack(\cI_{\ell+1}(\H))<\infty$.
\end{itemize}
Moreover, if $p=\dpack(\cI_{\ell+1}(\H))<\infty$, then
\[
\mone_\H(\eps,\delta)=O_{\ell,\gamma}\Bigl(\frac{p+1}{\eps^{\ell}}\log^2\frac{2}{\eps}\log\frac{2}{\delta}\Bigr).
\]
Finally, if $D=\dDS(\cI_{\ell+1}(\H))$ or $D=\dpair(\cI_{\ell+1}(\H))$ is finite, then $p=O_{\ell,\gamma}(D\log(D+2))$, so $\H$ is PAC learnable from one correct label.
\end{atheorem}

The reduction converts every full-set learner for $\cI_{\ell+1}(\H)$ with small constant error into a demonstration learner whose sample size is linear in that of the full-set learner (Corollary~\ref{cor:blackbox}). The packed dimension supplies such a learner with $O_{\ell,\gamma}(p+1)$ examples, which is why the bound is linear in $p+1$. The proof is in Section~\ref{sec:countable-one} and Appendix~\ref{app:comparison}. Whether finiteness of $\dDS(\cI_{\ell+1}(\H))$ is also necessary is left open (Section~\ref{sec:conclusion}).

\subsubsection*{Agnostic Learning from One Correct Label}
Under full-set feedback, and under bandit feedback with finite label spaces, realizable and agnostic learnability coincide (Theorems~\ref{thm:A} and~\ref{thm:D}), as in the multiclass case \cite{david2016supervised,pmlr-v178-hopkins22a,hanneke2025representation}. Under one-correct-label feedback they do not. A demonstration reveals only one label of the sampled set, so different distributions of correct-label sets may produce the same demonstrations and yet call for different predictions. In particular, every distribution of demonstrations is also produced by singleton sets, namely the demonstrated labels themselves, and the learner cannot rule this out.

Our characterization uses two conditions on $\H$ and an associated multiclass class. We call $\H$ \emph{nondegenerate} if it contains two hypotheses that are defined at every input, have singleton values, and disagree at every input. These hypotheses play the role of the two constant hypotheses in binary classification; for example, every class containing $x\mapsto\{1\}$ and $x\mapsto\{2\}$ is nondegenerate.

We call $\H$ \emph{locally compatible} if, at each input $x$, every hypothesis with at least two labels at $x$ contains the label of every hypothesis with a single label at $x$. Therefore, if a learner predicts only labels that
occur as singleton values at $x$, its prediction is accepted
by every hypothesis with at least two correct labels there. 

This motivates defining the \emph{collapsed class} $\widetilde\H$, which replaces each singleton $\{a\}$ with the label $a$ and each set of at least two labels with one new label $\diamond$, while undefined values stay undefined. It is a multiclass class over $\Y\cup\{\diamond\}$, so $\dDS(\widetilde\H)$ is its usual DS dimension.

\begin{atheorem}[Agnostic learning from one correct label]\label{thm:F}
Let $\Y$ be countable, and let $\H$ be a class whose correct-label sets have size at most $\ell$. Assume that $\X$ is countable, or that $\X$ is standard Borel and $\H$ satisfies the regularity assumptions of Appendix~\ref{app:measurability}.
\begin{itemize}
\item[--] If $\H$ is nondegenerate, then $\H$ is agnostically PAC learnable from one correct label if and only if $\dDS(\widetilde\H)<\infty$ and $\H$ is locally compatible.
\item[--] Nondegeneracy cannot be dropped. For $\ell=2$, there is a locally compatible class $\H$ on a single input with three labels such that $\dDS(\widetilde\H)=0$ and the four dimensions $\dN$, $\dDS$, $\dpair$ and $\dpack$ of $\cI_3(\H)$ all equal one, but every learner, at every sample size, has expected error at least $\Lset_\D(\H)+1/9$ under some distribution and demonstrator. In particular, $\H$ is not agnostically PAC learnable from one correct label.
\end{itemize}
\end{atheorem}

The class in the second part consists of the hypotheses with correct-label sets $\{1,2\}$, $\{1,3\}$ and $\{2,3\}$ at its only input, and three distributions of correct-label sets give uniform demonstrations but have different best predictions. The class of the three hypotheses $x\mapsto\{j\}$ on one input has an intersection class with the same four dimensions and is agnostically learnable (Remark~\ref{rem:agnostic-c0}), so these dimensions do not determine agnostic learnability.

For the upper bound in the first part, the learner treats separately the inputs where every defined hypothesis value is a singleton, on which an agnostic multiclass learner suffices, and the remaining inputs, on which it predicts only labels that occur as singleton values. Conversely, local compatibility is necessary for every class, and the DS condition follows from a multiclass lower bound and an indistinguishability argument. Section~\ref{sec:overview} explains both directions. The proofs are in Section~\ref{sec:agnostic} (Theorems~\ref{agn:thm:main} and~\ref{thm:agnostic-one-impossible}).

\subsubsection*{Limits of Dimensions}
All the results above assume a uniform bound $\ell$ on the number of correct labels, and our last result shows that this assumption cannot simply be dropped. Say that a parameter $\mathsf{D}$ of classes has \emph{finite witnesses} if $\mathsf{D}(\H)$ is the supremum of $\mathsf{D}$ over the finite subclasses of $\H$; all the dimensions above have this property.

\begin{atheorem}[Impossibility of scale-free and scale-sensitive dimensions]\label{thm:G}
Consider countable classes of total hypotheses over countable instance and label spaces.
\begin{itemize}
\item[(a)] \emph{Scale-free.} Under full-set or one-correct-label feedback, with correct-label sets that are finite but not uniformly bounded in size, no parameter with finite witnesses is finite exactly on the PAC learnable classes, and no parameter $\mathsf{D}$ taking integer values on all learnable classes admits a finite-valued $\Psi$ that bounds the sample complexity of every learnable class $\H$ by $\Psi(\mathsf{D}(\H),\eps,\delta)$.
\item[(b)] \emph{Scale-sensitive.} Under one-correct-label feedback, with correct-label sets that may be infinite, consider families $(\mathsf{D}_\alpha)_\alpha$ of relabeling-invariant parameters with finite witnesses. No such family is finite at all scales exactly on the learnable classes. Nor does any such family taking integer values on all learnable classes admit a finite-valued $\Psi$ and a scale map $\eps\mapsto\alpha(\eps)$ that bound the sample complexity of every learnable class $\H$ by $\Psi(\mathsf{D}_{\alpha(\eps)}(\H),\eps,\delta)$.
\end{itemize}
\end{atheorem}


The proofs are in Section~\ref{sec:limits} (Theorems~\ref{thm:scale-free} and~\ref{thm:scale-sensitive}). Related limitations of dimensions were found by \cite{lechner2024inherent} for learning classes of distributions.

\subsubsection*{Hypergraph Coloring}
The counting argument behind the transductive rule of Theorem~\ref{thm:A} is not specific to learning. Consider an $r$-uniform hypergraph whose edges are covered by $m$ complete $r$-partite hypergraphs with forbidden crossings, meaning that for any two of them, some part of one is disjoint from some part of the other. The argument shows that such a hypergraph has chromatic number $\exp(O_r(\log^2(m+1)))$ (Appendix~\ref{app:coloring}). This extends the quasipolynomial bound of \cite{mubayi2009bipartite}, for partitions of the edges of a graph into complete bipartite graphs, to $r$-uniform hypergraphs and to covers with forbidden crossings. For partitions of graphs, \cite{fox2026note} gives a sharper leading constant in the exponent.

\subsection{Organization}\label{sec:organization}
Section~\ref{sec:related} discusses related work, Section~\ref{sec:overview} gives an overview of the proofs, and Section~\ref{sec:conclusion} concludes with open problems. The technical part begins with the formal definitions in Section~\ref{sec:prelim} and four examples in Section~\ref{sec:examples}. Theorem~\ref{thm:A} is proved in Sections~\ref{sec:finite-full} and~\ref{sec:finite-bandit}, Theorem~\ref{thm:B} in Sections~\ref{sec:reduction} and~\ref{sec:finite-one}, Theorem~\ref{thm:C} in Section~\ref{sec:worst-case}, Theorem~\ref{thm:D} in Section~\ref{sec:countable-full}, Theorem~\ref{thm:E} in Section~\ref{sec:countable-one}, Theorem~\ref{thm:F} in Section~\ref{sec:agnostic}, and Theorem~\ref{thm:G} in Section~\ref{sec:limits}. The appendices contain the separating examples for Theorem~\ref{thm:D} (Appendix~\ref{app:separations}), the comparison of dimensions for Theorem~\ref{thm:E} (Appendix~\ref{app:comparison}), the hypergraph coloring bound (Appendix~\ref{app:coloring}), and general instance spaces (Appendix~\ref{app:measurability}).

\section{Related Work}\label{sec:related}

\paragraph{Multiclass PAC learning.}
When every example has exactly one correct answer ($\ell=1$), learning from full sets or from demonstrations is exactly multiclass PAC learning. For finite label spaces, learnability is characterized by the Natarajan dimension \cite{natarajan1989learning,ben1995characterizations}. Infinite label spaces behave differently: Daniely et al.~\cite{daniely2011multiclass} showed that empirical risk minimizers for the same class can have very different sample complexities, and Daniely and Shalev-Shwartz~\cite{daniely2014optimal} introduced the DS dimension and proved that its finiteness is necessary for learnability. Brukhim et al.~\cite{brukhim2022characterization} proved that it is also sufficient, and that finite Natarajan dimension does not suffice. Their learner combines the one-inclusion graph \cite{haussler1994predicting,rubinstein2006shifting,daniely2014optimal} with list learning and sample compression. List learning, in which the learner outputs a short list of labels, was also characterized in the PAC setting by Charikar and Pabbaraju~\cite{charikar2023characterization}; it is dual to our problem, in which the target, not the learner, provides several labels. Recently, building on an optimal Sauer lemma over $k$-ary alphabets \cite{hanneke2026optimalsauer}, Pabbaraju~\cite{pabbaraju2026multiclass} determined the optimal sample complexity of multiclass and list learning. Our results show which parts of this picture survive when an input has up to $\ell$ correct labels. For finite label spaces, the generalized Natarajan dimension still characterizes learnability from full sets (Theorem~\ref{thm:A}). For countable label spaces, however, the joint DS dimension no longer does so once $\ell\geq2$, and the packed dimension takes its place (Theorem~\ref{thm:D}).

\paragraph{Partial concept classes.}
Alon et al.~\cite{alon2021theory} developed the theory of PAC learning for partial concept classes, whose hypotheses may be undefined on part of the domain. We use this framework even for total classes $\H$, because the intersection class $\cI_{\ell+1}(\H)$ of Theorems~\ref{thm:B} and~\ref{thm:E} may be partial: an intersection of correct-label sets may be empty.

\paragraph{Online learning with several correct answers.}
Raman, Subedi and Tewari~\cite{raman2024online} study a model in which the learner predicts one label and then observes the whole set of correct labels, an online counterpart of full-set feedback. They show that deterministic and randomized online learnability differ, and they characterize both by dimensions in the realizable setting. Pour, Mansouri and Ben-David~\cite{pour2026learning} study online learning with multiple correct answers under three feedback models. For each model, they determine the optimal realizable mistake bound in terms of a combinatorial dimension, describe the possible agnostic regret rates, and derive batch sample complexity bounds from the same dimensions. The agnostic obstruction in the second part of Theorem~\ref{thm:F} adapts their indistinguishability argument. Shao et al.~\cite{shao2026partial} study an online model in which the learner observes one correct label per round, and Kosoy~\cite{kosoy2026ambiguous} studies one in which a prediction may consist of several labels. We work directly in the PAC setting and characterize both realizable and agnostic learnability, from full sets and from demonstrations for finite and countable label spaces, and from bandit feedback for finite ones. For agnostic learning from demonstrations, the characterization holds for nondegenerate classes.

\paragraph{Learning from correct demonstrations.}
Closest to our work is the statistical study of learning to answer from correct demonstrations by Joshi et al.~\cite{joshi2026learning}. For each training prompt, the learner observes one correct answer, and an unknown reward model from a known class specifies which answers are correct. For finite reward classes, their sample complexity is logarithmic in the class size and inverse-linear in accuracy, up to logarithmic factors, even for adaptive correct demonstrations. They also show that maximizing demonstration likelihood can fail. Theorems~\ref{thm:B} and~\ref{thm:E} characterize realizable learnability from i.i.d.\ demonstrations for possibly infinite classes with at most $\ell$ correct labels per input: such a class is learnable exactly when its intersection class $\cI_{\ell+1}(\H)$ is learnable from full sets. The dependence on the accuracy can then be far worse than inverse-linear: the classes of Theorem~\ref{thm:C} are infinite and have finite $\dN(\cI_{\ell+1}(\H))$, yet their sample complexity is of order at least $\eps^{-\ell}$.

\paragraph{Partial-label learning.}
Partial-label, or superset, learning \cite{cour2011learning,liu2014learnability} reverses the role of the label set: each example comes with a set of candidate labels of which only one is correct. In our model every label in the set is correct, and under one-correct-label feedback the learner observes a single element of this set.

\paragraph{Agnostic one-inclusion graphs.}
The packed dimension gives a combinatorial certificate of
difficulty in the weighted realizable one-inclusion prediction
game, using conflict blocks of at most $\ell+1$ hypotheses.
The analogous agnostic full-set game allows arbitrary assignments
of correct-label sets and subtracts their minimum exact-set
disagreement with $\H$, just as the agnostic one-inclusion
framework of \cite{pmlr-v247-asilis24a} subtracts
Hamming-distance credits. Our packed witnesses consist of
realizable assignments, for which these credits are zero.

\paragraph{Bandit feedback.}
Under bandit feedback, the learner proposes a label and observes only whether it is correct. For multiclass classification, this feedback has been studied in the online setting \cite{kakade2008efficient,daniely2013price,long2020new,raman2024multiclass,erez2024real,filmus2024bandit,hanneke2025equivalent,hanneke2025nonuniform} and in the statistical setting \cite{erez2024fast,hanneke2025universal,hanneke2025nonuniform,hanneke2026pac}. With several correct labels, a positive answer reveals one correct label and a negative answer excludes one label. For finite label spaces, this feedback still suffices: bandit learnability is characterized by the same dimension as full-set learnability (Theorem~\ref{thm:A}). Our bandit learners make independent uniform queries during training, as in the batch models of Daniely et al.~\cite{daniely2011multiclass} and Hanneke et al.~\cite{hanneke2026pac}.

\paragraph{Limits of dimensions.}
Ben-David et al.~\cite{bendavid2019learnability} showed that EMX (estimating the maximum) learnability can be independent of ZFC set theory, which rules out a characterization of EMX learnability by a dimension of finite character. Hanneke and Yang~\cite{hanneke2023bandit} established independence from ZFC for identifying near-optimal arms in structured bandits. For classes of distributions, Lechner and Ben-David~\cite{lechner2024inherent} established limitations of dimensions as characterizations of learnability, and they showed that no scale-sensitive dimension characterizes the learnability of such classes \cite{lechner2026no}. Theorem~\ref{thm:G} is a result of the same kind for multi-label PAC learning. When correct-label sets are finite but not uniformly bounded, no parameter with finite witnesses (a requirement in the spirit of finite character) is finite exactly on the learnable classes. This holds under full-set and under one-correct-label feedback, and here the obstruction is combinatorial rather than set-theoretic.


\section{Technical Overview}\label{sec:overview}

Most of the difficulty in our setting comes from one mismatch: the learner has to output only one of several correct labels, while the data may reveal all of them, only one of them, or only whether a guess is correct. For each upper bound we start from the natural first approach, show where this mismatch breaks it, and then describe what works. For the lower bounds and obstructions we describe the hard classes directly. Formal definitions are in Section~\ref{sec:prelim}.

Section~\ref{ov:finite} treats full-set feedback over a finite label space through a potential argument, and Section~\ref{ov:one} reduces learning from one demonstrated label to full-set learning of the intersection class. We then extend both results to countable label spaces in Section~\ref{ov:countable}, and show in Section~\ref{ov:lower} that the factor $\eps^{-\ell}$ in Theorem~\ref{thm:B} cannot be avoided. We end with agnostic learning from demonstrations in Section~\ref{ov:agnostic}, and with classes that have no uniform bound on the number of correct labels in Section~\ref{ov:limits}, where we show that no parameter with finite witnesses characterizes learnability.

\subsection{Full Sets and Finite Label Spaces}\label{ov:finite}

When $k=|\Y|$ is finite, a natural first idea is to view $\H$ as a multiclass class whose labels are the correct-label sets, and to learn the whole set $h^\star(x)$ by applying the characterization based on the Natarajan dimension of this class. This can be much harder than predicting one correct label. For example, on an infinite $\X$, let $\H$ contain every $h$ with $h(x)\in\{\{1,2\},\{1,3\}\}$ for all $x$. As a multiclass class, $\H$ has infinite Natarajan dimension, yet the constant predictor $1$ is always correct without requiring any training samples.

A dimension for our problem should therefore count only choices of sets with no common label. This is what the generalized Natarajan dimension $\dN(\H)$ introduced in Section~\ref{sec:results} does: product-shattering uses only families of sets with empty intersection. Product-shattered inputs force errors under full-set feedback, and a DS witness forces errors under each of our feedback models; these lower bounds complete the characterization (Propositions~\ref{prop:ds-lower} and~\ref{prop:nat-lower}). The rest of this subsection describes the upper bounds.

\paragraph{Counting realized products.}
Our main tool is a transductive rule that receives a pool $S=(x_1,\ldots,x_n)$ of inputs. The positions of
$S$ are then presented one at a time, in any order. At each step, the rule predicts a label and then observes the correct-label set. We show that it makes $O_\ell(d\log(nk))$ mistakes, where $d=\dN(\H)$.

For binary classes, where $\dN$ is the VC dimension, such a rule can be obtained by counting shattered sets. Let the version space $W$ be the set of patterns on $S$ that agree with the labels seen so far, let $\Phi(W)$ be the number of sets of positions shattered by $W$, the empty set included, and let $W_{i,\{y\}}$ consist of the patterns of $W$ with label $y$ at position $i$.

Every set shattered by $W_{i,\{0\}}$ or by $W_{i,\{1\}}$ omits $i$ and is shattered by $W$. If a set is shattered by both, then adding $i$ to it gives a set shattered by $W$ that neither half shatters. Thus every set counted twice in the sum $\Phi(W_{i,\{0\}})+\Phi(W_{i,\{1\}})$ is paired with a set counted zero times, and hence the sum is at most $\Phi(W)$. So at most one label $y$ has $\Phi(W_{i,\{y\}})>\Phi(W)/2$, and predicting such a label, or either label if none exists, at least halves $\Phi(W)$ on every mistake. Since only sets of at most $d$ positions can be shattered, initially $\Phi(W)\le(n+1)^d$, and the rule makes at most $d\log_2(n+1)$ mistakes.

We extend this rule to correct-label sets. The version space $W$ is now a set of tuples on $S$; initially it consists of the tuples $(h(x_1),\ldots,h(x_n))$ for all $h\in\H$. A \emph{product realized by} $W$ specifies a set $T$ of positions and, at each $i\in T$, a minimal family of sets with empty intersection (no proper subfamily has empty intersection), such that every combination of one set from each family occurs in $W$ at the positions of $T$. For binary classes, these products are exactly the shattered sets.

Let $\Phi(W)$ now be the number of products realized by $W$, including the empty product with $T=\emptyset$. Each of them product-shatters the inputs at the positions of $T$, so $|T|\le d$. A minimal family has at most $\ell+1$ members (Lemma~\ref{lem:small}), and as each member
has size at most $\ell$, there are at most $k^{O_\ell(1)}$ such
families. Hence,  $\log\Phi(W)=O_\ell(d\log(nk))$.

For a set $A$, write $W_{i,A}$ for the tuples of $W$ with value $A$ at position $i$. The corresponding inequality for
set-valued hypotheses (Lemma~\ref{lem:potential-inequality}) is that,  for every minimal family $\{A_1,\ldots,A_r\}$ with empty intersection,
\[
 \sum_{j=1}^r\Phi(W_{i,A_j})\le(r-1)\,\Phi(W).
\]
Indeed, no product realized by some $W_{i,A_j}$ uses position $i$, so a product of $W$ that uses position $i$ is counted zero times on the left. A product counted $r$ times is realized by every $W_{i,A_j}$, and adding the family $\{A_1,\ldots,A_r\}$ at position $i$ turns it into a product of $W$ that uses position $i$. Thus every product counted $r$ times is paired with a distinct product counted zero times, and every other product is counted at most $r-1$ times.

Since $r\le\ell+1$, this inequality implies that the sets $A$ with $\Phi(W_{i,A})>\frac{\ell}{\ell+1}\Phi(W)$ have a common label. Indeed, otherwise, by Lemma~\ref{lem:small}, some minimal subfamily of $r\le\ell+1$ of them has empty intersection, and the left side exceeds $r\cdot\frac{\ell}{\ell+1}\Phi(W)\ge(r-1)\Phi(W)$. The rule predicts such a common label, or any label if there are no such sets. After a mistake, the revealed set $A$ does not contain the prediction, so it is not one of these sets, and replacing $W$ by $W_{i,A}$ multiplies $\Phi(W)$ by at most $\ell/(\ell+1)$. The tuple of the target always remains in $W$, so $\Phi(W)\ge1$ throughout. Hence the rule makes $O_\ell(d\log(nk))$ mistakes on every pool and in every order of presentation (Theorem~\ref{thm:online}). Standard leave-one-out arguments turn this mistake bound into the realizable learners (Theorems~\ref{thm:full-expected} and~\ref{thm:hp}).

\paragraph{Finite covers.}
The rule changes its version space only after a mistake, so its run against a target is determined by a short list: the at most $O_\ell(d\log(nk))$ positions where it makes a mistake, together with the sets revealed there. For each such list, we simulate the rule with these answers and record its predictions, with the prediction at each listed position replaced by a label of the revealed set. These vectors form a \emph{finite cover} $F_S$ of $\H$ on $S$: every $h\in\H$ has a member that is correct at every input of $S$ where $h$ is defined. This cover can be built from the inputs $S=(x_1,\ldots,x_n)$ alone, without any correct-label sets, and $\log|F_S|=O_\ell(d\log^2(nk))$ (Lemma~\ref{lem:finite-cover}). Running exponential weights over its members on the training inputs and the test input, in a random order, gives the agnostic full-set learner (Theorem~\ref{thm:agnostic-full}).

\paragraph{Bandit feedback.}
On a known pool, the cover gives a bandit learner by elimination: predict the label recommended by the largest number of surviving members of $F_S$, and after a negative answer remove every member that recommended it. Each mistake removes at least a $1/k$ fraction of the survivors, since the predicted label has the most votes, and a member that is correct for the target is never removed (Proposition~\ref{prop:trans-bandit}). In the batch setting, the difficulty is that $F_S$ depends
on the test input, which is not known when the training
queries are made. We overcome this by querying an independent
uniformly random label at each training input and recording
the answer. Once the test input arrives, the learner
constructs the cover of the combined pool and processes the
recorded queries and answers in a uniformly random order.
A query reveals that an expert is wrong if the queried label
is the one recommended by that expert and the answer is
negative. Thus a wrong expert is detected with probability
$1/k$, while a correct expert is never detected as wrong.
In the realizable case, the learner eliminates the detected
experts. In the agnostic case, multiplying the indicator of
this event by $k$ gives an unbiased estimate of the expert's
loss, which is used in exponential weights. A leave-one-out
argument then gives the expected-risk bounds
(Section~\ref{sec:finite-bandit}).

\subsection{One Demonstrated Label}\label{ov:one}

The easy direction of Theorem~\ref{thm:B}, from demonstrations for $\H$ to full sets for $\cI_{\ell+1}(\H)$, was explained in Section~\ref{sec:results} (Lemma~\ref{lem:common-demonstrations}). We now describe the converse, which Section~\ref{sec:reduction} proves for every countable label space.

A natural approach is to choose a hypothesis consistent with the demonstrations and predict one of its labels, as in realizable PAC learning. Consistency alone does not
suffice, since observing one correct label does not determine
the others. For instance, if $h^\star(x)=\{1,2\}$ and $h(x)=\{2,3\}$ at every input, a demonstrator who always shows the label $2$ makes $h$ consistent with every sample, while the label $3$ of $h$ is always wrong.  Rather than
choosing a single consistent hypothesis, we retain a
collection of candidates and seek predictions with small
error against all of them. By Lemma~\ref{lem:small}, if the
correct-label sets of a collection of candidates have empty
intersection at some input, then this is already witnessed
by at most $\ell+1$ candidates. Thus it suffices to control
agreement among groups of $\ell+1$ candidates, which
explains the role of $\cI_{\ell+1}(\H)$.

The learner works on the pool $S=(x_1,\ldots,x_N)$ that consists of the $N-1$ training inputs and the test input, and its candidates are the hypotheses of $\H$ defined on all of $S$. It draws a random set $J$ of $\lfloor N/2\rfloor$ training positions, uses only the demonstrations at these positions to select the candidates, and predicts on the remaining positions $U=[N]\setminus J$, which include the test position. We say that $\ell+1$ candidates \emph{conflict} at a position if their sets have empty intersection there. The argument has two steps. First, if the learner retains a set of candidates in which any $\ell+1$ members conflict only on a small part of $U$, then there is a randomized prediction rule with small average error on $U$ against every retained
candidate. Second, the demonstrations on $J$ suffice to retain such a set of candidates that contains the target with high probability.

\paragraph{Agreement of small groups suffices.}
Suppose that the learner has a collection of candidates that contains the target and in which any $\ell+1$ members of the collection conflict on at most a $\theta$ fraction of the positions of $U$. We show that there is a randomized prediction rule
whose average error on $U$ is at most $\theta^{1/\ell}$
against every candidate in the collection.  

By the minimax theorem, it suffices to show that for every distribution $\mu$ on the candidates, there are labels with small expected error against $h\sim\mu$. Against $\mu$, at each position $i\in U$, the best label at $x_i$ is wrong with probability
\[
 \nu_i=1-\max_{y\in\Y}\Pr_{h\sim\mu}[y\in h(x_i)],
\]
so every label $y$ satisfies $\Pr_{h\sim\mu}[y\notin h(x_i)]\ge\nu_i$. Draw $h_1,\ldots,h_{\ell+1}$ independently from $\mu$ and condition on $h_1(x_i)$, which has at most $\ell$ labels. Assign each of these labels to a different candidate among $h_2,\ldots,h_{\ell+1}$ (some may stay unassigned). Each assigned candidate excludes its label with probability at least $\nu_i$, independently of the others, and if all of them do, the $\ell+1$ sets have empty intersection at $x_i$. Hence $h_1,\ldots,h_{\ell+1}$ conflict at $x_i$ with probability at least $\nu_i^\ell$. Since any $\ell+1$ candidates conflict on at most a $\theta$ fraction of $U$, averaging over $i\in U$ gives
\[
 \frac1{|U|}\sum_{i\in U}\nu_i^\ell\le\theta .
\]
By Jensen's inequality, the average of the $\nu_i$ over $U$ is at most $\theta^{1/\ell}$. Thus, to obtain average error at most $\eps$,
we should bound the fraction of conflicts
by $\eps^\ell$. This is the source of the
$\eps^{-\ell}$ dependence in the sample
complexity bound.

\paragraph{Making conflicts rare.}
The demonstrations show agreement only on $J$: all candidates consistent with them share the demonstrated label at every position of $J$, but their correct-label sets need not intersect on $U$. To control
these conflicts,  we use a random-split argument with a union bound. However, there may be infinitely many candidates, so we apply it to finitely many label vectors instead. For every candidate $h$ defined on $S$, the learner uses a family $F_{h,S}$ of at most $B_N$ label vectors on $S$, determined by $h$ and $S$ alone. The families are chosen so that we have the following
\emph{intersection property}: for any $\ell+1$ candidates
$h_1,\ldots,h_{\ell+1}$, some
$f\in\bigcap_{j=1}^{\ell+1}F_{h_j,S}$ predicts a label in
$\bigcap_{j=1}^{\ell+1}h_j(x_i)$ whenever this intersection
is nonempty (Definition~\ref{def:prediction-families}). For a threshold $\alpha$ fixed below, the learner retains a candidate $h$ if
\begin{enumerate}[(i)]
\item $h$ accepts every demonstrated label in $J$, and
\item every $f\in F_{h,S}$ that is correct for $h$ on all of $J$ is wrong for $h$ on at most an $\alpha$ fraction of $S$.
\end{enumerate}
Both conditions depend only on the demonstrations and the candidate $h$, not on the unobserved correct-label
sets of the target.

By exchangeability, we may analyze the expected loss by choosing $J$ uniformly among the $\lfloor N/2\rfloor$-subsets of $[N]$, then choosing the test position uniformly from $U=[N]\setminus J$.
We first show that the target is retained with high
probability. It satisfies~(i) because its demonstrated labels are correct. For~(ii), a fixed $f$ that is wrong for $h^\star$ on more than an $\alpha$ fraction of the pool is correct on all of the random set $J$ with probability at most $e^{-\alpha\lfloor N/2\rfloor}$. Since $F_{h^\star,S}$ is
determined independently of $J$, a union bound over the $B_N$ members of $F_{h^\star,S}$ shows that the target is retained with probability at least $1-1/N$ once we take $\alpha$ of order $(\log B_N+\log N)/N$.

 We next show that the retained candidates rarely
conflict. Any $\ell+1$ of them all accept the demonstrated labels, so their sets intersect at every position of $J$, and the intersection property gives a vector $f$ in all of their families that is correct for all of them on $J$.  By~(ii), $f$ makes at most $\alpha N$ mistakes
against each candidate on the entire pool. At every
position where the candidates conflict, $f$ must be
wrong for at least one of them. Thus there are at most
$(\ell+1)\alpha N$ conflicting positions, and since
$|U|\ge N/2$, the fraction of conflicts on $U$ is
at most $2(\ell+1)\alpha$.

\paragraph{The resulting bound.}
With $\theta=O(\ell\alpha)$, the minimax step gives randomized predictions on $U$ whose average error against every retained candidate, and in particular against the target, is $O((\ell\alpha)^{1/\ell})$. By exchangeability, the test position is a uniform position of $U$, so this average error, plus the probability $1/N$ of discarding the target, bounds the expected test error (Theorem~\ref{thm:family-conversion}). Thus, with this choice of $\alpha$, $N$ of order $\eps^{-\ell}(\log B_N+\log N)$ gives expected error $\eps$. Repeating the learner on independent blocks and taking
a most probable label of the average prediction
distribution gives a high-probability guarantee with
an additional $O_\ell(\log(2/\delta))$ factor in the sample
size (Lemma~\ref{lem:amplification}).

For a finite label space, a single family serves every candidate: the finite cover of $\cI_{\ell+1}(\H)$ on $S$ from Section~\ref{ov:finite} contains, for any $\ell+1$ candidates, a vector that is correct for their intersection wherever it is defined. Hence $\log B_N=O_\ell(d\log^2(Nk))$ with $d=\dN(\cI_{\ell+1}(\H))$, which gives Theorem~\ref{thm:B} (Section~\ref{sec:finite-one}).

\subsection{Countable Label Spaces}\label{ov:countable}

When $\Y$ is infinite, the bounds above are vacuous because they grow with $k$. For $\ell=1$, learnability is characterized by the DS dimension \cite{brukhim2022characterization}, so the joint DS dimension is the natural candidate for larger $\ell$. For $\ell\ge2$, however, it fails in both directions (Theorem~\ref{thm:D} and Appendix~\ref{app:separations}).


The packed dimension of Section~\ref{sec:results} asks for many disjoint conflicts. A conflict block at $x_i$ is a group of at most $\ell+1$ entries of a list of patterns that agree outside $x_i$ and have no common label at $x_i$. A \emph{$\gamma$-packed witness} on inputs $x_1,\ldots,x_n$ is a list of patterns, with repetitions that act as weights, such that at every $x_i$ disjoint conflict blocks cover at least a $\gamma$ fraction of the list. A predictor that sees the other coordinates of the hidden pattern makes an error on at least one entry of each block, and hence on at least a $\gamma/(\ell+1)$ fraction of the list.

\paragraph{A weak learner.}
The packed dimension yields a weak learner through a one-inclusion argument. Fix a pool of $n$ inputs, a finite set $V$ of patterns on it, and a distribution $\mu$ on $V$. For each position $i$, a predictor that sees the correct-label sets at all other inputs knows the hyperedge of the hidden pattern, and its best choice is the label contained in the largest $\mu$-mass of patterns of that hyperedge; let $b_i$ be the resulting error probability. Lemma~\ref{lem:ambiguity} shows that $b_1+\cdots+b_n\le p+n\gamma$, where $p=\dpack(\H)$.

To see this, let $\mu$ be rational, so that it is a list of patterns with repetitions. In each hyperedge at position $i$, greedily remove minimal groups of entries whose sets at $x_i$ have empty intersection. Each removed group is a conflict block, and the remaining entries share a label at $x_i$, so the removed blocks cover at least the mass $b_i$ that the best label misses (Lemma~\ref{lem:pack-family}). If more than $p$ positions had $b_i>\gamma$, their blocks would form a packed witness on more than $p$ distinct inputs, which is impossible. So at most $p$ positions contribute at most $1$ each, the others contribute at most $\gamma$ each, and the sum is at most $p+n\gamma$.

Thus every distribution on patterns has a response whose total error over the $n$ positions is at most $p+n\gamma$. By the minimax theorem, and by compactness when there are infinitely many patterns, a single randomized transductive rule achieves this bound on every pattern. A leave-one-out argument then gives a full-set learner with expected error at most $\gamma+p/(m+1)$ from $m$ examples, and since $\gamma<1/(\ell+1)$, this is below $1/(\ell+1)$ for $m=O_{\ell,\gamma}(p+1)$ (Lemma~\ref{lem:packed-weak}).

\paragraph{Boosting and compression.}
We boost the weak learner with multiplicative weights over the sample positions and turn the result into a compression scheme. The threshold $1/(\ell+1)$ plays the role of $1/2$ in binary boosting: if a distribution on $\Y$ puts less than $1/(\ell+1)$ of its mass outside a set of at most $\ell$ labels, then each of its most probable labels lies in that set (Lemma~\ref{lem:mode}). In each round, some multiset of $M=O_{\ell,\gamma}(p+1)$ sample examples gives the weak learner error below $1/(\ell+1)$ under the current weights. After $O(\log N)$ rounds, the average of the weak predictions puts less than $1/(\ell+1)$ of its mass on incorrect labels at every sample input, so its most probable label is correct on the whole sample (Lemma~\ref{lem:boosted-reconstruction}). Listing the multisets of all rounds gives a sample compression scheme of size $O((p+1)\log(2N))$, and the realizable bound follows from the usual generalization argument for compression schemes. This follows the approach of \cite{freund1997decision,moran2016sample,david2016supervised,cohen2025sample} (Section~\ref{sec:countable-full}).

For agnostic learning, the learner decodes every short list of sample examples and outputs a decoded predictor of smallest empirical error. For any $h\in\H$, the examples on which $h$ equals the observed set form a realizable subsample, so some decoded predictor has empirical error at most the empirical set disagreement of $h$, and the standard concentration argument for compression schemes finishes the proof (Lemma~\ref{lem:agnostic-reconstruction}).

\paragraph{Prediction families from compression.}
To learn from one demonstrated label we need prediction families with the intersection property. For infinite $\Y$, a single finite cover need not exist, even when one example identifies the target. For instance, suppose our class consists of all constant hypotheses $x\mapsto\{y\}$, $y\in\Y$. On a pool of data, every label vector is correct for at most one of them. We therefore build the family of each candidate separately, from short descriptions of predictors.

Start from any full-set learner for $\cI_{\ell+1}(\H)$ that uses $M$ examples and has expected error below $1/(\ell+1)$, and run the boosting procedure on a pool of $N\ge2M$ inputs, with an intersection $g\in\cI_{\ell+1}(\H)$ as the target. After $T=O(\log(N/M))$ rounds at most $M$ positions are still wrong, and we store them as corrections. The resulting description of $g$ consists of $O(MT)$ pairs $(i,A)$, each a position of the pool together with the correct-label set of $g$ there, and it decodes to a vector that is correct wherever $g$ is defined.

For a candidate $h$, let $F_{h,S}$ consist of the decoded vectors of all descriptions whose set at each position $i$ is a nonempty subset of $h(x_i)$. There are at most $2^\ell-1$ such sets at each position, so
\[
 \log|F_{h,S}|=O\bigl(M\log^2(eN/M)\bigr),
\]
with no dependence on $|\Y|$. If $g$ is the intersection of $\ell+1$ candidates, every set in the description of $g$ is contained in the corresponding set of each of them. The same description therefore yields one vector in the family of every candidate, correct wherever $g$ is defined, which is the intersection property.

The conversion of Section~\ref{ov:one} then gives a one-correct-label learner whose sample size is linear in $M$ (Theorem~\ref{thm:feedback}), and the packed weak learner supplies $M=O_{\ell,\gamma}(p+1)$ with $p=\dpack(\cI_{\ell+1}(\H))$. Since $\log|F_{h,S}|$ depends on $N$ only through the ratio $N/M$, the bound has no logarithmic factor in $p$ (Section~\ref{sec:countable-one}).

Finally, Appendix~\ref{app:comparison} bounds $\dpack(\cI_{\ell+1}(\H))$ by $O(D\log(D+2))$, where $D$ is the joint DS or the disjoint-neighbor dimension of $\cI_{\ell+1}(\H)$ (Theorem~\ref{thm:comparison}).

\subsection{The Exponent \texorpdfstring{$\ell$}{l} Is Necessary}\label{ov:lower}

The minimax step of Section~\ref{ov:one} loses an $\ell$th root when it passes from $\nu_i$ to $\nu_i^\ell$. To show that this loss cannot be avoided in general, we construct a class in which, at an unseen input, the best single label is wrong with probability of order $1/n$, whereas $\ell+1$ plausible targets have no common label only with probability of order $n^{-\ell}$.

Let $\ell\ge2$, $\X=\N^\ell$ and $\Y=[\ell]$. For $t\in\N^\ell$ and a finite set $F\subseteq\N^\ell$ let
\[
 f_F(x)=1+\ind\{x\in F\},\qquad h_{t,F}(x)=\{f_F(x)\}\cup\{i\in[\ell]:x_i\ne t_i\}.
\]
Thus $h_{t,F}$ always accepts the binary label $f_F(x)\in\{1,2\}$, and it excludes a label $i\ne f_F(x)$ exactly on the slice $x_i=t_i$. For the lower bound, put the inputs uniformly on the grid $[n]^\ell$, draw $F\subseteq[n]^\ell$ by including every grid point independently with probability $1/2$, draw $t$ uniformly from $[n]^\ell$, and let the demonstrator always show $f_F(x)$. The demonstrations reveal nothing about $t$, since the demonstrated label $f_F(x)$ does not depend on $t$.

At an unseen input, $f_F(x)$ is still equally likely to be $1$ or $2$, so each of these labels is incorrect with probability $1/(2n)$, and every other label with probability $1/n$. Therefore, every label is incorrect with probability at least $1/(2n)$.

With at most $cn^\ell$ examples, for a small constant $c$, a constant fraction of the test inputs is unseen, and the expected error there is of order $1/n$. A second-moment argument shows that, for some target, the loss is $\Omega(1/n)$ with probability greater than $1/8$. Taking $n$ of order $1/\eps$ gives the lower bound $\Omega(\eps^{-\ell})$. The dimension stays bounded: an intersection of $\ell+1$ targets excludes each label on at most $\ell+1$ slices, and $\dN(\cI_{\ell+1}(\H))=(\ell+1)^\ell$ (Lemma~\ref{lem:hard-base-dimension}). Independent copies of the construction give the linear factor in the dimension (Section~\ref{sec:worst-case}).

To see how this construction relates to the upper bound, fix an unseen
input $x$ and draw $\ell+1$ targets independently from the conditional
distribution given the sample. Their correct-label sets at $x$
have empty intersection exactly when each of the $\ell$ labels
is excluded by at least one of them. For each label $i$, one of
the $\ell+1$ targets must have $t_i=x_i$, and the coordinates of
the targets are independent, so the empty-intersection probability
is at most $(\ell+1)^\ell/n^\ell$. Assigning a different target
to exclude each label gives a lower bound of $\Omega(n^{-\ell})$,
for fixed $\ell$. The construction thus matches the relation between $\nu_i$ and $\nu_i^\ell$ in the upper bound.

\subsection{Agnostic Learning from One Correct Label}\label{ov:agnostic}

A demonstration $(x,y)$ has two readings: the hidden set was a large set containing $y$, or it was exactly $\{y\}$. In the realizable setting this does not matter, since every demonstrated label is correct. In the agnostic setting the benchmark $\Lset_\D(\H)$ compares whole sets, so two distributions of sets can induce the same law of demonstrations, hence the same law of the learner's output, while calling for different predictions. This mechanism is behind every impossibility result below, and the conditions of Theorem~\ref{thm:F} are what rule it out.

\paragraph{Where the conditions come from.}
The class of Theorem~\ref{thm:agnostic-one-impossible}, with values $\{1,2\}$, $\{1,3\}$ and $\{2,3\}$ at a single input, shows the mechanism at work: three distributions of sets demonstrate a uniform label, yet each makes a different label the worst prediction, so every learner has excess error at least $1/9$ under one of them. The root of the problem is that no label lies in all three large sets: the learner must bet on which large set is hidden, and demonstrations cannot inform the bet. The dimensions of $\cI_3(\H)$ do not see this; they coincide with those for the learnable class of the three singletons $x\mapsto\{j\}$ (Remark~\ref{rem:agnostic-c0}).

The conditions of Theorem~\ref{thm:F} make such bets unnecessary. Call an input \emph{mixed} if some hypothesis has two or more labels there, and let $\Y_x$ be the set of labels $a$ with $h(x)=\{a\}$ for some $h\in\H$. Local compatibility places $\Y_x$ inside every large value at $x$, so a prediction in $\Y_x$ is correct whenever the hidden set is a large value of the class, whichever it is. Local compatibility is necessary by the same mechanism: if a singleton $\{a\}$ and a large set $B\not\ni a$ occur at one input, a mixture of $\{a\}$ and $B$ and a distribution of singletons demonstrate identically, yet the first calls for a label of $B$ and the second for $a$ (Proposition~\ref{agn:prop:lc}). Nondegeneracy, two total hypotheses with everywhere different singleton values, makes $|\Y_x|\ge2$ at every input, so safe labels always exist (for the class above, $\Y_x=\emptyset$); it also supplies the anchors of the lower bound below.

A learner predicting in $\Y_x$ then needs to know about $h(x)$ only whether it is undefined, large, or a singleton $\{a\}$, and which $a$. This is the collapsed value $\widetilde h(x)$: the collapse forgets the content of large sets because local compatibility makes it irrelevant. At a mixed input, $\Y_x$ lies in a set of at most $\ell$ labels, so $\widetilde h(x)\in\Y_x\cup\{\diamond\}$ takes at most $\ell+1$ defined values: the mixed region is a multiclass problem over a bounded alphabet even when $\Y$ is infinite. On the remaining \emph{pure} inputs every defined value is a singleton, so a hypothesis's error on demonstrated labels is at most its exact-set mismatch error. An agnostic multiclass learner therefore suffices. Both regions are determined by $\H$ alone, and both are governed by $\dDS(\widetilde\H)$.

\paragraph{Necessity.}
If $\dDS(\widetilde\H)=\infty$, the DS dimension is infinite on one of the regions. On the pure region the realizable multiclass lower bound applies (Lemma~\ref{agn:lem:ds-lower}). On the mixed region the bounded alphabet turns large DS witnesses into large Natarajan cubes, from which nondegeneracy and local compatibility extract an \emph{anchored cube} (Lemma~\ref{agn:lem:extract}): inputs $x_1,\ldots,x_n$, labels $a_i\ne b_i$, an anchor with value $\{b_i\}$ at every $x_i$, and for each $T\subseteq[n]$ a hypothesis $h_T$ equal to $\{a_i\}$ on $T$ and containing $b_i$ off $T$. Put each $i$ into $T$ with probability $1/4$ and demonstrate $a_i$ on $T$ and $b_i$ off $T$. Read as the sets of $h_T$, the data are realizable, so the benchmark is zero and only $a_i$ is correct on $T$; read as singletons, the anchor is wrong only on $T$, so the benchmark is about $1/4$, while $a_i$ is wrong at three quarters of the inputs. At an unseen $x_j$ the learner cannot tell whether $j\in T$, and one of the two readings gives it excess error at least $1/8$ (Lemma~\ref{agn:lem:anchored}).

\paragraph{Sufficiency.}
The learner routes each example to the learner of its region; as $\Lset_\D(\H)$ is at least the mass-weighted sum of the regional benchmarks, it suffices to be competitive on each region of non-negligible mass (Lemma~\ref{agn:lem:glue}). On the pure region a hypothesis matches the hidden set only if it is the singleton of the demonstrated label, so an agnostic multiclass learner on the demonstrated pairs suffices (Theorems~\ref{thm:comparison} and~\ref{thm:D} with $\ell=1$; Section~\ref{agn:sec:pure}).

On the mixed region the learner predicts from $\Y_x$ and competes with a comparator $h$ whose set disagreement it cannot observe. The way out is to compare only where needed: where $h(x)$ is large, local compatibility makes the learner correct whenever $h$ is; elsewhere the comparison is the ordinary multiclass regret on the demonstrated pairs, with undefined values of $h$ counted as wrong. This observable regret bounds the excess error over $h$ (Lemma~\ref{agn:lem:observable}), and $h$ acts as a sleeping expert \cite{freund1997using} that abstains on its large values. Running specialists over a finite cover of $\widetilde\H$ on the pool, whose size is independent of $|\Y|$ by the bounded alphabet, and a leave-one-out argument make the expected regret against every $h$ vanish (Lemma~\ref{agn:lem:expected}). Validation cannot boost the confidence, since the regret involves the unknown $h$ and may be negative. Instead, the learner averages independent runs, predicts a most frequently demonstrated label of $\Y_x$ at the inputs in a fresh sample, and independently rounds the averaged predictions elsewhere. Averaging concentrates the regret; with high probability the fresh sample contains every input of large mass, so the rounding error also concentrates (Lemma~\ref{agn:lem:confidence}). Gluing the two regional learners proves Theorem~\ref{thm:F} (Section~\ref{agn:sec:proof}).

\subsection{Impossibility Results}\label{ov:limits}

\paragraph{Impossibility of scale-free dimension.} We partition the input space into infinitely many blocks, where block $j$ uses the labels $[s_j]$ with $s_j=j+2$. Each target in the class $\H_j$ excludes one label at every input of block $j$ and accepts all available labels on the other blocks; the excluded label is $1$ at all but finitely many inputs. Each $\H_j$ is unlearnable, because these finitely many exceptions can lie anywhere in the block, so at an unseen input of block $j$ the excluded label can be uniform on $[s_j]$. Uniform random guessing on block $j$, however, has expected error $1/s_j$. Consequently, every union $\bigcup_j\cV_j$ of finite subclasses $\cV_j\subseteq\H_j$ is learnable: at any prescribed accuracy and confidence, random guessing controls the error on the later blocks, and on the finitely many earlier blocks only finitely many exceptional inputs need to be handled. If a parameter $\mathsf D$ with finite witnesses characterized learnability, each unlearnable class $\H_j$ would have $\mathsf D(\H_j)=\infty$. The finite-witness property would then allow us to choose finite subclasses $\cV_j\subseteq\H_j$ with $\mathsf D(\cV_j)\ge j$, forcing the learnable union $\bigcup_j\cV_j$ to have infinite parameter, a contradiction.

\paragraph{Impossibility of scale-sensitive dimension.} Let $W=\{2,3,\ldots\}$ and consider the class $\cK_\infty$ of targets $h_{F,B}(x)=\{\ind\{x\in F\}\}\cup(W\setminus B)$, where $F\subseteq\N$ and $B\subseteq W$ are finite. Each target accepts one binary label at each input and every auxiliary label outside its forbidden set $B$. The class $\cK_\infty$ is unlearnable under one-correct-label feedback: demonstrations may reveal only binary labels, leaving unseen binary answers unpredictable, while a suitable finite forbidden set can invalidate almost all likely auxiliary predictions. The subclass $\cG$ of targets with $2\notin B$ is learnable by always predicting $2$. For every finite subclass $\cV\subseteq\cK_\infty$, the union of the forbidden sets of its targets is finite, so some auxiliary label is valid everywhere for all targets in $\cV$, and swapping this label with $2$ maps $\cV$ into $\cG$. Hence, if each $\mathsf D_\alpha$ has finite witnesses and is relabeling invariant, then $\mathsf D_\alpha(\cK_\infty)=\mathsf D_\alpha(\cG)$ at every scale $\alpha$. Such a family therefore assigns identical values to a learnable class and an unlearnable class, and cannot characterize learnability. This construction is inspired by \cite{lechner2024inherent}.

\section{Conclusion}\label{sec:conclusion}

We studied PAC learning when each example may come with several correct labels and predicting any one of them suffices. With at most $\ell$ correct labels per example, the generalized Natarajan dimension characterizes learnability from full sets and from bandit feedback when $\Y$ is finite (Theorem~\ref{thm:A}), and the packed dimension characterizes learnability from full sets when $\Y$ is countable (Theorem~\ref{thm:D}). In all these cases, realizable and agnostic learnability coincide, as in the multiclass case \cite{david2016supervised,pmlr-v178-hopkins22a,hanneke2025representation}.

Learning from demonstrations, where each training example reveals only one of its correct labels, is governed by a different class. In the realizable setting, for finite and countable $\Y$ alike, $\H$ is learnable from demonstrations if and only if its intersection class $\cI_{\ell+1}(\H)$ is learnable from full sets (Theorems~\ref{thm:B} and~\ref{thm:E}). This condition can fail already with three labels, for a class learnable from a single full-set example or even a single bandit answer (Example~\ref{ex:three-labels}). When it holds, the sample complexity is at most of order $\eps^{-\ell}$ up to logarithmic factors, and this exponent cannot be improved in general (Theorem~\ref{thm:C}).

Agnostic learning from demonstrations behaves differently. Already on a single input, two classes can have intersection classes with the same values of the dimensions that characterize realizable learning, while only one of them is agnostically learnable (Theorem~\ref{thm:F} and Remark~\ref{rem:agnostic-c0}). For nondegenerate classes, agnostic learnability is instead characterized by finiteness of the DS dimension of the collapsed class $\widetilde\H$, which merges every set of at least two correct labels into one new label, together with local compatibility (Theorem~\ref{thm:F}).

Finally, the uniform bound $\ell$ on the number of correct labels is essential. Without it, it is impossible to characterize learnability by any parameter with finite witnesses (Theorem~\ref{thm:G}).

We conclude with two open problems that arise naturally from our results and that we believe are worth exploring further.

\begin{itemize}
\item[$\star$] \emph{The DS dimension of intersection classes.} Over a countably infinite label space, must an intersection class $\cI_{\ell+1}(\H)$ that is learnable from full sets have finite joint DS dimension? By Theorem~\ref{thm:E}, this asks whether finiteness of $\dDS(\cI_{\ell+1}(\H))$, which suffices for learning from demonstrations, is also necessary. For finite label spaces it is (Theorem~\ref{thm:B}). The analogous statement fails for general classes, since the class $\H^+$ of Appendix~\ref{app:separations} is learnable from full sets but has infinite joint DS dimension (Theorem~\ref{thm:large-ds-learnable}). Intersection classes, however, are more constrained. If $\cI_{\ell+1}(\H)$ is learnable from full sets, then so is $\cI_R(\H)$ for every fixed $R$ (Corollary~\ref{cor:all-depths}), so a counterexample must remain learnable from full sets after the intersections of any fixed number of hypotheses of $\H$ are added.

\item[$\star$] \emph{Scale-sensitive characterizations.} Suppose that the correct-label sets are finite but not uniformly bounded. Is there a family $(\mathsf D_\eps)_{\eps>0}$ of parameters with finite witnesses such that, under full-set feedback or under demonstrations, $\H$ is PAC learnable if and only if $\mathsf D_\eps(\H)<\infty$ for every $\eps>0$? For real-valued classes, the fat-shattering dimension plays exactly this role. Theorem~\ref{thm:G} rules out a single such parameter. Its last part also rules out families of relabeling-invariant parameters with finite witnesses, but only when the correct-label sets may be infinite and the feedback consists of demonstrations, so the case of finite sets remains open.

\end{itemize}

\section*{AI Statement}
\phantomsection\addcontentsline{toc}{section}{AI Statement}
The results for finite label spaces under full-set and bandit feedback, in both the realizable and the agnostic setting (Theorem~\ref{thm:A}), were proved by the authors. The other results, as well as the writing of the paper, were developed with the help of Claude (Anthropic) and ChatGPT (OpenAI).

\section{Preliminaries}\label{sec:prelim}

We start with multi-label hypotheses and the three feedback models, continue with partial intersection classes and one-inclusion hypergraphs, and conclude with the four dimensions and a bound for plurality votes. We follow the notation of \cite{alon2021theory} for partial concepts and of \cite{brukhim2022characterization} for projections and one-inclusion graphs.

Write $\N=\{1,2,\ldots\}$, $[n]=\{1,\ldots,n\}$, and $[0]=\emptyset$; all logarithms are natural, and sample sizes are rounded up to integers. The letter $c$ denotes a positive constant that may increase from line to line. In a result whose statement declares some parameters fixed (for instance $\ell$ and $\gamma$), $c$ depends only on these parameters, and never on the class, the alphabet size, the dimension, or the sample size; the notation $O_\ell(\cdot)$ makes this dependence explicit.

\subsection{Multi-Label Hypotheses}\label{sec:prelim-hypotheses}
In the main text, the \emph{instance space} $\X$ and the \emph{label space} $\Y$ are nonempty countable sets with their discrete $\sigma$-algebras.\footnote{The hypothesis classes themselves need not be countable, and we impose no computational restrictions on learners. Appendix~\ref{app:measurability} extends the learning results to standard Borel instance spaces under explicit regularity assumptions, with the same sample-complexity bounds.} Throughout the positive results we fix an integer $\ell\geq1$, the maximum number of correct labels, and put
\begin{equation}\label{eq:allowed}
\cA=\{A\subseteq\Y : 1\leq|A|\leq\ell\}.
\end{equation}
When $\Y$ is finite we write $k=|\Y|$ and assume $\ell\leq k$; the choice $\ell=k$ puts no bound on the number of correct labels. The number of possible correct-label sets is then
\begin{equation}\label{eq:q}
q=|\cA|=\sum_{j=1}^{\ell}\binom{k}{j}.
\end{equation}

A \emph{multi-label hypothesis} is a function $h:\X\to\cA\cup\{\star\}$, and $h$ is \emph{undefined} at $x$ if $h(x)=\star$. The \emph{support} of $h$ is $\supp(h)=\{x\in\X: h(x)\neq\star\}$, and $h$ is \emph{total} if $\supp(h)=\X$. For $x\in\supp(h)$, the set $h(x)$ is the \emph{correct-label set} of $x$, and its elements are the \emph{correct labels} of $x$. A \emph{class} is a set $\H\subseteq(\cA\cup\{\star\})^{\X}$.

A \emph{predictor} is any function $g:\X\to\Y$ (learners may be improper), and it is \emph{valid} at $x$ for $h$ if $g(x)\in h(x)$. A \emph{randomized prediction} $p$ assigns to every $x\in\X$ a probability distribution $p(\cdot\mid x)$ on $\Y$, and we write $p(A\mid x)=\sum_{y\in A}p(y\mid x)$.

\subsection{Learning Models}\label{sec:prelim-models}
\paragraph{Realizable learning.}
Fix a \emph{target} $h^\star\in\H$ and a distribution $P$ on $\X$ with $P(\supp(h^\star))=1$. Since data are drawn only from the support of the target, undefined values let a class encode assumptions on the data, as for partial concept classes \cite{alon2021theory}. The learner outputs a single label, which counts as correct if it is one of the correct labels. Thus the \emph{errors} of a predictor $g$ and of a randomized prediction $p$ are
\begin{equation}\label{eq:risk}
L_{P,h^\star}(g)=\Pr_{x\sim P}\bigl[g(x)\notin h^\star(x)\bigr],
\qquad
L_{P,h^\star}(p)=\E_{x\sim P}\bigl[1-p(h^\star(x)\mid x)\bigr].
\end{equation}
The training inputs $x_1,\ldots,x_m$ are drawn independently from $P$, and the learner receives one of the following kinds of feedback.
\begin{enumerate}[(i)]
\item \emph{Full-set feedback.} The learner observes the sample $\sigma=((x_i,h^\star(x_i)))_{i=1}^m\sim\D_{P,h^\star}^m$, where $\D_{P,h}$ denotes the law of $(x,h(x))$ for $x\sim P$.
\item \emph{Bandit feedback.} After seeing $x_i$, the learner chooses a label $u_i\in\Y$, which may depend on all previous inputs and answers, and observes only $b_i=\ind\{u_i\in h^\star(x_i)\}$. Our bandit learners use a special case of this protocol: when $\Y$ is finite, they draw the queries uniformly at random, independently of the inputs and of each other, as in the batch model of \cite{hanneke2026pac}.
\item \emph{One-correct-label feedback.} The sample $\sigma=((x_1,y_1),\ldots,(x_m,y_m))$ consists of independent draws from a joint distribution $\D$ on $\X\times\Y$ whose marginal on $\X$ is $P$ and which satisfies $\Pr_{(x,y)\sim\D}[y\in h^\star(x)]=1$; we say that such a distribution is \emph{compatible} with $h^\star$, and we call $y_i$ a \emph{demonstrated label}. The conditional law of $y_i$ is otherwise arbitrary; for instance, the demonstrator may always show the same correct label.
\end{enumerate}
When $\ell=1$, full-set and one-correct-label feedback coincide, and the model is multiclass PAC learning of partial concept classes \cite{alon2021theory,brukhim2022characterization}.

Learners may be randomized: a learner may use an independent seed $\omega\sim\Unif([0,1])$, and its output is then a random predictor, drawn once. A learner may also depend on $\eps$ and $\delta$, and it may ignore surplus observations.\footnote{A randomized prediction on a countable instance space becomes a random predictor by drawing its value once at every input. This preserves the expected error but not the distribution of the error, so whenever we need a single predictor with a high-probability guarantee, we obtain it explicitly, by a most probable label, a plurality vote, or validation.}

\begin{definition}[PAC learnability]\label{def:pac}
Fix one of the three feedback models. A class $\H$ is \emph{realizably PAC learnable} if for every $\eps,\delta\in(0,1)$ there are an integer $m$ and a learner such that the following holds. Consider any target $h^\star\in\H$ and any distribution $P$ on $\X$ with $P(\supp(h^\star))=1$ (and, for one-correct-label feedback, any joint distribution $\D$ as above). With probability at least $1-\delta$ over $m$ training examples and the seed, the learner outputs a predictor $\widehat g$ with $L_{P,h^\star}(\widehat g)\leq\eps$. The least such $m$ is the \emph{sample complexity}, denoted $\mfull_\H(\eps,\delta)$, $\mband_\H(\eps,\delta)$, or $\mone_\H(\eps,\delta)$; it equals $\infty$ if no finite $m$ works.
\end{definition}

We pass between PAC guarantees and bounds on the expected error by two implications, valid for every random error $Z\in[0,1]$:
\begin{equation}\label{eq:pac-expect}
\Pr[Z>\eps]\leq\delta\ \Longrightarrow\ \E[Z]\leq\eps+\delta,
\qquad
\E[Z]\leq\eps\delta\ \Longrightarrow\ \Pr[Z>\eps]\leq\delta.
\end{equation}
The first holds because $Z\leq\eps+\ind\{Z>\eps\}$, and the second is Markov's inequality.

\paragraph{Agnostic learning.}
In the agnostic setting, examples $(x,A)$ are drawn from an arbitrary distribution $\D$ on $\X\times\cA$. Under full-set feedback the learner observes the pairs $(x_i,A_i)$, and under bandit feedback it observes $x_i$ and the answer $\ind\{u_i\in A_i\}$ to its query. Under one-correct-label feedback it observes only pairs $(x_i,y_i)$, where the conditional law of the demonstrated label $y$ given $(x,A)$ is supported on $A$ and otherwise arbitrary. The error of a predictor and the benchmark are
\begin{equation}\label{eq:agnostic}
L_\D(g)=\Pr_{(x,A)\sim\D}\bigl[g(x)\notin A\bigr],
\qquad
\Lset_\D(\H)=\inf_{h\in\H}\Pr_{(x,A)\sim\D}\bigl[h(x)\neq A\bigr].
\end{equation}
The benchmark compares the whole correct-label set of $h$ with the sampled set, and, as in \cite{alon2021theory}, it counts $h(x)=\star$ as a mismatch. It can therefore be much larger than the best achievable error. For example, with one input $x$, $\H=\{h\}$, $h(x)=\{1,2\}$, and $A=\{1\}$ almost surely, we have $\Lset_\D(\H)=1$, although predicting $1$ has zero error.

\begin{definition}[Agnostic PAC learnability]
Fix one of the three feedback models. A class $\H$ is \emph{agnostically PAC learnable} if for every $\eps,\delta\in(0,1)$ a finite number of examples suffices for some learner to ensure $\Pr[L_\D(\widehat g)>\Lset_\D(\H)+\eps]\leq\delta$ for every $\D$ (and, for one-correct-label feedback, every conditional law of the demonstrated label). Under full-set and bandit feedback, the least such number is denoted $\mfullag_\H(\eps,\delta)$ and $\mbandag_\H(\eps,\delta)$, respectively.
\end{definition}

Under one-correct-label feedback we need no notation for the sample complexity, since Section~\ref{sec:agnostic} characterizes learnability itself. A realizable experiment is the agnostic experiment with $A=h^\star(x)$, whose benchmark is zero. Hence agnostic PAC learnability implies realizable PAC learnability under each feedback model.

\subsection{Projections, Intersections, and One-Inclusion Hypergraphs}\label{sec:prelim-projections}
We use sequences of inputs rather than sets, because the inputs of a learning problem form a sequence \cite{brukhim2022characterization}. For a sequence $S=(x_1,\ldots,x_n)\in\X^n$, the \emph{projection} of $\H$ to $S$ records the correct-label sets on $S$ of the hypotheses defined on all of $S$:
\begin{equation}\label{eq:restriction}
\H|_S=\bigl\{(h(x_1),\ldots,h(x_n)) : h\in\H,\ \{x_1,\ldots,x_n\}\subseteq\supp(h)\bigr\}\subseteq\cA^n.
\end{equation}
We call the elements of $\H|_S$ \emph{patterns}. For a pattern $v=(v_1,\ldots,v_n)$ and $i\in[n]$, $v_{-i}$ denotes $v$ with its $i$th coordinate deleted, and for $V\subseteq\cA^n$ and $I\subseteq[n]$, $V|_I$ is the coordinate restriction of $V$ to $I$. We write $\sigma$ for labeled samples and $S$ for sequences of unlabeled inputs, which we call \emph{pools}; a pool may contain repeated inputs.

\paragraph{Partial intersection classes.}
If every demonstrated label of a sample lies in $h_1(x)\cap h_2(x)$, then the sample is consistent with both $h_1$ and $h_2$, and the learner cannot tell them apart. More generally, a label in $h_1(x)\cap\cdots\cap h_r(x)$ is a valid demonstration for each $h_a$, so a learner from one correct label must also succeed on samples generated by such intersections (Section~\ref{sec:reduction}). For an integer $R\geq1$, the \emph{partial intersection class} $\cI_R(\H)$ consists of all hypotheses $g$ of the form
\begin{equation}\label{eq:intersections}
g(x)=
\begin{cases}
\bigcap_{a=1}^{r}h_a(x), & x\in B,\\
\star, & x\notin B,
\end{cases}
\end{equation}
where $1\leq r\leq R$, $h_1,\ldots,h_r\in\H$, and $B\subseteq\bigcap_{a=1}^r\supp(h_a)$ is any set on which the intersection $\bigcap_a h_a(x)$ is nonempty. We call $(h_1,\ldots,h_r)$ a \emph{representing tuple} of $g$; repeating one of its members pads it to length exactly $R$. The class $\cI_R(\H)$ is closed under domain restrictions, that is, it contains the restriction of each of its members to every subset of its support, but in general it is not closed under further intersections. All one-correct-label results concern $\cI_{\ell+1}(\H)$, the intersections of at most $\ell+1$ hypotheses of $\H$. Its correct-label sets are nonempty subsets of correct-label sets of $\H$, so they still have at most $\ell$ labels.

\paragraph{One-inclusion hypergraphs.}
For a finite nonempty $V\subseteq\cA^n$, the \emph{one-inclusion hypergraph} of $V$ has vertex set $V$ and hyperedges
\begin{equation}\label{eq:edge}
e_{i,u}=\{v\in V: v_{-i}=u\},
\qquad i\in[n],\ u\in V|_{[n]\setminus\{i\}}.
\end{equation}
This is the one-inclusion graph of \cite{haussler1994predicting}, in the hypergraph form of \cite{rubinstein2006shifting} and \cite{brukhim2022characterization}, with patterns of correct-label sets in place of labels. The vertices of $e_{i,u}$ are the patterns that agree, as whole correct-label sets, at every coordinate other than $i$, and $i$ is the \emph{direction} of $e_{i,u}$. Hyperedges are indexed by the pair $(i,u)$, so equal vertex sets in different directions count as different hyperedges.

\subsection{Dimensions}\label{sec:prelim-dimensions}
We now define the four dimensions of this paper. Each of them is the largest $n$ such that some sequence of $n$ distinct inputs carries a certain finite witness; it is $0$ if no witness of positive size exists, and $\infty$ if witnesses of every size exist. In all four cases the witness consists of patterns, and it contains groups of patterns that agree at all inputs but one and have no common label at the remaining input, so that no single prediction there is correct for a whole group. The definitions differ in how many such groups they require and how the groups are arranged.

The first dimension extends the Natarajan dimension and asks for a full product of conflicting families.

\begin{definition}[Generalized Natarajan dimension]\label{def:nat}
A sequence $S=(x_1,\ldots,x_n)$ of distinct inputs is \emph{product-shattered} by $\H$ if there are finite nonempty families $\cA_1,\ldots,\cA_n\subseteq\cA$ such that
\[
\bigcap_{A\in\cA_i}A=\emptyset\quad(i\in[n])
\qquad\text{and}\qquad
\cA_1\times\cdots\times\cA_n\subseteq\H|_S.
\]
The product $\cA_1\times\cdots\times\cA_n$ is a \emph{product witness}. The \emph{generalized Natarajan dimension} $\dN(\H)$ is the largest size of a product-shattered sequence.
\end{definition}

In a product witness, every input carries a family of correct-label sets with no common label, and every combination of choices is realized. The next dimension asks for less, in the spirit of the pseudo-cubes of \cite{brukhim2022characterization}: whenever the correct-label sets at all other inputs are revealed, the remaining candidates at the hidden input still have no common label.

\begin{definition}[Joint DS dimension]\label{def:ds}
Let $S=(x_1,\ldots,x_n)\in\X^n$. A finite nonempty set $V\subseteq\H|_S$ is a \emph{DS witness} if every hyperedge of its one-inclusion hypergraph has an empty intersection in its own direction:
\[
\bigcap_{w\in e_{i,v_{-i}}}w_i=\emptyset
\qquad(v\in V,\ i\in[n]).
\]
The \emph{joint DS dimension} $\dDS(\H)$ is the largest $n$ for which some sequence of $n$ distinct inputs carries a DS witness.
\end{definition}

A stronger local condition asks that a single other candidate in the hyperedge already rules out every label of the current one. The resulting dimension is at most the joint DS dimension (Lemma~\ref{lem:orders}), so upper bounds in terms of it, as in Theorem~\ref{thm:E}, are stronger.

\begin{definition}[Disjoint-neighbor dimension]\label{def:pair}
Let $S=(x_1,\ldots,x_n)\in\X^n$. A finite nonempty set $V\subseteq\H|_S$ is a \emph{disjoint-neighbor witness} if for every $v\in V$ and $i\in[n]$ there is $w\in V$ with
\[
w_{-i}=v_{-i}
\qquad\text{and}\qquad
w_i\cap v_i=\emptyset.
\]
The \emph{disjoint-neighbor dimension} $\dpair(\H)$ is the largest $n$ for which some sequence of $n$ distinct inputs carries a disjoint-neighbor witness.
\end{definition}

When $\ell=1$, all correct-label sets are singletons: then $\dN$ is the Natarajan dimension \cite{natarajan1989learning}, and $\dDS$ and $\dpair$ both equal the DS dimension \cite{daniely2014optimal}. Already in this case $\dN$ can be smaller than the other two.

\begin{example}
Let $\X=\{x_1,x_2\}$, and let $\H$ consist of the six total hypotheses whose patterns on $(x_1,x_2)$ are $(1,2)$, $(3,2)$, $(3,4)$, $(5,4)$, $(5,6)$, $(1,6)$, where each label stands for a singleton. The one-inclusion hypergraph of $\H|_{(x_1,x_2)}$ is a hexagon, and every pattern has a neighbor with a different label in each direction, so $\dDS(\H)=\dpair(\H)=2$. On the other hand, the first labels $1$, $3$ and $5$ occur with the second labels $\{2,6\}$, $\{2,4\}$ and $\{4,6\}$, respectively. No two of these sets share two labels, so $\H|_{(x_1,x_2)}$ contains no product of two pairs, and $\dN(\H)=1$.
\end{example}

For larger $\ell$, the joint DS dimension differs from the DS dimension of the class obtained by viewing each correct-label set as a single label: our hyperedges must have empty intersections, while the ordinary DS dimension only requires at least two distinct sets in each hyperedge.

With sets of size two or more, an empty intersection in a hyperedge can force very few errors: if all members of a large hyperedge but one contain the label $1$, then predicting $1$ is wrong for only one of them. The last dimension, designed for countable label spaces, therefore asks for many disjoint small conflicts, which must cover a constant fraction of a weighted list of patterns in every direction. The fraction is a coverage parameter $\gamma$, which we fix once and for all in the range
\begin{equation}\label{eq:gamma-range}
0<\gamma<\frac{1}{\ell+1}.
\end{equation}

\begin{definition}[Packed dimension]\label{def:pack}
Let $S=(x_1,\ldots,x_n)\in\X^n$. A \emph{$\gamma$-packed witness} on $S$ consists of a nonempty indexed list $v^1,\ldots,v^M\in\H|_S$, with repetitions allowed, and, for every direction $i\in[n]$, a collection $\cB_i$ of pairwise disjoint subsets of $[M]$, called \emph{conflict blocks}, such that every $B\in\cB_i$ satisfies
\[
2\leq|B|\leq\ell+1,
\qquad
v^a_j=v^b_j\ \ (a,b\in B,\ j\neq i),
\qquad
\bigcap_{a\in B}v^a_i=\emptyset,
\]
and the blocks of each direction cover at least a $\gamma$ fraction of the list:
\[
\Bigl|\bigcup_{B\in\cB_i}B\Bigr|\geq\gamma M
\qquad(i\in[n]).
\]
The \emph{packed dimension} $\dpack(\H)$ is the largest $n$ for which some sequence of $n$ distinct inputs carries a $\gamma$-packed witness. For another coverage parameter $0<\rho\leq1$, the dimension $\dpackr{\rho}(\H)$ is defined in the same way, with $\gamma$ replaced by $\rho$.
\end{definition}

Blocks in different directions may overlap. The repetitions in the list encode rational weights, and they are kept when a witness is projected to fewer coordinates. The upper bound on $\gamma$ serves the weak learner of Section~\ref{sec:countable-full}: when $\dpack(\H)<\infty$, its expected error from $m$ examples is at most $\gamma+\dpack(\H)/(m+1)$ (Lemma~\ref{lem:packed-weak}), which is below the threshold $1/(\ell+1)$ of Lemma~\ref{lem:mode} for sufficiently large $m$. Other coverage levels $\rho$ appear mainly in comparisons with DS-type dimensions (Theorem~\ref{thm:comparison}).

We now compare the four dimensions. The comparison, and many later arguments, rest on the following simple fact. A finite family of sets is a \emph{minimal empty-intersection family} if its intersection is empty and every proper subfamily has a nonempty intersection.

\begin{lemma}[Small empty-intersection families]\label{lem:small}
If a family of nonempty sets of size at most $\ell$ has an empty intersection, then some subfamily of at most $\ell+1$ members has an empty intersection. Consequently, every minimal empty-intersection family of such sets has between two and $\ell+1$ members.
\end{lemma}

\begin{proof}
Choose one member $A$ of the family, and for each $y\in A$ choose a member that does not contain $y$. These at most $1+|A|\leq\ell+1$ sets have an empty intersection. A minimal family has at least two members, because each set is nonempty.
\end{proof}

\begin{lemma}[Comparisons and projections]\label{lem:orders}
For every class $\H$,
\begin{equation}\label{eq:dimension-order}
\dN(\H)\leq\dDS(\H),
\qquad
\dpair(\H)\leq\dDS(\H),
\qquad
\dN(\H)\leq\dpack(\H).
\end{equation}
The last inequality holds for $\dpackr{\rho}$ with every $0<\rho\leq1$. Moreover, if $V$ is a DS witness on $S$ and $I$ is a nonempty set of coordinates, then $V|_I$ is a DS witness on the corresponding subsequence; and a packed witness restricted to $I$, with all indices of the list kept, is a packed witness on that subsequence.
\end{lemma}

\begin{proof}
Take a product witness $\cA_1\times\cdots\times\cA_n$ and shrink each $\cA_i$ to a minimal empty-intersection family, which by Lemma~\ref{lem:small} has between two and $\ell+1$ members. The product $V=\cA_1\times\cdots\times\cA_n$ is a DS witness, because the hyperedge of $v\in V$ in direction $i$ is $\{w\in V: w_{-i}=v_{-i}\}$ and its $i$th coordinates range over all of $\cA_i$; this proves the first inequality. A disjoint-neighbor witness is a DS witness, because the hyperedge $e_{i,v_{-i}}$ contains $v$ and a pattern whose $i$th coordinate is disjoint from $v_i$; this proves the second inequality. Finally, if we list every element of the product $V$ once, the hyperedges in direction $i$ are disjoint blocks of size at most $\ell+1$ with empty intersections that cover the whole list, so $V$ is a packed witness for every coverage parameter.

For the projection of a DS witness, let $u\in V|_I$ and $i\in I$, and choose $v\in V$ with $v|_I=u$. The hyperedge of $u$ in direction $i$ contains the projection of the hyperedge of $v$ in direction $i$, which has the same sets at coordinate $i$. An intersection over more sets can only be smaller, so it remains empty. For packed witnesses, the equalities away from each retained direction, the empty intersections, and the covered fractions are all preserved when the list is restricted to $I$.
\end{proof}

\paragraph{Disjoint products.}
For lower-bound constructions we combine classes on disjoint instance spaces (Section~\ref{sec:worst-case}). Let $\X$ be the disjoint union of $\X_1,\ldots,\X_b$, and for each $j\in[b]$ let $\H_j\subseteq(\cA\cup\{\star\})^{\X_j}$. The \emph{disjoint product} of $\H_1,\ldots,\H_b$ is the class
\[
\bigl\{h\in(\cA\cup\{\star\})^{\X} : h|_{\X_j}\in\H_j\ \text{for all}\ j\in[b]\bigr\};
\]
thus a member is obtained by choosing a member of each $\H_j$ independently, undefined values included. The four dimensions add up under this operation.

\begin{lemma}[Disjoint products]\label{lem:disjoint-product}
Let $\H$ be the disjoint product of nonempty classes $\H_1,\ldots,\H_b$, $b\ge1$, over a common label space, with the same bound $\ell$. For each $\mathsf{D}\in\{\dN,\dDS,\dpair,\dpack\}$,
\[
\mathsf{D}(\H)=\sum_{j=1}^{b}\mathsf{D}(\H_j).
\]
If moreover $\H_1,\ldots,\H_b$ are total, then $\mathsf{D}(\cI_{\ell+1}(\H))=\sum_{j=1}^{b}\mathsf{D}(\cI_{\ell+1}(\H_j))$.
\end{lemma}

\begin{proof}
For the upper bound, take a witness on $S$ and split its coordinates according to the component that contains each input. Every pattern of $\H$ on $S$ restricts to a pattern of $\H_j$ on the part of $S$ in component $j$, so it suffices to check that each type of witness projects to a witness of the same type. A product witness projects to a product witness, and DS and packed witnesses (the latter with all indices kept) project correctly by Lemma~\ref{lem:orders}. For a disjoint-neighbor witness $V$ and coordinates $I$, if $w$ is a disjoint neighbor of $v$ in a direction $i\in I$, then $w|_I$ is a disjoint neighbor of $v|_I$ in direction $i$. Thus the part of a witness in component $j$ has at most $\mathsf{D}(\H_j)$ coordinates, and summing over $j$ gives the upper bound.

For the lower bound, take finite witnesses in the components (a single empty-coordinate pattern for a zero-dimensional component) and form their Cartesian product, whose elements are patterns of $\H$ because the components of a member of $\H$ are chosen independently. Fix a coordinate $i$ in component $j$. A hyperedge of the product in direction $i$ fixes the other components, so its sets at $i$ are those of a hyperedge of the witness of component $j$. Hence products, empty intersections and disjoint neighbors at $i$ are supplied by that witness. For packed witnesses, index the product list by tuples of indices, and in direction $i$ copy every block of component $j$ for each choice of the indices of the other components; these blocks are disjoint and cover the same fraction of the list. The product witness has the sum of the component sizes, and taking arbitrarily large witnesses handles infinite values.

For the intersection classes, it suffices to show that $\cI_{\ell+1}(\H)$ is the disjoint product of the classes $\cI_{\ell+1}(\H_j)$. Pad all representing tuples to length $\ell+1$. A tuple of members of $\H$ restricts to a tuple of members of $\H_j$ in each component, and conversely, tuples chosen in the components combine position by position into a tuple of members of $\H$. Since the components are total, the only constraint on the domain $B$ in \eqref{eq:intersections} is that the intersection be nonempty on $B$, and this holds exactly when it holds on every $B\cap\X_j$. So the domains also factor across the components.
\end{proof}

\subsection{Plurality Votes}\label{sec:prelim-conventions}
A binary majority vote is wrong only if at least half of the votes are wrong; with up to $\ell$ correct labels, the fraction one half becomes $1/(\ell+1)$. A \emph{plurality} label is a label that receives the largest number of votes,\footnote{Plurality votes in multiclass boosting are studied, for example, by \cite{brukhim2023improper}.} and a \emph{most probable} label of a distribution on $\Y$ is a label of maximum probability (one exists, since only finitely many labels have mass larger than half of the supremum of the masses); ties are broken by a fixed enumeration of $\Y$.

\begin{lemma}[Most probable labels]\label{lem:mode}
Let $A$ be a set of at most $\ell$ labels. If a most probable label of a distribution on $\Y$ lies outside $A$, then the distribution assigns mass at least $1/(\ell+1)$ to $\Y\setminus A$. Consequently, the error of a most probable label is at most $\ell+1$ times the error of the randomized prediction. Likewise, a plurality vote is invalid at an input only if at least a $1/(\ell+1)$ fraction of the votes there are invalid, so its error is at most $\ell+1$ times the average error of the voters.
\end{lemma}

\begin{proof}
Let $a$ be the mass of $\Y\setminus A$. The most probable label lies outside $A$, so its mass is at most $a$, and each of the at most $\ell$ labels in $A$ has mass at most that of the most probable label. Thus $1-a\leq\ell a$, that is, $a\geq1/(\ell+1)$. Applying this at every input with $A=h^\star(x)$ and taking expectations over $x\sim P$ gives the consequence for a most probable label; for plurality votes, apply the same argument to the empirical distribution of the votes.
\end{proof}

\section{Examples}\label{sec:examples}

We now compute the dimensions of Section~\ref{sec:prelim-dimensions} for four classes. In all of them every correct-label set has at most two labels, so $\ell=2$ and the parameter $\gamma$ of $\dpack$ ranges over $(0,1/3)$. We start with two geometric classes on the real line, in which two labels are correct exactly at boundary points.\footnote{Both classes satisfy the assumptions of Appendix~\ref{app:measurability}, so Theorems~\ref{thm:A}--\ref{thm:E} apply to them by Theorem~\ref{thm:standard-borel-extension}. The verification is routine: each hypothesis is constant on the pieces of a partition of $\mathbb R$ into finitely many intervals and points, and for each pattern the input tuples that realize it form a finite union of $\sigma$-compact sets.} For both, $\dN$, $\dDS$ and $\dpair$ equal the corresponding parameter of the classical single-label class, so the extra correct labels at the boundaries cost nothing.

\begin{example}[Unions of intervals]\label{ex:intervals}
Let $\X=\mathbb R$, $\Y=\{0,1\}$, and $r\ge1$. For a union $U$ of at most $r$ disjoint closed intervals, the empty union included, let $h_U(x)=\{1\}$ if $x$ is an interior point of $U$, $h_U(x)=\{0,1\}$ if $x$ is an endpoint of one of its intervals, and $h_U(x)=\{0\}$ otherwise. The class $\H_{\mathrm{int}}$ of all these hypotheses satisfies
\[
\dN(\H_{\mathrm{int}})=\dDS(\H_{\mathrm{int}})=\dpair(\H_{\mathrm{int}})=2r.
\]
For the lower bound, take points $x_1<\cdots<x_{2r}$. A labeling of these points by $0$ and $1$ has at most $r$ runs of ones, and choosing for each run a closed interval that contains the run in its interior and no other $x_i$ (these intervals are disjoint, since distinct runs are separated by a point labeled $0$) realizes the corresponding pattern of singletons. Hence $x_1,\ldots,x_{2r}$ are product-shattered with the families $\{\{0\},\{1\}\}$, and the realized patterns form a disjoint-neighbor witness.

For the upper bound, select the label $0$ at every endpoint, turning each $h_U$ into the indicator of the interior of $U$. In a DS witness of $\H_{\mathrm{int}}$, the sets along each hyperedge have no common label, so they include both $\{0\}$ and $\{1\}$. Patterns that agree outside a coordinate still agree there after the selection, so the selection maps the witness to a DS witness of the binary class of these indicators. For binary classes the DS dimension is the VC dimension, and the VC dimension of interiors of unions of at most $r$ intervals is at most $2r$, because the alternating labeling $1,0,1,\ldots,0,1$ of $2r+1$ points needs $r+1$ intervals. Hence $\dDS(\H_{\mathrm{int}})\le2r$, and by Lemma~\ref{lem:orders} the same bound holds for $\dN$ and $\dpair$.
\end{example}

\begin{example}[Nearest-site classification]\label{ex:sites}
Let $\X=\mathbb R$ and $\Y=[k]$ with $k\ge2$. For sites $s_1<\cdots<s_k$, let $h_s(x)$ be the set of indices $j$ that minimize $|x-s_j|$. Thus every input has a single nearest site, except the midpoint of two adjacent sites, where both of their indices are correct. The class $\H_{\mathrm{site}}$ of all these hypotheses satisfies
\[
\dN(\H_{\mathrm{site}})=\dDS(\H_{\mathrm{site}})=\dpair(\H_{\mathrm{site}})=k-1.
\]
For the lower bound, fix sites $s_1<\cdots<s_k$ and let $x_j=(s_j+s_{j+1})/2$ for $j\in[k-1]$. Sites are recovered from $s_1$ and their midpoints $m_j$ by $s_{j+1}=2m_j-s_j$. Given signs, keep $s_1$, move the $j$th midpoint to $m_j=x_j\pm\delta$ for every $j$, and recompute the sites. Each recomputed site moves by at most $2\delta$ more than the previous one, hence by at most $2k\delta$, so for small $\delta>0$ the sites stay in order and $x_j$ remains closest to the $j$th or the $(j+1)$st site. If the $j$th midpoint moved right, then $x_j$ lies to its left and the correct-label set at $x_j$ is $\{j\}$; if it moved left, this set is $\{j+1\}$. Hence $x_1,\ldots,x_{k-1}$ are product-shattered with the families $\{\{j\},\{j+1\}\}$, and the realized patterns form a disjoint-neighbor witness.

For the upper bound, take a DS witness of $\H_{\mathrm{site}}$; reordering its inputs, we may assume they are $x_1<\cdots<x_n$. Select the smallest label of each correct-label set. The sets of a hyperedge cannot all have the same smallest label, since that label would be common to all of them, so the selected sequences form a DS witness of singletons. Moreover, each selected sequence is nondecreasing, because the smallest index of a nearest site does not decrease as the input moves to the right.

Choose a selected sequence $f=(f_1,\ldots,f_n)$ with minimum sum of entries. In each direction $i$, $f$ has a neighbor $f'$ that differs from it only at coordinate $i$, and $f'_i>f_i$ by the choice of $f$. Since $f'$ is also nondecreasing and agrees with $f$ off coordinate $i$, $f_i<f'_i\le f'_{i+1}=f_{i+1}$ for $i<n$, and $f_n<f'_n\le k$. Hence $1\le f_1<\cdots<f_n<k$, so $n\le k-1$, and by Lemma~\ref{lem:orders} the same bound holds for $\dN$ and $\dpair$.
\end{example}

The next two examples separate the roles of the different dimensions.
The first gives classes with disjoint-neighbor dimension zero while the other three
dimensions can be arbitrarily large or infinite. In particular, this shows that
finite disjoint-neighbor dimension does not imply learnability from full sets.
This contrasts with the fact that
finite disjoint-neighbor dimension of the intersection class implies learnability from one
correct label. The second gives a three-label class whose four dimensions are at most one. It is learnable from a single full-set example, and even from a single bandit answer, but not from any number of demonstrations. Thus it separates full-set and bandit learnability from one-correct-label learnability.

\begin{example}[Triangle products]\label{ex:triangle}
Let $\Y=[3]$ and $\mathcal T=\{\{1,2\},\{1,3\},\{2,3\}\}$, and let $\H_\triangle$ be the class of all total hypotheses on $\X$ whose correct-label sets lie in $\mathcal T$. If $\X=[d]$, then
\[
\dN(\H_\triangle)=\dDS(\H_\triangle)=\dpack(\H_\triangle)=\dpair(\cI_2(\H_\triangle))=d,
\qquad
\dpair(\H_\triangle)=0.
\]
Indeed, the three sets of $\mathcal T$ have no common label, so $\mathcal T^d$ is a product witness, and by Lemma~\ref{lem:orders} the first three dimensions are at least $d$; none of them exceeds $|\X|=d$. On the other hand, any two sets of $\mathcal T$ share a label, so no pattern has a disjoint neighbor, and $\dpair(\H_\triangle)=0$.

Intersections behave differently, because every singleton in $\Y$ is the intersection of two sets of $\mathcal T$. Choosing these intersections independently at the $d$ inputs shows that $\cI_2(\H_\triangle)$ contains every total hypothesis with singleton correct-label sets, and the patterns of these hypotheses form a disjoint-neighbor witness. Thus $\dpair(\cI_2(\H_\triangle))=d$, and the same holds for $\cI_3(\H_\triangle)\supseteq\cI_2(\H_\triangle)$.

If $\X=\N$, the same arguments give $\dpair(\H_\triangle)=0$ and infinite values of the other dimensions, so by Theorem~\ref{thm:A} the class is not learnable even from full sets. Thus finite disjoint-neighbor dimension of the original class
does not imply full-set learnability. In this example,
$\dpair(\cI_3(\H_\triangle))=\infty$, so the sufficient condition
for one-correct-label learnability in Theorem~\ref{thm:E}
does not hold.
\end{example}

\begin{example}[A three-label separation]\label{ex:three-labels}
Let $\X=\N$ and $\Y=[3]$. For every $b:\N\to\{0,1\}$, let $h_b(x)=\{1,2+b(x)\}$, so that $h_b(x)$ is $\{1,2\}$ where $b(x)=0$ and $\{1,3\}$ where $b(x)=1$, and let $g(x)=\{2,3\}$ for all $x$. Thus every $h_b$ accepts the label $1$ at every input, and $g$ rejects it everywhere. The class $\H_{\mathrm{bit}}=\{h_b: b\in\{0,1\}^{\N}\}\cup\{g\}$, a subclass of $\H_\triangle$ on $\X=\N$, satisfies
\[
\dN(\H_{\mathrm{bit}})=\dDS(\H_{\mathrm{bit}})=\dpack(\H_{\mathrm{bit}})=1,
\qquad
\dpair(\H_{\mathrm{bit}})=0.
\]
Indeed, the three sets of $\mathcal T$ occur at every input and have no common label, so every input is product-shattered with the family $\mathcal T$, and by Lemma~\ref{lem:orders} the first three dimensions are at least one. On two or more inputs, however, patterns that agree at all inputs but one still agree at some input. Since only $g$ takes the value $\{2,3\}$, such patterns either all equal the pattern of $g$ or all come from hypotheses $h_b$ and share the label $1$. Hence no hyperedge has an empty intersection and there are no conflict blocks, so there are no DS or packed witnesses on two or more inputs, and $\dN(\H_{\mathrm{bit}})\le\dDS(\H_{\mathrm{bit}})=1$ by Lemma~\ref{lem:orders}. As for $\H_\triangle$, any two sets of $\mathcal T$ share a label, so $\dpair(\H_{\mathrm{bit}})=0$.

The class is easy to learn from full sets and from bandit feedback. The correct-label set of one example, or the answer to the query $1$ at one example, reveals whether the target is $g$ or one of the hypotheses $h_b$. The learner that then predicts $2$ at every input if the target is $g$, and $1$ at every input otherwise, has error zero. Thus $\mfull_{\H_{\mathrm{bit}}}(\eps,\delta)\le1$ and $\mband_{\H_{\mathrm{bit}}}(\eps,\delta)\le1$ for all $\eps,\delta\in(0,1)$.

However, $\H_{\mathrm{bit}}$ is not learnable from one correct label. The label $2+b(x)$ is correct for both $h_b$ and $g$, so a demonstrator who always shows it produces the same sample for both targets. Fix $m$ and $n$, choose $b(1),\ldots,b(n)$ as independent fair bits and $b(x)=0$ for $x>n$, let $P$ be uniform on $[n]$, and let the demonstrator always show $2+b(x)$. Let the target be $g$ with probability $1/3$ and $h_b$ with probability $2/3$, independently of $b$. Since the sample does not depend on this choice, the target is still $g$ with probability $1/3$ given the sample and the learner's seed, and at a test input absent from the $m$ training inputs, $b(x)$ is still a fair bit, independent of this choice. At such an input, the label $1$ is wrong with probability $1/3$, namely when the target is $g$, and each of the labels $2$ and $3$ is wrong with probability $\frac23\cdot\frac12=\frac13$, since it is wrong for $h_b$ at exactly one value of $b(x)$. The test input is absent from the sample with probability $(1-\frac1n)^m$, so, averaged over $b$ and the choice of the target, every learner using $m$ examples has expected error at least $\frac13(1-\frac1n)^m$, and the same holds for some fixed $b$ and target, with the demonstrator showing $2+b(x)$. By \eqref{eq:pac-expect}, a learner whose error is at most $1/8$ with probability at least $7/8$ has expected error at most $1/4$, which is less than $\frac13(1-\frac1n)^m$ once $n$ is large. Hence $\mone_{\H_{\mathrm{bit}}}(1/8,1/8)=\infty$.

All four dimensions of the intersection class are infinite, already for $\cI_2(\H_{\mathrm{bit}})$. Indeed, $h_b(x)\cap g(x)=\{2+b(x)\}$ at every input, so $\cI_2(\H_{\mathrm{bit}})$ contains every total hypothesis with values in $\{\{2\},\{3\}\}$. Thus every sequence of distinct inputs is product-shattered by $\cI_2(\H_{\mathrm{bit}})$ with the families $\{\{2\},\{3\}\}$, and the realized patterns form a disjoint-neighbor witness. By Lemma~\ref{lem:orders}, all four dimensions of $\cI_2(\H_{\mathrm{bit}})$ are infinite, and so are those of $\cI_3(\H_{\mathrm{bit}})\supseteq\cI_2(\H_{\mathrm{bit}})$. In particular, $\dN(\cI_3(\H_{\mathrm{bit}}))=\infty$, so by Theorem~\ref{thm:A}, $\cI_3(\H_{\mathrm{bit}})$ is not learnable from full sets: the condition of Theorem~\ref{thm:B} fails, although the four dimensions of $\H_{\mathrm{bit}}$ itself are at most one.
\end{example}

\section{Full-Set Learning with Finite Labels}\label{sec:finite-full}

In this section the label space is finite, $k=|\Y|<\infty$, and $1\le\ell\le k$; the class $\H$ is nonempty and may be partial, and we write $d=\dN(\H)$. When the label space is finite, realizable and agnostic learnability with full-set feedback coincide, as in the multiclass case \cite{david2016supervised,pmlr-v178-hopkins22a,hanneke2025representation}, and both are characterized by either dimension.

\begin{theorem}[Full-set learning with finite labels]\label{thm:finite-full}
Let $k=|\Y|<\infty$ and let $\H\subseteq(\cA\cup\{\star\})^{\X}$ be a nonempty class. The following conditions are equivalent:
\begin{enumerate}[label=(\roman*)]
\item $\dDS(\H)<\infty$;
\item $\dN(\H)<\infty$;
\item $\H$ is realizably PAC learnable with full-set feedback;
\item $\H$ is agnostically PAC learnable with full-set feedback, with the benchmark \eqref{eq:agnostic}.
\end{enumerate}
\end{theorem}

The main tool is a deterministic rule that predicts the labels of a known pool of $n$ inputs one at a time and makes $O_\ell(d\log(nk))$ mistakes, whatever the target and the order of presentation (Theorem~\ref{thm:online}). It is an analogue of the halving algorithm, and its potential counts the products of conflicting sets that the remaining hypotheses still realize. This technique was initially inspired by work on transductive and adversary-directed online learning \cite{hanneke2024multiclass,hanneke2025list,hanneke2026adversary}.

All our upper bounds come from this rule. A leave-one-out argument turns it into realizable learners, both in expectation and with high probability (Section~\ref{ff:sec:realizable}). Since the rule changes its state only after a mistake, a run is determined by the few positions where it makes a mistake. Listing the possible runs therefore gives a small finite cover of the class on every pool (Section~\ref{ff:sec:covers}), and exponential weights over this cover give the agnostic learner (Section~\ref{ff:sec:agnostic}). For the lower bounds (Section~\ref{ff:sec:lower}), a DS witness or a product witness leaves the correct-label set of an unseen test input undetermined, so that every prediction there is wrong with constant probability. Section~\ref{ff:sec:proof} assembles the proof of Theorem~\ref{thm:finite-full}, and Section~\ref{sec:finite-bandit} adds two further equivalent conditions for bandit feedback.

\subsection{A Transductive Mistake Bound}\label{ff:sec:counts}\label{ff:sec:products}

As in the one-inclusion graph approach of \cite{haussler1994predicting}, we first solve a transductive problem, in which the inputs are known in advance. In a \emph{transductive prediction problem} the learner receives an unlabeled pool $S=(x_1,\ldots,x_n)\in\X^n$, possibly with repetitions. Positions $i\in[n]$ are then presented one at a time, in any order and possibly more than once. When position $i$ is presented, the learner predicts a label in $\Y$, and then, if $x_i\in\supp(h^\star)$, it observes the correct-label set $h^\star(x_i)$. A \emph{mistake} is a prediction outside $h^\star(x_i)$ at a position with $x_i\in\supp(h^\star)$.

Our rule keeps a set of candidate patterns on the pool. Let
\[
 W_0=\{(h(x_1),\ldots,h(x_n)):h\in\H\}\subseteq(\cA\cup\{\star\})^n ,
\]
and call its elements \emph{partial patterns}. Unlike $\H|_S$, the set $W_0$ also keeps the patterns of hypotheses that are undefined on part of the pool; the covers of Section~\ref{ff:sec:covers} need them. For $W\subseteq W_0$, a position $i$ and a set $A\in\cA$, put
\begin{equation}\label{ff:eq:update}
 W_{i,A}=\{v\in W:v_i=A\}.
\end{equation}

The halving algorithm counts the hypotheses that are still consistent, and every mistake removes at least half of them. Here the number of patterns can be exponential in $n$, so counting them would give a mistake bound linear in $n$. We count instead the products of minimal empty-intersection families, placed at a few positions of the pool, that $W$ still realizes. Each such product is a product witness of $\H$, so, as in the Sauer--Shelah lemma, their number $\Phi(W)$ is at most a polynomial of degree $d$ in $n$. The main inequality (Lemma~\ref{lem:potential-inequality}) says that if $\cB$ is a minimal empty-intersection family of $r$ sets, then the potentials $\Phi(W_{i,A})$, $A\in\cB$, sum to at most $(r-1)\Phi(W)$. Since $r\le\ell+1$, the sets that keep more than an $\ell/(\ell+1)$ fraction of the potential share a label, and a mistake on this label multiplies the potential by at most $\ell/(\ell+1)$.

We now make this precise. Let $\mathfrak C$ be the collection of all minimal empty-intersection families contained in $\cA$. By Lemma~\ref{lem:small}, each of them has between $2$ and $\ell+1$ members, and therefore, with $q=|\cA|$,
\begin{equation}
 \Gamma=|\mathfrak C|\le\sum_{r=2}^{\min\{\ell+1,q\}}\binom qr.
 \label{eq:finite-Gamma}
\end{equation}
Fix a pool $S=(x_1,\ldots,x_n)$. For $W\subseteq W_0$, a \emph{product specification} of $W$ consists of a set $T\subseteq[n]$, its \emph{support}, and families $\cB_i\in\mathfrak C$ for $i\in T$, such that the product $\prod_{i\in T}\cB_i$ is realized by patterns in $W$ that are defined on $T$:
\begin{equation}\label{ff:eq:products}
 \prod_{i\in T}\cB_i\subseteq
 \{(v_i)_{i\in T}:v\in W,\ v_i\ne\star\text{ for every }i\in T\}.
\end{equation}
The \emph{potential} $\Phi(W)$ is the number of product specifications of $W$; the empty specification is counted once when $W\ne\emptyset$, and $\Phi(\emptyset)=0$. Define
\begin{equation}
 \Pi_d(n)=\sum_{j=0}^{\min\{d,n\}}\binom nj\Gamma^j,
 \label{eq:finite-Pi}
\end{equation}
where $\Gamma^0=1$. If $d<\infty$, then
\begin{equation}\label{ff:eq:phi}
 1\le\Phi(W)\le\Pi_d(n)\qquad(W\ne\emptyset).
\end{equation}
Indeed, the lower bound counts the empty specification. For the upper bound, note first that repeated inputs cannot both lie in a support: if $i\ne i'$ and $x_i=x_{i'}$, then every $v\in W_0$ has $v_i=v_{i'}$, whereas the product contains vectors with different sets at $i$ and $i'$, since every family in $\mathfrak C$ has at least two members. So the inputs $x_i$, $i\in T$, are distinct, and the right side of \eqref{ff:eq:products} is contained in $\H|_{(x_i)_{i\in T}}$. A specification with $|T|>d$ would thus show that these inputs are product-shattered, contradicting $\dN(\H)=d$. Hence $|T|\le\min\{d,n\}$, and there are at most $\binom nj\Gamma^j$ specifications with $|T|=j$.

\begin{lemma}[Product-counting inequality]\label{lem:potential-inequality}
Let $W\subseteq W_0$, let $i\in[n]$, and let $\cB\in\mathfrak C$ have $r$ members. Then
\begin{equation}\label{ff:eq:phi-ineq}
 \sum_{A\in\cB}\Phi(W_{i,A})\le(r-1)\Phi(W).
\end{equation}
\end{lemma}

\begin{proof}
Every specification of $W_{i,A}$ is a specification of $W$. Its support does not contain $i$, because all vectors of $W_{i,A}$ have the same set $A$ at $i$, while each family in $\mathfrak C$ has at least two members. Let $\cU$ be the set of specifications of $W$ whose support does not contain $i$, and for $p\in\cU$ let $r_p$ be the number of sets $A\in\cB$ such that $p$ is a specification of $W_{i,A}$. Since $r_p\le r-1+\ind\{r_p=r\}$,
\[
 \sum_{A\in\cB}\Phi(W_{i,A})
 =\sum_{p\in\cU}r_p
 \le(r-1)|\cU|+|\{p\in\cU:r_p=r\}|.
\]
Now let $p\in\cU$ have support $T$ and $r_p=r$. Then every vector of the product specified by $p$, combined with every $A\in\cB$ at coordinate $i$, is realized by some $v\in W$ that is defined on $T\cup\{i\}$. So adding the family $\cB$ at position $i$ extends $p$ to a specification of $W$ with support $T\cup\{i\}$. Distinct $p$ give distinct extensions, and none of them lies in $\cU$. Therefore $\Phi(W)\ge|\cU|+|\{p\in\cU:r_p=r\}|$, and since $r\ge2$, the right side of the display is at most $(r-1)\Phi(W)$.
\end{proof}

The rule predicts a label shared by all the sets that keep a large part of the potential. For $W\subseteq W_0$ and $i\in[n]$, let
\[
 \mathcal G=\Bigl\{A\in\cA:\Phi(W_{i,A})>\frac{\ell}{\ell+1}\Phi(W)\Bigr\},
\]
and let $y_i(W)$ be the first label of $\bigcap_{A\in\mathcal G}A$ in the fixed enumeration of $\Y$, or the first label of $\Y$ if $\mathcal G=\emptyset$ (for instance, if $W=\emptyset$). The next lemma shows that the intersection is nonempty, so $y_i(W)$ is well defined.

\begin{lemma}[A prediction that shrinks the potential on a mistake]\label{lem:predict}
Let $W\subseteq W_0$ be nonempty and let $i\in[n]$. If $\mathcal G\ne\emptyset$, then $\bigcap_{A\in\mathcal G}A\ne\emptyset$. Moreover,
\begin{equation}\label{ff:eq:predict}
 y_i(W)\notin A\ \Longrightarrow\
 \Phi(W_{i,A})\le\frac\ell{\ell+1}\Phi(W)
 \qquad(A\in\cA).
\end{equation}
\end{lemma}

\begin{proof}
Suppose that $\mathcal G\ne\emptyset$ and $\bigcap_{A\in\mathcal G}A=\emptyset$. By Lemma~\ref{lem:small}, $\mathcal G$ contains a family $\cB\in\mathfrak C$ with some number $r\le\ell+1$ of members. Then $r\ell/(\ell+1)\ge r-1$, and so
\[
 \sum_{A\in\cB}\Phi(W_{i,A})>\frac{r\ell}{\ell+1}\Phi(W)\ge(r-1)\Phi(W),
\]
which contradicts Lemma~\ref{lem:potential-inequality}. Hence the intersection is nonempty. If $y_i(W)\notin A$, then $A\notin\mathcal G$, which is \eqref{ff:eq:predict}.
\end{proof}

\begin{algbox}
\centerline{\textbf{Algorithm 1: Conservative transductive rule}}
\smallskip
\noindent\textbf{Input:} a pool $S=(x_1,\ldots,x_n)$; positions of $S$ presented one at a time.
\begin{enumerate}
\item Set $W=W_0$.
\item When position $i$ is presented, predict $y_i(W)$.
\item If a correct-label set $A=h^\star(x_i)$ is revealed and $y_i(W)\notin A$, replace $W$ by $W_{i,A}$. Otherwise keep $W$.
\end{enumerate}
\end{algbox}

We call $W$ the \emph{version space} of the rule. A rule is \emph{conservative} if it changes its state only after a mistake, as Algorithm~1 does. Its mistake bound is the logarithm of the initial potential in base $1+1/\ell$:
\begin{equation}
 M_d(n)=\left\lceil\frac{\log\Pi_d(n)}{\log(1+1/\ell)}\right\rceil.
 \label{eq:finite-M}
\end{equation}
Both $\Pi_d(n)$ and $M_d(n)$ are nondecreasing in $n$. For finite $d$, since $\binom nj\le n^j\le\binom dj n^j$ for $j\le d$, the binomial theorem gives $\Pi_d(n)\le\sum_{j\le d}\binom dj(n\Gamma)^j=(1+n\Gamma)^d$, and hence
\begin{equation}\label{ff:eq:M-log}
 M_d(n)\le\left\lceil\frac{d\log(1+n\Gamma)}{\log(1+1/\ell)}\right\rceil.
\end{equation}

\begin{theorem}[Conservative transductive mistake bound]\label{thm:online}
Assume $d<\infty$, and let $S$ be a pool of $n$ inputs. Algorithm~1 is deterministic and conservative. For every target $h^\star\in\H$ and every order of presentation, with repetitions allowed, it makes at most $M_d(n)$ mistakes.
\end{theorem}

\begin{proof}
The rule is clearly deterministic and conservative. Let $v^\star=(h^\star(x_1),\ldots,h^\star(x_n))\in W_0$. An update after a mistake at $i$ uses $A=h^\star(x_i)=v^\star_i$, so $v^\star$ stays in $W$. Hence $W$ is never empty, and $\Phi(W)\ge1$ by \eqref{ff:eq:phi}. On a mistake we have $y_i(W)\notin A$, so by \eqref{ff:eq:predict} the update multiplies $\Phi(W)$ by at most $\ell/(\ell+1)$, and all other steps leave $W$ unchanged. Thus after $s$ mistakes
\[
 1\le\Phi(W)\le\Pi_d(n)\left(\frac{\ell}{\ell+1}\right)^s,
\]
where the second inequality uses \eqref{ff:eq:phi} for $W_0$. Taking logarithms, and using that $s$ is an integer, gives $s\le M_d(n)$.
\end{proof}

\begin{remark}[Labeling the one-inclusion hypergraph]\label{prop:hypergraph-labeling}
When all correct-label sets are singletons, a labeling of the hyperedges of the one-inclusion hypergraph by labels of their vertices is an orientation \cite{haussler1994predicting,daniely2014optimal,brukhim2022characterization}, and Theorem~\ref{thm:online} gives the analogue of an orientation with small out-degree. Precisely, assume $d<\infty$ and let $S\in\X^n$. For every finite nonempty $V\subseteq\H|_S$ there is a map $\phi$ from the hyperedges of the one-inclusion hypergraph of $V$ to $\Y$ such that
\begin{equation}\label{ff:eq:hyperedge-labeling}
 \max_{v\in V}\sum_{i=1}^n
 \ind\{\phi(e_{i,v_{-i}})\notin v_i\}\le M_d(n).
\end{equation}
Indeed, all vertices of $e_{i,v_{-i}}$ agree at the coordinates $1,\ldots,i-1$; let $\phi(e_{i,v_{-i}})$ be the prediction of Algorithm~1 at position $i$ after positions $1,\ldots,i-1$ are presented in this order with these common sets. For each $v\in V$, the labels $\phi(e_{i,v_{-i}})$, $i\in[n]$, are then the predictions of Algorithm~1 in the order $1,\ldots,n$ against a target realizing $v$, and Theorem~\ref{thm:online} gives \eqref{ff:eq:hyperedge-labeling}.
\end{remark}

\subsection{Finite Covers}\label{ff:sec:covers}

A \emph{finite cover} of $\H$ on a pool $S=(x_1,\ldots,x_n)$ is a finite family $F\subseteq\Y^{n}$ with the following \emph{covering property}: for every $h\in\H$ some $f\in F$ satisfies $f(i)\in h(x_i)$ at every position $i$ with $x_i\in\supp(h)$. Because Algorithm~1 is conservative, its run against a target is determined by the positions of its mistakes and the sets revealed there, and there are few such runs; correcting the wrong predictions of each run gives the cover.

\begin{lemma}[Finite cover of a partial class]\label{lem:finite-cover}
Assume $d<\infty$. For every pool $S=(x_1,\ldots,x_n)$ there is a finite cover $F_S\subseteq\Y^n$ of $\H$ on $S$, which depends only on $S$ and $\H$, such that
\begin{equation}\label{eq:cover-bound}
 |F_S|\le
 \sum_{j=0}^{\min\{n,M_d(n)\}}\binom nj(qk)^j
 \le(1+nqk)^{M_d(n)}.
\end{equation}
\end{lemma}

\begin{proof}
We simulate Algorithm~1 with the positions presented in the order $1,\ldots,n$, guessing where it makes mistakes. For every set $I\subseteq[n]$ with $|I|\le M_d(n)$ and every choice of $A_i\in\cA$ and $b_i\in A_i$ for $i\in I$, we form one vector $f\in\Y^n$ as follows. Start with $W=W_0$. At a position $i\in I$, set $f(i)=b_i$ and replace $W$ by $W_{i,A_i}$. At a position $i\notin I$, set $f(i)=y_i(W)$ and keep $W$. Let $F_S$ be the set of all vectors obtained in this way; the construction uses only $S$ and $\H$. For $|I|=j$ there are $\binom nj$ choices of $I$ and at most $(qk)^j$ choices of the pairs $(A_i,b_i)$, which gives the first bound in \eqref{eq:cover-bound}. The second follows from $\binom nj\le n^j$ and $\sum_{j=0}^M(nqk)^j\le(1+nqk)^M$ for $M=M_d(n)$.

Now fix $h\in\H$. Run Algorithm~1 against the target $h$ in the order $1,\ldots,n$, and let $I$ be the set of positions where it makes a mistake; $|I|\le M_d(n)$ by Theorem~\ref{thm:online}. Choose this $I$, the sets $A_i=h(x_i)$, and any labels $b_i\in h(x_i)$ for $i\in I$. The simulation and the run then update at the same positions with the same sets, so by induction on the position they have the same version space at every step. At a position $i\in I$ we have $f(i)=b_i\in h(x_i)$. At a position $i\notin I$ with $x_i\in\supp(h)$, the value $f(i)$ is the prediction of Algorithm~1, which is not a mistake. Hence $f(i)\in h(x_i)$ at every position with $x_i\in\supp(h)$.
\end{proof}

\subsection{Lower Bounds}\label{ff:sec:lower}

Both bounds draw the target at random from a witness, so that the correct-label set at a test input missing from the sample is not determined by the sample. The first bound is reused in Sections~\ref{sec:finite-bandit} and~\ref{sec:finite-one} for the other two feedback models.

\begin{proposition}[DS lower bound]\label{prop:ds-lower}
Let $k<\infty$ and let $\H$ be a partial class with $\dDS(\H)\ge n\ge2$. For every learner that uses $m$ full-set examples there exist a distribution $P$ and $h^\star\in\H$ with $P(\supp(h^\star))=1$ such that
\begin{equation}\label{ff:eq:lower-ds}
 \E_{\substack{\sigma\sim\D_{P,h^\star}^{m}\\\omega\sim\Unif([0,1])}}\bigl[L_{P,h^\star}(\widehat g)\bigr]\ge\frac1q\left(1-\frac1n\right)^m,
\end{equation}
where $\widehat g$ is the returned predictor. The same holds for every learner that uses $m$ examples with bandit feedback, and for every learner that uses $m$ examples with one-correct-label feedback, where the demonstrated label is drawn uniformly from $h^\star(x)$.
\end{proposition}

\begin{proof}
Since $\dDS(\H)\ge n$, the projection part of Lemma~\ref{lem:orders} gives distinct inputs $x_1,\ldots,x_n$ and a DS witness $V\subseteq\H|_{(x_1,\ldots,x_n)}$. For each $v\in V$ fix $h_v\in\H$ with $(h_v(x_1),\ldots,h_v(x_n))=v$. Let $P$ be uniform on $\{x_1,\ldots,x_n\}$, so that $P(\supp(h_v))=1$. We choose $v$ uniformly from $V$, use the target $h_v$, and prove \eqref{ff:eq:lower-ds} on average over $v$; then it holds for some $h_v$.

Draw a test point $x_i\sim P$ independently of the sample, and condition on the event that $x_i$ is not among the training inputs. Give the learner $v_{-i}$ as well; this can only help it. The training observations are functions of $v_{-i}$, the training inputs and the seed, so the conditional law of $v$ is uniform on the hyperedge $e_{i,v_{-i}}$. Distinct vertices of this hyperedge differ at coordinate $i$, so it has at most $q$ vertices, and by the definition of a DS witness every label $y$ is missing from $w_i$ for some $w\in e_{i,v_{-i}}$. Hence any prediction at $x_i$, deterministic or randomized, is wrong with conditional probability at least $1/|e_{i,v_{-i}}|\ge1/q$. Finally, a uniform test point is missing from $m$ independent uniform training points with probability $(1-1/n)^m$. Averaging over the target, the sample and the test point gives \eqref{ff:eq:lower-ds}.

For the other feedback models, note that a full-set learner can answer every membership query $\ind\{u\in h^\star(x_j)\}$ and can draw a label uniformly from $h^\star(x_j)$ with its own seed. So every bandit learner, and every one-correct-label learner under the uniform law of demonstrations, is simulated by a full-set learner with the same output, and the bound transfers. A lower bound for one admissible law of the demonstrations is enough, since a learner must succeed under all of them.
\end{proof}

The second bound replaces the factor $1/q$ by $1/(\ell+1)$ and gives the linear dependence of the sample complexity on $d$.

\begin{proposition}[Product-witness lower bound]\label{prop:nat-lower}
If $\dN(\H)\ge n\ge2$, then for every learner that uses $m$ full-set examples there are $P$ and $h^\star\in\H$ with $P(\supp(h^\star))=1$ such that
\begin{equation}\label{ff:eq:nat-expected}
 \E\bigl[L_{P,h^\star}(\widehat g)\bigr]\ge
 \frac1{\ell+1}\left(1-\frac1n\right)^m.
\end{equation}
Consequently, if $d\ge2$, then for $0<\eps\le1/[8(\ell+1)]$ and $0<\delta<1/[4(\ell+1)]$,
\begin{equation}\label{ff:eq:nat-pac}
 \mfull_{\H}(\eps,\delta)>
 \frac{d-1}{8(\ell+1)\eps},
\end{equation}
and if $d=\infty$, this holds with every integer $n\ge2$ in place of $d$.
\end{proposition}

\begin{proof}
Take a product-shattered sequence $x_1,\ldots,x_n$ with families $\cB_1,\ldots,\cB_n$, and replace each family by a minimal empty-intersection subfamily; by Lemma~\ref{lem:small}, these have between $2$ and $\ell+1$ members. Choose the sets $v_i\in\cB_i$ independently and uniformly, fix one target $h_v\in\H$ realizing each product vector $v$, and let $P$ be uniform on $\{x_1,\ldots,x_n\}$. Since the sets at distinct inputs are independent, the set $v_i$ at an input $x_i$ absent from the training sample is independent of all observed sets, and it is uniform on $\cB_i$. Every label is excluded by some member of $\cB_i$, so the conditional probability of an error at $x_i$ is at least $1/|\cB_i|\ge1/(\ell+1)$. The test input is missing with probability $(1-1/n)^m$, which gives \eqref{ff:eq:nat-expected}.

For the PAC statement we make the unseen inputs rare. Fix the set at $x_n$ to one member of $\cB_n$ and keep the independent uniform choices at $x_1,\ldots,x_{n-1}$. Let $a=4(\ell+1)\eps\le1/2$, put mass $a/(n-1)$ on each of $x_1,\ldots,x_{n-1}$ and the remaining mass $1-a$ on $x_n$, and let $Z$ be the part of the risk contributed by $x_1,\ldots,x_{n-1}$. The argument above gives
\[
 0\le Z\le a,\qquad
 \E[Z]\ge\frac a{\ell+1}\left(1-\frac a{n-1}\right)^m.
\]
If $m\le(n-1)/[8(\ell+1)\eps]=(n-1)/(2a)$, then $(1-a/(n-1))^m\ge1-ma/(n-1)\ge1/2$, and so $\E[Z]\ge2\eps$. The reverse Markov inequality for $Z\in[0,a]$, which follows from $Z\le\eps+a\ind\{Z>\eps\}$, then gives
\[
 \Pr[Z>\eps]\ge\frac{\E[Z]-\eps}a\ge\frac\eps a=\frac1{4(\ell+1)}>\delta.
\]
The risk is at least $Z$, and some fixed target attains at least this average failure probability. Taking $n=d$ gives \eqref{ff:eq:nat-pac}.
\end{proof}

Proposition~\ref{prop:nat-lower} holds for every label space, since its proof does not use the finiteness of $\Y$. The term $d-1$ cannot be replaced by $d$ for all classes: a class of two distinct constant singleton hypotheses has $\dN=1$ and is identified from one example.

\subsection{Realizable Full-Set Learning}\label{ff:sec:realizable}

We first turn Algorithm~1 into an inductive learner by the leave-one-out argument of \cite{haussler1994predicting}: the learner places the query among the training inputs in a random order, and its expected risk is the average number of mistakes per position of Algorithm~1.

\begin{theorem}[Expected risk]\label{thm:full-expected}
Assume $d<\infty$. For every $m\ge0$ there is a full-set learner such that for every $h^\star\in\H$ and every $P$ with $P(\supp(h^\star))=1$,
\begin{equation}\label{ff:eq:full-expect}
 \E_{\substack{\sigma\sim\D_{P,h^\star}^{m}\\\omega\sim\Unif([0,1])}}\bigl[L_{P,h^\star}(\widehat g)\bigr]\le\frac{M_d(m+1)}{m+1}.
\end{equation}
\end{theorem}

\begin{proof}
We describe the learner and then compare it with a single run of Algorithm~1 on $m+1$ examples.

\paragraph{The learner.}
Given $\sigma=((x_1,A_1),\ldots,(x_m,A_m))$ and a query $x$, let $n=m+1$ and add $x$ to the training inputs as position $n$. Use the seed to draw a uniformly random ordering of the $n$ positions. Run Algorithm~1 on the pool listed in this order, revealing the observed sets $A_j$ at the training positions that come before the query, and return the prediction at the query. The ordering is a function of the seed and the query, so for fixed $\sigma$ and $\omega$ the learner returns a fixed predictor $\widehat g$; the agnostic learners below use the seed in the same way.

\paragraph{Analysis.}
Draw $n$ independent examples $(x_j,h^\star(x_j))$; almost surely all $x_j$ lie in $\supp(h^\star)$. List them in a uniformly random order and run Algorithm~1 on the whole sequence, revealing every set. The prediction at the $t$-th listed position depends on the observed sets only through those revealed before it, so it is the prediction of the learner trained on the other $n-1$ examples with the $t$-th one as query. By exchangeability, the loss of the learner at a test point has the same law as the loss at a uniformly chosen listed position. For each fixed sequence and order, Algorithm~1 makes at most $M_d(n)$ mistakes by Theorem~\ref{thm:online}, so the average loss over a uniform position is at most $M_d(n)/n$. Averaging over the examples and the order gives \eqref{ff:eq:full-expect}.
\end{proof}

Markov's inequality turns Theorem~\ref{thm:full-expected} into a PAC bound with a factor $1/\delta$. To get $\log(1/\delta)$ instead, we use the leave-one-out conversion of \cite{aden2023optimal}. Write $\sigma^{-i}$ for a sample $\sigma$ of size $n$ with its $i$-th example deleted, and $\sigma_{\le t}$ for its first $t$ examples. The conversion takes a deterministic, permutation-invariant rule $R(\sigma;x)$ that makes few leave-one-out mistakes, that is, such that on every realizable sample $\sigma=((x_i,h^\star(x_i)))_{i=1}^n$,
\begin{equation}\label{ff:eq:loo-condition}
 \sum_{i=1}^n\ind\{R(\sigma^{-i};x_i)\notin h^\star(x_i)\}\le M_n ,
\end{equation}
and it shows that, with high probability, the rules $R(\sigma_{\le t};\cdot)$, $n/4\le t<n$, have small average risk. Their theorem allows a general $\{0,1\}$-valued loss, so it applies to the loss $\ind\{y\notin A\}$. We obtain a suitable rule by running Algorithm~1 on the distinct sample inputs and the query, listed in a fixed order on $\X$.

\begin{theorem}[High-probability bound]\label{thm:hp}
Fix any total order on $\X$. Assume $d<\infty$, let $\delta\in(0,1)$, and let $n$ be a positive multiple of $4$. There is a deterministic learner such that for every $h^\star\in\H$ and every $P$ with $P(\supp(h^\star))=1$, with probability at least $1-\delta$ over $\sigma\sim\D_{P,h^\star}^n$, the returned predictor $\widehat g$ satisfies
\begin{equation}\label{ff:eq:hp}
 L_{P,h^\star}(\widehat g)\le
 \frac{4.82(\ell+1)}n\left(M_d(n)+\log\frac2\delta\right).
\end{equation}
\end{theorem}

\begin{proof}
We build a rule that satisfies \eqref{ff:eq:loo-condition} with $M_n=M_d(n)$, apply the conversion, and take a plurality vote.

\paragraph{A leave-one-out rule.}
Given a sample $\sigma$ and a query $x$, define $R(\sigma;x)$ as follows. If $\sigma$ is not realizable by $\H$, return a fixed label. If $x$ is one of the sample inputs, return the first label of its observed set. Otherwise, list the distinct sample inputs together with $x$ in increasing order and run Algorithm~1 on this pool, presenting only the inputs smaller than $x$, in increasing order and with their observed sets; return its prediction at $x$. This rule depends only on the sample as an unordered collection, so it is permutation invariant.

\paragraph{The leave-one-out condition.}
Fix a realizable sample $\sigma=((x_i,h^\star(x_i)))_{i=1}^n$. If $x_i$ occurs at least twice in $\sigma$, then it occurs in $\sigma^{-i}$, and $R(\sigma^{-i};x_i)$ is correct. If $x_i$ occurs once, the pool used by $R(\sigma^{-i};x_i)$ is the list of distinct inputs of $\sigma$, the same for all such $i$, and the inputs smaller than $x_i$ all remain in $\sigma^{-i}$. So $R(\sigma^{-i};x_i)$ is the prediction at $x_i$ of a single run of Algorithm~1 on the distinct inputs of $\sigma$ in increasing order, with every set revealed. This pool has at most $n$ positions, so by Theorem~\ref{thm:online} and the monotonicity of $M_d$, the run makes at most $M_d(n)$ mistakes. Thus \eqref{ff:eq:loo-condition} holds with $M_n=M_d(n)$.

\paragraph{Applying the conversion.}
For a fixed target, we apply Theorem~2.1 of \cite{aden2023optimal} on the input space $\supp(h^\star)$, with the loss $\ind\{y\notin A\}$ and the rule $R$ restricted there.\footnote{Formally, the label alphabet is the disjoint union of $\Y$ and $\cA$; the loss of $y\in\Y$ against $A\in\cA$ is $\ind\{y\notin A\}$, the loss of a set against itself is zero, and the loss of the remaining pairs, which never arise, is an arbitrary value in $\{0,1\}$.} Every realizable sample there satisfies \eqref{ff:eq:loo-condition} with $M_n=M_d(n)$, and the rule itself is defined on all of $\X$ and does not depend on $h^\star$. The theorem gives, with probability at least $1-\delta$,
\begin{equation}\label{ff:eq:loo-suffix}
 \frac{4}{3n}\sum_{t=n/4}^{n-1}L_{P,h^\star}\bigl(R(\sigma_{\le t};\cdot)\bigr)
 \le\frac{4.82}{n}\left(M_d(n)+\log\frac2\delta\right),
\end{equation}
where the left side is the average risk of the $3n/4$ rules.

\paragraph{Plurality.}
Let $\widehat g(x)$ be the plurality of the $3n/4$ labels $R(\sigma_{\le t};x)$, $n/4\le t<n$, and let $N(x)$ be the number of these labels outside $h^\star(x)$. By Lemma~\ref{lem:mode} applied to the votes, $\ind\{\widehat g(x)\notin h^\star(x)\}\le(\ell+1)N(x)/(3n/4)$. Integrating over $x\sim P$ and using \eqref{ff:eq:loo-suffix} proves \eqref{ff:eq:hp}.
\end{proof}

\begin{corollary}[Realizable sample complexity]\label{cor:finite-full-rate}
If $d=\dN(\H)<\infty$, then
\[
\mfull_\H(\eps,\delta)\le\frac{c}{\eps}\Bigl(d\log\frac{(d+1)(k+1)}{\eps}+\log\frac{2}{\delta}\Bigr),
\]
where $c$ depends only on $\ell$.
\end{corollary}

\begin{proof}
We have $q\le(k+1)^\ell$ and $\Gamma\le(q+1)^{\ell+1}$, so \eqref{ff:eq:M-log} gives $M_d(n)\le c\,d\log(2n(k+1))$ for $d\ge1$, while $M_0(n)=0$. Thus the right side of \eqref{ff:eq:hp} is at most $\eps$ as soon as $n\ge c\,(d\log(2n(k+1))+\log(2/\delta))/\eps$. Since $n\ge2a\log(2ab)$ implies $n\ge a\log(bn)$ whenever $ab\ge e$, a routine calculation shows that this holds, with a larger $c$, for every $n$ divisible by $4$ that is at least the bound of the corollary.
\end{proof}

\subsection{Agnostic Full-Set Learning}\label{ff:sec:agnostic}

Recall that $\D$ is a distribution on $\X\times\cA$ and that the benchmark $\Lset_\D(\H)$ in \eqref{eq:agnostic} counts an undefined value as a mismatch. The cover of Lemma~\ref{lem:finite-cover} depends only on the inputs, so its members can serve as experts before any set is seen, and exponential weights over them compete with every hypothesis in $\H$.

\begin{theorem}[Expected excess risk]\label{thm:agnostic-full}
Assume $d<\infty$. For every $m\ge0$ there is a randomized full-set learner such that for every $\D$,
\begin{equation}\label{ff:eq:agnostic-full}
 \E_{\substack{\sigma\sim\D^{m}\\\omega\sim\Unif([0,1])}}\bigl[L_\D(\widehat g)\bigr]
 \le \Lset_\D(\H)+\sqrt{\frac{2M_d(m+1)\log(1+(m+1)qk)}{m+1}}.
\end{equation}
\end{theorem}

\begin{proof}
Let $n=m+1$. We first bound the regret on a known pool and then pass to the inductive learner.

\paragraph{Exponential weights on a known pool.}
Fix a pool $S$ of $n$ inputs with sets $A_1,\ldots,A_n\in\cA$, revealed one at a time in the order of the pool, and let $F_S$ be the cover of Lemma~\ref{lem:finite-cover}. Run exponential weights with the vectors $f\in F_S$ as experts and the losses $\lambda_i(f)=\ind\{f(i)\notin A_i\}$: at position $i$, the learner draws an expert with probability proportional to its current weight and predicts its label $\widehat Y_i$. The standard Hedge bound for losses in $[0,1]$, with a suitable learning rate, gives for every $f\in F_S$
\begin{equation}\label{ff:eq:full-regret}
 \E_{\omega\sim\Unif([0,1])}\Bigl[\sum_{i=1}^n\ind\{\widehat Y_i\notin A_i\}\Bigr]
 \le\sum_{i=1}^n\lambda_i(f)+\sqrt{2n\log|F_S|}.
\end{equation}
For $h\in\H$, let $f_h\in F_S$ be given by the covering property. Then for every position $i$,
\[
 \ind\{f_h(i)\notin A_i\}\le\ind\{h(x_i)\ne A_i\},
\]
because $h(x_i)=A_i$ implies $x_i\in\supp(h)$ and hence $f_h(i)\in h(x_i)=A_i$.

\paragraph{The inductive learner.}
Given $\sigma$ and a query $x$, add $x$ to the training inputs and list the $n$ positions in a uniformly random order. Run exponential weights with the cover of this ordered pool, processing the training examples before the query, and return its prediction at the query. Since the cover depends only on the inputs and their order, the argument of Theorem~\ref{thm:full-expected} shows that the expected test loss equals the expected average loss of exponential weights over $n$ independent examples from $\D$. Applying \eqref{ff:eq:full-regret} with $f=f_h$ for a fixed $h\in\H$, dividing by $n$ and taking expectations gives
\[
 \E\bigl[L_\D(\widehat g)\bigr]\le\Pr_{(x,A)\sim\D}[h(x)\ne A]+\sqrt{\frac{2\log\sup_{S\in\X^n}|F_S|}n},
\]
and the theorem follows from \eqref{eq:cover-bound} by taking the infimum over $h\in\H$.
\end{proof}

A standard confidence-boosting step converts this expected bound into a high-probability one: we train independent candidates and keep the one with the smallest validation loss.

\begin{proposition}[From expected to high-probability excess risk]\label{prop:agnostic-full-pac}
Let $\eps,\delta\in(0,1)$, and suppose a full-set learner trained on $m$ examples satisfies $\E[L_\D(\widehat g)]\le\Lset_\D(\H)+\eps/8$ for every $\D$. Run it independently on $b$ blocks of $m$ training examples, and select the candidate with the smallest empirical loss on a common independent validation set of $v$ examples, where
\[
 b=\left\lceil\frac8{3\eps}\log\frac2\delta\right\rceil,
 \qquad
 v=\left\lceil\frac8{\eps^2}\log\frac{4b}{\delta}\right\rceil.
\]
With probability at least $1-\delta$, the selected predictor has risk at most $\Lset_\D(\H)+\eps$.
\end{proposition}

\begin{proof}
Let $a=\Lset_\D(\H)$; we may assume $a+\eps<1$, since otherwise there is nothing to prove. By Markov's inequality, one run has risk at most $a+\eps/2$ with probability at least
\[
 1-\frac{a+\eps/8}{a+\eps/2}
 =\frac{3\eps/8}{a+\eps/2}\ge\frac{3\eps}{8},
\]
since $a+\eps/2<1$. Hence, except with probability at most $(1-3\eps/8)^b\le\delta/2$, some candidate has risk at most $a+\eps/2$. Conditionally on the candidates $\widehat g_1,\ldots,\widehat g_b$, the validation losses of each candidate are independent Bernoulli variables. Let $\widehat L_j$ be the empirical validation loss of $\widehat g_j$. By Hoeffding's inequality and a union bound over $j\le b$,
\[
 \max_{j\le b}\bigl|\widehat L_j-L_\D(\widehat g_j)\bigr|\le\eps/4
\]
with probability at least $1-\delta/2$. On both events, the risk of the selected candidate exceeds the smallest risk among the candidates by at most $\eps/2$, so it is at most $a+\eps$.
\end{proof}

\subsection{Proof of Theorem~\ref{thm:finite-full}}\label{ff:sec:proof}

\begin{proof}[Proof of Theorem~\ref{thm:finite-full}]
(i)$\Rightarrow$(ii) follows from $\dN(\H)\le\dDS(\H)$ in Lemma~\ref{lem:orders}.

(ii)$\Rightarrow$(iii) follows from Corollary~\ref{cor:finite-full-rate}, which bounds $\mfull_\H(\eps,\delta)$ for all $\eps,\delta\in(0,1)$.

(ii)$\Rightarrow$(iv). By \eqref{ff:eq:M-log}, the excess term in \eqref{ff:eq:agnostic-full} tends to zero as $m\to\infty$. So for $m$ large the learner of Theorem~\ref{thm:agnostic-full} satisfies the assumption of Proposition~\ref{prop:agnostic-full-pac}, which gives agnostic PAC learnability.

(iv)$\Rightarrow$(iii). Every realizable distribution $\D_{P,h^\star}$ has benchmark zero and satisfies $L_{\D}(g)=L_{P,h^\star}(g)$ for every predictor $g$ (Section~\ref{sec:prelim-models}), so an agnostic learner is a realizable learner with the same sample size.

(iii)$\Rightarrow$(i). Suppose $\dDS(\H)=\infty$, and let $\eps=\delta=1/(5q)$, so that $\eps+\delta<1/(2q)$. Fix any learner with $m$ examples. By Proposition~\ref{prop:ds-lower} with $n$ large enough that $(1-1/n)^m\ge1/2$, some target has expected risk $\E[L]\ge1/(2q)$. Since the risk $L$ is at most $1$, we have $\E[L]\le\eps+\Pr[L>\eps]$, and so
\[
 \Pr[L>\eps]\ge\frac1{2q}-\frac1{5q}>\delta.
\]
So no finite sample size works, and $\H$ is not realizably PAC learnable.
\end{proof}


\section{Bandit Learning with Finite Labels}\label{sec:finite-bandit}

We now show that when the label space is finite, $k=|\Y|<\infty$, learning from bandit feedback (Section~\ref{sec:prelim-models}) is characterized by the same conditions as learning from full sets.

\begin{theorem}[Bandit learning with finite labels]\label{thm:finite-bandit}
Let $k=|\Y|<\infty$ and let $\H\subseteq(\cA\cup\{\star\})^{\X}$ be a nonempty partial class. Conditions (i)--(iv) of Theorem~\ref{thm:finite-full} are equivalent to each of the following conditions:
\begin{enumerate}[label=(\roman*)]
\setcounter{enumi}{4}
\item $\H$ is realizably PAC learnable with bandit feedback;
\item $\H$ is agnostically PAC learnable with bandit feedback.
\end{enumerate}
Moreover, suppose that $d=\dN(\H)<\infty$, and let $q=|\cA|$ be the number of possible correct-label sets. For every $m\ge0$, with $n=m+1$, there are bandit learners with $m$ examples whose expected risks are at most
\begin{equation}\label{fb:eq:rates}
 \min\left\{1,\frac{kM_d(n)\log(1+nqk)}{n}\right\}
 \qquad\text{and}\qquad
 \Lset_\D(\H)+\sqrt{\frac{2kM_d(n)\log(1+nqk)}n}
\end{equation}
in the realizable and the agnostic case, respectively, where $M_d(n)$ is the mistake bound \eqref{eq:finite-M} of Theorem~\ref{thm:online} and $\Lset_\D(\H)$ is the exact-set mismatch benchmark in~\eqref{eq:agnostic}. Consequently, $\mband_\H$ and $\mbandag_\H$ satisfy the bounds \eqref{fb:eq:amp-real} and \eqref{fb:eq:amp-ag} of Proposition~\ref{prop:bandit-amplify}, where $m_{\mathrm B}(r)$ is the least $m$ for which the corresponding excess term in \eqref{fb:eq:rates} is at most $r$.
\end{theorem}

All our learners use the finite cover $F_S$ of Lemma~\ref{lem:finite-cover} as a set of experts. Since the cover is built from the inputs $S=(x_1,\ldots,x_n)$ alone, the experts are available before any correct-label set is seen. The difficulty is that bandit feedback reveals their losses only partly. A query $u$ at position $i$ tests only the experts recommending $u$, and it tests them all at once: if the answer is negative, all of them are wrong at $i$, and if it is positive, all of them are right there.

We first use this on a known pool, where the learner chooses its queries after seeing the whole pool (Section~\ref{fb:sec:trans}). We then treat the inductive problem, in which the training queries are made before the test input is known (Section~\ref{fb:sec:ind}), and finally turn the bounds on the expected risk into PAC bounds by validation (Section~\ref{fb:sec:amp}).

\subsection{Transductive Bounds}\label{fb:sec:trans}

In the transductive problem of Section~\ref{ff:sec:counts} with bandit feedback, the learner knows the pool $S=(x_1,\ldots,x_n)$ in advance; when position $i$ is presented, it predicts $\hat y_i\in\Y$ and observes only $\ind\{\hat y_i\in A_i\}$. On such a pool, the cover turns the halving algorithm into a bandit learner: predict the most popular label among the surviving experts, and discard its supporters after a mistake.

\begin{proposition}[Transductive mistake bound]\label{prop:trans-bandit}
Assume $d<\infty$ and $k\ge2$. For every pool $S$ of size $n$ there is a deterministic online bandit learner such that, for every $h^\star\in\H$ with $x_1,\ldots,x_n\in\supp(h^\star)$ and $A_i=h^\star(x_i)$, the number of mistakes is at most
\begin{equation}\label{fb:eq:mistake-bandit}
 \left\lceil \frac{\log|F_S|}{-\log(1-1/k)}\right\rceil
 \le\left\lceil kM_d(n)\log(1+nqk)\right\rceil.
\end{equation}
\end{proposition}

\begin{proof}
The learner keeps a set of surviving experts, initially $F_S$. At each round it predicts a label recommended by the largest number of surviving experts. After a negative answer it eliminates every surviving expert $f$ with $f(i)=\hat y_i$, and after a positive answer it changes nothing. Since $F_S$ is a cover, some $f\in F_S$ satisfies $f(i)\in h^\star(x_i)$ for every $i$, and this expert is never eliminated. On a mistake, all the surviving experts that recommended $\hat y_i$ are eliminated, and since $\hat y_i$ is a plurality label among $k$ labels, they form at least a $1/k$ fraction of the survivors. Thus after $r$ mistakes we have $1\le|F_S|(1-1/k)^r$. Taking logarithms gives the first inequality in \eqref{fb:eq:mistake-bandit}, and the second follows from \eqref{eq:cover-bound} and $-\log(1-1/k)\ge1/k$.
\end{proof}

Without realizability, the same cover gives a regret bound through the EXP4 algorithm of \cite{auer2002nonstochastic}.

\begin{remark}[Transductive regret]\label{rem:trans-regret}
For any pool $S=(x_1,\ldots,x_n)$ and nonempty finite cover $F_S$ of $\H$ on $S$, there is a randomized online bandit learner such that, for every sequence $A_1,\ldots,A_n\in\cA$ fixed independently of its randomness,
\begin{equation}\label{fb:eq:trans-regret}
 \E\left[\sum_{i=1}^n\ind\{\hat y_i\notin A_i\}\right]
 \le\inf_{h\in\H}\sum_{i=1}^n\ind\{h(x_i)\ne A_i\}
 +C\sqrt{kn\log|F_S|},
\end{equation}
where $C$ is universal. Indeed, EXP4 run with the members of $F_S$ and the uniform expert as experts and with rewards $\ind\{y\in A_i\}$ has regret at most $C\sqrt{kn\log|F_S|}$ against the best member of $F_S$ \cite[Theorem~7.1]{auer2002nonstochastic} (when $|F_S|=1$, the learner simply follows its only member). Since $F_S$ is a cover, every $h\in\H$ has a member $f_h$ with $\ind\{f_h(i)\notin A_i\}\le\ind\{h(x_i)\ne A_i\}$ for every $i\in[n]$, as in the proof of Theorem~\ref{thm:agnostic-full}, so the regret against $f_h$ bounds the regret against $h$.
\end{remark}

\subsection{Inductive Bounds}\label{fb:sec:ind}

The bounds above assume that the whole pool, including the test input, is known in advance: the experts depend on it, and in Proposition~\ref{prop:trans-bandit} so do the queries. In the inductive problem the test input arrives only after training, so the queries can no longer be chosen by the experts. Our learners therefore query an independent uniform label at each training input and record the answers. Once the test input arrives, they build the cover of the whole pool, replay the recorded answers in a uniformly random order, and predict when the test input is reached. As in the proof of Theorem~\ref{thm:full-expected}, the random order reduces the test loss to the average loss of a complete run.

A uniform query $u$ at an input with correct-label set $A$ \emph{exposes} an expert $f$ if $u$ is the label of $f$ at this input and $u\notin A$. This happens with probability $1/k$ if $f$ is wrong at this input, and never if $f$ is right. In the realizable case we eliminate the exposed experts. Each input then removes, in expectation, a $1/k$ fraction of the surviving experts that are wrong there, while an expert that is right at every input is never removed, so the expected number of mistakes along a run is at most $k\log|F|$. In the agnostic case, $k$ times the indicator of exposure is an unbiased estimate of the loss of $f$, and we run exponential weights with these estimates.

\begin{theorem}[Inductive bandit learners]\label{thm:inductive-bandit}
Suppose that for each $n$ and each pool $S$ of size $n$ there is a finite cover $F_S\subseteq\Y^n$ of $\H$ on $S$ that depends only on $S$ and satisfies $|F_S|\le L_n$. Let $n=m+1$. There are randomized bandit learners with $m$ examples, whose training queries are independent, uniform on $\Y$, and independent of the training inputs and of the test input, such that the returned predictor $\widehat g$ satisfies
\begin{align}
 \E\bigl[L_{P,h^\star}(\widehat g)\bigr]
 &\le\min\left\{1,\frac{k\log L_n}{n}\right\}&&\text{(realizable)},\label{fb:eq:ind-real}\\
 \E\bigl[L_\D(\widehat g)\bigr]
 &\le\Lset_\D(\H)+\sqrt{\frac{2k\log L_n}{n}}&&\text{(agnostic)},\label{fb:eq:ind-agnostic}
\end{align}
for every distribution $P$ and $h^\star\in\H$ with $P(\supp(h^\star))=1$, and for every distribution $\D$ on $\X\times\cA$, respectively. The expectations are over the $m$ examples, drawn from $\D_{P,h^\star}$ and from $\D$, and over the learner's seed.
\end{theorem}

\begin{proof}
Throughout, $A_i$ denotes the correct-label set of $x_i$, which the learner never sees.

\paragraph{The learners.}
On each training input $x_i$, $i\in[m]$, the learner draws $u_i\sim\Unif(\Y)$ and observes $b_i=\ind\{u_i\in A_i\}$. Given a test input $x$, it puts $x_n=x$, draws a uniformly random permutation $\pi$ of $[n]$, and lists the inputs in this order, $z_t=x_{\pi(t)}$ for $t\in[n]$. Let $\tau=\pi^{-1}(n)$ be the rank of the test input, so that $z_\tau=x$, let $F=F_{(z_1,\ldots,z_n)}$, and for $t\ne\tau$ write $(\tilde u_t,\tilde b_t)=(u_{\pi(t)},b_{\pi(t)})$. We say that $f\in F$ is exposed at rank $t$ if $\tilde u_t=f(t)$ and $\tilde b_t=0$. Both learners process the ranks $t=1,\ldots,\tau-1$ in this order and then predict at rank $\tau$.

In the realizable case, let $V_1=F$, and obtain $V_{t+1}$ by removing from $V_t$ every expert exposed at rank $t$. At rank $\tau$, draw $f$ uniformly from $V_\tau$ and output $f(\tau)$.

In the agnostic case, let $w_1(f)=1$ for $f\in F$, and at rank $t$ update
\[
 w_{t+1}(f)=w_t(f)\exp\bigl(-\eta\hat\lambda_t(f)\bigr),\qquad
 \hat\lambda_t(f)=k\,\ind\{\tilde u_t=f(t),\ \tilde b_t=0\},
\]
so that $\hat\lambda_t(f)$ is $k$ times the indicator that $f$ is exposed at rank $t$, with $\eta=\sqrt{2\log L_n/(kn)}$ (if $L_n=1$, the family has a single member and the learner outputs its label). At rank $\tau$, draw $f$ with probability proportional to $w_\tau(f)$ and output $f(\tau)$.

\paragraph{Reduction to a complete run.}
We claim that the expected test loss equals the average, over $t\in[n]$, of the expected loss at rank $t$ of the rule run through all $n$ ranks, where the loss at rank $t$ is computed before the update at rank $t$. To see this, draw the $n$ examples, training and test, together with independent uniform queries $u_1,\ldots,u_n$; the query $u_n$ is never used by the learner, and we define $(\tilde u_\tau,\tilde b_\tau)$ as for the other ranks. The triples $(x_i,A_i,u_i)$ are i.i.d., so the permuted triples are again i.i.d.\ and independent of $\pi$, and in particular of the rank $\tau$, which is uniform on $[n]$. Up to rank $\tau-1$ the learner coincides with the complete run, which does not depend on $\tau$, and averaging over $\tau$ proves the claim. Below we write $A'_t$ for the correct-label set of $z_t$ and argue for a fixed sequence $z_1,\ldots,z_n$, $A'_1,\ldots,A'_n$, averaging over it only at the end. Then $F$ is fixed, and the queries $\tilde u_1,\ldots,\tilde u_n$ are independent and uniform on $\Y$.

\paragraph{The realizable bound.}
Here $A'_t=h^\star(z_t)$ for every $t$, since $P(\supp(h^\star))=1$. Since $F$ is a cover, some expert $f$ satisfies $f(t)\in A'_t$ at every rank, and it is never exposed, because $\tilde u_t=f(t)$ forces $\tilde b_t=1$. Thus every $V_t$ is nonempty. Let
\[
 \rho_t=\frac1{|V_t|}\sum_{f\in V_t}\ind\{f(t)\notin A'_t\}
\]
be the fraction of $V_t$ that is wrong at rank $t$; given $V_t$, the rule makes a mistake at rank $t$ with probability exactly $\rho_t$. An expert in $V_t$ is exposed at rank $t$ exactly when it is wrong there and its label equals $\tilde u_t$. Since $\tilde u_t$ is uniform and independent of $V_t$, the expected fraction of $V_t$ eliminated at rank $t$ is $\rho_t/k$. Applying $-\log(1-s)\ge s$ to $s=1-|V_{t+1}|/|V_t|<1$ and taking conditional expectations gives
\[
 \E\bigl[\log|V_t|-\log|V_{t+1}|\bigm| V_t\bigr]\ge\frac{\rho_t}k.
\]
Summing over $t$, the left sides telescope, and since $|V_1|\le L_n$ and $|V_{n+1}|\ge1$,
\begin{equation}\label{fb:eq:telescope}
 \E\left[\sum_{t=1}^n\rho_t\right]\le k\log L_n.
\end{equation}
By the reduction, the expected test loss is $\E[\frac1n\sum_t\rho_t]$, and \eqref{fb:eq:telescope} proves \eqref{fb:eq:ind-real}.

\paragraph{The agnostic bound.}
Write $\lambda_t(f)=\ind\{f(t)\notin A'_t\}$ for the loss of $f$ at rank $t$, and let $p_t(f)=w_t(f)/W_t$ with $W_t=\sum_{f\in F}w_t(f)$. The weights $p_t$ are determined by the past, that is, by $\tilde u_1,\ldots,\tilde u_{t-1}$, and given the past the rule makes a mistake at rank $t$ with probability $\sum_{f}p_t(f)\lambda_t(f)$. Since $\tilde b_t=0$ exactly when $\tilde u_t\notin A'_t$, and $\tilde u_t$ is uniform and independent of the past,
\begin{equation}\label{fb:eq:importance}
 \E\bigl[\hat\lambda_t(f)\bigm|\text{past}\bigr]=\lambda_t(f),\qquad
 \E\bigl[\hat\lambda_t(f)^2\bigm|\text{past}\bigr]=k\lambda_t(f)\le k.
\end{equation}
Using $e^{-x}\le1-x+x^2/2$ for $x\ge0$ and then $\log(1+y)\le y$, we get
\[
 \log\frac{W_{t+1}}{W_t}=\log\sum_fp_t(f)e^{-\eta\hat\lambda_t(f)}
 \le-\eta\sum_fp_t(f)\hat\lambda_t(f)
 +\frac{\eta^2}{2}\sum_fp_t(f)\hat\lambda_t(f)^2.
\]
On the other hand, $\log(W_{n+1}/W_1)\ge-\eta\sum_t\hat\lambda_t(f)-\log|F|$ for each fixed $f\in F$. We sum the upper bound over $t$ and compare it with this lower bound. Because $p_t$ is determined by the past, \eqref{fb:eq:importance} gives $\E[\sum_fp_t(f)\hat\lambda_t(f)]=\E[\sum_fp_t(f)\lambda_t(f)]$ and $\E[\sum_fp_t(f)\hat\lambda_t(f)^2]\le k$, so taking expectations and using the choice of $\eta$ we obtain
\begin{equation}\label{fb:eq:ew}
 \E\left[\sum_{t=1}^n\sum_{f'\in F}p_t(f')\lambda_t(f')\right]
 \le\sum_{t=1}^n\lambda_t(f)+\frac{\log L_n}{\eta}+\frac{\eta kn}{2}
 =\sum_{t=1}^n\lambda_t(f)+\sqrt{2kn\log L_n}.
\end{equation}
Now fix $h\in\H$, and let $f_h\in F$ satisfy $f_h(t)\in h(z_t)$ at every rank $t$ with $z_t\in\supp(h)$. As in the proof of Theorem~\ref{thm:agnostic-full}, $\lambda_t(f_h)\le\ind\{h(z_t)\ne A'_t\}$ for every $t$. By the reduction, dividing \eqref{fb:eq:ew} with $f=f_h$ by $n$ and averaging over the i.i.d.\ pool gives
\[
 \E[L_\D(\widehat g)]\le\Pr_{(x,A)\sim\D}[h(x)\ne A]+\sqrt{\frac{2k\log L_n}n}.
\]
Taking the infimum over $h\in\H$ proves \eqref{fb:eq:ind-agnostic}.
\end{proof}

With the covers of Lemma~\ref{lem:finite-cover}, this gives the rates of Theorem~\ref{thm:finite-bandit}.

\begin{corollary}[Bandit rates]\label{cor:bandit-nat}
If $d=\dN(\H)<\infty$, then the bounds \eqref{fb:eq:rates} of Theorem~\ref{thm:finite-bandit} hold. For fixed $k$, $\ell$ and $d$, where $\ell$ bounds the sizes of the correct-label sets, both excess terms tend to zero as $n\to\infty$.
\end{corollary}

\begin{proof}
Apply Theorem~\ref{thm:inductive-bandit} with the covers of Lemma~\ref{lem:finite-cover} and $L_n=(1+nqk)^{M_d(n)}$, which is allowed by \eqref{eq:cover-bound}. By \eqref{ff:eq:M-log}, $M_d(n)=O(\log(n+1))$ for fixed $k$, $\ell$ and $d$, so the two excess terms are $O(\log^2(n+1)/n)$ and $O(\log(n+1)/\sqrt n)$.
\end{proof}

\subsection{Confidence Amplification}\label{fb:sec:amp}

We now turn the bounds on the expected risk into PAC bounds by repetition and validation, as in Proposition~\ref{prop:agnostic-full-pac} for full sets: we run the learner on several independent blocks of examples and select one of the resulting candidates on a fresh validation block. Bandit feedback suffices for the selection. For candidates produced by any algorithm, one uniform query per validation example gives an unbiased estimate of the risk of every candidate at once, at the price of a factor $k$ in the variance.

\begin{lemma}[Validation with uniform queries]\label{lem:bandit-validation}
Let $s,v\ge1$ be integers and $\eps,\delta\in(0,1)$. Let $g_1,\ldots,g_s$ be predictors, possibly random, that are independent of the validation examples $(x_i,A_i)\sim\D$, $i\in[v]$. At each validation example query an independent $u_i\sim\Unif(\Y)$ and observe $b_i=\ind\{u_i\in A_i\}$. Define
\begin{equation}\label{fb:eq:validate}
 \widehat L_j=\frac{k}{v}\sum_{i=1}^v\ind\{u_i=g_j(x_i),\ b_i=0\}\qquad(j\in[s]).
\end{equation}
Then $\E[\widehat L_j\mid g_1,\ldots,g_s]=L_\D(g_j)$, and there is an absolute constant $c$ with the following two properties. With probability at least $1-\delta$,
\begin{equation}\label{fb:eq:validation-bound}
 \max_{j\le s}\bigl|\widehat L_j-L_\D(g_j)\bigr|
 \le c\left(\sqrt{\frac{k\log(2s/\delta)}{v}}+\frac{k\log(2s/\delta)}{v}\right).
\end{equation}
Moreover, if $v\ge c(k/\eps)\log(2s/\delta)$, then with probability at least $1-\delta$ every $j$ with $L_\D(g_j)\le\eps/4$ has $\widehat L_j<\eps/2$, and every $j$ with $L_\D(g_j)\ge\eps$ has $\widehat L_j>\eps/2$.
\end{lemma}

\begin{proof}
Condition on $g_1,\ldots,g_s$; since the claims hold for every value of the predictors, they also hold unconditionally. Fix $j$ and put $Z_{i,j}=\ind\{u_i=g_j(x_i),\ b_i=0\}$. This is the indicator of the event that $g_j(x_i)\notin A_i$ and $u_i=g_j(x_i)$, so $Z_{1,j},\ldots,Z_{v,j}$ are independent Bernoulli variables with mean $L_\D(g_j)/k$. Hence $kZ_{i,j}$ has mean $L_\D(g_j)$, lies in $[0,k]$, and has variance at most $kL_\D(g_j)\le k$, and Bernstein's inequality with a union bound over the $s$ candidates gives \eqref{fb:eq:validation-bound}. For the threshold test, note that the threshold $\eps/(2k)$ for the average $\frac1v\sum_iZ_{i,j}$ is at least twice its mean when $L_\D(g_j)\le\eps/4$, and at most half its mean when $L_\D(g_j)\ge\eps$. In both cases the multiplicative Chernoff bound shows that the average falls on the wrong side of the threshold with probability at most $\exp(-v\eps/(ck))$, after enlarging $c$ if necessary, and a union bound over the $s$ candidates finishes the proof.
\end{proof}

\begin{proposition}[Confidence amplification]\label{prop:bandit-amplify}
Let $m_{\mathrm B}$ be any function such that, for every $r>0$, some bandit learner with $m_{\mathrm B}(r)$ examples has expected risk at most $r$ in the realizable case, or at most $\Lset_\D(\H)+r$ in the agnostic case. Then, respectively,
\begin{align}
 \mband_\H(\eps,\delta)&=O\left(m_{\mathrm B}(\eps/16)\log\frac2\delta
 +\frac{k}{\eps}\log\frac{2+\log(2/\delta)}\delta\right),\label{fb:eq:amp-real}\\
 \mbandag_\H(\eps,\delta)&=O\left(\frac{m_{\mathrm B}(\eps/8)}\eps\log\frac2\delta
 +\frac{k}{\eps^2}\log\frac{2+\eps^{-1}\log(2/\delta)}\delta\right),\label{fb:eq:amp-ag}
\end{align}
where the constants in $O(\cdot)$ are universal.
\end{proposition}

\begin{proof}
In both cases the learner runs the given learner on $s$ disjoint blocks of $m_{\mathrm B}(\cdot)$ examples and selects one of the resulting candidates $g_1,\ldots,g_s$ with Lemma~\ref{lem:bandit-validation}, applied with failure probability $\delta/2$ on one further block of $v$ examples. The total number of examples is $s\,m_{\mathrm B}(\cdot)+v$.

\paragraph{Realizable case.}
By Markov's inequality, a run with expected risk at most $\eps/16$ has risk at most $\eps/4$ with probability at least $3/4$. With $s=\lceil\log_4(2/\delta)\rceil$ independent runs, some candidate has risk at most $\eps/4$ with probability at least $1-\delta/2$. Take $v=O((k/\eps)\log(4s/\delta))$ as in the threshold test of Lemma~\ref{lem:bandit-validation}, and output a candidate with $\widehat L_j<\eps/2$. On both events such a candidate exists, and every such candidate has risk less than $\eps$.

\paragraph{Agnostic case.}
Put $a=\Lset_\D(\H)$. If $a+\eps\ge1$ there is nothing to prove, so assume $a+\eps<1$. By Markov's inequality, a run with expected risk at most $a+\eps/8$ has risk at most $a+\eps/2$ with probability at least $1-(a+\eps/8)/(a+\eps/2)=(3\eps/8)/(a+\eps/2)\ge3\eps/8$, as in the proof of Proposition~\ref{prop:agnostic-full-pac}. With $s=\lceil\frac{8}{3\eps}\log\frac2\delta\rceil$ independent runs, some candidate $g_{j^\star}$ has risk at most $a+\eps/2$ with probability at least $1-\delta/2$. Take $v=O((k/\eps^2)\log(4s/\delta))$, so that \eqref{fb:eq:validation-bound} holds with accuracy $\eps/4$, and output a candidate $g_{\hat j}$ minimizing $\widehat L_j$. On both events,
\[
 L_\D(g_{\hat j})\le\widehat L_{\hat j}+\frac\eps4\le\widehat L_{j^\star}+\frac\eps4\le L_\D(g_{j^\star})+\frac\eps2\le a+\eps.
\]

In both cases $\log(4s/\delta)$ is at most a constant multiple of the logarithm in \eqref{fb:eq:amp-real} and \eqref{fb:eq:amp-ag}, respectively, which proves the two bounds.
\end{proof}

The training term in \eqref{fb:eq:amp-real} depends multiplicatively on $\log(2/\delta)$, because the given learner is run on $s=\lceil\log_4(2/\delta)\rceil$ blocks. In contrast, the realizable full-set bound of Theorem~\ref{thm:hp} depends additively on $\log(1/\delta)$, and so does elimination with uniform queries on a known pool. To see this, fix a pool $(x_1,\ldots,x_N)$ with correct-label sets $A_1,\ldots,A_N$ and a family $F\subseteq\Y^N$ of experts. Let $I$ be a uniformly random subset of $[N]$ of size $m$, query an independent $u_i\sim\Unif(\Y)$ at each $i\in I$, and observe $b_i=\ind\{u_i\in A_i\}$. Call $f\in F$ a \emph{survivor} if no $i\in I$ has $f(i)=u_i$ and $b_i=0$, that is, if no query exposes $f$.

\begin{proposition}[Survivors on a fixed pool]\label{prop:bandit-hp-pool}
In the setting above, let $1\le m\le N$ and $\delta\in(0,1)$, suppose that $A_i=h^\star(x_i)$ for all $i$, where $h^\star\in\H$ satisfies $x_1,\ldots,x_N\in\supp(h^\star)$, and suppose that $F$ contains an expert $f^\star$ with $f^\star(i)\in A_i$ for all $i$. Then $f^\star$ survives, and with probability at least $1-\delta$ every survivor $f$ satisfies
\begin{equation}\label{fb:eq:hp-pool}
 \frac1N\sum_{i=1}^N\ind\{f(i)\notin A_i\}
 \le\frac{k}{m}\left(\log|F|+\log\frac1\delta\right).
\end{equation}
\end{proposition}

\begin{proof}
If $f(i)=u_i$ and $f(i)\in A_i$, then $b_i=1$; hence $f^\star$ survives. Fix $f\in F$, let $p$ be the left side of \eqref{fb:eq:hp-pool}, and put $a_i=1-1/k$ if $f(i)\notin A_i$ and $a_i=1$ otherwise, so that the average of $a_1,\ldots,a_N$ is $1-p/k$. Given $I$, the expert $f$ survives exactly when $u_i\ne f(i)$ at every $i\in I$ with $f(i)\notin A_i$, which has probability $\prod_{i\in I}a_i$. By Maclaurin's inequality, the average of the products of $m$ distinct entries of a list of nonnegative numbers is at most the $m$th power of their average, and therefore
\[
 \Pr[f\text{ survives}]=\binom Nm^{-1}\sum_{|I|=m}\prod_{i\in I}a_i
 \le\left(\frac1N\sum_{i=1}^Na_i\right)^m=(1-p/k)^m\le e^{-mp/k}.
\]
Thus every $f$ with $p>\frac km(\log|F|+\log\frac1\delta)$ survives with probability less than $\delta/|F|$, and a union bound over $F$ proves the claim.
\end{proof}

We leave open whether an inductive bandit learner can achieve an additive dependence on $\log(1/\delta)$ in the realizable case. Even with singleton correct-label sets, the confidence parameter has nontrivial effects in bandit PAC learning \cite{zhang2026bandit}.

\subsection{Proof of Theorem~\ref{thm:finite-bandit}}

\begin{proof}[Proof of Theorem~\ref{thm:finite-bandit}]
By Theorem~\ref{thm:finite-full}, (i)--(iv) are equivalent. We show (ii)$\Rightarrow$(vi)$\Rightarrow$(v)$\Rightarrow$(i).

Assume (ii), and let $d=\dN(\H)$. By Corollary~\ref{cor:bandit-nat}, the bounds \eqref{fb:eq:rates} hold, and their excess terms tend to zero, so $m_{\mathrm B}(r)$ is finite for every $r>0$. Proposition~\ref{prop:bandit-amplify} then gives \eqref{fb:eq:amp-real} and \eqref{fb:eq:amp-ag}, and in particular (vi).

As noted in Section~\ref{sec:prelim-models}, agnostic PAC learnability implies realizable PAC learnability under every feedback model, so (vi) implies (v). Finally, a learner with bandit feedback sees less than a full-set learner, and Proposition~\ref{prop:ds-lower} covers bandit learners explicitly; hence the proof of (iii)$\Rightarrow$(i) in Theorem~\ref{thm:finite-full} shows that (v) implies (i).
\end{proof}
\section{A Reduction between Full-Set and One-Correct-Label Feedback}\label{sec:reduction}

In this section we relate one-correct-label learning of a class $\H$ to full-set learning of its partial intersection class $\cI_{\ell+1}(\H)$, defined in~\eqref{eq:intersections}. One direction follows from a union bound: a correct label of $g=\bigcap_{a=1}^r h_a$ at $x$ is a valid demonstration for every $h_a$, so a one-correct-label learner for $\H$ also learns the intersection class from full sets (Lemma~\ref{lem:common-demonstrations}).

The converse occupies the rest of the section. Theorem~\ref{thm:family-conversion} turns finite families of single-label predictions with an intersection property into a one-correct-label learner with small expected loss, and Lemma~\ref{lem:amplification} turns this into a high-probability guarantee. The families themselves are constructed for finite label spaces in Section~\ref{sec:finite-one} and for countable label spaces in Section~\ref{sec:countable-one}.

\begin{lemma}[Common demonstrations]\label{lem:common-demonstrations}
For every integer $R\ge1$ and $\eps,\delta\in(0,1)$,
\[
  \mfull_{\cI_R(\H)}(\eps,\delta)\le\mone_{\H}(\eps/R,\delta/R).
\]
\end{lemma}

\begin{proof}
Let $m=\mone_{\H}(\eps/R,\delta/R)$; if $m=\infty$ there is nothing to prove. Fix a one-correct-label learner for $\H$ that uses $m$ examples and has accuracy $\eps/R$ and confidence $1-\delta/R$. Our full-set learner receives $(x_1,A_1),\ldots,(x_m,A_m)$, replaces each $A_i$ by its first element $y_i$ in the fixed enumeration of $\Y$, and runs the one-correct-label learner on $(x_1,y_1),\ldots,(x_m,y_m)$.

Let $g\in\cI_R(\H)$ and let $P$ be a distribution with $P(\supp(g))=1$. By~\eqref{eq:intersections}, $g(x)=\bigcap_{a=1}^r h_a(x)$ for every $x\in\supp(g)$, where $r\le R$, $h_1,\ldots,h_r\in\H$ and $\supp(g)\subseteq\bigcap_{a=1}^r\supp(h_a)$. The pairs $(x_i,y_i)$ are i.i.d.\ draws from the law $\D$ of $(x,y(x))$, where $x\sim P$ and $y(x)$ is the first element of $g(x)$. This law has marginal $P$, and for every $a\in[r]$ we have $P(\supp(h_a))=1$ and $y(x)\in g(x)\subseteq h_a(x)$ for every $x\in\supp(g)$. Thus $\D$ is a valid one-correct-label law for each of the targets $h_1,\ldots,h_r$, so by a union bound over $a\in[r]$, with probability at least $1-r\delta/R\ge1-\delta$ the output $\widehat f$ satisfies $L_{P,h_a}(\widehat f)\le\eps/R$ for every $a$. Since $\widehat f(x)\notin g(x)$ implies $\widehat f(x)\notin h_a(x)$ for some $a$, we get
\[
  L_{P,g}(\widehat f)\le\sum_{a=1}^r L_{P,h_a}(\widehat f)\le\frac{r\eps}{R}\le\eps.
\]
\end{proof}

We now turn to the converse. A demonstrated label $y\in h^\star(x)$ reveals nothing about the other labels of $h^\star(x)$, so consistency with the demonstrations does not force good predictions. Our learner works with a pool $S$ of $N$ inputs, the training inputs together with the test input, and uses the demonstrations at a random half of the positions. With each candidate $h$ it associates a small family $F_{h,S}$ of single-label predictions on the pool. Among hypotheses consistent with the demonstrations, it retains $h$ only if every member of $F_{h,S}$ that is correct for $h$ on the labeled half has small loss against $h$ on the whole pool. A member with large loss is rarely correct on a random half, so a union bound over $F_{h^\star,S}$ shows that the target is retained with high probability.

The families must have an intersection property: any $\ell+1$ candidates share a member that predicts a common label wherever their correct-label sets intersect. For $\ell+1$ retained candidates, the shared member is correct for all of them on the labeled half and hence has small loss against each on the whole pool. Their correct-label sets therefore intersect at all but a small fraction of the unlabeled positions. A minimax argument on these positions turns this into randomized predictions with small average loss against every retained candidate, and in particular against the target.

For a multiset $S=(x_1,\ldots,x_N)$ of inputs we write $\supp(S)=\{x_1,\ldots,x_N\}$ for the set of its distinct elements, and we say that a hypothesis $h$ is \emph{defined on $S$} if $\supp(S)\subseteq\supp(h)$.

\begin{definition}[Prediction families with the intersection property]\label{def:prediction-families}
Fix $N\ge2$. Suppose that for every multiset $S$ of $N$ inputs and every $h\in\H$ defined on $S$ we are given a family
\[
  F_{h,S}\subseteq\{f:\supp(S)\to\Y\},\qquad |F_{h,S}|\le B_N,
\]
where $B_N\ge1$ does not depend on $S$ and $h$. These families have the \emph{intersection property} if for every $S$ and all $h_1,\ldots,h_{\ell+1}\in\H$ defined on $S$ there is a function $f\in\bigcap_{a=1}^{\ell+1}F_{h_a,S}$ such that
\begin{equation}\label{eq:red-intersection-property}
  \bigcap_{a=1}^{\ell+1}h_a(x)\ne\emptyset
  \quad\Longrightarrow\quad
  f(x)\in\bigcap_{a=1}^{\ell+1}h_a(x)
  \qquad(x\in\supp(S)).
\end{equation}
\end{definition}

The main step of the converse is the following theorem.

\begin{theorem}[Conversion from prediction families]\label{thm:family-conversion}
Let $\H$ be a nonempty class and $N\ge2$, and suppose that families $F_{h,S}$ with size bound $B_N$ have the intersection property of Definition~\ref{def:prediction-families}. Then there is a one-correct-label learner that uses $N-1$ examples and outputs a randomized prediction rule $\widehat p$ with the following property. For every $h^\star\in\H$, every distribution $P$ on $\X$ with $P(\supp(h^\star))=1$, and every law $\D$ on $\X\times\Y$ with marginal $P$ and $\Pr_{(x,y)\sim\D}[y\in h^\star(x)]=1$,
\begin{equation}\label{eq:red-conversion-bound}
  \E\bigl[L_{P,h^\star}(\widehat p)\bigr]
  \le\min\left\{1,\;2\left(\frac{(\ell+1)(\log B_N+\log N)}{\lfloor N/2\rfloor}\right)^{1/\ell}+\frac1N\right\},
\end{equation}
where the expectation is over the training examples drawn from $\D$ and the seed of the learner.
\end{theorem}

\begin{proof}
We first define the retained class and the learner, and then analyze an experiment on $N$ examples in which the test position is uniform among the unlabeled ones. Let
\[
  t=\lfloor N/2\rfloor,
  \qquad\alpha=\frac{\log B_N+\log N}{t},
  \qquad\beta=\left(\frac{N(\ell+1)\alpha}{N-t}\right)^{1/\ell}.
\]
Here $t$ is the number of labeled positions, $\alpha$ is chosen so that $B_Ne^{-t\alpha}=1/N$, and $\beta$ is chosen so that $\beta^\ell$ is the bound in~\eqref{eq:red-empty} below. For a multiset $S=(x_1,\ldots,x_N)$, a function $f:\supp(S)\to\Y$ and a hypothesis $h$ defined on $S$, let
\[
  L_S(f,h)=\frac1N\sum_{i=1}^N\ind\{f(x_i)\notin h(x_i)\}.
\]

\paragraph{The retained class.}
Fix a pool $S=(x_1,\ldots,x_N)$, a set $J\subseteq[N]$ of $t$ observed positions, and labels $(y_i)_{i\in J}$. Write $U=[N]\setminus J$ for the unobserved positions and $S_U=\{x_i:i\in U\}$ for their inputs. Let $\H_J$ be the set of $h\in\H$ that are defined on $S$, satisfy $y_i\in h(x_i)$ for every $i\in J$, and satisfy
\begin{equation}\label{eq:red-retained}
  \text{for every }f\in F_{h,S}:\qquad
  f(x_i)\in h(x_i)\text{ for all }i\in J
  \quad\Longrightarrow\quad
  L_S(f,h)\le\alpha.
\end{equation}
The class $\H_J$ depends only on the multiset $S$ and the multiset of observed examples $\{(x_i,y_i):i\in J\}$. We show below that if $\H_J\ne\emptyset$, there are distributions $(p_z)_{z\in S_U}$ on $\Y$ such that
\begin{equation}\label{eq:red-goal}
  \frac1{|U|}\sum_{i\in U}\bigl(1-p_{x_i}(h(x_i))\bigr)\le\beta
  \qquad\text{for every }h\in\H_J,
\end{equation}
where $p_z(A)=\sum_{y\in A}p_z(y)$. Fix a rule that selects such distributions as a function of $S$ and the multiset of observed examples (when $\H_J=\emptyset$, every $p_z$ is a point mass on the first label of $\Y$); since $\X$ and $\Y$ are countable, this rule is measurable.

\paragraph{The learner.}
Algorithm~2 places the test input at the last position of the pool and uses the demonstrations at a random set of $t$ training positions.

\begin{algbox}
\centerline{\textbf{Algorithm 2: One-correct-label learner of Theorem~\ref{thm:family-conversion}}}
\smallskip
\noindent\textbf{Input:} training examples $(x_1,y_1),\ldots,(x_{N-1},y_{N-1})$; a test input $x$.
\begin{enumerate}
\item Set $x_N=x$ and $S=(x_1,\ldots,x_N)$.
\item Using the seed, draw $J$ uniformly among the $t$-element subsets of $[N-1]$. Let $U=[N]\setminus J$.
\item Form the retained class $\H_J$ from $S$ and $(x_i,y_i)_{i\in J}$, as in~\eqref{eq:red-retained}.
\item Select distributions $(p_z)_{z\in S_U}$ satisfying~\eqref{eq:red-goal} by the fixed rule.
\item Output $\widehat p(\cdot\mid x)=p_{x}$.
\end{enumerate}
\end{algbox}

\noindent
Since $J\subseteq[N-1]$, the test position $N$ lies in $U$, so $p_x$ is defined in step~5. Moreover, the seed determines $J$ before the test input is seen, so for a fixed seed and fixed training examples the map $x\mapsto\widehat p(\cdot\mid x)$ is a randomized prediction.

\paragraph{Reduction to a transductive experiment.}
For the analysis, give the test input $x_N\sim P$ a label $y_N$ drawn from the conditional law of $\D$ given $x_N$. The learner does not use $y_N$, and the pairs $(x_1,y_1),\ldots,(x_N,y_N)$ are now i.i.d.\ from $\D$; almost surely, $h^\star$ is defined on $S$ and $y_i\in h^\star(x_i)$ for every $i\in[N]$. The learner observes the positions in a uniform $t$-subset $J$ of $[N-1]$ and is tested at position $N$. For a set $J$ of $t$ observed positions and a test position $I\notin J$, the loss $1-p_{x_I}(h^\star(x_I))$ depends on the examples only through the multiset $S$, the multiset $\{(x_i,y_i):i\in J\}$ and $x_I$. Since the $N$ pairs are i.i.d., relabeling the positions does not change their joint law, so the expected loss is the same for every such pair $(J,I)$. We may therefore draw $J$ uniformly among the $t$-subsets of $[N]$, independently of the examples, and then $I$ uniformly from $U$, without changing the expected loss. From now on, $J$ is drawn in this way. Conditional on the examples and on $J$, the predictions $(p_z)_{z\in S_U}$ are fixed, and
\begin{equation}\label{eq:red-transductive}
  \E_I\bigl[1-p_{x_I}(h^\star(x_I))\bigr]
  =\frac1{|U|}\sum_{i\in U}\bigl(1-p_{x_i}(h^\star(x_i))\bigr).
\end{equation}
It remains to bound this average.

\paragraph{The true target is retained.}
We first show that
\begin{equation}\label{eq:red-survival}
  \Pr_J\bigl[h^\star\notin\H_J\mid(x_i,y_i)_{i=1}^N\bigr]\le\frac1N.
\end{equation}
Since all demonstrated labels are correct for $h^\star$, the target is discarded only if some $f\in F_{h^\star,S}$ with $L_S(f,h^\star)>\alpha$ is correct for $h^\star$ at every position of $J$. Fix such an $f$, and let $b=N\,L_S(f,h^\star)$ be the number of positions $i\in[N]$ with $f(x_i)\notin h^\star(x_i)$. Since $J$ is a uniform $t$-subset of $[N]$,
\[
  \Pr_J\bigl[f(x_i)\in h^\star(x_i)\text{ for all }i\in J\bigr]
  =\frac{\binom{N-b}{t}}{\binom Nt}
  \le\Bigl(1-\frac bN\Bigr)^t
  \le\exp\bigl(-t\,L_S(f,h^\star)\bigr)
  <e^{-t\alpha}.
\]
The family $F_{h^\star,S}$ is determined by $S$ alone, independently of $J$ and of the labels, and has at most $B_N$ members. A union bound over it bounds the probability of discarding the target by $B_Ne^{-t\alpha}=1/N$, which proves~\eqref{eq:red-survival}.

\paragraph{Empty intersections are rare on $U$.}
Fix the examples and $J$, and let $h_1,\ldots,h_{\ell+1}\in\H_J$, repetitions allowed. All of them accept every demonstrated label, so $y_i\in\bigcap_{a=1}^{\ell+1}h_a(x_i)$ for every $i\in J$. Hence the function $f\in\bigcap_{a=1}^{\ell+1}F_{h_a,S}$ given by the intersection property~\eqref{eq:red-intersection-property} is correct for every $h_a$ at every position of $J$, and~\eqref{eq:red-retained} gives $L_S(f,h_a)\le\alpha$ for every $a\in[\ell+1]$. At a position where $\bigcap_{a}h_a(x_i)=\emptyset$, the label $f(x_i)$ is incorrect for at least one $h_a$. Therefore
\begin{equation}\label{eq:red-empty}
\begin{aligned}
  \frac1{|U|}\sum_{i\in U}\ind\Bigl\{\bigcap_{a=1}^{\ell+1}h_a(x_i)=\emptyset\Bigr\}
  &\le\frac1{|U|}\sum_{i\in U}\sum_{a=1}^{\ell+1}\ind\{f(x_i)\notin h_a(x_i)\}\\
  &\le\frac N{|U|}\sum_{a=1}^{\ell+1}L_S(f,h_a)
  \le\frac{N(\ell+1)\alpha}{N-t}=\beta^\ell,
\end{aligned}
\end{equation}
where the second inequality extends the sum from $U$ to $[N]$.

\paragraph{The minimax game on $U$.}
We prove~\eqref{eq:red-goal} first for a finite nonempty subset $\H_0$ of $\H_J$. Only the finitely many labels in the sets $h(z)$, $h\in\H_0$, $z\in S_U$, can be correct, so we obtain a finite zero-sum game: the predictor chooses a function $g$ from $S_U$ to these labels, the adversary chooses $h\in\H_0$, and the payoff is the average loss $\frac1{|U|}\sum_{i\in U}\ind\{g(x_i)\notin h(x_i)\}$. Let $\mu$ be a mixed strategy of the adversary, that is, a distribution on $\H_0$. Against $\mu$, the best label at $z\in S_U$ is wrong with probability
\[
  \nu_z=1-\max_{y\in\Y}\Pr_{h\sim\mu}[y\in h(z)],
\]
so the function that predicts such a label at every $z$ has expected payoff $\frac1{|U|}\sum_{i\in U}\nu_{x_i}$.

To bound $\nu_z$, draw $h_1,\ldots,h_{\ell+1}$ independently from $\mu$, and condition on $h_1(z)=\{y^1,\ldots,y^r\}$; here $1\le r\le\ell$, since every correct-label set has at most $\ell$ elements. For each $j\in[r]$, the hypothesis $h_{j+1}$ excludes $y^j$ with probability $\Pr_{h\sim\mu}[y^j\notin h(z)]\ge\nu_z$. These $r$ events are independent, and if all of them occur, no label of $h_1(z)$ lies in $\bigcap_a h_a(z)$. Conditional on $h_1(z)$, the empty-intersection probability is therefore at least $\nu_z^r\ge\nu_z^\ell$. Averaging over $h_1$ gives
\begin{equation}\label{eq:red-power}
  \Pr_{h_1,\ldots,h_{\ell+1}\sim\mu}\Bigl[\bigcap_{a=1}^{\ell+1}h_a(z)=\emptyset\Bigr]\ge\nu_z^\ell.
\end{equation}
The bound~\eqref{eq:red-empty} holds for every choice of $h_1,\ldots,h_{\ell+1}\in\H_0$. Taking its expectation over $h_1,\ldots,h_{\ell+1}\sim\mu$ and using~\eqref{eq:red-power} gives $\frac1{|U|}\sum_{i\in U}\nu_{x_i}^\ell\le\beta^\ell$, and by Jensen's inequality,
\begin{equation}\label{eq:red-average}
  \frac1{|U|}\sum_{i\in U}\nu_{x_i}\le\beta.
\end{equation}
So every mixed strategy of the adversary has a response with expected payoff at most $\beta$. By the minimax theorem~\cite{vonneumann1928theorie}, some mixed strategy of the predictor has expected payoff at most $\beta$ against every $h\in\H_0$. Its marginals $p_z$, $z\in S_U$, therefore satisfy~\eqref{eq:red-goal} for every $h\in\H_0$, and taking $\H_0=\H_J$ proves~\eqref{eq:red-goal} when $\H_J$ is finite.

\paragraph{Infinite retained classes.}
If $\H_J$ is infinite, we use compactness. The set of subprobability distributions
\[
  \Delta_{\le1}(\Y)=\Bigl\{\theta\in[0,1]^{\Y}:\sum_{y\in F}\theta(y)\le1\text{ for every finite }F\subseteq\Y\Bigr\}
\]
is closed in the product topology, hence compact, and so is its finite product over $S_U$. In this product impose, for each $h\in\H_J$, the constraint
\begin{equation}\label{eq:red-compactness}
  \frac1{|U|}\sum_{i\in U}\sum_{y\in h(x_i)}p_{x_i}(y)\ge1-\beta.
\end{equation}
Each constraint involves finitely many coordinates and is therefore closed, and every finite collection of constraints is satisfiable by the finite case above. By compactness, some $(p_z)_{z\in S_U}$ satisfies all the constraints. Assigning the missing mass $1-\sum_yp_z(y)$ of each $p_z$ to a fixed label does not decrease any $p_z(h(z))$, and gives distributions satisfying~\eqref{eq:red-goal}.

\paragraph{Putting the pieces together.}
By~\eqref{eq:red-goal}, the average in~\eqref{eq:red-transductive} is at most $\beta$ when $h^\star\in\H_J$, and at most one otherwise. Taking expectations and using~\eqref{eq:red-survival}, we obtain
\begin{align*}
  \E\bigl[L_{P,h^\star}(\widehat p)\bigr]
  \le\beta+\frac1N
  &=\left(\frac{N(\ell+1)(\log B_N+\log N)}{(N-t)\,t}\right)^{1/\ell}+\frac1N\\
  &\le2\left(\frac{(\ell+1)(\log B_N+\log N)}{t}\right)^{1/\ell}+\frac1N,
\end{align*}
where we used $N/(N-t)\le2$ and $2^{1/\ell}\le2$. Since the loss is also at most one, this proves~\eqref{eq:red-conversion-bound}.
\end{proof}

Theorem~\ref{thm:family-conversion} bounds the expected loss of a randomized prediction rule. To obtain a high-probability guarantee for a single predictor, we run the learner on independent blocks of examples and predict a most probable label of the average output.

\begin{lemma}[Confidence amplification]\label{lem:amplification}
Let $\eps,\delta\in(0,1)$. Suppose that a one-correct-label learner for $\H$ uses $n$ examples and outputs a randomized prediction rule $\widehat p$ with $\E[L_{P,h^\star}(\widehat p)]\le\eps/[32(\ell+1)^2]$ for every $h^\star\in\H$, every distribution $P$ with $P(\supp(h^\star))=1$, and every law $\D$ on $\X\times\Y$ with marginal $P$ and $\Pr_{(x,y)\sim\D}[y\in h^\star(x)]=1$. Then
\[
  \mone_{\H}(\eps,\delta)\le n\,\bigl\lceil24(\ell+1)\log(2/\delta)\bigr\rceil=O_\ell\bigl(n\log(2/\delta)\bigr).
\]
\end{lemma}

\begin{proof}
Put $s=\ell+1$ and $b=\lceil24s\log(2/\delta)\rceil$. Run the learner on $b$ disjoint blocks of $n$ examples with independent seeds, and let $p_1,\ldots,p_b$ be the outputs. Call $p_j$ \emph{good} if $L_{P,h^\star}(p_j)\le\eps/(4s)$, and \emph{bad} otherwise. By Markov's inequality, each $p_j$ is bad with probability at most $1/(8s)$, independently of the others, so the number of bad $p_j$ is stochastically dominated by $\Bin(b,1/(8s))$, whose mean is $b/(8s)$. By a multiplicative Chernoff bound,
\[
  \Pr\bigl[\text{at least }b/(4s)\text{ of the }p_j\text{ are bad}\bigr]\le e^{-b/(24s)}\le\delta.
\]
Let $\bar p=\frac1b\sum_{j=1}^bp_j$, let $\widehat g(x)$ be a most probable label of $\bar p(\cdot\mid x)$, and assume that fewer than $b/(4s)$ of the $p_j$ are bad. If $\widehat g(x)\notin h^\star(x)$, then $\bar p(\Y\setminus h^\star(x)\mid x)\ge1/s$ by Lemma~\ref{lem:mode}. The bad $p_j$ contribute less than $1/(4s)$ to this mass, so the good ones contribute more than $3/(4s)$. Hence
\[
  \ind\{\widehat g(x)\notin h^\star(x)\}\le\frac{4s}{3b}\sum_{j:\,p_j\text{ good}}\bigl(1-p_j(h^\star(x)\mid x)\bigr).
\]
Taking expectations over $x\sim P$ and using $L_{P,h^\star}(p_j)\le\eps/(4s)$ for every good $p_j$ gives $L_{P,h^\star}(\widehat g)\le\frac{4s}{3b}\cdot b\cdot\frac{\eps}{4s}<\eps$. This holds with probability at least $1-\delta$, and the learner uses $nb$ examples.
\end{proof}


\section{One-Correct-Label Learning with Finite Labels}\label{sec:finite-one}\label{fo:sec:realizable}

Here we prove Theorem~\ref{thm:B}, in the following stronger form that adds two equivalent conditions and a bound on the expected loss; throughout, $k=|\Y|<\infty$. By the conversion of Section~\ref{sec:reduction}, it suffices to attach to every hypothesis a small family of single-label predictions with the intersection property. With finitely many labels, one family serves all hypotheses at once: a finite cover of $\cI_{\ell+1}(\H)$ on the pool, as constructed in Section~\ref{sec:finite-full}. Its size grows with $k$, which is why Section~\ref{sec:countable-one} gives each hypothesis a family of its own when $\Y$ is infinite.

\begin{theorem}[One-correct-label learning with finite labels]\label{thm:finite-one}
Let $|\Y|=k<\infty$, and let $\H\subseteq(\cA\cup\{\star\})^{\X}$ be nonempty. The following are equivalent:
\begin{enumerate}[(i)]
 \item $\H$ is realizably one-correct-label PAC learnable;
 \item $\cI_{\ell+1}(\H)$ is realizably full-set PAC learnable;
 \item $\dN(\cI_{\ell+1}(\H))<\infty$;
 \item $\dDS(\cI_{\ell+1}(\H))<\infty$;
 \item $\dpair(\cI_{\ell+1}(\H))<\infty$;
 \item $\dpack(\cI_{\ell+1}(\H))<\infty$.
\end{enumerate}
Moreover, let $d=\dN(\cI_{\ell+1}(\H))<\infty$. There is a constant $c$, depending only on $\ell$, such that the following hold.
\begin{enumerate}[(a)]
 \item For every $m\ge1$ there is a learner that uses $m$ one-correct-label examples and outputs a randomized prediction rule $\widehat p_m$ such that
 \begin{equation}\label{eq:finite-one-expected}
 \E\bigl[L_{P,h^\star}(\widehat p_m)\bigr]
 \le\min\left\{1,\ c\left[\frac{(d+1)\log^2(2(m+1)(k+1))}{m+1}\right]^{1/\ell}\right\}
 \end{equation}
 for every $h^\star\in\H$ and every law $\D$ compatible with $h^\star$ whose marginal $P$ satisfies $P(\supp(h^\star))=1$, where the expectation is over the examples and the seed of the learner.
 \item The sample complexity satisfies
 \begin{equation}\label{eq:finite-one-rate}
 \mone_\H(\eps,\delta)\le\left\lceil c\,(d+1)\,\eps^{-\ell}\log^2\!\left(\frac{(d+2)(k+1)}{\eps}\right)\log\frac2\delta\right\rceil.
 \end{equation}
\end{enumerate}
Both bounds also hold with $\dDS(\cI_{\ell+1}(\H))$ in place of $d$, and if $d=0$, then $\mone_\H(\eps,\delta)=0$.
\end{theorem}

The proof rests on one observation: the partial intersection of any $\ell+1$ hypotheses of $\H$ on a pool is itself a member of $\cI_{\ell+1}(\H)$. A cover of $\cI_{\ell+1}(\H)$ therefore contains, for any $\ell+1$ hypotheses, a vector that is correct for all of them wherever their correct-label sets intersect, which is exactly the intersection property. By Lemma~\ref{lem:finite-cover} the cover has $\log B_N=O_\ell((d+1)\log^2(Nk))$ on a pool of $N$ inputs, Theorem~\ref{thm:family-conversion} turns this into \eqref{eq:finite-one-expected}, and Lemma~\ref{lem:amplification} turns \eqref{eq:finite-one-expected} into \eqref{eq:finite-one-rate}.

\begin{proof}
Throughout the proof, $c$ denotes a constant that depends only on $\ell$; its value may change between occurrences.

\paragraph{The equivalences.}
By Lemma~\ref{lem:common-demonstrations} with $R=\ell+1$, (i) implies (ii). Theorem~\ref{thm:finite-full} applied to $\cI_{\ell+1}(\H)$ shows that (ii), (iii) and (iv) are equivalent, and since a finite label space is countable, Theorem~\ref{thm:countable-full} shows that (ii) and (vi) are equivalent. Moreover, (iv) implies (v) because $\dpair\le\dDS$ by Lemma~\ref{lem:orders}, and (v) implies (vi) by Theorem~\ref{thm:comparison}. It remains to show that (iii) implies (i), and this follows from the learner constructed below, which also gives the two bounds.

\paragraph{Prediction families from a finite cover.}
Assume $d<\infty$, and fix $N\ge2$ and a multiset $S$ of $N$ inputs. We list $S$ in a fixed total order on $\X$, so that the ordered pool depends only on $S$, and let $F_S$ be the finite cover of $\cI_{\ell+1}(\H)$ on this pool given by Lemma~\ref{lem:finite-cover}. Reading each vector at the first occurrence of every input, we regard $F_S$ as a family of functions on $\supp(S)$. By \eqref{eq:cover-bound},
\begin{equation}\label{fo:eq:family}
 |F_S|\le B_N:=(1+Nqk)^{M_d(N)},
\end{equation}
where $q=|\cA|\le(k+1)^\ell$ is the number of possible correct-label sets and $M_d$ is the mistake bound \eqref{eq:finite-M} of Algorithm~1 for the class $\cI_{\ell+1}(\H)$.

Set $F_{h,S}=F_S$ for every $h\in\H$ defined on $S$. Given $h_1,\ldots,h_{\ell+1}\in\H$ defined on $S$, let $g$ be their intersection restricted to the inputs $x\in\supp(S)$ at which $\bigcap_ah_a(x)\ne\emptyset$. Then $g\in\cI_{\ell+1}(\H)$ by \eqref{eq:intersections}, so the covering property gives $f\in F_S=\bigcap_aF_{h_a,S}$ with $f(x)\in\bigcap_ah_a(x)$ at every such input. This is \eqref{eq:red-intersection-property}, so the families $F_{h,S}$ have the intersection property with size bound $B_N$.

\paragraph{Expected loss.}
By \eqref{eq:finite-Gamma}, $\Gamma\le(q+1)^{\ell+1}$, so \eqref{ff:eq:M-log} and $q\le(k+1)^\ell$ give $M_d(N)\le c\,(d+1)\log(2N(k+1))$. Hence, by \eqref{fo:eq:family},
\[
 \log B_N\le M_d(N)\log(1+Nqk)\le c\,(d+1)\log^2(2N(k+1)).
\]
Apply Theorem~\ref{thm:family-conversion} with $N=m+1$, which gives a learner that uses $m$ examples. Since $\lfloor N/2\rfloor\ge N/3$ and $\log N\le\log^2(2N(k+1))$, the term with the $\ell$th root in \eqref{eq:red-conversion-bound} is at most $c\,[(d+1)\log^2(2N(k+1))/N]^{1/\ell}$. This quantity is at least $N^{-1/\ell}\ge1/N$, so increasing $c$ absorbs the additive term $1/N$, and \eqref{eq:finite-one-expected} follows.

\paragraph{Confidence.}
Lemma~\ref{lem:amplification} needs a learner with expected loss at most $\eps/[32(\ell+1)^2]$, so it remains to find a sample size $n$ at which the right side of \eqref{eq:finite-one-expected} is this small. Let $a=\log((d+2)(k+1)/\eps)$, which bounds each of $\log(d+2)$, $\log(k+1)$ and $\log(1/\eps)$ and exceeds $1$, and let $n=\lceil C(d+1)\eps^{-\ell}a^2\rceil$ for a parameter $C\ge1$. Then $n\le2C(d+1)\eps^{-\ell}a^2$, and since $\log a\le a$, we get $\log(2(n+1)(k+1))\le\log(8C)+(\ell+4)a\le c\,(1+\log C)\,a$. Hence
\[
 \frac{(d+1)\log^2(2(n+1)(k+1))}{n+1}\le\frac{c\,(1+\log C)^2a^2(d+1)}{C(d+1)\eps^{-\ell}a^2}=\frac{c\,(1+\log C)^2}{C}\,\eps^\ell .
\]
Choosing $C$ large enough in terms of $\ell$ therefore makes the right side of \eqref{eq:finite-one-expected} with $m=n$ at most $\eps/[32(\ell+1)^2]$. Lemma~\ref{lem:amplification} now gives $\mone_\H(\eps,\delta)\le n\lceil24(\ell+1)\log(2/\delta)\rceil$, which gives \eqref{eq:finite-one-rate} because $n\le2C(d+1)\eps^{-\ell}a^2$. Finally, $\dN\le\dDS$ by Lemma~\ref{lem:orders} and both bounds increase with $d$, so they also hold with $\dDS(\cI_{\ell+1}(\H))$ in place of $d$.

\paragraph{The case $d=0$.}
Let $d=0$ and fix $x\in\X$. We claim that the sets $h(x)$, over all $h\in\H$ defined at $x$, have a common label. Otherwise, by Lemma~\ref{lem:small}, at most $\ell+1$ of them already have an empty intersection. Since $\H\subseteq\cI_{\ell+1}(\H)$, these sets form a product witness of size one, that is, a family of correct-label sets at the single input $x$ with no common label, which contradicts $d=0$. The predictor that outputs at $x$ the first common label in the fixed enumeration of $\Y$, and a fixed label if no hypothesis is defined at $x$, uses no examples and is correct at every input of $\supp(h^\star)$, for every $h^\star\in\H$. This finishes the proof.
\end{proof}
\section{Worst-Case One-Correct-Label Sample Complexity}\label{sec:worst-case}

The upper bound of Theorem~\ref{thm:finite-one} grows as $\eps^{-\ell}$ in the accuracy. In this section we show that, for $\ell\ge2$, this exponent cannot be improved in general. The construction has two steps. We first define a base class $\cG_\ell$ on the grid $\N^\ell$, independent of the accuracy, whose partial intersection class $\cI_{\ell+1}(\cG_\ell)$ has generalized Natarajan dimension $(\ell+1)^\ell$. Independent copies of $\cG_\ell$ then give a lower bound that is linear in the dimension.

\paragraph{The base class.}
Fix $\ell\ge2$. The instance space is $\N^\ell$ and the label space is $[\ell]$; we write $x=(x_1,\ldots,x_\ell)$ for $x\in\N^\ell$. For $t\in\N^\ell$ and a finite set $F\subseteq\N^\ell$ define
\begin{equation}\label{wc:base}
 f_F(x)=1+\ind\{x\in F\},\qquad
 h_{t,F}(x)=\{f_F(x)\}\cup\{i\in[\ell]:x_i\ne t_i\},
\end{equation}
and let $\cG_\ell=\{h_{t,F}:t\in\N^\ell,\ F\subseteq\N^\ell\text{ finite}\}$. Since $f_F(x)\in\{1,2\}\subseteq[\ell]$, every $h_{t,F}(x)$ is a nonempty subset of $[\ell]$, so $\cG_\ell$ is a total class whose correct-label sets have at most $\ell$ labels. Thus each target has a binary label $f_F(x)$, which is always correct, and any other label $i$ is excluded only on the slice where the $i$th coordinate equals $t_i$:
\begin{equation}\label{wc:slice}
 i\notin h_{t,F}(x)\quad\Longrightarrow\quad x_i=t_i.
\end{equation}

\begin{lemma}[Dimension of the base class]\label{lem:hard-base-dimension}
$\dN(\cI_{\ell+1}(\cG_\ell))=(\ell+1)^\ell$.
\end{lemma}

\begin{proof}
For the lower bound, let $S=[\ell+1]^\ell$, fix any labeling $u:S\to\{1,2\}$, and let $F=\{x\in S:u(x)=2\}$, so that $f_F=u$ on $S$. For $a\in[\ell+1]$ let $t^{(a)}=(a,\ldots,a)$. Every $h_{t^{(a)},F}(x)$ contains $u(x)$, and a label $i\ne u(x)$ is excluded by the constituent with $a=x_i$, because then $x_i=t^{(a)}_i$ and $f_F(x)\ne i$. Therefore
\[
 \bigcap_{a=1}^{\ell+1} h_{t^{(a)},F}(x)=\{u(x)\}\qquad(x\in S).
\]
Thus $\cI_{\ell+1}(\cG_\ell)$ realizes every pattern of the singletons $\{1\},\{2\}$ on the $(\ell+1)^\ell$ inputs of $S$, and $S$ is product-shattered with the families $\{\{1\},\{2\}\}$.

For the upper bound, we show that a product-shattered set takes at most $\ell+1$ values in each coordinate. Let a finite set $S$ be product-shattered by $\cI_{\ell+1}(\cG_\ell)$ with families $\cA_x$, $x\in S$, and fix a label $i\in[\ell]$. Since the sets in $\cA_x$ have no common label, some $C_x\in\cA_x$ excludes $i$, and product shattering gives $g\in\cI_{\ell+1}(\cG_\ell)$ with $g(x)=C_x$ for all $x\in S$. Write $g(x)=\bigcap_{a=1}^{r}h_{t^{(a)},F^{(a)}}(x)$ for $x\in S$, with $r\le\ell+1$. For each $x\in S$ we have $i\notin g(x)$, so some constituent $h_{t^{(a)},F^{(a)}}$ excludes $i$ at $x$, and \eqref{wc:slice} gives $x_i=t^{(a)}_i$. Hence
\[
 \{x_i:x\in S\}\subseteq\{t^{(1)}_i,\ldots,t^{(r)}_i\}.
\]
This holds for every $i\in[\ell]$, so $S$ lies in a product of $\ell$ sets of size at most $\ell+1$, and $|S|\le(\ell+1)^\ell$.
\end{proof}

\paragraph{Independent copies.}
For an integer $b\ge1$, the instance space is $[b]\times\N^\ell$ and the label space is $[\ell]$. A target is given by parameters $\mathbf t=(t^{(1)},\ldots,t^{(b)})\in(\N^\ell)^b$ and a finite set $F\subseteq[b]\times\N^\ell$:
\begin{equation}\label{wc:block}
 f_F(j,x)=1+\ind\{(j,x)\in F\},\qquad
 h_{\mathbf t,F}(j,x)=\{f_F(j,x)\}\cup\{i\in[\ell]:x_i\ne t^{(j)}_i\}.
\end{equation}
Let $\H_b$ be the class of all these targets. A finite set $F\subseteq[b]\times\N^\ell$ amounts to a finite set $\{x:(j,x)\in F\}\subseteq\N^\ell$ for each block $j$, so $\H_b$ is the disjoint product of $b$ copies of the total class $\cG_\ell$, and Lemmas~\ref{lem:disjoint-product} and~\ref{lem:hard-base-dimension} give
\begin{equation}\label{wc:dimension}
 \dN(\cI_{\ell+1}(\H_b))=b\,(\ell+1)^\ell.
\end{equation}

\begin{theorem}[Worst-case lower bound]\label{thm:worst-case}
Fix $\ell\ge2$. For every $b\ge1$ and $0<\eps\le1/256$,
\[
 \mone_{\H_b}(\eps,1/8)>\frac{b}{4(32\eps)^\ell}.
\]
In particular, for every integer $d\ge1$ and $b=\lceil d/(\ell+1)^\ell\rceil$,
\[
 d\le\dN(\cI_{\ell+1}(\H_b))<d+(\ell+1)^\ell,
 \qquad
 \mone_{\H_b}(\eps,1/8)>\frac{d}{4(32(\ell+1)\eps)^\ell}.
\]
\end{theorem}

The idea of the proof is that demonstrations reveal the binary labels but nothing about the slices. We put the inputs uniformly on the grid $[b]\times[n]^\ell$, draw the binary labels as fair coins and the slice parameters $T^{(j)}_i$ uniformly from $[n]$, and let the demonstrator always show the binary label. A learner that sees at most a quarter of the grid must guess the binary label at most grid points, and about half of these guesses are wrong. A wrong guess $i$ at $(j,x)$ is an error only if $x_i=T^{(j)}_i$, and since the sample carries no information about $T^{(j)}_i$, this happens with probability $1/n$. The expected loss is therefore of order $1/n$, so with $n$ of order $1/\eps$ the learner needs a constant fraction of the $bn^\ell$ grid points, that is, of order $b\eps^{-\ell}$ examples; the exponent $\ell$ is the dimension of the grid. A second-moment argument over the $b\ell$ slices, which are independent given the sample, turns this into a bound that holds with constant probability.

\begin{proof}
Fix an integer $n\ge1$, put $G=[b]\times[n]^\ell$ and $N=|G|=bn^\ell$, and fix a learner, possibly randomized, that uses $m\le N/4$ examples. We construct a random target and a compatible one-correct-label distribution under which the loss of this learner exceeds $1/(16n)$ with probability more than $1/8$.

\paragraph{The hard distribution.}
Choose $F\subseteq G$ at random by including each point of $G$ independently with probability $1/2$, so that the values $f_F(j,x)$, $(j,x)\in G$, are independent and uniform on $\{1,2\}$. Independently of $F$, let the entries of $\mathbf T=(T^{(j)}_i)_{j\in[b],\,i\in[\ell]}$ be independent and uniform on $[n]$. The target is $h_{\mathbf T,F}\in\H_b$, the marginal $P$ is uniform on $G$, and the demonstrated label at $(j,x)$ is always $f_F(j,x)$, which belongs to $h_{\mathbf T,F}(j,x)$; so for every realization of $F$ and $\mathbf T$ this is a one-correct-label distribution for the target. The learner receives a sample $\sigma$ of $m$ independent examples, and its internal randomness is a seed $\omega\sim\Unif([0,1])$, independent of $F$, $\mathbf T$ and $\sigma$.

Since the demonstrations depend on $F$ but not on $\mathbf T$, the entries of $\mathbf T$ remain independent and uniform on $[n]$ conditional on $(F,\sigma,\omega)$. Given $(\sigma,\omega)$, the output of the learner is a fixed predictor, possibly randomized; let $p_i(j,x)$ be the probability that it predicts $i$ at $(j,x)$, so that $p_i(j,x)\ge0$ and $\sum_{i=1}^\ell p_i(j,x)=1$.

\paragraph{Disagreement with the binary labels.}
We first show that the output disagrees with $f_F$ on a constant fraction of $G$. Let
\[
 Q=\frac1N\sum_{(j,x)\in G}\bigl(1-p_{f_F(j,x)}(j,x)\bigr),\qquad \mu=\E[Q],
\]
so that $Q$ is the expected fraction of $G$ on which the output disagrees with $f_F$. At most $m$ points of $G$ appear in $\sigma$, and at every other point $(j,x)$ the value $f_F(j,x)$ is still uniform on $\{1,2\}$ given $(\sigma,\omega)$, so that
\[
 \E\bigl[1-p_{f_F(j,x)}(j,x)\,\big|\,\sigma,\omega\bigr]
 =1-\frac{p_1(j,x)+p_2(j,x)}2\ge\frac12.
\]
Since every summand of $Q$ lies in $[0,1]$ and $m\le N/4$,
\begin{equation}\label{wc:mu}
 0\le Q\le1,\qquad \mu\ge\frac{N-m}{2N}\ge\frac38.
\end{equation}

\paragraph{From disagreement to loss.}
Let $L$ be the loss of the learner's output against $h_{\mathbf T,F}$ under $P$. By \eqref{wc:block}, a label $i\in[\ell]$ is invalid at $(j,x)$ exactly when $f_F(j,x)\ne i$ and $x_i=T^{(j)}_i$. We group the errors according to the slice that causes them: $nL=Z=\sum_{j=1}^b\sum_{i=1}^\ell Z_{j,i}$, where
\[
 Z_{j,i}=\frac nN\sum_{x\in[n]^\ell:\ x_i=T^{(j)}_i} p_i(j,x)\,\ind\{f_F(j,x)\ne i\}.
\]
Conditional on $(F,\sigma,\omega)$, the numbers $p_i(j,x)$ and $f_F(j,x)$ are fixed, so $Z_{j,i}$ is a function of $T^{(j)}_i$ alone, and the variables $Z_{j,i}$ are conditionally independent. Each slice $\{x\in[n]^\ell:x_i=T^{(j)}_i\}$ has $n^{\ell-1}$ points, so $0\le Z_{j,i}\le n\cdot n^{\ell-1}/N=1/b$. Moreover, $\Pr[T^{(j)}_i=x_i\mid F,\sigma,\omega]=1/n$ for every fixed $x\in[n]^\ell$, and summing over $j$ and $i$ gives
\[
 \E\bigl[Z\,\big|\,F,\sigma,\omega\bigr]=\frac1N\sum_{(j,x)\in G}\ \sum_{i\ne f_F(j,x)}p_i(j,x)=Q.
\]
In particular $\E[Z]=\mu\ge3/8$.

\paragraph{A second-moment bound.}
By conditional independence, and since $0\le Z_{j,i}\le1/b$ implies $\Var(Z_{j,i}\mid F,\sigma,\omega)\le\frac1b\E[Z_{j,i}\mid F,\sigma,\omega]$,
\[
 \E\bigl[Z^2\,\big|\,F,\sigma,\omega\bigr]=Q^2+\sum_{j=1}^b\sum_{i=1}^\ell\Var\bigl(Z_{j,i}\,\big|\,F,\sigma,\omega\bigr)\le Q^2+\frac Qb.
\]
Taking expectations and using $Q^2\le Q$ and $b\ge1$ gives $\E[Z^2]\le2\mu$. Since $\mu/6\ge1/16$, the Paley--Zygmund inequality yields
\[
 \Pr\Bigl[Z>\frac1{16}\Bigr]\ge\Pr\Bigl[Z>\frac\mu6\Bigr]\ge\Bigl(\frac56\Bigr)^2\frac{\mu^2}{\E[Z^2]}\ge\frac{25}{72}\,\mu>\frac18.
\]
As $Z=nL$, this means $\Pr[L>1/(16n)]>1/8$, where the probability is over $F$, $\mathbf T$, the sample $\sigma$ and the seed $\omega$. Since $F$ and $\mathbf T$ take finitely many values, some fixed realization $(F,\mathbf T)$ satisfies $\Pr_{\sigma,\omega}[L>1/(16n)]>1/8$, and this realization is a target in $\H_b$ together with a compatible one-correct-label distribution.

\paragraph{Choice of $n$.}
Now let $0<\eps\le1/256$ and choose $n=\lfloor1/(16\eps)\rfloor$. Then $n\ge16$, $1/(16n)\ge\eps$, and $n\ge1/(32\eps)$. We have shown that every learner using $m\le bn^\ell/4$ examples has $\Pr[L>\eps]>1/8$ for some target in $\H_b$ and some compatible distribution, and therefore
\[
 \mone_{\H_b}(\eps,1/8)>\frac{bn^\ell}4\ge\frac b{4(32\eps)^\ell}.
\]
For the last assertion, let $b=\lceil d/(\ell+1)^\ell\rceil$. Then $d\le b(\ell+1)^\ell<d+(\ell+1)^\ell$, so \eqref{wc:dimension} gives $d\le\dN(\cI_{\ell+1}(\H_b))<d+(\ell+1)^\ell$. Since $b\ge d/(\ell+1)^\ell$, the bound just proved gives $\mone_{\H_b}(\eps,1/8)>d/(4(32(\ell+1)\eps)^\ell)$. This finishes the proof.
\end{proof}

\begin{remark}[The accuracy exponent is a worst-case statement]\label{rem:worst-only}
The exponent $\ell$ in Theorem~\ref{thm:worst-case} is forced for the worst class of a given dimension, not by the dimension alone: for $\ell\ge2$, no lower bound proportional to $d\eps^{-\ell}$ holds for every class with $\dN(\cI_{\ell+1}(\H))\ge d$. Indeed, let $\X=\{x_0,x_1,\ldots,x_d\}$ and $\Y=[\ell]$, and let $\H$ consist of all $h$ with $h(x_j)\in\{\{1\},\{2\}\}$ for $j\in[d]$ and $h(x_0)=[\ell]$; its largest correct-label set has $\ell$ labels. The inputs $x_1,\ldots,x_d$ are product-shattered with the families $\{\{1\},\{2\}\}$, while $x_0$ lies in no product-shattered set, because every member of $\cI_{\ell+1}(\H)$ defined at $x_0$ equals $[\ell]$ there. Thus $\dN(\cI_{\ell+1}(\H))=d$.

This class is no harder than a binary class on $d$ points: every label is correct at $x_0$, and at $x_1,\ldots,x_d$ the correct label is unique and is the one demonstrated. Let the learner output a predictor $g$ with $g(x_0)=1$ and $g(x_j)\in\{1,2\}$ for $j\in[d]$ that agrees with every demonstrated label at the inputs $x_1,\ldots,x_d$ of the sample; such a predictor exists, since the target's labels at $x_j$ are singletons. A predictor of this form is wrong only at inputs $x_j$ with $j\in[d]$, and there exactly when it disagrees with the demonstrated label; so if its loss exceeds $\eps$, it agrees with all $m$ demonstrations at these inputs with probability at most $(1-\eps)^m$. A union bound over the $2^d$ predictors gives
\[
 \mone_{\H}(\eps,\delta)\le\left\lceil\frac{d\log2+\log(1/\delta)}\eps\right\rceil,
\]
which is of order $d/\eps$ rather than $d\eps^{-\ell}$.
\end{remark}


\section{Full-Set Learning with Countably Many Labels}\label{sec:countable-full}

We now let the label space $\Y$ be countable, and possibly infinite. The rates of Section~\ref{sec:finite-full} grow with $k=|\Y|$ and become vacuous when $\Y$ is infinite, so the question is which dimension takes the place of the generalized Natarajan dimension. The answer is the packed dimension of Definition~\ref{def:pack}: it characterizes full-set learning, in both the realizable and the agnostic setting, with rates that do not depend on $|\Y|$. Throughout the section we fix $\ell\ge1$ and $\gamma$ as in \eqref{eq:gamma-range}.

\begin{theorem}[Full-set learning with countably many labels]\label{thm:countable-full}
For every nonempty class $\H\subseteq(\cA\cup\{\star\})^{\X}$ the following are equivalent:
\begin{enumerate}[(i)]
\item $\H$ is realizably full-set PAC learnable;
\item $\H$ is agnostically full-set PAC learnable, with the benchmark $\Lset$ of \eqref{eq:agnostic};
\item $\dpack(\H)<\infty$.
\end{enumerate}
Moreover, if $d=\dpack(\H)<\infty$ and $B=d+1+\log(2/\delta)$, then
\begin{align}
 \mfull_{\H}(\eps,\delta)&\le
 \left\lceil c\,\frac{B}{\eps}\log^2\!\left(\frac{2B}{\eps}\right)\right\rceil,
 \label{eq:full-rate}\\
 \mfullag_{\H}(\eps,\delta)&\le
 \left\lceil c\,\frac{B}{\eps^2}\log^2\!\left(\frac{2B}{\eps}\right)\right\rceil,
 \label{eq:full-agnostic-rate}
\end{align}
where $c$ depends only on $\ell$ and $\gamma$. These conditions imply $\dN(\H)<\infty$. Finally, for $\ell\ge2$, finiteness of $\dDS(\H)$ neither implies nor is implied by realizable full-set PAC learnability, even for total classes with at most two correct labels per input.
\end{theorem}

The last assertion is proved in Appendix~\ref{app:separations} (Theorems~\ref{thm:rank-example} and~\ref{thm:large-ds-learnable}). Necessity of a finite packed dimension is a no-free-lunch argument: the conflict blocks of a packed witness on many inputs act like a shattered set and force a constant expected error (Section~\ref{cf:sec-lower}).

Sufficiency follows the boosting-to-compression route, in three steps.
\begin{itemize}
\item \emph{A weak learner} (Section~\ref{cf:sec-weak}). If $d=\dpack(\H)$ is finite, then for every finite set of patterns and every distribution on it, all but at most $d$ coordinates can be predicted from the other coordinates with error at most $\gamma$. A one-inclusion graph argument turns this into a learner with $O(d+1)$ examples whose expected error is below $1/(\ell+1)$.
\item \emph{Compression} (Section~\ref{cf:sec-compression}). Boosting the weak learner shows that every realizable sample of size $N$ contains $O((d+1)\log N)$ examples from which a predictor that is valid on the whole sample can be reconstructed. The generalization bound for sample compression schemes then gives \eqref{eq:full-rate}; its union bound over the $N^{O((d+1)\log N)}$ short lists is the source of the second logarithm.
\item \emph{The agnostic case} (Section~\ref{cf:sec-agnostic}). The learner decodes every short list of examples and outputs a decoded predictor with the least empirical error.
\end{itemize}

\subsection{The Lower Bound}\label{cf:sec-lower}

When the test input is missing from the training sample, the learner cannot distinguish the targets of a conflict block, and no single label is correct for all of them. Since the blocks of each direction cover a $\gamma$ fraction of the witness, this costs a constant error.

\begin{proposition}\label{prop:pack-lower}
Suppose that $\H$ has a $\gamma$-packed witness on $n\ge2$ inputs. Then for every $m$ and every full-set learner that uses $m$ examples, possibly with randomized predictions, there are $h^\star\in\H$ and a distribution $P$ with $P(\supp(h^\star))=1$ such that the output $\widehat g$ satisfies
\begin{equation}\label{cf:pack-lower}
 \E\bigl[L_{P,h^\star}(\widehat g)\bigr]\ge
 \frac{\gamma}{\ell+1}\left(1-\frac1n\right)^m.
\end{equation}
\end{proposition}

\begin{proof}
Let $S=(x_1,\ldots,x_n)$ be the inputs of the witness, and let $v^1,\ldots,v^M\in\H|_S$ and $\cB_1,\ldots,\cB_n$ be as in Definition~\ref{def:pack}. For each index $a\in[M]$ fix $h_a\in\H$ that is defined on $S$ and satisfies $(h_a(x_1),\ldots,h_a(x_n))=v^a$. Let $P$ be uniform on $\{x_1,\ldots,x_n\}$, so that $P(\supp(h_a))=1$ for every $a$, and choose the target $h_a$ with $a$ uniform in $[M]$.

Fix a direction $i$ and a sequence of training inputs that avoids $x_i$, and condition on it. Let $B\in\cB_i$. All $h_a$ with $a\in B$ agree at the witness inputs other than $x_i$, so they produce the same training sample, and the learner predicts at $x_i$ with the same distribution $p(\cdot\mid x_i)$ for all of them. Since $\bigcap_{a\in B}v_i^a=\emptyset$, every label $y$ lies outside $v_i^a$ for some $a\in B$, and averaging over $y\sim p(\cdot\mid x_i)$ gives
\[
 \sum_{a\in B}\bigl[1-p(v_i^a\mid x_i)\bigr]\ge1 .
\]
Thus each block contributes at least $1/M$ to the error at $x_i$ averaged over the uniform index $a$, and the indices outside the blocks contribute nonnegatively. The blocks in $\cB_i$ are disjoint, have at most $\ell+1$ members each, and cover at least $\gamma M$ indices, so there are at least $\gamma M/(\ell+1)$ of them, and the averaged error at $x_i$ is at least $\gamma/(\ell+1)$.

The test input is $x_i$ with probability $1/n$, and $x_i$ is absent from the training sample with probability $(1-1/n)^m$; neither event depends on $a$. Averaging over $i$ shows that the expected loss, averaged over $a$, is at least the right side of \eqref{cf:pack-lower}. Hence some fixed index $a$ attains at least this average, and we take $h^\star=h_a$.
\end{proof}

\begin{corollary}\label{cf:necessity}
If $\dpack(\H)=\infty$, then $\H$ is not realizably full-set PAC learnable.
\end{corollary}

\begin{proof}
Fix $m$ and take a $\gamma$-packed witness on $n\ge\max\{2,2m\}$ inputs. Since $(1-1/n)^m\ge1-m/n\ge1/2$, Proposition~\ref{prop:pack-lower} gives a target with expected loss at least $\gamma/[2(\ell+1)]$. On the other hand, if a learner with $m$ examples had an $(\eps,\delta)$ guarantee for $\eps=\delta=\gamma/[8(\ell+1)]$, then by \eqref{eq:pac-expect} its expected loss would be at most $\eps+\delta=\gamma/[4(\ell+1)]$, a contradiction.
\end{proof}

\subsection{A Weak Learner from the Packed Dimension}\label{cf:sec-weak}

The weak learner is a randomized one-inclusion graph algorithm: it treats the test input as one more position of a pool and predicts its label from the correct-label sets at the other positions. Its error is governed by the following quantities. Let $S=(x_1,\ldots,x_n)$ be an ordered pool, possibly with repeated inputs, and let $V\subseteq\H|_S$ be finite and nonempty. Write $\cE_i(V)$ for the set of hyperedges of $V$ in direction $i$ (see \eqref{eq:edge}); each $v\in V$ lies in exactly one of them. For a distribution $\mu$ on $V$ put
\begin{align}
 b_i(V,\mu)&=\sum_{e\in\cE_i(V)}
  \Bigl[\mu(e)-\max_{y\in\Y}\mu\bigl(\{v\in e:y\in v_i\}\bigr)\Bigr],
 \label{cf:bi}\\
 \Lambda(V,\mu)&=\sum_{i=1}^n b_i(V,\mu).
 \label{cf:Lambda}
\end{align}
Only finitely many labels occur in $V$, so each maximum exists. If $v\sim\mu$, then $b_i(V,\mu)$ is the smallest expected error of a rule that predicts a label at coordinate $i$ from $v_{-i}$. Indeed, such a rule chooses one label for each hyperedge in direction $i$, and the best choice is a label that lies in $v_i$ for the largest mass of patterns $v$ of the hyperedge. In particular, $0\le b_i(V,\mu)\le1$.

The next two lemmas bound $\Lambda$ by the packed dimension. The main idea is that, once rational weights are written as repetitions, each hyperedge in direction $i$ contains disjoint conflict blocks that cover at least as much mass as its best label misses (Lemma~\ref{lem:pack-family}). Thus, if $b_i(V,\mu)>\gamma$ held in more than $\dpack(\H)$ directions, the blocks of these directions would form a $\gamma$-packed witness, which is impossible (Lemma~\ref{lem:ambiguity}).

\begin{lemma}\label{lem:pack-family}
Let $A_1,\ldots,A_K\in\cA$. There are pairwise disjoint sets $B_1,\ldots,B_s\subseteq[K]$ such that $2\le|B_j|\le\ell+1$ and $\bigcap_{a\in B_j}A_a=\emptyset$ for every $j$, and
\[
 \Bigl|\bigcup_{j=1}^s B_j\Bigr|\ge K-\max_{y\in\Y}\bigl|\{a\in[K]:y\in A_a\}\bigr|.
\]
\end{lemma}

\begin{proof}
As long as the sets at the remaining indices have empty intersection, remove a set of indices whose sets form a minimal family with empty intersection. By Lemma~\ref{lem:small} each removed set has between $2$ and $\ell+1$ indices, and the removed sets are disjoint by construction. When the process stops, either no index remains or the remaining sets share a label $y$, and in the second case at most $|\{a:y\in A_a\}|$ indices remain. In both cases at least the claimed number of indices has been removed.
\end{proof}

\begin{lemma}\label{lem:ambiguity}
Suppose $d=\dpack(\H)<\infty$. Let $S$ be a pool of $n$ inputs, let $V\subseteq\H|_S$ be finite and nonempty, and let $\mu$ be a distribution on $V$. Then at most $d$ coordinates $i$ satisfy $b_i(V,\mu)>\gamma$. In particular,
\begin{equation}\label{cf:ambiguity}
 \Lambda(V,\mu)\le d+n\gamma.
\end{equation}
\end{lemma}

\begin{proof}
We first dispose of repeated inputs: if $x_i=x_j$ for some $j\ne i$, then $b_i(V,\mu)=0$. Indeed, every $v\in V$ is the pattern of a hypothesis, so $v_i=v_j$. The members of a hyperedge in direction $i$ agree at $j$, hence also at $i$, so any label of their common set at $i$ is correct for all of them, and every term of \eqref{cf:bi} vanishes.

Next, let $\mu$ take rational values with common denominator $K$. We show that $d+1$ coordinates $i$ with $b_i(V,\mu)>\gamma$ would yield a $\gamma$-packed witness on $d+1$ distinct inputs. Form a list of length $K$ that contains $K\mu(v)$ copies of each $v\in V$, and fix $i$. Applying Lemma~\ref{lem:pack-family}, separately in each hyperedge $e\in\cE_i(V)$, to the sets $v_i$ of the copies in $e$ gives disjoint blocks of between $2$ and $\ell+1$ copies that agree outside $i$ and have empty intersection at $i$. In $e$ these blocks cover at least $K\mu(e)-K\max_y\mu(\{v\in e:y\in v_i\})$ copies, so the blocks of direction $i$ together cover at least $K\,b_i(V,\mu)$ copies.

Now suppose that a set $I$ of $d+1$ coordinates satisfies $b_i(V,\mu)>\gamma$ for all $i\in I$. By the first step, the inputs $x_i$ with $i\in I$ are distinct. Project every copy onto $I$, keeping all $K$ copies, and keep the blocks of the directions in $I$. The projected copies lie in $\H|_{S_I}$, where $S_I$ is the subpool indexed by $I$. The blocks are sets of copies, so projection leaves them disjoint and of the same sizes; the members of a block of direction $i$ still agree at every coordinate of $I$ other than $i$, and their sets at $i$, and hence their empty intersection, are unchanged. Since the blocks of each direction in $I$ still cover more than $\gamma K$ copies, this is a $\gamma$-packed witness on $d+1$ distinct inputs, a contradiction.

For a general $\mu$, each $b_i(V,\mu)$ is a continuous function of the finitely many masses of $\mu$, being a sum of linear terms minus maxima of finitely many linear terms. So if $b_i(V,\mu)>\gamma$ held for $d+1$ coordinates, it would also hold for them under a sufficiently close rational distribution, which we have just excluded. Finally, $b_i\le1$ for the at most $d$ exceptional coordinates and $b_i\le\gamma$ for the others, which gives \eqref{cf:ambiguity}.
\end{proof}

Lemma~\ref{lem:ambiguity} says that against every distribution on patterns some prediction rule makes few mistakes on the pool. To obtain a learner we need the quantifiers in the other order, a single rule that is good for every pattern; the minimax theorem provides it, and the leave-one-out argument of the one-inclusion graph algorithm turns its bound on the pool into a bound on the expected error.

\begin{lemma}\label{lem:packed-weak}
Suppose $d=\dpack(\H)<\infty$. For every $m\ge1$ there is a full-set learner that maps each sample $\sigma$ of $m$ examples to a randomized prediction $p_\sigma$ such that
\begin{equation}\label{cf:weak-pack}
 \E_{\sigma\sim\D_{P,h^\star}^{m}}\bigl[L_{P,h^\star}(p_\sigma)\bigr]\le\gamma+\frac{d}{m+1}
\end{equation}
for every $h^\star\in\H$ and every $P$ with $P(\supp(h^\star))=1$. Moreover, $p_\sigma$ is a deterministic function of $\sigma$ that does not change when the examples of $\sigma$ are permuted.
\end{lemma}

\begin{proof}
Let $n=m+1$ and fix an ordered pool $S=(x_1,\ldots,x_n)$. A \emph{transductive rule} on $S$ assigns to each direction $i\in[n]$ and each $u\in\cA^{n-1}$ a distribution $p_{i,u}$ on $\Y$, and at position $i$ of a pattern $v$ it predicts with $p_{i,v_{-i}}$. Its loss on $v$ is the expected number of positions at which it is wrong,
\[
 \sum_{i=1}^n\bigl[1-p_{i,v_{-i}}(v_i)\bigr].
\]
We first show that some transductive rule on $S$ has loss at most $d+n\gamma$ on every $v\in\H|_S$.

Let $V\subseteq\H|_S$ be finite and nonempty, and let $\Y_V$ be the finite set of labels that occur in $V$. Consider the finite zero-sum game in which the learner chooses a label in $\Y_V$ for every hyperedge of $V$, the adversary chooses $v\in V$, and the loss is the number of positions $i$ at which the label chosen for the hyperedge of $v$ in direction $i$ is not in $v_i$. Against a distribution $\mu$ on $V$, the learner can choose the labels of different hyperedges separately, and the label $y$ chosen for $e\in\cE_i(V)$ contributes $\mu(e)-\mu(\{v\in e:y\in v_i\})$ to the expected loss. Labels outside $\Y_V$ are correct for no pattern, so the maximum in \eqref{cf:bi} may be taken over $\Y_V$, and the best response to $\mu$ has expected loss exactly $\Lambda(V,\mu)\le d+n\gamma$ (Lemma~\ref{lem:ambiguity}). By the minimax theorem \cite{vonneumann1928theorie}, the learner has a mixed strategy whose expected loss is at most $d+n\gamma$ against every $v\in V$. This loss depends only on the law $p_{i,u}$ of the label that the strategy chooses for each hyperedge $e_{i,u}$, so these laws, with the remaining $p_{i,u}$ set arbitrarily, form a transductive rule with loss at most $d+n\gamma$ on every $v\in V$.

The set $\H|_S$ may be infinite, and we pass to it by compactness. When $\Y$ is infinite, the probability distributions on $\Y$ do not form a closed subset of $[0,1]^{\Y}$, so we use the compact set $\cP_{\le1}(\Y)$ of all $z\in[0,1]^{\Y}$ with $\sum_{y\in F}z_y\le1$ for every finite $F\subseteq\Y$. Assign a vector $z^{i,u}\in\cP_{\le1}(\Y)$ to every pair $(i,u)$, and let $Z$ be the compact product of these spaces. For $v\in\H|_S$, the constraint $\sum_{i=1}^n[1-\sum_{y\in v_i}z^{i,v_{-i}}_y]\le d+n\gamma$ involves finitely many coordinates, since each $v_i$ is finite, so it defines a closed subset of $Z$. By the finite case, every finitely many of these constraints are satisfied by some point of $Z$, and by the finite intersection property some point satisfies all of them. Adding the missing mass $1-\sum_y z^{i,u}_y$ of each $z^{i,u}$ to one fixed label gives distributions $p_{i,u}$ and does not decrease the mass on any correct-label set, so the resulting rule has loss at most $d+n\gamma$ on every $v\in\H|_S$.

Now fix such a rule for every ordered pool of $n$ inputs; since $\X$ is countable, there are countably many pools, so the resulting learner is measurable. Replace the rule on each pool $S$ by its average over all permutations of the positions of $S$, applied to the correspondingly permuted pool and patterns. This preserves the bound on the loss, which is a sum over all positions, and makes the rules permutation-equivariant. Given $\sigma=((x_1,A_1),\ldots,(x_m,A_m))$ and a test input $x$, the learner forms the pool $(x_1,\ldots,x_m,x)$ and predicts with $p_{n,(A_1,\ldots,A_m)}$. By equivariance, this prediction does not change when the examples of $\sigma$ are permuted.

Finally, let $x_1,\ldots,x_n$ be i.i.d.\ from $P$ and $v=(h^\star(x_1),\ldots,h^\star(x_n))$. Almost surely all $x_j$ lie in $\supp(h^\star)$, so $v\in\H|_S$. The expected loss of the learner is the expected error of the rule at position $n$, and by exchangeability and equivariance it equals the expected error at every position $i$. It is therefore $1/n$ times the expected loss of the transductive rule on $v$, which is at most $(d+n\gamma)/n=\gamma+d/(m+1)$.
\end{proof}

\subsection{Compression and the Realizable Upper Bound}\label{cf:sec-compression}

A full-set sample $\sigma=((x_1,A_1),\ldots,(x_n,A_n))$ is \emph{realizable} by $\H$ if some $h\in\H$ satisfies $x_i\in\supp(h)$ and $h(x_i)=A_i$ for all $i$. We now boost the weak learner on a realizable sample until its most probable label is correct on the whole sample, following the boosting-to-compression argument of \cite{freund1997decision}, \cite{moran2016sample}, \cite{david2016supervised} and \cite{cohen2025sample}. As in the last two papers, training labels are sets and the loss is validity; Lemma~\ref{lem:mode} replaces the majority vote. Since the boosted predictor is determined by few examples of the sample, the generalization argument for sample compression schemes \cite{littlestone1986relating} then gives the realizable bound.

\paragraph{Boosting.}
We prove the boosting step once, for an arbitrary class $\H_{\mathrm{aux}}\subseteq(\cA\cup\{\star\})^{\X}$ and with the number of rounds as a parameter, because Section~\ref{sec:countable-one} applies it to an intersection class. Fix $M\ge1$ and a full-set learner for $\H_{\mathrm{aux}}$ that uses $M$ examples and whose randomized prediction has expected loss at most $\alpha$, where $0\le\alpha<1/(\ell+1)$, for every target $h\in\H_{\mathrm{aux}}$ and every $P$ with $P(\supp(h))=1$. Averaging its prediction over its seed and over all orderings of its sample gives a rule $\sigma\mapsto p_\sigma$ with the same guarantee that is \emph{symmetric}, that is, $p_\sigma$ depends only on the multiset of examples in $\sigma$. Put $\tau=1/(\ell+1)$, $\Delta=\tau-\alpha>0$ and
\begin{equation}\label{eq:boost-c}
 \eta=\frac{\Delta}{2(\alpha+1)},\qquad
 \zeta=\eta\tau-\alpha(e^\eta-1).
\end{equation}
Both constants depend only on $\ell$ and $\alpha$, and $\zeta>0$ (see the proof below).

The next lemma is the multiplicative-weights analysis of boosting \cite{freund1997decision}, run with weights on the positions of a pool. In each round the weak learner has small error on the weighted pool, so the total weight grows slowly, while by Lemma~\ref{lem:mode} the weight of a position at which the final most probable label is invalid grows fast. Here we only need long runs, after which no such position is left; the correction step at the end of the lemma is used in Section~\ref{sec:countable-one}, where runs are shorter.

\begin{lemma}\label{lem:boosted-reconstruction}
Fix integers $N,T\ge1$. Let $S=(x_1,\ldots,x_n)$ be an ordered pool with $n\le N$, let $h\in\H_{\mathrm{aux}}$, and suppose that $D=\{i\in[n]:x_i\in\supp(h)\}$ is nonempty. Then there are multisets $\sigma_1,\ldots,\sigma_T$, each consisting of $M$ examples $(x_i,h(x_i))$ with $i\in D$, such that the predictor
\[
 f_0(x)=\text{a most probable label of }\ \frac1T\sum_{t=1}^T p_{\sigma_t}(\cdot\mid x)
\]
satisfies $f_0(x_i)\notin h(x_i)$ for at most $Ne^{-\zeta T}$ positions $i\in D$. Moreover, let $Q\subseteq D$ be the set of these positions, and let $f(x)$ be the least label of $h(x_i)$ in the fixed enumeration of $\Y$ if $x=x_i$ for some $i\in Q$, and $f(x)=f_0(x)$ otherwise. Then $f(x_i)\in h(x_i)$ for every $i\in D$, and $f$ is determined by $\sigma_1,\ldots,\sigma_T$ and the examples $(x_i,h(x_i))$ with $i\in Q$.
\end{lemma}

\begin{proof}
We first check that $\zeta>0$. Since $0<\eta<1$, we have $e^\eta-1\le\eta+\eta^2$, and since $\alpha\eta\le\Delta/2$, this gives $\zeta\ge\eta\Delta-\alpha\eta^2\ge\eta\Delta/2>0$.

Give every position $i\in D$ the weight $w_1(i)=1$. At round $t$, let $P_t$ be the distribution of $x_i$ when $i\in D$ is drawn with probability proportional to $w_t(i)$. Then $P_t(\supp(h))=1$, and for every sample $\sigma$ the loss $L_{P_t,h}(p_\sigma)$ is the weighted average of $1-p_\sigma(h(x_i)\mid x_i)$ over $i\in D$. A sample from $\D_{P_t,h}^M$ consists of $M$ examples $(x_i,h(x_i))$ with $i\in D$, and by the weak-learning guarantee its expected loss is at most $\alpha$. Hence some multiset $\sigma_t$ of $M$ such examples has $L_{P_t,h}(p_{\sigma_t})\le\alpha$. Put
\[
 \lambda_t(i)=1-p_{\sigma_t}(h(x_i)\mid x_i)\in[0,1]
 \qquad\text{and}\qquad
 w_{t+1}(i)=w_t(i)\,e^{\eta\lambda_t(i)}.
\]
Since $e^{\eta z}\le1+(e^\eta-1)z$ for $0\le z\le1$, the total weight grows in each round by a factor at most $1+(e^\eta-1)\alpha\le\exp(\alpha(e^\eta-1))$. As it starts at $|D|\le N$, after $T$ rounds it is at most $N\exp(T\alpha(e^\eta-1))$.

Now suppose that $f_0(x_i)\notin h(x_i)$ for some $i\in D$. Since $|h(x_i)|\le\ell$, Lemma~\ref{lem:mode} gives $T^{-1}\sum_{t}\lambda_t(i)\ge\tau$, so the final weight of $i$ is at least $e^{\eta T\tau}$. If there are $e_T$ such positions, then
\[
 e_T\,e^{\eta T\tau}\le N e^{T\alpha(e^\eta-1)},
 \qquad\text{that is,}\qquad
 e_T\le N e^{-\zeta T}.
\]
For the correction step, note that equal inputs have equal correct-label sets under $h$ and equal values under $f_0$, so a position whose input is also the input of a position in $Q$ lies in $Q$ itself. Hence $f(x_i)=f_0(x_i)\in h(x_i)$ for $i\in D\setminus Q$, while $f(x_i)\in h(x_i)$ for $i\in Q$ by definition. The values of $f$ depend only on the stated examples.
\end{proof}

\paragraph{A compression scheme.}
Assume $d=\dpack(\H)<\infty$, and apply the construction above with $\H_{\mathrm{aux}}=\H$ and the learner of Lemma~\ref{lem:packed-weak}, where
\[
 \alpha=\frac12\Bigl(\gamma+\frac1{\ell+1}\Bigr),\qquad
 M=\max\Bigl\{1,\Bigl\lceil\frac d{\alpha-\gamma}\Bigr\rceil\Bigr\}.
\]
By \eqref{cf:weak-pack}, the expected loss of this learner is at most $\gamma+d/(M+1)\le\alpha$, and $\alpha<1/(\ell+1)$, so $\eta$ and $\zeta$ depend only on $\ell$ and $\gamma$. For a public bound $N$ on the sample size put
\begin{equation}\label{cf:rN}
 T_N=\Bigl\lceil\frac{2\log(2N)}{\zeta}\Bigr\rceil,
 \qquad r_N=MT_N .
\end{equation}
The decoder $\Dec_N$ splits an ordered list of $r_N$ full-set examples into $T_N$ consecutive groups $\sigma_1,\ldots,\sigma_{T_N}$ of $M$ examples, and its output, the \emph{decoded predictor} of the list, is a most probable label of $T_N^{-1}\sum_tp_{\sigma_t}(\cdot\mid x)$ at every $x$.

\begin{lemma}\label{def:compression}
Every nonempty realizable sample $\sigma$ of size at most $N$ has an ordered list of $r_N$ of its examples, with repetitions allowed, whose decoded predictor is valid on all of $\sigma$. Moreover, for fixed $\ell$ and $\gamma$,
\begin{equation}\label{cf:rN-bound}
 r_N\le c\,(d+1)\log(2N).
\end{equation}
\end{lemma}

\begin{proof}
Apply Lemma~\ref{lem:boosted-reconstruction} with $T=T_N$ rounds to the inputs of $\sigma$ and a hypothesis $h$ that realizes $\sigma$. Then $D$ consists of all positions, so the multisets $\sigma_t$ consist of examples of $\sigma$, and since $Ne^{-\zeta T_N}\le1/(4N)<1$, the predictor $f_0$ is valid on all of $\sigma$. List the multisets one after the other, in any order within each. Since each $p_{\sigma_t}$ depends only on the multiset $\sigma_t$, the decoder returns $f_0$ on this list. Finally, $M\le c(d+1)$ and $T_N\le c\log(2N)$, which gives \eqref{cf:rN-bound}.
\end{proof}

In other words, the lists and $\Dec_N$ form a sample compression scheme of size $r_N$, in which the stored examples are ordered and may repeat.

\begin{proof}[Proof of Theorem~\ref{thm:countable-full}, realizable part]
Let $\sigma\sim\D_{P,h^\star}^N$. The decoded predictor of an index list in $[N]^{r_N}$ is that of the list of examples it indexes. The learner goes through these index lists in a fixed order and outputs the decoded predictor $\widehat g$ of the first list whose decoded predictor is valid on all of $\sigma$. Such a list exists by Lemma~\ref{def:compression}, since $\sigma$ is realizable almost surely.

Fix a list, let $J$ be its set of distinct indices, so that $|J|\le r_N$, and let $g$ be its decoded predictor. Conditional on the examples indexed by $J$, the predictor $g$ is fixed and the other $N-|J|$ examples are i.i.d. If $L_{P,h^\star}(g)>\eps$, then $g$ is valid on all of them with probability at most $e^{-\eps(N-r_N)}$. A union bound over the at most $N^{r_N}$ lists gives
\begin{equation}\label{cf:compression-tail}
 \Pr\bigl[L_{P,h^\star}(\widehat g)>\eps\bigr]
 \le\exp\bigl(r_N\log N-\eps(N-r_N)\bigr),\qquad N>r_N.
\end{equation}
Now take $N=\lceil cB\eps^{-1}\log^2(2B/\eps)\rceil$. By \eqref{cf:rN-bound}, the term $r_N\log N$ is of order $(d+1)\log^2(2N)$, and this is where the squared logarithm in \eqref{eq:full-rate} comes from. For a large enough $c$, a routine calculation gives $r_N(\log N+1)\le\eps N/2$ and $\log(1/\delta)\le\eps N/2$. The first inequality implies $N>2r_N$, and together they bound the exponent in \eqref{cf:compression-tail} by $-\eps N/2\le-\log(1/\delta)$. This proves \eqref{eq:full-rate} and (iii)$\Rightarrow$(i). Corollary~\ref{cf:necessity} gives (i)$\Rightarrow$(iii), and $\dN(\H)\le\dpack(\H)$ by Lemma~\ref{lem:orders}.
\end{proof}

\subsection{Agnostic Full-Set Learning}\label{cf:sec-agnostic}

The benchmark $\Lset_\D(\H)$ in \eqref{eq:agnostic} compares the whole set $h(x)$ with the observed set, so the examples on which a fixed $h\in\H$ equals the observed set are realizable by $h$. The compression scheme therefore applies to them: some decoded predictor is valid on these examples, and it is wrong on the sample only where $h$ disagrees with the observed set. It remains to show that the empirical errors of the finitely many decoded predictors concentrate. This is the set-valued form of the compression-based agnostic reduction of \cite{david2016supervised} and \cite{cohen2025sample}.

\begin{lemma}\label{lem:agnostic-reconstruction}
For every $n\ge1$, let $r_n\ge1$ and let $\Dec_n$ map ordered lists of $r_n$ full-set examples to predictors. Assume that for every nonempty realizable sample $\sigma'$ of size at most $n$ there is an ordered list of $r_n$ examples of $\sigma'$, with repetitions allowed, whose decoded predictor is valid on all of $\sigma'$. Then for $n>r_n$ there is an agnostic full-set learner with $n$ examples whose output satisfies, with probability at least $1-\delta$,
\begin{equation}\label{cf:agnostic-compression}
 L_\D(\widehat g)\le\Lset_\D(\H)
 +\frac{r_n}{n}
 +\sqrt{\frac{r_n\log n+\log(4/\delta)}{2(n-r_n)}}
 +\sqrt{\frac{\log(2/\delta)}{2n}}.
\end{equation}
\end{lemma}

For $n\ge2r_n$, the excess in \eqref{cf:agnostic-compression} is $O(\sqrt{(r_n\log n+\log(1/\delta))/n})$, the familiar form of agnostic compression bounds.

\begin{proof}
Let $\sigma=((x_i,A_i))_{i=1}^n\sim\D^n$ and write $\widehat L_\sigma(g)=\frac1n\sum_{i=1}^n\ind\{g(x_i)\notin A_i\}$. The candidates are the decoded predictors of all index lists in $[n]^{r_n}$ and one fixed fallback predictor, at most $n^{r_n}+1$ predictors in all. The learner outputs a candidate $\widehat g$ that minimizes $\widehat L_\sigma$, with ties broken by a fixed order. Note that $\widehat g$ does not depend on any hypothesis in $\H$.

Fix $h\in\H$. The positions $i$ with $h(x_i)=A_i$ form a realizable subsample (recall that $h(x_i)\ne A_i$ when $h(x_i)=\star$). If this subsample is nonempty, some candidate is valid on it; if it is empty, every candidate has empirical loss at most $1$. In both cases
\begin{equation}\label{cf:comparator}
 \widehat L_\sigma(\widehat g)
 \le\frac1n\sum_{i=1}^n\ind\{h(x_i)\ne A_i\}.
\end{equation}

Next, fix an index list, let $J$ be its set of distinct indices, $t=|J|\le r_n$, and let $g$ be its decoded predictor. Conditional on the examples indexed by $J$, the predictor $g$ is fixed and the other $n-t$ examples are i.i.d.\ from $\D$. By Hoeffding's inequality \cite{hoeffding1963probability}, for every $u>0$,
\[
 \Pr\Bigl[L_\D(g)>
 \frac1{n-t}\sum_{i\notin J}\ind\{g(x_i)\notin A_i\}+u
 \Bigm|(x_i,A_i)_{i\in J}\Bigr]
 \le e^{-2(n-t)u^2}.
\]
Since $\widehat L_\sigma(g)$ is the average of the mean on $J$ and the mean outside $J$ with weights $t/n$ and $(n-t)/n$, the mean outside $J$ exceeds $\widehat L_\sigma(g)$ by at most $t/n$. Take
\[
 u=\sqrt{\frac{r_n\log n+\log(4/\delta)}{2(n-r_n)}},
\]
so that $e^{-2(n-t)u^2}\le\delta/(4n^{r_n})$; the same bound holds for the fallback, with $J=\emptyset$. A union bound over the candidates shows that, with probability at least $1-\delta/2$, every candidate $g$ satisfies
\[
 L_\D(g)\le\widehat L_\sigma(g)+\frac{r_n}n+u .
\]
By Hoeffding's inequality again, with probability at least $1-\delta/2$,
\[
 \frac1n\sum_{i=1}^n\ind\{h(x_i)\ne A_i\}
 \le\Pr_{(x,A)\sim\D}[h(x)\ne A]
       +\sqrt{\frac{\log(2/\delta)}{2n}}.
\]
Combining the last two displays with \eqref{cf:comparator} gives \eqref{cf:agnostic-compression} with $\Pr_{(x,A)\sim\D}[h(x)\ne A]$ in place of $\Lset_\D(\H)$. Applying this to some $h\in\H$ with $\Pr_{(x,A)\sim\D}[h(x)\ne A]\le\Lset_\D(\H)+1/j$ shows that the event on which $L_\D(\widehat g)$ exceeds the right side of \eqref{cf:agnostic-compression} plus $1/j$ has probability at most $\delta$. As $j\to\infty$ these events increase to the event on which \eqref{cf:agnostic-compression} fails, so it too has probability at most $\delta$.
\end{proof}

\begin{proof}[Proof of Theorem~\ref{thm:countable-full}, agnostic part]
Assume $d=\dpack(\H)<\infty$. By Lemma~\ref{def:compression}, the decoders $\Dec_n$ of Section~\ref{cf:sec-compression} satisfy the assumption of Lemma~\ref{lem:agnostic-reconstruction}, with $r_n\le c(d+1)\log(2n)$. Take
\[
 n=\Bigl\lceil c\,\frac{B}{\eps^2}\log^2\!\Bigl(\frac{2B}{\eps}\Bigr)\Bigr\rceil .
\]
For a large enough $c$, depending only on $\ell$ and $\gamma$, we have $n>2r_n$, and the three excess terms in \eqref{cf:agnostic-compression} sum to at most $\eps$; as in the realizable case, the dominant term $\sqrt{r_n\log n/n}$, with $r_n\log n$ of order $(d+1)\log^2(2n)$, is what requires the factor $\log^2(2B/\eps)/\eps^2$. This proves \eqref{eq:full-agnostic-rate} and (iii)$\Rightarrow$(ii). Finally, agnostic learnability implies realizable learnability (Section~\ref{sec:prelim-models}), which proves (ii)$\Rightarrow$(i).
\end{proof}

\begin{remark}\label{rem:fixed-gamma}
By Theorem~\ref{thm:countable-full}, finiteness of $\dpack(\H)$ for one $\gamma$ in the range \eqref{eq:gamma-range} is equivalent to finiteness for every such $\gamma$. It is not equivalent to finiteness of $\sup_\gamma\dpack(\H)$, as the following learnable class shows. Let $\ell=1$, let $\X$ be infinite, and let $\H$ consist of the function that equals $\{0\}$ everywhere and the functions that equal $\{1\}$ at one input and $\{0\}$ elsewhere. The class is learnable: predict $1$ at an input seen with label set $\{1\}$, and $0$ elsewhere. On the other hand, $\dDS(\H)=1$ and
\[
 \dpackr{\rho}(\H)=\lfloor2/\rho\rfloor-1\qquad(0<\rho\le1).
\]
Indeed, on $n$ inputs a block in direction $i$ must pair a copy of the zero pattern with a copy of the pattern that is $\{1\}$ only at $i$. If these patterns have proportions $\mu_0,\mu_1,\ldots,\mu_n$ in the list, then coverage $\rho$ requires $\mu_0,\mu_i\ge\rho/2$ for all $i$, and thus $(n+1)\rho/2\le1$. Conversely, the uniform list of the $n+1$ patterns has coverage $2/(n+1)$ in every direction. This construction remains valid when the same class is viewed with a larger bound $\ell\ge2$, so the supremum is infinite in that case as well.
\end{remark}

\section{One-Correct-Label Learning with Countably Many Labels}\label{sec:countable-one}

We now return to one-correct-label feedback and drop the assumption that $\Y$ is finite; as everywhere in the paper, $\Y$ is countable. The main result of this section shows that one-correct-label learnability of $\H$ is still equivalent to full-set learnability of $\cI_{\ell+1}(\H)$, and hence, by Theorem~\ref{thm:countable-full}, to finiteness of $\dpack(\cI_{\ell+1}(\H))$.

\begin{theorem}\label{thm:countable-one}
Let $\H\subseteq(\cA\cup\{\star\})^{\X}$ be nonempty, and fix $\gamma$ as in \eqref{eq:gamma-range}. The following are equivalent:
\begin{enumerate}[(i)]
\item $\H$ is realizably one-correct-label PAC learnable;
\item $\cI_{\ell+1}(\H)$ is realizably full-set PAC learnable;
\item $\dpack(\cI_{\ell+1}(\H))<\infty$.
\end{enumerate}
These conditions imply $\dN(\cI_{\ell+1}(\H))<\infty$. Moreover, if $p=\dpack(\cI_{\ell+1}(\H))<\infty$, then
\begin{equation}\label{co:pack-rate}
 \mone_{\H}(\eps,\delta)
 \le\left\lceil c\,(p+1)\,\eps^{-\ell}\log^2(2/\eps)\log(2/\delta)\right\rceil,
\end{equation}
where $c$ depends only on $\ell$ and $\gamma$. If $p=0$, then $\mone_\H(\eps,\delta)=0$.
\end{theorem}

Unlike the finite-label rate \eqref{eq:finite-one-rate}, the bound \eqref{co:pack-rate} does not depend on $|\Y|$, and it is linear in $p+1$. Call a full-set learner for $\cI_{\ell+1}(\H)$ \emph{weak} if its expected loss is below $1/(\ell+1)$. Theorem~\ref{thm:countable-one} is a consequence of the following conversion, which turns any weak learner into a one-correct-label learner whose sample size is linear in that of the weak learner.

\begin{theorem}[Conversion of a weak full-set learner]\label{thm:feedback}
Let $\H\subseteq(\cA\cup\{\star\})^{\X}$ be nonempty. Let $M\ge1$ and $0\le\alpha<1/(\ell+1)$, and suppose there is a full-set learner for $\cI_{\ell+1}(\H)$ that uses $M$ examples and whose randomized prediction has expected loss at most $\alpha$ for every target $g\in\cI_{\ell+1}(\H)$ and every $P$ with $P(\supp(g))=1$. Then for every $N\ge2M$ there is a one-correct-label learner that uses $N-1$ examples and whose randomized prediction $\widehat p$ satisfies, for every $h^\star\in\H$, every $P$ with $P(\supp(h^\star))=1$, and every law $\D$ on $\X\times\Y$ with marginal $P$ and $\Pr_{(x,y)\sim\D}[y\in h^\star(x)]=1$,
\begin{equation}\label{co:feedback-expected}
 \E\bigl[L_{P,h^\star}(\widehat p)\bigr]
 \le\min\left\{1,\ c\left[\frac{M\log^2(eN/M)}{N}\right]^{1/\ell}\right\}.
\end{equation}
Consequently,
\begin{equation}\label{co:feedback-pac}
 \mone_{\H}(\eps,\delta)
 \le\left\lceil c\,M\eps^{-\ell}\log^2(2/\eps)\log(2/\delta)\right\rceil.
\end{equation}
Here $c$ depends only on $\ell$ and $\alpha$.
\end{theorem}

The main idea is as follows. By Theorem~\ref{thm:family-conversion}, it suffices to construct prediction families with the intersection property whose size bound satisfies $\log B_N=O(M\log^2(eN/M))$. The single cover of $\cI_{\ell+1}(\H)$ used in Section~\ref{sec:finite-one} has size growing with $|\Y|$, so we build a separate family $F_{h,S}$ for each hypothesis $h$ instead.

These families come from short descriptions. A short run of the boosting procedure of Section~\ref{sec:countable-full}, followed by a few corrections, describes each $g\in\cI_{\ell+1}(\H)$ on a pool of $N$ inputs by $O(M\log(eN/M))$ pairs, each consisting of a position of the pool and a set of labels. The family $F_{h,S}$ consists of the decoded predictors of all descriptions whose stored sets are nonempty subsets of the corresponding sets $h(x_i)$; there are at most $2^\ell-1$ such subsets at each position, whatever the size of $\Y$. The description of $g=h_1\cap\cdots\cap h_{\ell+1}$ stores only the sets $g(x_i)$, which are subsets of every $h_a(x_i)$, and this gives the intersection property. Finally, the count depends on $N$ only through $N/M$, so the sample size is linear in $M$.

\subsection{Short Descriptions of Partial Intersections}\label{co:sec-descriptions}

We first encode each $g\in\cI_{\ell+1}(\H)$, on a pool of $N$ inputs, by a short message, and decode the message to a predictor that is valid at every position of the pool where $g$ is defined. Throughout, $M$, $\alpha$ and the learner are those of Theorem~\ref{thm:feedback}, and we apply the construction preceding Lemma~\ref{lem:boosted-reconstruction} with $\H_{\mathrm{aux}}=\cI_{\ell+1}(\H)$. This gives a symmetric rule $\sigma\mapsto p_\sigma$ and the constants $\eta,\zeta$ of \eqref{eq:boost-c}, which depend only on $\ell$ and $\alpha$. Recall that the lemma gives, after $T$ rounds, $T$ multisets of $M$ examples whose averaged prediction is wrong at no more than $Ne^{-\zeta T}$ positions of the pool, and that correcting these positions gives a predictor that is valid wherever the target is defined. Fix $N\ge2M$ and put
\begin{equation}\label{co:rounds}
 T=\left\lceil\zeta^{-1}\log(N/M)\right\rceil.
\end{equation}
Since $N\ge2M$, we have $T\ge1$ and $Ne^{-\zeta T}\le M$, so at most $M$ corrections are needed.

\paragraph{Messages and the decoder.}
Fix an ordered pool $S=(x_1,\ldots,x_N)$. A \emph{message} is either the empty message or a tuple $(z_1,\ldots,z_T;Z)$, where
\begin{itemize}
\item each $z_t$ is a multiset of $M$ pairs $(i,A)\in[N]\times\cA$, and
\item $Z$ is a set of at most $M$ such pairs with distinct positions $i$.
\end{itemize}
The decoder $\MsgDec_S$ maps the empty message to a fixed fallback predictor. It decodes a tuple $(z_1,\ldots,z_T;Z)$ in two steps. First, replacing every pair $(i,A)$ of $z_t$ by the example $(x_i,A)$ gives a multiset $\sigma_t$ of $M$ examples, and $f_0(x)$ is a most probable label of $T^{-1}\sum_{t}p_{\sigma_t}(\cdot\mid x)$. Second, the pairs of $Z$ override $f_0$: if $x=x_i$ for some $(i,A)\in Z$, the decoder takes the least such $i$ and returns the least label of its set $A$ in the fixed enumeration of $\Y$; otherwise it returns $f_0(x)$. Thus every message decodes to a total predictor.

\paragraph{The message of a partial intersection.}
Let $g\in\cI_{\ell+1}(\H)$ and $D=\{i\in[N]:x_i\in\supp(g)\}$. If $D$ is empty, the \emph{message of $g$} is the empty message. Otherwise Lemma~\ref{lem:boosted-reconstruction} with $T$ rounds gives multisets $\sigma_1,\ldots,\sigma_T$ of $M$ examples $(x_i,g(x_i))$, $i\in D$, and a set $Q\subseteq D$ of at most $Ne^{-\zeta T}\le M$ positions, such that the corrected predictor $f$ of that lemma is valid at every position of $D$. The message writes these examples in terms of positions: $z_t$ is obtained from $\sigma_t$ by replacing each example $(x,g(x))$ by the pair $(i,g(x_i))$, where $i$ is the least position in $D$ with $x_i=x$. With $Z=\{(i,g(x_i)):i\in Q\}$, the message of $g$ is $(z_1,\ldots,z_T;Z)$.

This message decodes to $f$. Indeed, the decoder turns each pair $(i,g(x_i))$ of $z_t$ back into the example $(x_i,g(x_i))=(x,g(x))$, so it recovers $\sigma_1,\ldots,\sigma_T$ and hence $f_0$. Since $g$ is a function of the input, all positions of $Q$ with a given input $x_i$ carry the same set $g(x_i)$, so at such an input the decoder returns the least label of $g(x_i)$, as $f$ does; at every other input both predictors equal $f_0$. We record for later that every pair $(i,A)$ in the message of $g$ satisfies $A=g(x_i)$.

\subsection{Message Families of Single Hypotheses}\label{co:sec-families}

Let $h\in\H$ be defined on the pool, that is, $x_i\in\supp(h)$ for all $i\in[N]$. Let $\cM_{h,N}$ consist of the empty message and of all messages $(z_1,\ldots,z_T;Z)$ in which every pair $(i,A)$ satisfies $\emptyset\ne A\subseteq h(x_i)$. Since Definition~\ref{def:prediction-families} uses functions on $\supp(S)$, we restrict the decoded predictors to the pool and set
\[
 F_{h,S}=\bigl\{\MsgDec_S(z)|_{\supp(S)}:z\in\cM_{h,N}\bigr\}.
\]
Allowing subsets of $h(x_i)$, rather than $h(x_i)$ itself, is what makes the message of an intersection admissible for every constituent hypothesis.

Put $b_\ell=2^\ell-1$. Since $|h(x_i)|\le\ell$, at most $b_\ell$ sets $A$ are admissible at each position, so at most $Nb_\ell$ pairs are admissible. Counting the multisets of $M$ admissible pairs for each $z_t$, and the sets of $j\le M$ admissible pairs with distinct positions for $Z$, gives
\begin{equation}\label{co:candidate-count}
 |F_{h,S}|\le|\cM_{h,N}|\le B_N
 :=1+\binom{Nb_\ell+M-1}{M}^{T}\sum_{j=0}^M\binom Nj b_\ell^j .
\end{equation}
The elementary estimates $\binom{Nb_\ell+M-1}{M}\le(e(b_\ell+1)N/M)^M$ and $\sum_{j\le M}\binom Nj b_\ell^j\le\sum_{j\le M}\binom{Nb_\ell}{j}\le(eNb_\ell/M)^M$, together with $T\le c\log(eN/M)$ from \eqref{co:rounds} and $N\ge2M$, give
\begin{equation}\label{co:log-count}
 \log B_N\le c\,M\log^2(eN/M).
\end{equation}

\subsection{Proof of the Conversion Theorem}\label{co:sec-proof}

\begin{proof}[Proof of Theorem~\ref{thm:feedback}]
Let $S$ be a multiset of $N$ inputs, and list its elements, with multiplicity, in the order of a fixed enumeration of the countable set $\X$. This gives an ordered pool determined by $S$, and we use the families $F_{h,S}$ of Section~\ref{co:sec-families} for this pool. By \eqref{co:candidate-count}, they have size bound $B_N$.

\paragraph{The intersection property.}
Let $h_1,\ldots,h_{\ell+1}\in\H$ be defined on $S$, and let $B$ be the set of $x\in\supp(S)$ with $\bigcap_ah_a(x)\ne\emptyset$. Let $g(x)=\bigcap_ah_a(x)$ for $x\in B$ and $g(x)=\star$ otherwise. Then $g\in\cI_{\ell+1}(\H)$, since this class is closed under domain restrictions. Let $z$ be the message of $g$. Every pair $(i,A)$ in $z$ has $A=g(x_i)$, a nonempty subset of every $h_a(x_i)$, so $z\in\cM_{h_a,N}$ for every $a$ (this is trivial if $z$ is empty). Hence $f=\MsgDec_S(z)|_{\supp(S)}$ lies in $\bigcap_aF_{h_a,S}$, and $f(x)\in g(x)=\bigcap_ah_a(x)$ for every $x\in B$. This is the intersection property of Definition~\ref{def:prediction-families}.

\paragraph{Expected loss.}
By Theorem~\ref{thm:family-conversion} there is a learner with $N-1$ examples whose expected loss is at most
\[
 \min\left\{1,\;2\left(\frac{(\ell+1)(\log B_N+\log N)}{\lfloor N/2\rfloor}\right)^{1/\ell}+\frac1N\right\}.
\]
We have $\log N\le M\log(eN/M)\le M\log^2(eN/M)$, since $z\mapsto z\log(eN/z)$ is nondecreasing on $[1,N]$ and exceeds $\log N$ at $z=1$, and since $\log(eN/M)\ge1$. Together with \eqref{co:log-count} and $\lfloor N/2\rfloor\ge N/3$, this bounds the term with the root by $c\,[M\log^2(eN/M)/N]^{1/\ell}$. Since $M\log^2(eN/M)/N\ge1/N$, the additive term $1/N$ is absorbed by increasing $c$. This proves \eqref{co:feedback-expected}.

\paragraph{Confidence.}
Lemma~\ref{lem:amplification} needs expected loss at most $\eps/[32(\ell+1)^2]$, and we choose $N$ to achieve this. Let $a=\log(2/\eps)$, and use $n=\lceil C M\eps^{-\ell}a^2\rceil$ examples, so that $N=n+1$. For $C\ge6$ we have $N/M\ge x_0:=C\eps^{-\ell}a^2\ge e$, and in particular $N\ge2M$. Moreover, $\log(ex_0)=1+\log C+\ell\log(1/\eps)+2\log a\le c'(1+\log C)\,a$, where $c'$ depends only on $\ell$. Since $x\mapsto\log^2(ex)/x$ is nonincreasing for $x\ge e$, the expression in brackets in \eqref{co:feedback-expected} therefore satisfies
\[
 \frac{M\log^2(eN/M)}{N}\le\frac{\log^2(ex_0)}{x_0}\le\frac{c'^2(1+\log C)^2\eps^{\ell}}{C}.
\]
Choosing $C$ large enough in terms of $\ell$ and $\alpha$ therefore makes the expected loss of the $n$-example learner at most $\eps/[32(\ell+1)^2]$. Lemma~\ref{lem:amplification} now gives a learner with $n\lceil24(\ell+1)\log(2/\delta)\rceil$ examples, which proves \eqref{co:feedback-pac}.
\end{proof}

\subsection{The Feedback Equivalence and the Packed Rate}\label{co:sec-equivalence}

Theorem~\ref{thm:feedback} needs only a black-box weak learner, so it already gives the equivalence of the two feedback models. For the packed rate we feed it the learner of Lemma~\ref{lem:packed-weak}, which is weak already with $O(p+1)$ examples.

\begin{corollary}[One-correct-label versus full-set feedback]\label{cor:blackbox}
Let $\H\subseteq(\cA\cup\{\star\})^{\X}$ be nonempty. Then $\H$ is realizably one-correct-label PAC learnable if and only if $\cI_{\ell+1}(\H)$ is realizably full-set PAC learnable. Moreover, let
\[
 M=\max\left\{1,\ \mfull_{\cI_{\ell+1}(\H)}\left(\frac1{8(\ell+1)},\frac1{8(\ell+1)}\right)\right\}.
\]
If $M<\infty$, then
\begin{equation}\label{co:blackbox-rate}
 \mone_{\H}(\eps,\delta)\le c\,M\eps^{-\ell}\log^2(2/\eps)\log(2/\delta),
\end{equation}
where $c$ depends only on $\ell$.
\end{corollary}

\begin{proof}
By Lemma~\ref{lem:common-demonstrations} with $R=\ell+1$, one-correct-label learnability of $\H$ implies full-set learnability of $\cI_{\ell+1}(\H)$. Conversely, if $M<\infty$, a full-set PAC learner for $\cI_{\ell+1}(\H)$ with accuracy and confidence parameters $1/[8(\ell+1)]$ uses $M$ examples and, by \eqref{eq:pac-expect}, has expected loss at most $1/[4(\ell+1)]$. Theorem~\ref{thm:feedback} with $\alpha=1/[4(\ell+1)]$ gives \eqref{co:blackbox-rate}, and in particular one-correct-label learnability.
\end{proof}

\begin{proof}[Proof of Theorem~\ref{thm:countable-one}]
Corollary~\ref{cor:blackbox} shows that (i) and (ii) are equivalent, and Theorem~\ref{thm:countable-full}, applied to the class $\cI_{\ell+1}(\H)$, shows that (ii) and (iii) are equivalent. Since $\dN\le\dpack$ by Lemma~\ref{lem:orders}, (iii) implies $\dN(\cI_{\ell+1}(\H))<\infty$.

For the rate, let $p=\dpack(\cI_{\ell+1}(\H))<\infty$ and put
\[
 \alpha=\frac12\Bigl(\gamma+\frac1{\ell+1}\Bigr),\qquad
 M=\max\left\{1,\left\lceil\frac{p}{\alpha-\gamma}\right\rceil\right\}.
\]
The learner of Lemma~\ref{lem:packed-weak} for $\cI_{\ell+1}(\H)$ with $M$ examples has expected loss at most $\gamma+p/(M+1)\le\alpha<1/(\ell+1)$. Since $M\le c\,(p+1)$ and $\alpha$ depends only on $\ell$ and $\gamma$, \eqref{co:feedback-pac} gives \eqref{co:pack-rate}.

Finally, let $p=0$. Then $\dN(\cI_{\ell+1}(\H))=0$ by Lemma~\ref{lem:orders}. The argument for the case $d=0$ in the proof of Theorem~\ref{thm:finite-one} uses only this fact and Lemma~\ref{lem:small}, so it applies to every countable label space. It shows that at every input $x$ the sets $h(x)$, over the $h\in\H$ defined at $x$, share a label. Predicting the first such label requires no examples and is valid on $\supp(h^\star)$ for every $h^\star\in\H$, so $\mone_\H(\eps,\delta)=0$.
\end{proof}

\begin{corollary}[Every fixed intersection depth]\label{cor:all-depths}
If $\H$ is realizably one-correct-label PAC learnable, then $\cI_R(\H)$ is realizably full-set PAC learnable for every $R\ge1$. For $R\ge\ell+1$ the converse also holds. Moreover, if $p=\dpack(\cI_{\ell+1}(\H))<\infty$, then for every $R\ge1$ and $\beta>0$ there is a full-set learner for $\cI_R(\H)$ with expected loss at most $\beta$ that uses at most $c\,(p+1)$ examples, where $c$ depends only on $\ell$, $R$, $\gamma$ and $\beta$.
\end{corollary}

\begin{proof}
The first assertion is Lemma~\ref{lem:common-demonstrations}. For $R\ge\ell+1$ we have $\cI_{\ell+1}(\H)\subseteq\cI_R(\H)$, so a full-set learner for $\cI_R(\H)$ is also one for $\cI_{\ell+1}(\H)$, and Corollary~\ref{cor:blackbox} gives the converse. For the last assertion we may assume $\beta<1$. By Lemma~\ref{lem:common-demonstrations} and by \eqref{co:pack-rate} with $\eps=\delta=\beta/(2R)$, we have $\mfull_{\cI_R(\H)}(\beta/2,\beta/2)\le\mone_\H(\beta/(2R),\beta/(2R))\le c'(p+1)$, where $c'$ depends only on $\ell$, $R$, $\gamma$ and $\beta$. The corresponding learner has expected loss at most $\beta$ by \eqref{eq:pac-expect}.
\end{proof}

\begin{remark}[Intersection depth and the accuracy exponent]\label{rem:depth}
The characterization involves intersections of $\ell+1$ hypotheses, while the rate has exponent $\ell$. Both come from the same step in the proof of Theorem~\ref{thm:family-conversion}: the correct-label set of a first candidate has at most $\ell$ labels, and each of $\ell$ further independent candidates may exclude one of them. This gives the $\ell$th power in \eqref{eq:red-power}, and hence the factor $\eps^{-\ell}$.
\end{remark}

\subsection{Joint DS and Disjoint-Neighbor Bounds}\label{co:sec-ds}

We end the section with sufficient conditions in terms of the joint DS dimension and the disjoint-neighbor dimension of $\cI_{\ell+1}(\H)$. They follow from Theorem~\ref{thm:countable-one} and the following comparison, proved in Appendix~\ref{app:comparison}, which bounds the packed dimension of $\cI_{\ell+1}(\H)$ by either of these dimensions up to a logarithmic factor.

\begin{theorem}[Comparison for partial intersection classes]\label{thm:comparison}
Let $\H\subseteq(\cA\cup\{\star\})^{\X}$ be nonempty, and fix $0<\rho\le1$. Let $d$ be either $\dDS(\cI_{\ell+1}(\H))$ or $\dpair(\cI_{\ell+1}(\H))$. There is a constant $c$, depending only on $\ell$ and $\rho$, such that if $d<\infty$, then
\begin{equation}\label{co:compare}
 \dpackr{\rho}(\cI_{\ell+1}(\H))\le c\,d\log(d+2).
\end{equation}
The bound holds for arbitrary sets $\X$ and $\Y$, countable or not.
\end{theorem}

\begin{corollary}[A joint DS or disjoint-neighbor bound]\label{cor:main-ds}
Let $d$ be either $\dDS(\cI_{\ell+1}(\H))$ or $\dpair(\cI_{\ell+1}(\H))$. If $d<\infty$, then $\H$ is realizably one-correct-label PAC learnable and
\begin{equation}\label{co:ds-rate}
 \mone_{\H}(\eps,\delta)\le
 \left\lceil c\,(d+1)\log(d+2)\,\eps^{-\ell}\log^2(2/\eps)\log(2/\delta)\right\rceil,
\end{equation}
where $c$ depends only on $\ell$. If $d=0$, then $\mone_\H(\eps,\delta)=0$.
\end{corollary}

\begin{proof}
Put $\rho_0=1/[8(\ell+1)]$, which satisfies \eqref{eq:gamma-range}. By Theorem~\ref{thm:comparison} with $\rho=\rho_0$,
\[
 p_0=\dpackr{\rho_0}(\cI_{\ell+1}(\H))\le c\,d\log(d+2).
\]
Hence $p_0+1\le c\,(d+1)\log(d+2)$ after enlarging $c$, and Theorem~\ref{thm:countable-one} with $\gamma=\rho_0$ gives \eqref{co:pack-rate} with $p=p_0$. This proves \eqref{co:ds-rate}. If $d=0$, then $p_0=0$, and Theorem~\ref{thm:countable-one} gives $\mone_\H(\eps,\delta)=0$.
\end{proof}

\section{Agnostic Learning from One Correct Label}\label{sec:agnostic}

Recall the agnostic model from Section~\ref{sec:prelim-models}. Examples $(x,A)$ are drawn from an arbitrary distribution $\D$ on $\X\times\cA$, but the learner observes only $x$ and a label $y\in A$ chosen by the \emph{demonstrator}, a conditional law of $y$ given $(x,A)$ that is supported on $A$. The learner competes with the benchmark $\Lset_\D(\H)$ of \eqref{eq:agnostic}.

In the realizable setting, learnability from one correct label is governed by the intersection class $\cI_{\ell+1}(\H)$ (Theorems~\ref{thm:B} and~\ref{thm:E}). In the agnostic setting it is not, because a demonstration does not reveal the set it came from: Section~\ref{sec:agnostic-one} gives a class whose intersection class has all four dimensions equal to one and which is still not learnable. For nondegenerate classes we show that learnability is characterized by a local compatibility condition together with finiteness of the DS dimension of the \emph{collapsed class}, an ordinary multiclass class obtained by replacing every correct-label set with two or more labels by one new label. For $\ell=1$ this recovers the characterization of agnostic learnability of partial concept classes \cite{alon2021theory} (Remark~\ref{agn:rem:binary}).

We state the characterization in Section~\ref{agn:sec:statement} and show in Section~\ref{sec:agnostic-one} that nondegeneracy cannot be dropped; after some notation (Section~\ref{agn:sec:regions}), we prove necessity (Section~\ref{agn:sec:necessity}), give learners for the two regions of the input space (Sections~\ref{agn:sec:pure} and~\ref{agn:sec:mixed}), and glue them together (Section~\ref{agn:sec:proof}).

\subsection{Nondegenerate Classes and the Characterization}\label{agn:sec:statement}

In binary classification, the two constant hypotheses disagree at every input. Our standing assumption asks for two hypotheses that play the same role.

\begin{definition}[Nondegenerate classes]\label{agn:def:nondegenerate}
A class $\H$ is \emph{nondegenerate} if it contains two total hypotheses $g_0,g_1$ with singleton values that differ at every input, that is, there are $b_0,b_1:\X\to\Y$ with
\[
 g_j(x)=\{b_j(x)\}\quad(j\in\{0,1\})
 \qquad\text{and}\qquad
 b_0(x)\neq b_1(x)\qquad(x\in\X).
\]
Otherwise $\H$ is \emph{degenerate}.
\end{definition}

For instance, every class that contains two distinct constant singleton hypotheses is nondegenerate, regardless of its other members, which may be partial.

The characterization uses the following multiclass class. Fix a new label $\diamond\notin\Y$. For $h\in\H$, define
\begin{equation}\label{agn:eq:collapse}
 \widetilde h(x)=
 \begin{cases}
 a, & h(x)=\{a\},\\
 \diamond, & h(x)\neq\star\text{ and }|h(x)|\geq2,\\
 \star, & h(x)=\star,
 \end{cases}
 \qquad\qquad
 \widetilde\H=\{\widetilde h:h\in\H\}.
\end{equation}
The collapsed class $\widetilde\H$ keeps singleton and undefined values and turns every correct-label set with two or more labels into the label $\diamond$, which counts as defined. Reading each value $a$ as the singleton $\{a\}$, we view $\widetilde\H$ as a class with $\ell=1$ over the countable label space $\Y\cup\{\diamond\}$, with Natarajan and DS dimensions $\dN(\widetilde\H)$ and $\dDS(\widetilde\H)$ as in Section~\ref{sec:prelim-dimensions}.

The collapse forgets which labels a large correct-label set contains. The next condition makes this information unnecessary.

\begin{definition}[Local compatibility]\label{agn:def:lc}
 A class $\H$ is \emph{locally compatible} if all $h, h' \in \H$ and $x \in \supp(h)\cap\supp(h')$ such that $|h(x)| = 1$ and $|h'(x)| \geq 2$ satisfy $h(x) \subseteq h'(x)$.

\end{definition}

In words, at every input, each correct-label set with two or more labels contains all singleton values of the class at that input. The condition is needed because demonstrations do not tell the learner whether a label came from a singleton or from a larger set. Under local compatibility, a singleton value of the class at $x$ is a safe prediction whenever the hidden set at $x$ is any value of the class with two or more labels, so the learner never needs to know which one.

Our main result in this section is the following characterization.

\begin{theorem}[Agnostic learning from one correct label]\label{agn:thm:main}
Let $\X$ and $\Y$ be countable, and let $\H\subseteq(\cA\cup\{\star\})^{\X}$ be nondegenerate and locally compatible. Then $\H$ is agnostically PAC learnable from one correct label if and only if
\[
 \dDS(\widetilde\H)<\infty.
\]
\end{theorem}

Assuming local compatibility costs nothing, because it is necessary for every class.

\begin{proposition}[Local compatibility is necessary]\label{agn:prop:lc}
Let $\H$ be a class that is not locally compatible. For every $m$ and every learner that uses $m$ one-correct-label examples, there are a distribution $\D$ on $\X\times\cA$ and a demonstrator such that the learner's output $\widehat g$ satisfies
\[
 \E\bigl[L_\D(\widehat g)\bigr]\geq\Lset_\D(\H)+\frac1{15}.
\]
In particular, $\H$ is not agnostically PAC learnable from one correct label.
\end{proposition}

Together, Theorem~\ref{agn:thm:main} and Proposition~\ref{agn:prop:lc} give the first part of Theorem~\ref{thm:F}: a nondegenerate class is agnostically PAC learnable from one correct label if and only if it is locally compatible and $\dDS(\widetilde\H)<\infty$.

Both directions of Theorem~\ref{agn:thm:main} treat two regions of the input space separately. The \emph{mixed region} $\X_{\mathrm{mix}}$ consists of the inputs at which some hypothesis has a correct-label set with at least two labels, and the \emph{pure region} $\X_{\mathrm{pure}}$ is the rest. On the pure region every defined value is a singleton, and an agnostic multiclass learner suffices (Section~\ref{agn:sec:pure}). On the mixed region the learner predicts only singleton values of the class, which local compatibility makes safe, and runs specialists \cite{freund1997using} over a finite cover of the collapsed class (Section~\ref{agn:sec:mixed}). Section~\ref{agn:sec:proof} glues the two learners, and necessity follows from a multiclass lower bound on the pure region and an indistinguishability argument on the mixed region (Section~\ref{agn:sec:necessity}).

\paragraph{A binary reformulation.}
When every input belongs to the mixed region, the condition can be stated with a partial binary class on $\X\times\Y$. For $h\in\H$ define
\[
 J_h(x,a)=
 \begin{cases}
 1, & h(x)=\{a\},\\
 0, & h(x)\neq\star\text{ and }h(x)\neq\{a\},\\
 \star, & h(x)=\star,
 \end{cases}
 \qquad\qquad
 \mathcal J=\{J_h:h\in\H\}.
\]
As for partial concept classes \cite{alon2021theory}, $\mathrm{VC}(\mathcal J)$ is the largest $n$ for which some $n$ points of $\X\times\Y$ are shattered by the members of $\mathcal J$ that are defined at all of them.

\begin{corollary}[A VC criterion]\label{agn:cor:vc}
Let $\X$ and $\Y$ be countable, let $\H$ be nondegenerate and locally compatible, and suppose that at every input some member of $\H$ has a correct-label set with at least two labels. Then $\H$ is agnostically PAC learnable from one correct label if and only if $\mathrm{VC}(\mathcal J)<\infty$.
\end{corollary}

The proof is in Section~\ref{agn:sec:proof}. Undefined values must stay undefined in $J_h$ (Remark~\ref{agn:rem:zero}).

\subsection{The Degenerate Obstruction}\label{sec:agnostic-one}

We now show that nondegeneracy cannot be dropped from Theorem~\ref{agn:thm:main}, and that the dimensions of $\cI_{\ell+1}(\H)$ do not determine agnostic learnability. The reason is statistical: different distributions of correct-label sets can induce the same law of demonstrations, and hence the same law of the learner's output, while a prediction that is good under one of them is bad under another. The following quantitative form of the second part of Theorem~\ref{thm:F} adapts the indistinguishability argument of \cite{pour2026learning}, given there for online learning with multiple correct answers, to the i.i.d.\ setting and the benchmark $\Lset_\D$ of \eqref{eq:agnostic}.

\begin{theorem}[A degenerate class that is not agnostically learnable]\label{thm:agnostic-one-impossible}
Let $\ell=2$. There is a class $\H$ on a one-point input space with $|\Y|=3$, in which every correct-label set has size two, that is not agnostically PAC learnable from one correct label: for every sample size and every learner there are a law $\D$ on $\X\times\cA$ and a demonstrator such that the learner's output $\widehat g$ satisfies
\[
 \E\bigl[L_\D(\widehat g)\bigr]\ge\Lset_\D(\H)+\frac19
 \qquad\text{and}\qquad
 \Pr\Bigl[L_\D(\widehat g)>\Lset_\D(\H)+\frac1{18}\Bigr]\ge\frac15.
\]
Moreover, $\H$ is degenerate and locally compatible, $\dDS(\widetilde\H)=0$, and
\[
 \dN(\cI_3(\H))=\dDS(\cI_3(\H))=\dpair(\cI_3(\H))=\dpackr{\rho}(\cI_3(\H))=1
 \qquad(0<\rho\le1).
\]
\end{theorem}

\begin{proof}
Let $\X=\{x\}$, $\Y=\{1,2,3\}$, and $\H=\{h_1,h_2,h_3\}$ with $h_i(x)=\Y\setminus\{i\}$. We construct three distributions under which the demonstrated label is uniform on $\Y$, while under the $i$th distribution the label $i$ is the worst prediction. For $i\in[3]$, let $\D_i$ assign probability $2/3$ to $(x,\Y\setminus\{i\})$ and probability $1/3$ to $(x,\{i\})$, and let the demonstrated label be a uniformly random element of the sampled set $A$. Under every $\D_i$ the demonstrated label is indeed uniform: label $i$ is demonstrated with probability $1/3$, and each other label with probability $(2/3)(1/2)=1/3$.

Thus the law of the demonstrated sample, and hence of the learner's output, is the same under $\D_1$, $\D_2$ and $\D_3$. Let $p_j$ be the probability that the learner predicts $j$ at $x$, over the sample and the seed. Since $p_1+p_2+p_3=1$, there is $i^\star$ with $p_{i^\star}\ge1/3$. Under $\D_{i^\star}$, predicting $i^\star$ has loss $2/3$ and predicting either other label has loss $1/3$, so
\[
 \E\bigl[L_{\D_{i^\star}}(\widehat g)\bigr]
 =\frac23p_{i^\star}+\frac13(1-p_{i^\star})
 =\frac13+\frac{p_{i^\star}}3\ge\frac49.
\]
On the other hand, $h_{i^\star}$ disagrees with the sampled set exactly when $A=\{i^\star\}$, which has probability $1/3$, and the other two hypotheses always disagree with it. Hence $\Lset_{\D_{i^\star}}(\H)=1/3$, which proves the bound in expectation.

For the bound in probability, condition on the sample and the seed, and let $r$ be the probability that the resulting prediction rule assigns to $i^\star$ at $x$ (so $r\in\{0,1\}$ for a deterministic predictor). Its excess loss under $\D_{i^\star}$ is $Z=r/3\in[0,1/3]$, and $\E[Z]\ge1/9$ by the computation above. If $u=\Pr[Z>1/18]$, then
\[
 \frac19\le\E[Z]\le\frac1{18}(1-u)+\frac13u,
\]
so $u\ge1/5$.

It remains to compute the dimensions. For $i\ne j$ the set $h_i(x)\cap h_j(x)$ is a singleton, so $\cI_3(\H)$ contains the three hypotheses with singleton values at $x$. Any two of them form a witness of size one for each of the four dimensions; in the packed witness the single block covers both indices of the list, so this works for every $0<\rho\le1$. Since there is only one input, no witness has size larger than one. Finally, no member of $\H$ has a singleton value, so $\H$ is degenerate and locally compatible. The collapsed class consists of the single hypothesis $x\mapsto\diamond$. Its only pattern has no neighbor, so $\dDS(\widetilde\H)=0$.
\end{proof}

The obstruction is specific to the agnostic setting: in the realizable case every demonstrated label is correct at the only input, so predicting it has zero loss. Characterizing agnostic learnability from one correct label for degenerate classes is an interesting open direction.

\begin{remark}[The intersection class does not decide agnostic learnability]\label{rem:agnostic-c0}
Let $\H_0$ consist of the three hypotheses $x\mapsto\{j\}$ on $\X=\{x\}$, $\Y=\{1,2,3\}$. Every member of $\cI_3(\H_0)$ defined at $x$ belongs to $\H_0$, so, as in the proof above, all four dimensions of $\cI_3(\H_0)$ equal one. Yet $\H_0$ is agnostically learnable from one correct label, because the most frequently demonstrated label competes with the best constant. Indeed, let $t_j$ be the probability that $j$ is demonstrated and $s_j=\Pr[A=\{j\}]$. Every demonstrated label lies in $A$, so $t_j\ge s_j$, and the constant prediction $j$ has loss at most $1-t_j$. By Hoeffding's inequality and a union bound over the three labels, with probability at least $1-\delta$ the most frequent demonstrated label among $O(\eps^{-2}\log(1/\delta))$ examples has loss at most $1-\max_jt_j+\eps\le1-\max_js_j+\eps=\Lset_\D(\H_0)+\eps$. This agrees with Theorem~\ref{agn:thm:main}: the class $\H_0$ is nondegenerate and locally compatible, and $\dDS(\widetilde\H_0)=1$.
\end{remark}

\subsection{Regions and Local Alphabets}\label{agn:sec:regions}

This subsection collects the notation and the combinatorial facts used in both directions of the proof of Theorem~\ref{agn:thm:main}.

\paragraph{The two regions.}
We write $|h(x)|\ge2$ to mean that $h$ is defined at $x$ and $h(x)$ has at least two labels. The mixed region and the pure region are
\begin{equation}\label{agn:eq:regions}
 \X_{\mathrm{mix}}=\bigl\{x\in\X:\ |h(x)|\ge2\text{ for some }h\in\H\bigr\},
 \qquad
 \X_{\mathrm{pure}}=\X\setminus\X_{\mathrm{mix}}.
\end{equation}
On the pure region every defined value of every hypothesis is a singleton, so $\widetilde h(x)=a$ means $h(x)=\{a\}$ there, and the label $\diamond$ occurs only on the mixed region.

\paragraph{Local alphabets.}
The \emph{local alphabet} $\Y_x$ of an input $x$ is the set of labels $a$ with $h(x)=\{a\}$ for some $h\in\H$. If $\H$ is nondegenerate and locally compatible, $x\in\X_{\mathrm{mix}}$, and $|h(x)|\ge2$, then
\begin{equation}\label{agn:eq:local-alphabet}
 \{b_0(x),b_1(x)\}\subseteq\Y_x\subseteq h(x),
\end{equation}
where the first inclusion holds because $g_0,g_1\in\H$, and the second is local compatibility. Hence every label of $\Y_x$ is correct whenever the hidden set is a value of a class member with at least two labels, and $2\le|\Y_x|\le\ell$ on the mixed region. In particular, the collapsed values at a mixed input lie in $\Y_x\cup\{\diamond\}$, a set of at most $\ell+1$ labels; this bound uses only local compatibility.

\paragraph{Restrictions and DS witnesses.}
For $Z\subseteq\X$, the \emph{restriction} of a class to $Z$ consists of the hypotheses that agree with a member of the class on $Z$ and are undefined outside $Z$. The collapsed class and its restrictions have singleton values. For such a class we identify a set $\{a\}$ with the label $a$, so patterns are tuples of labels, and a DS witness is a pseudo-cube in the sense of \cite{brukhim2022characterization}: a finite set $V$ of patterns in which every $v\in V$ has, in every direction $i$, a neighbor in $V$ that differs from $v$ exactly at coordinate $i$. We need two simple facts about such classes.

\begin{lemma}[Splitting and size of DS witnesses]\label{agn:lem:ds-facts}
Let $\mathcal C$ be a class with singleton values, and let $\X=Z_1\cup Z_2$ with $Z_1\cap Z_2=\emptyset$. Let $\mathcal C_1$ and $\mathcal C_2$ be the restrictions of $\mathcal C$ to $Z_1$ and $Z_2$. Then
\[
 \max\{\dDS(\mathcal C_1),\dDS(\mathcal C_2)\}\le\dDS(\mathcal C)\le\dDS(\mathcal C_1)+\dDS(\mathcal C_2).
\]
Moreover, every DS witness of $\mathcal C$ on $n$ inputs has at least $2^n$ patterns.
\end{lemma}

\begin{proof}
The patterns of $\mathcal C_j$ on a sequence of inputs from $Z_j$ are exactly those of $\mathcal C$, which gives the first inequality. For the second, let $V$ be a DS witness of $\mathcal C$ on a sequence $S$, and let $I_j$ be the set of coordinates of the inputs of $S$ that lie in $Z_j$. By Lemma~\ref{lem:orders}, if $I_j$ is nonempty then $V|_{I_j}$ is a DS witness on the corresponding subsequence, and its patterns are patterns of $\mathcal C_j$. Hence $|I_j|\le\dDS(\mathcal C_j)$ for $j=1,2$.

For the size, we induct on $n$; a witness on zero inputs is nonempty. Split a witness $V$ on $n\ge1$ inputs according to the label at the last coordinate. Every nonempty part, projected to the first $n-1$ coordinates, is a DS witness, because neighbors in the first $n-1$ directions have the same last label and so lie in the same part. There are at least two nonempty parts, since every pattern has a neighbor in the last direction. By induction, $|V|\ge2\cdot2^{n-1}$.
\end{proof}

When only boundedly many labels occur at each input, as on the mixed region, Natarajan's lemma \cite{natarajan1989learning} bounds the number of patterns, and the Natarajan and DS dimensions are finite together.

\begin{lemma}[Bounded local alphabets]\label{agn:lem:natarajan}
Let $\mathcal C$ be a class with singleton values such that, at every input $x$, the defined values $c(x)$, $c\in\mathcal C$, lie in a set of at most $\ell+1$ labels. If $\dN(\mathcal C)\le d$, then
\[
 |\mathcal C|_S|\le n^d(\ell+1)^{2d}
\]
for every sequence $S$ of $n\ge1$ distinct inputs. Consequently, $\dN(\mathcal C)<\infty$ if and only if $\dDS(\mathcal C)<\infty$.
\end{lemma}

\begin{proof}
Map the possible labels at each input of $S$ injectively into $[\ell+1]$. This turns $\mathcal C|_S$ into a set of functions from $[n]$ to $[\ell+1]$, and its Natarajan dimension is at most $d$, since a product-shattered set of coordinates is a subsequence of $S$ product-shattered by $\mathcal C$. Natarajan's lemma \cite{natarajan1989learning} bounds the size of such a set by $n^d(\ell+1)^{2d}$. For the equivalence, $\dN(\mathcal C)\le\dDS(\mathcal C)$ by Lemma~\ref{lem:orders}. Conversely, if $\dN(\mathcal C)\le d<\infty$, a DS witness on $n$ inputs has at least $2^n$ patterns by Lemma~\ref{agn:lem:ds-facts}, so $2^n\le n^d(\ell+1)^{2d}$, which bounds $n$.
\end{proof}

\paragraph{Learning on a region.}
The proof of sufficiency learns on each region separately. For a set $R\subseteq\X$, we say that $\H$ is \emph{agnostically learnable on $R$} if the requirement of agnostic PAC learnability from one correct label holds for all distributions $\D$ on $\X\times\cA$ with $\D(R\times\cA)=1$. The benchmark is still $\Lset_\D(\H)$, which then depends only on the values of the hypotheses on $R$. We write $\D_{\mathrm{obs}}$ for the law of the observed pair $(x,y)$, where $(x,A)\sim\D$ and $y$ is the demonstrated label.

\subsection{Necessity}\label{agn:sec:necessity}

We now prove Proposition~\ref{agn:prop:lc} and the necessity part of Theorem~\ref{agn:thm:main}. The two indistinguishability arguments (Proposition~\ref{agn:prop:lc} and Lemma~\ref{agn:lem:anchored}) follow the pattern of Theorem~\ref{thm:agnostic-one-impossible}: two distributions of correct-label sets induce the same law of demonstrations, but no prediction is good under both. We first show that local compatibility is necessary. For the DS condition, Lemma~\ref{agn:lem:ds-facts} shows that if $\dDS(\widetilde\H)=\infty$, then $\widetilde\H$ has arbitrarily large DS witnesses within one of the two regions. On the pure region it gives a realizable lower bound (Lemma~\ref{agn:lem:ds-lower}); on the mixed region it gives large Natarajan cubes, which nondegeneracy and local compatibility turn into the anchored cubes defined below (Lemmas~\ref{agn:lem:anchored} and~\ref{agn:lem:extract}). Since excess errors are at most one, the first implication in \eqref{eq:pac-expect} shows that an $(\eps,\delta)$ agnostic guarantee gives expected excess error at most $\eps+\delta$, so each lower bound below rules out agnostic learnability.

\begin{proof}[Proof of Proposition~\ref{agn:prop:lc}]
We use one input at which a singleton value lies outside a large correct-label set. Since $\H$ is not locally compatible, there are $x\in\X$, $h,h'\in\H$ and a label $s$ such that $h(x)=\{s\}$, the set $h'(x)$ has at least two labels, and $s\notin h'(x)$. Choose distinct $u,v\in h'(x)$. Both distributions below put all their mass on the input $x$.
\begin{itemize}
\item[--] Under $\D_1$, the set $A$ is $h'(x)$ with probability $3/5$ and $\{s\}$ with probability $2/5$; when $A=h'(x)$, the demonstrator shows $u$ or $v$ with probability $1/2$ each.
\item[--] Under $\D_2$, the set $A$ is $\{s\}$, $\{u\}$ or $\{v\}$ with probabilities $2/5$, $3/10$ and $3/10$, and the demonstrator shows its only label.
\end{itemize}
In both cases the demonstrated label is $s$, $u$ or $v$ with probabilities $2/5$, $3/10$ and $3/10$, so the sample, and hence the learner's prediction $\widehat g(x)$, has the same law under $\D_1$ and $\D_2$. The hypothesis $h'$ disagrees with $A$ under $\D_1$ only when $A=\{s\}$, and $h$ disagrees with $A$ under $\D_2$ only when $A\neq\{s\}$, so $\Lset_{\D_1}(\H)\le2/5$ and $\Lset_{\D_2}(\H)\le3/5$.

Let $p_s$, $p_{\mathrm{in}}$ and $p_{\mathrm{out}}$ be the probabilities that $\widehat g(x)$ equals $s$, lies in $h'(x)$, or lies outside $h'(x)\cup\{s\}$. Under $\D_1$, the prediction $s$ is wrong with probability $3/5$, a label of $h'(x)$ with probability $2/5$, and any other label always. Under $\D_2$, the prediction $s$ is wrong with probability $3/5$, a label of $h'(x)$ with probability at least $7/10$, and any other label always. Hence the expected excess errors $e_1$ and $e_2$ under $\D_1$ and $\D_2$ satisfy
\[
 e_1\ge\frac15p_s+\frac35p_{\mathrm{out}},
 \qquad
 e_2\ge\frac1{10}p_{\mathrm{in}}+\frac25p_{\mathrm{out}},
\]
and therefore $\frac13e_1+\frac23e_2\ge\frac1{15}(p_s+p_{\mathrm{in}}+p_{\mathrm{out}})=\frac1{15}$. One of $e_1,e_2$ is at least $1/15$.
\end{proof}

\paragraph{Anchored cubes.}
The following configuration defeats every agnostic learner: a cube in which each hypothesis takes a prescribed singleton or contains a baseline label, together with one hypothesis that takes all baseline labels as singletons.

\begin{definition}[Anchored cubes]\label{agn:def:anchored}
An \emph{anchored cube} of size $n$ in $\H$ consists of distinct inputs $x_1,\ldots,x_n$, labels $a_i\neq b_i$ for $i\in[n]$, an \emph{anchor} $g\in\H$ with $g(x_i)=\{b_i\}$ for all $i$, and, for every $T\subseteq[n]$, a hypothesis $h_T\in\H$ such that
\[
 h_T(x_i)=\{a_i\}\quad(i\in T)
 \qquad\text{and}\qquad
 h_T(x_i)\neq\star,\ \ b_i\in h_T(x_i)\quad(i\notin T).
\]
\end{definition}

The same demonstrations admit two explanations. In the first, the hidden sets are given by $h_T$, so the benchmark is zero and the learner must predict $a_i$ on $T$. In the second, the hidden sets are the demonstrated singletons: predicting $a_i$ is then wrong outside $T$, while the anchor keeps the benchmark small. We choose $T$ at random so that the sample does not reveal whether an unseen input belongs to $T$.

\begin{lemma}[Anchored cubes defeat every learner]\label{agn:lem:anchored}
If $\H$ has an anchored cube of size $n\ge\max\{1,6m\}$, then for every learner that uses $m$ one-correct-label examples there are a distribution $\D$ on $\X\times\cA$ and a demonstrator such that
\[
 \E\bigl[L_\D(\widehat g)\bigr]\ge\Lset_\D(\H)+\frac18.
\]
Consequently, if $\H$ has anchored cubes of every size, then it is not agnostically PAC learnable from one correct label.
\end{lemma}

\begin{proof}
Choose a random set $T\subseteq[n]$ that contains each $i$ independently with probability $1/4$. In both distributions below, the input is uniform on $\{x_1,\ldots,x_n\}$ and the demonstrated label at $x_i$ is $a_i$ if $i\in T$ and $b_i$ otherwise. Under $\D_T$, the correct-label set at $x_i$ is $h_T(x_i)$; this is a realizable experiment with target $h_T$, so $\Lset_{\D_T}(\H)=0$. Under $\D'_T$, the correct-label set is the demonstrated singleton; the anchor $g$ disagrees with it only at the inputs $x_i$ with $i\in T$, so $\Lset_{\D'_T}(\H)\le|T|/n$. Both demonstrators are valid by the definition of an anchored cube, and the samples have the same law under $\D_T$ and $\D'_T$.

Draw a test index $j$ uniformly from $[n]$, and let $E$ be the event that $x_j$ does not occur in the sample. Its probability is $u=(1-1/n)^m\ge1-m/n\ge5/6$. On $E$, the event $j\in T$ is independent of the sample, the test index and the seed, and it has probability $1/4$. Let $r$ be the conditional probability, given $E$, that $\widehat g(x_j)=a_j$. Under $\D_T$, if $j\in T$ the only correct label is $a_j$, so
\[
 \E\bigl[L_{\D_T}(\widehat g)\bigr]\ge\frac u4(1-r).
\]
Under $\D'_T$, at an unseen input the prediction $a_j$ is wrong with probability $3/4$ and every other prediction with probability at least $1/4$, so the conditional error on $E$ is at least $1/4+r/2$. Since $\E|T|/n=1/4$,
\[
 \E\bigl[L_{\D'_T}(\widehat g)-\Lset_{\D'_T}(\H)\bigr]\ge u\Bigl(\frac14+\frac r2\Bigr)-\frac14=\frac{ur}2-\frac{1-u}4.
\]
These expectations include the choice of $T$. We weight the two bounds so that $r$ cancels:
\[
 \frac23\cdot\frac u4(1-r)+\frac13\Bigl(\frac{ur}2-\frac{1-u}4\Bigr)=\frac u4-\frac1{12}\ge\frac5{24}-\frac2{24}=\frac18.
\]
Hence for some fixed $T$, one of $\D_T$ and $\D'_T$ has expected excess error at least $1/8$. The consequence follows by taking $n\ge6m$ and $\eps+\delta<1/8$.
\end{proof}

Nondegeneracy and local compatibility turn every Natarajan cube of the collapsed class into an anchored cube of a third of its size. The two anchors $g_0,g_1$ handle the coordinates at which the cube compares a singleton label with $\diamond$.

\begin{lemma}[Extracting anchored cubes]\label{agn:lem:extract}
Let $\H$ be nondegenerate and locally compatible. If $\widetilde\H$ product-shatters a sequence of $N$ distinct inputs, then $\H$ has an anchored cube of size at least $N/3$.
\end{lemma}

\begin{proof}
For a class with singleton values, a product witness on $x_1,\ldots,x_N$ contains a cube $\prod_{i=1}^N\{c_i,c'_i\}$ with $c_i\ne c'_i$ in $\Y\cup\{\diamond\}$. Every vertex is the pattern of some $\widetilde h$ with $h$ defined at all $x_i$, and at most one of $c_i,c'_i$ is $\diamond$. Partition the coordinates into the three groups below; a largest group has at least $N/3$ coordinates, and each group yields an anchored cube. To define its hypotheses $h_T$, choose the indicated vertex on that group and fix arbitrary choices on all remaining coordinates.
\begin{itemize}
\item[--] \emph{Both $c_i,c'_i$ are labels of $\Y$.} Put $a_i=c_i$ and $b_i=c'_i$. Let $h_T$ take $\{a_i\}$ on $T$ and $\{b_i\}$ off $T$, and let the anchor be $h_\emptyset$.
\item[--] \emph{The symbols are $a_i\in\Y$ and $\diamond$, with $a_i\ne b_0(x_i)$.} Put $b_i=b_0(x_i)$, and let the anchor be $g_0$. Let $h_T$ take $\{a_i\}$ on $T$ and a set with at least two labels off $T$; by local compatibility with $g_0(x_i)=\{b_i\}$, this set contains $b_i$.
\item[--] \emph{The symbols are $b_0(x_i)$ and $\diamond$.} Put $a_i=b_0(x_i)$ and $b_i=b_1(x_i)$, which differ by nondegeneracy, and let the anchor be $g_1$. Let $h_T$ take $\{a_i\}$ on $T$ and a set with at least two labels off $T$, which contains $b_i$ by local compatibility with $g_1$.
\end{itemize}
In each group the conditions of Definition~\ref{agn:def:anchored} hold on the retained inputs.
\end{proof}

On the pure region, large DS witnesses may use unboundedly many labels, so we need a realizable lower bound that does not depend on the number of labels. The bound $1/q$ of Proposition~\ref{prop:ds-lower} gives no positive constant when $\Y$ is infinite; for singleton values, the same argument gives $1/2$.

\begin{lemma}[DS lower bound with singleton values]\label{agn:lem:ds-lower}
Let $S$ be a sequence of $n$ distinct inputs, and let $V\subseteq\H|_S$ be a DS witness whose patterns consist of singletons. For every learner that uses $m$ one-correct-label examples there are $h^\star\in\H$ and a distribution $P$ with $P(\supp(h^\star))=1$ such that
\[
 \E\bigl[L_{P,h^\star}(\widehat g)\bigr]\ge\frac12\Bigl(1-\frac1n\Bigr)^m.
\]
\end{lemma}

\begin{proof}
With singleton targets every demonstration is the correct label itself, so this is ordinary realizable multiclass learning. We follow the proof of Proposition~\ref{prop:ds-lower}, with $P$ uniform on the inputs of $S$ and a uniformly random target pattern $v\in V$. At a test input $x_i$ missing from the sample, given $v_{-i}$, the target is uniform on the hyperedge $e_{i,v_{-i}}$. This hyperedge has at least two patterns, and they have distinct labels at coordinate $i$, so every prediction is wrong with conditional probability at least $1-1/|e_{i,v_{-i}}|\ge1/2$. The rest of the proof is unchanged, and it does not use the finiteness of $\Y$.
\end{proof}

We can now prove the necessity part of Theorem~\ref{agn:thm:main}.

\begin{proof}[Proof of necessity in Theorem~\ref{agn:thm:main}]
Let $\H$ be nondegenerate and locally compatible, and suppose that $\dDS(\widetilde\H)=\infty$. By Lemma~\ref{agn:lem:ds-facts}, the restriction of $\widetilde\H$ to $\X_{\mathrm{pure}}$ or its restriction to $\X_{\mathrm{mix}}$ has infinite DS dimension. We show that $\H$ is not agnostically PAC learnable in either case.

\emph{The pure region.} Let $V$ be a DS witness of $\widetilde\H$ on $n$ inputs of $\X_{\mathrm{pure}}$. There $\widetilde h(x)=a$ means $h(x)=\{a\}$, so $V$, read as a set of singleton patterns, is also a DS witness in $\H|_S$, where $S$ is the sequence of these inputs. A realizable experiment is an agnostic experiment with benchmark zero, so Lemma~\ref{agn:lem:ds-lower} with $n\ge2m$ gives, for every learner with $m$ examples, a distribution under which the expected excess error is at least $\frac12(1-\frac1n)^m\ge\frac12(1-\frac mn)\ge\frac14$. Since such witnesses exist for every $n$, $\H$ is not agnostically learnable.

\emph{The mixed region.} By \eqref{agn:eq:local-alphabet}, the restriction of $\widetilde\H$ to $\X_{\mathrm{mix}}$ has at most $\ell+1$ possible values at each input. Its DS dimension is infinite, so by Lemma~\ref{agn:lem:natarajan} its Natarajan dimension is infinite as well. A sequence product-shattered by this restriction is product-shattered by $\widetilde\H$, so Lemma~\ref{agn:lem:extract} gives anchored cubes of every size, and Lemma~\ref{agn:lem:anchored} shows that $\H$ is not agnostically learnable.
\end{proof}

\subsection{Learning on the Pure Region}\label{agn:sec:pure}

We now prove sufficiency, one region at a time. On the pure region every defined value is a singleton, so a hypothesis can match the hidden set only by predicting the demonstrated label. Its error on the demonstrated labels is therefore at most its exact-set mismatch error. An ordinary agnostic multiclass learner suffices; we obtain one from our full-set results with $\ell=1$.

\begin{lemma}[Learning on the pure region]\label{agn:lem:pure}
If $\dDS(\widetilde\H)<\infty$, then $\H$ is agnostically learnable on $\X_{\mathrm{pure}}$.
\end{lemma}

\begin{proof}
Let $\mathcal C$ be the class of restrictions of the members of $\H$ to $\X_{\mathrm{pure}}$, viewed as a class on the countable instance space $\X_{\mathrm{pure}}$. Its values are singletons or $\star$, so $\mathcal C$ is a class with $\ell=1$, and on $\X_{\mathrm{pure}}$ the collapsed hypothesis $\widetilde h$ is $h$ with each value $\{a\}$ written as $a$. Hence $\dDS(\mathcal C)\le\dDS(\widetilde\H)<\infty$ by Lemma~\ref{agn:lem:ds-facts}.

With $\ell=1$ the intersection class adds nothing: two singletons intersect only if they are equal, so every member of $\cI_2(\mathcal C)$ agrees on its support with a member of $\mathcal C$, and $\mathcal C\subseteq\cI_2(\mathcal C)$. The two classes therefore have the same projections, and in particular $\dDS(\cI_2(\mathcal C))<\infty$. Theorem~\ref{thm:comparison} with $\rho=\gamma=1/4$ (this $\gamma$ satisfies \eqref{eq:gamma-range}) gives $\dpack(\mathcal C)=\dpack(\cI_2(\mathcal C))<\infty$, and Theorem~\ref{thm:countable-full} then gives an agnostic full-set learner for $\mathcal C$. With $\ell=1$ a full-set example is a pair $(x,\{y\})$, so this is an ordinary agnostic multiclass learner: for all $\eps,\delta$ there is a finite sample size such that for every distribution of pairs $(x,y)$ on $\X_{\mathrm{pure}}\times\Y$, with probability at least $1-\delta$, the output $\widehat g$ satisfies
\[
 \Pr[\widehat g(x)\ne y]\le\inf_{h\in\H}\Pr\bigl[h(x)\ne\{y\}\bigr]+\eps,
\]
where an undefined value counts as an error. We run it on the observed pairs, drawn from $\D_{\mathrm{obs}}$.

It remains to pass from the demonstrated label to the hidden set. For every $(x,A)$ with $x\in\X_{\mathrm{pure}}$, every $y\in A$, every $h\in\H$ and every label $z$,
\[
 \ind\{h(x)\ne\{y\}\}\le\ind\{h(x)\ne A\}
 \qquad\text{and}\qquad
 \ind\{z\notin A\}\le\ind\{z\ne y\}.
\]
For the first inequality, if $h(x)=A$ then $A$ is a defined value at a pure input, hence a singleton, and so $A=\{y\}$. The second holds because $y\in A$. Taking expectations, with probability at least $1-\delta$,
\[
 L_\D(\widehat g)\le\Pr[\widehat g(x)\ne y]\le\inf_{h\in\H}\Pr\bigl[h(x)\ne\{y\}\bigr]+\eps\le\Lset_\D(\H)+\eps.
 \qedhere
\]
\end{proof}

\subsection{Learning on the Mixed Region}\label{agn:sec:mixed}

On the mixed region a demonstrated label may come from a correct-label set with several elements, and the learner cannot tell which. By \eqref{agn:eq:local-alphabet}, every label of $\Y_x$ is correct whenever the hidden set is a value of a class member with at least two labels, so we restrict predictions to $\Y_x$. The argument has three steps. First, we bound the hidden excess risk against any $h\in\H$ by a comparison $\Delta_h$ that depends only on the law of the observed pairs (Lemma~\ref{agn:lem:observable}). Second, specialists over a finite cover of the collapsed class, combined with a leave-one-out argument, make this comparison small in expectation (Lemma~\ref{agn:lem:expected}). Third, we obtain a single predictor with high confidence (Lemma~\ref{agn:lem:confidence}). Ordinary validation does not work in this step, since the comparison can be negative and evaluating it requires the unknown comparator $h$; instead we average independent runs, predict a most frequently demonstrated label at inputs of large mass, and round randomly elsewhere.

Throughout this subsection $\H$ is nondegenerate and locally compatible, and $\D$ is a distribution on $\X_{\mathrm{mix}}\times\cA$. A randomized prediction $p$ is \emph{admissible} if $p(\cdot\mid x)$ is supported on $\Y_x$ for every $x\in\X_{\mathrm{mix}}$, and a predictor $g$ is admissible if $g(x)\in\Y_x$; we identify $g$ with the randomized prediction that puts mass one on $g(x)$. As in \eqref{eq:risk}, the risk of $p$ is $L_\D(p)=\E_{(x,A)\sim\D}[1-p(A\mid x)]$.

\paragraph{An observable comparison.}
For $h\in\H$, an admissible $p$, and an observed pair $(x,y)$, let
\[
 r_h(p;x,y)=
 \begin{cases}
 \ind\{y=a\}-p(y\mid x), & h(x)=\{a\},\\
 0, & |h(x)|\ge2,\\
 -p(y\mid x), & h(x)=\star,
 \end{cases}
 \qquad
 \Delta_h(p)=\E_{(x,y)\sim\D_{\mathrm{obs}}}\bigl[r_h(p;x,y)\bigr].
\]
At a singleton value, $r_h$ compares how often $h$ and $p$ agree with the demonstrated label, as in ordinary multiclass learning. At a value with two or more labels it is zero, since an admissible prediction is correct whenever this value is. At an undefined value, $h$ counts as wrong and the learner's agreement $p(y\mid x)$ is subtracted. The \emph{comparison} $\Delta_h(p)$ is affine in $p$ and lies in $[-1,1]$; it can be negative.

\begin{lemma}[Observable comparison]\label{agn:lem:observable}
For every $h\in\H$ and every admissible $p$,
\[
 L_\D(p)-\Pr_{(x,A)\sim\D}[h(x)\ne A]\le\Delta_h(p).
\]
\end{lemma}

\begin{proof}
It suffices to show that $1-p(A\mid x)-\ind\{h(x)\ne A\}\le r_h(p;x,y)$ whenever $y\in A$. If $h(x)=\{a\}=A$, then $y=a$ and both sides equal $1-p(a\mid x)$. If $h(x)$ is a singleton $\{a\}\ne A$, or $h(x)=\star$, the left side is $-p(A\mid x)\le-p(y\mid x)$, which is at most the right side. If $|h(x)|\ge2$ and $h(x)=A$, then $\Y_x\subseteq A$ by \eqref{agn:eq:local-alphabet}, so $p(A\mid x)=1$ and both sides are zero. If $|h(x)|\ge2$ and $h(x)\ne A$, the left side is $-p(A\mid x)\le0$.
\end{proof}

We will also need the comparison input by input. Let $\mu_x$ be the probability of $x$ under $\D_{\mathrm{obs}}$, let $\pi_x$ be the conditional law of the demonstrated label given $x$ (any law if $\mu_x=0$), and let $\pi_x(p)=\sum_a\pi_x(a)p(a\mid x)$ be the probability that a prediction drawn from $p(\cdot\mid x)$ equals the demonstrated label. Let $c_h(x)=\pi_x(a)$ if $h(x)=\{a\}$ and $c_h(x)=0$ otherwise, and let $w_h(x)=0$ if $|h(x)|\ge2$ and $w_h(x)=1$ otherwise. Then, directly from the definition of $r_h$,
\begin{equation}\label{agn:eq:conditional}
 \Delta_h(p)=\sum_{x}\mu_x\bigl(c_h(x)-w_h(x)\,\pi_x(p)\bigr).
\end{equation}
Only $\pi_x(p)$ depends on the prediction, and its coefficient is nonpositive, so increasing agreement with the demonstrated label cannot increase the comparison against any $h$. This lets us choose predictions at inputs of large mass without knowing the comparator.

\paragraph{Specialists over a finite cover.}
To make the comparison small we need finitely many candidate predictors on each pool of inputs, and the finite cover of Section~\ref{sec:finite-full} provides them. Let $d$ be the Natarajan dimension of the restriction of $\widetilde\H$ to $\X_{\mathrm{mix}}$; it is at most $\dDS(\widetilde\H)$ by Lemma~\ref{lem:orders}. At each $x\in\X_{\mathrm{mix}}$ the collapsed values lie in $\Y_x\cup\{\diamond\}$, so an injection $\iota_x$ of this set into $[\ell+1]$ turns the restriction into a class with singleton values over $[\ell+1]$, still of Natarajan dimension $d$. Lemma~\ref{lem:finite-cover}, applied to this class,\footnote{With $1$ in place of $\ell$ and $k=q=\ell+1$. Then $\Gamma\le\binom{\ell+1}2$, and \eqref{agn:eq:Lambda} follows from \eqref{eq:cover-bound} and \eqref{ff:eq:M-log}.} gives for every pool $S=(x_1,\ldots,x_n)$ of inputs in $\X_{\mathrm{mix}}$ a set $F_S\subseteq[\ell+1]^n$, depending only on $S$ and $\H$, with the following covering property: for every $h\in\H$ some $f\in F_S$ satisfies $f(i)=\iota_{x_i}(\widetilde h(x_i))$ at every position $i$ with $x_i\in\supp(h)$. Its size satisfies
\begin{equation}\label{agn:eq:Lambda}
 \log|F_S|\le M_d(n)\log\bigl(1+n(\ell+1)^2\bigr)\le\Lambda_n,
 \qquad
 \Lambda_n=c\,(d+1)\log^2(n+1),
\end{equation}
where $c$ depends only on $\ell$.

Each $f\in F_S$ acts as a \emph{specialist}: at a position $i$ with $f(i)=\iota_{x_i}(a)$ for some $a\in\Y_{x_i}$, it is \emph{awake} and predicts $a$; at all other positions it is \emph{asleep}. We run a multiplicative-weights version of the specialists algorithm of \cite{freund1997using}. As in Theorem~\ref{thm:agnostic-full}, the learner predicts by weighting experts, but here only awake experts contribute, and each weight is updated by the expert's gain relative to the learner. Each $f\in F_S$ starts with weight $1/|F_S|$, and the positions are processed in the order of the pool. At position $i$, if some specialist is awake, $p_i(\cdot\mid x_i)$ is the weighted distribution of the labels of the awake specialists; otherwise it puts mass one on a fixed label of $\Y_{x_i}$, which is nonempty. After $y_i$ is revealed, an awake specialist with label $a$ receives the payoff $r_{f,i}=\ind\{y_i=a\}-p_i(y_i\mid x_i)$, and a sleeping specialist receives $r_{f,i}=0$. For a learning rate $0<\beta\le1/2$, every weight is then multiplied by $1+\beta r_{f,i}$. All predictions $p_i$ are admissible.

The next lemma transfers the usual specialist regret bound to $r_h$, using a member of $F_S$ that covers $h$.

\begin{lemma}[Specialist bound]\label{agn:lem:specialists}
For every pool $S=(x_1,\ldots,x_n)$ of inputs in $\X_{\mathrm{mix}}$, all labels $y_1,\ldots,y_n\in\Y$, and every $h\in\H$,
\[
 \sum_{i=1}^n r_h(p_i;x_i,y_i)\le\frac{\Lambda_n}\beta+\beta n.
\]
\end{lemma}

\begin{proof}
By the definition of $p_i$, the payoffs at every position have weighted sum zero, so the weights always sum to one and each of them is at most one. Since $|\beta r_{f,i}|\le1/2$ and $\log(1+z)\ge z-z^2$ for $|z|\le1/2$, the final weight $W_f$ of a specialist $f$ satisfies
\[
 0\ge\log W_f\ge-\log|F_S|+\beta\sum_{i=1}^nr_{f,i}-\beta^2n .
\]
Thus every specialist has total payoff at most $\log|F_S|/\beta+\beta n$. Now let $f$ cover $h$. We claim that $r_h(p_i;x_i,y_i)\le r_{f,i}$ at every position. Where $h(x_i)=\{a\}$, the specialist $f$ is awake and predicts $a$, so the two payoffs are equal. Where $h(x_i)$ has two or more labels, $f(i)=\iota_{x_i}(\diamond)$, so $f$ is asleep and both payoffs are zero. Where $h$ is undefined, $r_h=-p_i(y_i\mid x_i)$, while $r_{f,i}$ is either $0$ or $\ind\{y_i=a\}-p_i(y_i\mid x_i)$. Summing over $i$ and using \eqref{agn:eq:Lambda} proves the lemma.
\end{proof}

\paragraph{An expected comparison bound.}
A leave-one-out argument turns the specialist bound into an inductive learner, exactly as in the proofs of Theorems~\ref{thm:full-expected} and~\ref{thm:agnostic-full}.

\begin{lemma}[Expected comparison]\label{agn:lem:expected}
Let $n\ge4(\Lambda_n+1)$. There is a learner that uses $n-1$ observed examples and returns an admissible randomized prediction $p$ such that for every distribution of pairs on $\X_{\mathrm{mix}}\times\Y$ and every $h\in\H$,
\[
 \E\bigl[\Delta_h(p)\bigr]\le2\sqrt{\frac{\Lambda_n+1}n}.
\]
\end{lemma}

\begin{proof}
Let $\beta=\sqrt{(\Lambda_n+1)/n}\le1/2$. Given $n-1$ observed pairs and a query $x\in\X_{\mathrm{mix}}$, add $x$ to the training inputs as an $n$-th position. For each ordering of the $n$ positions, run the specialists algorithm on the pool listed in this order, revealing the labels of the training positions that come before the query, and take its prediction at the query. Let $p(\cdot\mid x)$ be the average of these predictions over all $n!$ orderings; it is admissible and is a deterministic function of the sample.

Now draw $n$ independent pairs, fix an ordering, and run the specialists algorithm once on all positions, revealing every label. The prediction at a position depends only on labels revealed earlier, so it is the prediction used by the learner for this ordering when that position is the query and the other $n-1$ pairs form the sample. Let $p^{(j)}$ be the output of the learner trained on all pairs but the $j$-th. Averaging over orderings, and using that $r_h$ is affine in the prediction, Lemma~\ref{agn:lem:specialists} gives
\[
 \sum_{j=1}^n r_h\bigl(p^{(j)};x_j,y_j\bigr)\le\frac{\Lambda_n}\beta+\beta n ,
\]
where $(x_j,y_j)$ is the $j$-th pair. The $j$-th pair is independent of the others, so each summand has expectation $\E[\Delta_h(p)]$. Dividing by $n$ gives $\E[\Delta_h(p)]\le\Lambda_n/(\beta n)+\beta\le2\beta$.
\end{proof}

Since $\Lambda_n=O_\ell((d+1)\log^2(n+1))$, the bound of Lemma~\ref{agn:lem:expected} tends to zero, so for every $\alpha>0$ a finite sample size gives $\E[\Delta_h(p)]\le\alpha$ for all distributions and all $h$ at once.

\paragraph{High confidence for a single predictor.}
It remains to pass from expectation to high probability. The learner of the next lemma does not depend on $h$, but its guarantee holds for each fixed $h$ separately rather than uniformly over $\H$; this suffices, because the proof of Proposition~\ref{agn:prop:mixed} applies it to a single near-optimal $h$.

\begin{lemma}[Confidence]\label{agn:lem:confidence}
Let $d<\infty$. For all $\eps,\delta\in(0,1)$ there is a learner that uses finitely many observed examples and returns a random admissible predictor $\widehat g$ such that for every distribution of pairs on $\X_{\mathrm{mix}}\times\Y$ and every $h\in\H$,
\[
 \Pr\bigl[\Delta_h(\widehat g)>\eps\bigr]\le\delta.
\]
\end{lemma}

\begin{proof}
Let $\alpha=\eps/4$, and let $m_0$ be the sample size for which the learner of Lemma~\ref{agn:lem:expected} satisfies $\E[\Delta_h(p)]\le\alpha$. Set
\[
 K=\Bigl\lceil\frac2{\alpha^2}\log\frac4\delta\Bigr\rceil,
 \quad
 \tau=\min\Bigl\{\frac12,\frac{2\alpha^2}{\log(4/\delta)}\Bigr\},
 \quad
 u=\Bigl\lceil\frac1\tau\log\frac4{\delta\tau}\Bigr\rceil,
 \quad
 v=\Bigl\lceil\frac2{\alpha^2}\Bigl(u\log\ell+\log\frac8\delta\Bigr)\Bigr\rceil.
\]
The learner uses three independent blocks of $Km_0$, $u$ and $v$ examples, in this order. Fix a distribution and $h\in\H$. We show that each of the four steps below fails with probability at most $\delta/4$.

\emph{Step 1 (averaging).} Train $K$ randomized predictions $p_1,\ldots,p_K$ on separate parts of the first block, each of $m_0$ examples, and let $\bar p$ be their average. The values $\Delta_h(p_j)$ are independent, lie in $[-1,1]$, and have mean at most $\alpha$. Since $\Delta_h$ is affine, Hoeffding's inequality gives $\Delta_h(\bar p)\le2\alpha$ except with probability $e^{-K\alpha^2/2}\le\delta/4$.

\emph{Step 2 (heavy inputs).} Let $U$ be the set of inputs of the second block. At most $1/\tau$ inputs have $\mu_x>\tau$, and each of them is missed with probability at most $e^{-u\tau}$. So, except with probability $\tau^{-1}e^{-u\tau}\le\delta/4$, every input outside $U$ has mass at most $\tau$.

\emph{Step 3 (most frequent labels on $U$).} For $x\in U$, let $\widehat a_x$ be the label of $\Y_x$ that is demonstrated most often at $x$ in the third block, with ties broken by the fixed enumeration of $\Y$. This choice maximizes the empirical frequency of $\{x\in U,\ y=a(x)\}$ over the at most $\ell^{u}$ maps $a$ with $a(x)\in\Y_x$. For a fixed map $a$, the event has probability $\sum_{x\in U}\mu_x\pi_x(a(x))$. By Hoeffding's inequality and a union bound, conditionally on $U$, all these frequencies are within $\alpha/2$ of their probabilities except with probability $2\ell^ue^{-v\alpha^2/2}\le\delta/4$. On this event, comparing $\widehat a$ with a map that picks a most probable label of $\pi_x$ in $\Y_x$ at each $x\in U$ gives
\[
 \sum_{x\in U}\mu_x\Bigl(\max_{a\in\Y_x}\pi_x(a)-\pi_x(\widehat a_x)\Bigr)\le\alpha .
\]
Now let $p^U$ agree with $\bar p$ outside $U$ and put mass one on $\widehat a_x$ at $x\in U$. Since $\bar p$ is admissible, $\pi_x(\bar p)\le\max_{a\in\Y_x}\pi_x(a)$, and since $w_h(x)\in\{0,1\}$, \eqref{agn:eq:conditional} gives
\[
 \Delta_h(p^U)-\Delta_h(\bar p)=\sum_{x\in U}\mu_x\,w_h(x)\bigl(\pi_x(\bar p)-\pi_x(\widehat a_x)\bigr)\le\alpha .
\]

\emph{Step 4 (rounding).} Let $\widehat g(x)=\widehat a_x$ for $x\in U$, and for every other $x\in\X_{\mathrm{mix}}$ draw $\widehat g(x)$ from $\bar p(\cdot\mid x)$, independently across inputs and once for all, using the seed. Conditionally on the examples, \eqref{agn:eq:conditional} shows that $\Delta_h(\widehat g)-\Delta_h(p^U)$ is a sum over $x\notin U$ of independent terms of mean zero, and the term at $x$ lies in an interval of length at most $\mu_x$. If every $x\notin U$ has $\mu_x\le\tau$, then $\sum_{x\notin U}\mu_x^2\le\tau$, and Hoeffding's inequality\footnote{For a countable sum, apply Hoeffding's lemma to finite partial sums and pass to the limit in their moment generating functions.} gives $\Delta_h(\widehat g)\le\Delta_h(p^U)+\alpha$ except with probability $e^{-2\alpha^2/\tau}\le\delta/4$.

By a union bound over the four steps, with probability at least $1-\delta$ we have $\Delta_h(\widehat g)\le2\alpha+\alpha+\alpha=\eps$.
\end{proof}

For standard Borel instance spaces, Section~\ref{app:meas-agnostic} proves the same guarantee by rounding on finitely many Borel sets in place of individual inputs.

\begin{proposition}[Learning on the mixed region]\label{agn:prop:mixed}
Let $\H$ be nondegenerate and locally compatible, and suppose the restriction of $\widetilde\H$ to $\X_{\mathrm{mix}}$ has finite Natarajan dimension, for instance because $\dDS(\widetilde\H)<\infty$. Then $\H$ is agnostically learnable on $\X_{\mathrm{mix}}$.
\end{proposition}

\begin{proof}
Given $\eps,\delta$, run the learner of Lemma~\ref{agn:lem:confidence} with accuracy $\eps/2$ on the observed pairs. For the hidden distribution $\D$ at hand, fix $h\in\H$ with $\Pr_{(x,A)\sim\D}[h(x)\ne A]\le\Lset_\D(\H)+\eps/2$; the learner does not use it. With probability at least $1-\delta$, $\Delta_h(\widehat g)\le\eps/2$, and Lemma~\ref{agn:lem:observable}, applied to the admissible predictor $\widehat g$, gives $L_\D(\widehat g)\le\Pr_{(x,A)\sim\D}[h(x)\ne A]+\eps/2\le\Lset_\D(\H)+\eps$.
\end{proof}

\subsection{Proof of the Characterization}\label{agn:sec:proof}

The two regions depend only on $\H$, so the learner can assign each input to its region and pass each regional learner the examples from its region.

\begin{lemma}[Gluing]\label{agn:lem:glue}
If $\H$ is agnostically learnable on $\X_{\mathrm{pure}}$ and on $\X_{\mathrm{mix}}$, then $\H$ is agnostically PAC learnable from one correct label.
\end{lemma}

\begin{proof}
Fix $\eps,\delta$. For each region $R$, let $m_R$ be the sample size of a learner on $R$ with accuracy $\eps/4$ and confidence $\delta/4$, let $m_*=\max\{1,m_{\mathrm{pure}},m_{\mathrm{mix}}\}$, and draw
\[
 N=\Bigl\lceil\frac{32}\eps\Bigl(m_*+\log\frac8\delta\Bigr)\Bigr\rceil
\]
examples. For each region $R$, pass its first $m_R$ examples to the regional learner; if fewer are available, use an arbitrary predictor on $R$. The two learners use independent seeds, and the final predictor uses the output for the region containing its input. Conditional on which sample positions fall in $R$, the retained examples are independent draws from the conditional law $\D_R$ of $(x,A)$ given $x\in R$, with the same demonstrator, so the regional guarantee applies whenever enough examples are retained.

Let $\mu_R=\Pr_\D[x\in R]$. A region of mass below $\eps/4$ costs at most its mass, so only regions of larger mass need a guarantee. If $\mu_R\ge\eps/4$, the number of examples in $R$ has mean $N\mu_R\ge8(m_*+\log(8/\delta))>2m_R$, so by a multiplicative Chernoff bound fewer than $m_R$ examples are retained with probability at most $e^{-N\mu_R/8}\le e^{-\log(8/\delta)}=\delta/8$. Each regional learner fails with probability at most $\delta/4$. Hence, with probability at least $1-2\cdot\delta/8-2\cdot\delta/4\ge1-\delta$, every region with $\mu_R\ge\eps/4$ satisfies $L_{\D_R}(\widehat g)\le\Lset_{\D_R}(\H)+\eps/4$.

The infimum of a sum is at least the sum of the infima, so $\sum_R\mu_R\Lset_{\D_R}(\H)\le\Lset_\D(\H)$, where regions of mass zero contribute zero. On the same event, therefore,
\[
 L_\D(\widehat g)\le\sum_{R:\,\mu_R\ge\eps/4}\mu_R\Bigl(\Lset_{\D_R}(\H)+\frac\eps4\Bigr)+\sum_{R:\,\mu_R<\eps/4}\mu_R
 \le\Lset_\D(\H)+\frac{3\eps}4 .
 \qedhere
\]
\end{proof}

\begin{proof}[Proof of Theorem~\ref{agn:thm:main}]
Necessity was proved in Section~\ref{agn:sec:necessity}. For sufficiency, let $\H$ be nondegenerate and locally compatible with $\dDS(\widetilde\H)<\infty$. Then $\H$ is agnostically learnable on $\X_{\mathrm{pure}}$ by Lemma~\ref{agn:lem:pure} and on $\X_{\mathrm{mix}}$ by Proposition~\ref{agn:prop:mixed}, and Lemma~\ref{agn:lem:glue} combines the two learners.
\end{proof}

Finally, we prove Corollary~\ref{agn:cor:vc} by comparing the collapsed class with the binary class $\mathcal J$.

\begin{lemma}[Collapsed and binary classes]\label{agn:lem:vc}
For every class $\H$, $\dN(\widetilde\H)\le\mathrm{VC}(\mathcal J)$. If $\H$ is locally compatible and $\X_{\mathrm{mix}}=\X$, then $\mathrm{VC}(\mathcal J)<\infty$ if and only if $\dDS(\widetilde\H)<\infty$.
\end{lemma}

\begin{proof}
Take a cube $\prod_{i=1}^n\{c_i,c'_i\}$ realized by $\widetilde\H$ on distinct inputs $x_1,\ldots,x_n$, as in the proof of Lemma~\ref{agn:lem:extract}. At each coordinate at least one of $c_i,c'_i$ is a label of $\Y$; call it $a_i$. A hypothesis $h$ realizing a vertex has $J_h(x_i,a_i)=1$ if the vertex takes $a_i$ at $i$ and $J_h(x_i,a_i)=0$ otherwise. So the points $(x_i,a_i)$ are shattered by $\mathcal J$, which proves the inequality.

Now let $\H$ be locally compatible with $\X_{\mathrm{mix}}=\X$, and suppose that $\mathcal J$ shatters $(x_1,a_1),\ldots,(x_n,a_n)$. The inputs are distinct, because $J_h(x,a)=J_h(x,a')=1$ forces $a=a'$. The $2^n$ binary patterns come from hypotheses defined at all $x_i$, and two of them with different binary patterns have different collapsed patterns on $S=(x_1,\ldots,x_n)$, since exactly one of them takes $a_i$ at a coordinate $i$ where they differ. So $|\widetilde\H|_S|\ge2^n$. On the other hand, since every input is mixed, local compatibility puts the collapsed values at each input in $\Y_x\cup\{\diamond\}$, a set of at most $\ell+1$ labels (Section~\ref{agn:sec:regions}), so Lemma~\ref{agn:lem:natarajan} gives $2^n\le n^d(\ell+1)^{2d}$ if $\dN(\widetilde\H)\le d$. Hence $\mathrm{VC}(\mathcal J)$ is finite if and only if $\dN(\widetilde\H)$ is, and by Lemma~\ref{agn:lem:natarajan} if and only if $\dDS(\widetilde\H)$ is.
\end{proof}

\begin{proof}[Proof of Corollary~\ref{agn:cor:vc}]
Under the assumption of the corollary every input has a value with at least two labels, so $\X_{\mathrm{mix}}=\X$. Lemma~\ref{agn:lem:vc} then shows that $\mathrm{VC}(\mathcal J)<\infty$ if and only if $\dDS(\widetilde\H)<\infty$, and Theorem~\ref{agn:thm:main} completes the proof.
\end{proof}

\begin{remark}[Partial concept classes]\label{agn:rem:binary}
For $\ell=1$ every defined value is a singleton, so the mixed region is empty, local compatibility holds vacuously, and no anchors are needed. The proof of Lemma~\ref{agn:lem:pure} with $\X_{\mathrm{pure}}=\X$ gives a learner when $\dDS(\H)<\infty$, and Lemma~\ref{agn:lem:ds-lower}, applied to DS witnesses of $\H$ on the whole of $\X$, rules out learning when $\dDS(\H)=\infty$. Thus, for $\ell=1$ and without nondegeneracy, $\H$ is agnostically learnable if and only if $\dDS(\H)<\infty$. For $\Y=\{0,1\}$ a DS witness is a full binary cube, so the DS dimension is the VC dimension, and we recover the characterization of agnostic learnability of partial concept classes by finite VC dimension \cite{alon2021theory}.
\end{remark}

\begin{remark}[Undefined values cannot be coded as zero]\label{agn:rem:zero}
In Corollary~\ref{agn:cor:vc}, an undefined value of $h$ must stay undefined in $J_h$. Let $\X=\N$, $\Y=\{0,1\}$ and $\ell=2$, and let $\H$ consist of the constant hypotheses $\{0\}$, $\{1\}$ and $\{0,1\}$, and, for every finite $F\subseteq\N$, the hypothesis that equals $\{0\}$ on $F$ and is undefined elsewhere. This class is nondegenerate, locally compatible, and every input has a value with two labels. On $n$ distinct inputs the only patterns of $\widetilde\H$ are the constant ones $0^n$, $1^n$ and $\diamond^n$, so $\dDS(\widetilde\H)=1$ and $\H$ is agnostically learnable. Coding undefined values as zero, however, turns $x\mapsto J_h(x,0)$ into the indicators of all finite sets, a class of infinite VC dimension.
\end{remark}


\section{Limits of Dimension Characterizations}\label{sec:limits}

All other sections assume that every correct-label set has at most $\ell$ labels. In this section we drop this uniform bound and prove two impossibility results for characterizations of PAC learnability by dimensions. The first rules out \emph{scale-free} characterizations, which use a single parameter independent of the accuracy and the confidence, as well as scale-free sample complexity bounds; it holds under both full-set and one-correct-label feedback. The second concerns \emph{scale-sensitive profiles}, which are families of parameters indexed by a scale $\alpha\in(0,1)$. Under one-correct-label feedback, we show that every such profile whose members are determined by their finite subclasses and ignore the names of the labels (Definition~\ref{def:finite-witness}) takes the same value on a class learnable from zero examples, on learnable classes of arbitrarily large sample complexity, and on a class that is not learnable. Our approach is related to \cite{lechner2024inherent}, who showed that no notion of dimension characterizes the learnability of classes of distributions under natural requirements (see also \cite{lechner2026no}).

\paragraph{Setting.}
In this section a hypothesis is a total function $h$ from $\X$ to the nonempty subsets of $\Y$. In Section~\ref{sec:limits-scale-free} every correct-label set is finite, but there is no common bound on their sizes; in Section~\ref{sec:limits-scale-sensitive} every correct-label set has finite complement in $\Y$. The errors \eqref{eq:risk}, the feedback models, and Definition~\ref{def:pac} apply without change. For a bijection $\pi:\Y\to\Y$ and a hypothesis $h$, let $(\pi h)(x)=\{\pi(y):y\in h(x)\}$ and $\pi\H=\{\pi h:h\in\H\}$. The parameters that we rule out are of the following kind.

\begin{definition}\label{def:finite-witness}
A \emph{parameter} $\mathrm D$ assigns a value $\mathrm D(\H)\in\N_0\cup\{\infty\}$, where $\N_0=\{0\}\cup\N$, to every nonempty class $\H$ under consideration.
\begin{enumerate}[(i)]
\item $\mathrm D$ has the \emph{finite-witness property} if for every nonempty class $\H$,
\begin{equation}\label{lim:finite-witness}
\mathrm D(\H)=\sup\bigl\{\mathrm D(\cV):\cV\subseteq\H,\ 1\leq|\cV|<\infty\bigr\}.
\end{equation}
Such a parameter is monotone: $\mathrm D(\H')\leq\mathrm D(\H)$ whenever $\H'\subseteq\H$.
\item $\mathrm D$ is \emph{relabeling invariant} if $\mathrm D(\pi\H)=\mathrm D(\H)$ for every class $\H$ and every bijection $\pi:\Y\to\Y$.
\item A \emph{scale-sensitive profile} is a family $\boldsymbol{\mathrm D}=(\mathrm D_\alpha)_{\alpha\in(0,1)}$ of parameters, which need not be related to each other.
\end{enumerate}
\end{definition}

\paragraph{Minimax risk.}
Fix the feedback model, and write $m_\H(\eps,\delta)$ for the corresponding sample complexity $\mfull_\H(\eps,\delta)$ or $\mone_\H(\eps,\delta)$. A \emph{data law} for a target $h^\star$ is a distribution $P$ on $\X$ under full-set feedback, and a distribution $\D$ with marginal $P$ that is compatible with $h^\star$ under one-correct-label feedback. For an integer $m\geq0$, the minimax risk with $m$ examples is
\[
\RiskMin_m(\H)=\inf_{\text{learners}}\ \sup_{h^\star\in\H}\ \sup_{\text{data laws}}\ \E\bigl[L_{P,h^\star}(\widehat g)\bigr],
\]
where the infimum is over learners that use $m$ examples and the expectation is over the sample and the seed. By \eqref{eq:pac-expect}, an $(\eps,\delta)$-PAC guarantee with $m$ examples gives $\RiskMin_m(\H)\leq\eps+\delta$; hence a class with $\RiskMin_m(\H)\geq c>0$ for every $m$ is not realizably PAC learnable.

Both constructions below use labels in $\N_0$, and then the two feedback models are related by a simple reduction. A full-set learner can pass the least label of each observed set, which is a valid demonstration, to a one-correct-label learner. Hence every one-correct-label guarantee is also a full-set guarantee, and every full-set lower bound is also a one-correct-label lower bound.

\subsection{Scale-Free Impossibility}\label{sec:limits-scale-free}

\begin{theorem}[Scale-free impossibility]\label{thm:scale-free}
Fix either full-set or one-correct-label feedback.
Let $\X$ and $\Y$ be countably infinite, and consider countable classes $\H$ of total hypotheses with $1\leq|h(x)|<\infty$ for all $h\in\H$ and $x\in\X$.
\begin{enumerate}[(i)]
\item There is no parameter $\mathrm D$ with the finite-witness property such that $\mathrm D(\H)<\infty$ if and only if $\H$ is realizably PAC learnable.
\item There is no pair $(\mathrm D,\Psi)$ of a parameter $\mathrm D$ that is finite on every realizably learnable class and a function $\Psi:\N_0\times(0,1)^2\to[0,\infty)$ such that
\begin{equation}\label{lim:integer-upper}
m_\H(\eps,\delta)\leq\Psi\bigl(\mathrm D(\H),\eps,\delta\bigr)
\end{equation}
for every realizably learnable class $\H$ and all $\eps,\delta\in(0,1)$.
\end{enumerate}
\end{theorem}

The proof uses classes $\H_j$, each on its own block of inputs, whose minimax error is $1/s_j$ at every sample size, where $s_j=j+2$; in particular, no $\H_j$ is learnable. A finite subclass of $\H_j$, however, departs from a default target at only finitely many inputs, and a learner that memorizes its sample is wrong on block $j$ only at the departures that it has not seen. Since the error floors $1/s_j$ tend to $0$, at any given accuracy all but finitely many blocks can be handled by uniform guessing, so every union of finite subclasses $\cV_j\subseteq\H_j$ is learnable. A parameter with finite witnesses that is finite exactly on learnable classes is infinite on each $\H_j$, hence at least $j$ on some finite $\cV_j$, and therefore infinite on the learnable union; this proves part~(i). For part~(ii) we choose the finite subclasses diagonally, so that at accuracy of order $1/s_j$ the subclass $\cV_j$ needs more examples than $\Psi$ allows for parameter values up to $j$.

\paragraph{Construction: blocks with decreasing error floors.}
Since $\X$ and $\Y$ are countably infinite, we may take $\X=\N\times\N$ and $\Y=\N$, and call $\{k\}\times\N$ the $k$th block. For $j\in\N$ and each function $a:\N\to[s_j]$ that equals $1$ outside a finite set, define
\begin{equation}\label{lim:block-target}
h_{j,a}(k,t)=
\begin{cases}
[s_j]\setminus\{a(t)\}, & k=j,\\
[s_k], & k\neq j,
\end{cases}
\end{equation}
and let $\H_j$ be the class of these targets. Thus a target in $\H_j$ excludes the label $a(t)$ at each input $(j,t)$ of its active block and accepts all of $[s_k]$ on every other block $k$. Each $\H_j$ is countable, and its correct-label sets are finite and nonempty. By the reduction above, it suffices to prove lower bounds under full-set feedback and learning guarantees under one-correct-label feedback; each then holds under both models.

\begin{lemma}[Error floors and finite unions]\label{lem:blocks}
\leavevmode
\begin{enumerate}[(i)]
\item For every $j\geq1$ and every integer $m\geq0$,
\begin{equation}\label{lim:error-floor}
\RiskMin_m(\H_j)=\frac1{s_j}.
\end{equation}
In particular, $\H_j$ is not realizably PAC learnable.
\item Every nonempty class $\H=\bigcup_{j\geq1}\cV_j$ with finite $\cV_j\subseteq\H_j$ is realizably PAC learnable.
\end{enumerate}
\end{lemma}

\begin{proof}
\textbf{The lower bound in \eqref{lim:error-floor}.}\ Fix a full-set learner that uses $m$ examples, and an integer $M>m$. Choose $a(1),\ldots,a(M)$ independently and uniformly from $[s_j]$, set $a(t)=1$ for $t>M$, and let $P$ be the uniform distribution on $(j,1),\ldots,(j,M)$. The sample reveals the values of $a$ only at the sampled inputs. Hence, at a test input that does not appear in the sample, the excluded label is uniform on $[s_j]$ and independent of the sample and the seed; a prediction in $[s_j]$ is then invalid with probability $1/s_j$, and a prediction outside $[s_j]$ is always invalid. A test input is absent from the sample with probability $(1-1/M)^m$, so averaging over the random choice of $a$ gives
\begin{equation}\label{lim:finite-block-lower}
\sup_a\E\bigl[L_{P,h_{j,a}}(\widehat g)\bigr]\geq\frac1{s_j}\Bigl(1-\frac1M\Bigr)^m.
\end{equation}
Letting $M\to\infty$ proves $\RiskMin_m(\H_j)\geq1/s_j$ for every $m$, and in particular $\H_j$ is not realizably PAC learnable.

\paragraph{The upper bound in \eqref{lim:error-floor}.}
The learner that ignores the sample and outputs $g_\omega(k,t)=1+\lfloor s_k\omega\rfloor$, with the seed $\omega$ uniform on $[0,1)$, predicts a uniform label of $[s_k]$ on block $k$. For a target $h_{j,a}\in\H_j$, this label is always valid outside block $j$ and is invalid at $(j,t)$ with probability $1/s_j$. Hence its expected error is at most $1/s_j$.

\paragraph{Unions of finite subclasses.}
Let each $\cV_j\subseteq\H_j$ be finite. The targets in $\cV_j$ exclude a label other than $1$ only at inputs $(j,t)$ with $t$ in the finite set
\[
E_j=\{t\in\N:\ a(t)\neq1\text{ for some }h_{j,a}\in\cV_j\},
\qquad\text{and we put}\qquad
\cE_J=\bigcup_{j\leq J}\bigl(\{j\}\times E_j\bigr).
\]
Fix $J$ and let the seed $\omega$ be uniform on $[0,1)$. The learner predicts
\[
\widehat g(j,t)=
\begin{cases}
\text{the first label demonstrated at }(j,t), & \text{if }(j,t)\text{ appears in the sample},\\
2, & \text{if }(j,t)\text{ is unseen and }j\leq J,\\
1+\lfloor s_j\omega\rfloor, & \text{if }(j,t)\text{ is unseen and }j>J.
\end{cases}
\]
Let the target be $h_{j',a}\in\cV_{j'}$. Demonstrated labels are correct. At an unseen input of a block $k\neq j'$, the target accepts all of $[s_k]$, and the prediction lies in $[s_k]$ (for the label $2$ because $s_k\geq3$). On block $j'$ there are two cases. If $j'\leq J$, the label $2$ is correct at $(j',t)$ unless $a(t)=2$, and then $(j',t)\in\cE_J$. If $j'>J$, the expected error on block $j'$ is at most $1/s_{j'}\leq1/s_{J+1}$, as in the upper bound above. Hence, for every $h^\star\in\H$ and every data law,
\begin{equation}\label{lim:assembly-risk}
\E\bigl[L_{P,h^\star}(\widehat g)\bigr]
\leq\sum_{x\in\cE_J}p_x(1-p_x)^m+\frac1{s_{J+1}}
\leq\frac{|\cE_J|}{m+1}+\frac1{s_{J+1}},
\end{equation}
where $p_x=P(\{x\})$, and the second inequality uses $p(1-p)^m\leq1/(m+1)$ for $p\in[0,1]$. Given $\eps,\delta$, choose $J$ with $1/s_{J+1}\leq\eps\delta/2$, and then $m$ with $|\cE_J|/(m+1)\leq\eps\delta/2$. The expected error is then at most $\eps\delta$, and the second implication in \eqref{eq:pac-expect} proves realizable PAC learnability under one-correct-label feedback, and hence under full-set feedback.
\end{proof}

\begin{proof}[Proof of Theorem~\ref{thm:scale-free}]
For part~(i), suppose that $\mathrm D$ has the stated properties. By Lemma~\ref{lem:blocks}(i), no $\H_j$ is learnable, so $\mathrm D(\H_j)=\infty$, and the finite-witness property gives a finite nonempty subclass $\cV_j\subseteq\H_j$ with $\mathrm D(\cV_j)\geq j$. By Lemma~\ref{lem:blocks}(ii), the class $\H=\bigcup_j\cV_j$ is realizably PAC learnable. However, monotonicity gives $\mathrm D(\H)\geq\sup_j\mathrm D(\cV_j)=\infty$, a contradiction.

For part~(ii), suppose that $\mathrm D$ and $\Psi$ satisfy \eqref{lim:integer-upper}, and set
\begin{equation}\label{lim:diagonal-parameters}
\eps_j=\delta_j=\frac1{8s_j},\qquad
q_j=1+\Bigl\lceil\max_{d\in\{0,\ldots,j\}}\Psi(d,\eps_j,\delta_j)\Bigr\rceil,\qquad
M_j=2q_j,
\end{equation}
so that $q_j>\Psi(d,\eps_j,\delta_j)$ for every $d\leq j$. Let $\cV_j$ consist of the targets $h_{j,a}\in\H_j$ with arbitrary values of $a$ on $[M_j]$ and $a(t)=1$ for $t>M_j$. These classes are finite, so $\H=\bigcup_j\cV_j$ is realizably PAC learnable by Lemma~\ref{lem:blocks}(ii). For a learner that uses $m\leq q_j$ examples, the argument for \eqref{lim:finite-block-lower} with $M=M_j$ uses only targets in $\cV_j$, and gives
\[
\sup_{h^\star\in\cV_j}\ \sup_{\text{data laws}}\ \E\bigl[L_{P,h^\star}(\widehat g)\bigr]
\geq\frac1{s_j}\Bigl(1-\frac1{M_j}\Bigr)^m
\geq\frac1{s_j}\Bigl(1-\frac m{M_j}\Bigr)
\geq\frac1{2s_j}.
\]
An $(\eps_j,\delta_j)$-PAC guarantee would instead give expected error at most $\eps_j+\delta_j=1/(4s_j)$, by \eqref{eq:pac-expect}. Hence $m_\H(\eps_j,\delta_j)>q_j$. Since $d_0=\mathrm D(\H)$ is finite, choosing $j\geq\max\{1,d_0\}$ yields
\[
m_\H(\eps_j,\delta_j)>q_j>\Psi(d_0,\eps_j,\delta_j),
\]
which contradicts \eqref{lim:integer-upper}.
\end{proof}

\subsection{Scale-Sensitive Impossibility}\label{sec:limits-scale-sensitive}



\begin{theorem}[Scale-sensitive impossibility]\label{thm:scale-sensitive}
Under one-correct-label feedback, there are countable input and label spaces and countable classes $\cG\subseteq\cK_N\subseteq\cK_\infty$ ($N\in\N$) with the following properties.
\begin{enumerate}[(i)]
\item A single predictor has error zero for every target in $\cG$, so $\cG$ is learnable from zero examples.
\item Each $\cK_N$ is realizably PAC learnable, and $m_{\cK_N}(1/16,1/16)>\lfloor N/2\rfloor$ for $N\geq2$.
\item The class $\cK_\infty$ is not realizably PAC learnable, and $\RiskMin_m(\cK_\infty)=1/2$ for every integer $m\geq0$.
\item Every scale-sensitive profile $\boldsymbol{\mathrm D}$ whose coordinates have the finite-witness property and are relabeling invariant satisfies $\boldsymbol{\mathrm D}(\cG)=\boldsymbol{\mathrm D}(\cK_N)=\boldsymbol{\mathrm D}(\cK_\infty)$ for all $N$.
\end{enumerate}
Consequently, no such profile characterizes realizable PAC learnability, and no finite-valued function of such a profile bounds the sample complexity on learnable classes.
\end{theorem}

Each target accepts, at every input $x$, one binary label $\ind\{x\in F\}$, for a finite set $F$ of inputs, together with all but finitely many auxiliary labels $2,3,\ldots$, and the accepted auxiliary labels are the same at every input. Hence any finite set of targets has a common auxiliary label that is correct everywhere, and the relabeling that swaps this label with $2$ maps the finite set into the class $\cG$ of targets that accept $2$. The class $\cG$ is learnable from zero examples, and by relabeling invariance and finite witnesses every coordinate of the profile takes the same value on $\cG$ as on the whole class.

The whole class, in contrast, is not learnable from demonstrations. The demonstrator may show only binary labels, which are fair bits at unseen inputs, so a binary prediction there is wrong half of the time. An auxiliary prediction does not help either, because the target may exclude, after the learner is fixed, the auxiliary labels that the learner is likely to predict. Adding to $\cG$ the targets with $F\subseteq[N]$ gives the intermediate classes $\cK_N$, which are learnable but whose sample complexity grows with $N$.

\paragraph{Construction: cofinite correct-label sets.}
Take $\X=\N$ and $\Y=\N_0$, and let $W=\{2,3,\ldots\}$ be the set of \emph{auxiliary} labels. For a finite set $F\subseteq\N$ and a finite set $B\subseteq W$, define
\begin{equation}\label{lim:cofinite-target}
h_{F,B}(x)=\bigl\{\ind\{x\in F\}\bigr\}\cup(W\setminus B),
\end{equation}
and set
\begin{equation}\label{lim:cofinite-classes}
\begin{aligned}
\cK_\infty&=\{h_{F,B}:\ F\subseteq\N\text{ and }B\subseteq W\text{ both finite}\},\\
\cG&=\{h_{F,B}\in\cK_\infty:\ 2\notin B\},\\
\cK_N&=\cG\cup\{h_{F,B}\in\cK_\infty:\ F\subseteq[N]\}.
\end{aligned}
\end{equation}
These classes are countable, every correct-label set has finite complement, and $\bigcup_{N\geq1}\cK_N=\cK_\infty$. The pair $(F,B)$ is determined by $h_{F,B}$, since $F=\{x:1\in h_{F,B}(x)\}$ and $B=W\setminus h_{F,B}(x)$ for every $x$.

\begin{proof}[Proof of Theorem~\ref{thm:scale-sensitive}]
\textbf{Identical profiles.}\ Let $\cV=\{h_{F_1,B_1},\ldots,h_{F_r,B_r}\}\subseteq\cK_\infty$ be finite and nonempty. Choose $y\in W\setminus(\{2\}\cup\bigcup_iB_i)$, and let $\pi$ swap $2$ and $y$ and fix all other labels. Then $\pi h_{F_i,B_i}=h_{F_i,\pi(B_i)}$ and $2\notin\pi(B_i)$, so $\pi\cV\subseteq\cG$. For every relabeling invariant parameter $\mathrm D$ with the finite-witness property, it follows that
\[
\mathrm D(\cV)=\mathrm D(\pi\cV)\leq\mathrm D(\cG).
\]
Taking the supremum over all finite nonempty $\cV\subseteq\cK_\infty$ and using \eqref{lim:finite-witness} gives $\mathrm D(\cK_\infty)\leq\mathrm D(\cG)$, and the inclusions $\cG\subseteq\cK_N\subseteq\cK_\infty$ together with monotonicity give equality for every $\cK_N$. Applying this to each coordinate $\mathrm D_\alpha$ proves part~(iv).

\paragraph{Learnability of $\cG$ and $\cK_N$.}
The constant predictor $2$ has error zero for every target in $\cG$, which proves part~(i). For $\cK_N$, consider the deterministic learner that outputs
\[
\begin{cases}
\text{the constant }a, & \text{if $a$ is the first auxiliary label in the sample},\\
\text{the constant }2, & \text{if no auxiliary label appears, but some $(x,1)$ with $x>N$ does},\\
g_T, & \text{otherwise},
\end{cases}
\]
where $g_T(x)=\ind\{x\in T\}$ and $T\subseteq[N]$ is the set of inputs demonstrated with label $1$. Let the target be $h_{F,B}\in\cK_N$. In the first case, $a$ is correct at some input, so $a\notin B$ and $a$ is correct at every input. In the second case, $x\in F$ and $x>N$, so $F\not\subseteq[N]$; hence $h_{F,B}\in\cG$, and $2$ is correct at every input. In the last case, every demonstrated label is binary and hence equals $\ind\{x\in F\}$, so $g_T$ agrees with all demonstrations, and $g_T$ lies in the fixed family $\cQ_N=\{g_T:T\subseteq[N]\}$ of size $2^N$. Since a demonstrated label is correct, an invalid prediction never agrees with it; so a function $g\in\cQ_N$ of error $r>\eps$ agrees with each demonstration with probability at most $1-r$, and
\[
\Pr\bigl[g(x_i)=y_i\text{ for all }i\in[m]\bigr]\leq(1-r)^m\leq e^{-m\eps}.
\]
A union bound over $\cQ_N$ bounds the failure probability by $2^Ne^{-m\eps}$, uniformly over all targets in $\cK_N$ and all data laws. Hence $m_{\cK_N}(\eps,\delta)\leq\lceil(N\log2+\log(1/\delta))/\eps\rceil<\infty$.

\paragraph{Lower bounds for $\cK_N$ and $\cK_\infty$.}
Fix a learner that uses $m$ examples, an integer $N\geq2$, and $\eta>0$. Choose $F\subseteq[N]$ by independent fair membership bits, and let $\D_F$ be the joint law of $(x,\ind\{x\in F\})$ for $x$ uniform on $[N]$; for every finite $B\subseteq W$, $\D_F$ is compatible with $h_{F,B}$. Draw $\sigma\sim\D_F^m$, an independent test input $x\sim\Unif([N])$, and the seed $\omega$, and write $Z=\widehat g(x)$ for the learner's prediction at $x$. Since the joint law of $(F,\sigma,x,\omega)$ does not depend on $B$, we may choose $B$ after the learner. As $Z$ takes values in $\N_0$, countable additivity gives an integer $L\geq2$ with $\Pr[Z>L]\leq\eta$, and we set $B=\{2,\ldots,L\}$, so that
\begin{equation}\label{lim:auxiliary-tail}
\Pr[Z\in W\setminus B]\leq\eta.
\end{equation}

Let $E$ be the event that the test input does not appear in $\sigma$; then $\Pr[E]=(1-1/N)^m$. Conditional on $E$, $\sigma$, $x$, and $\omega$, the unseen bit $\ind\{x\in F\}$ is fair, so $Z$ differs from it with probability at least $1/2$ (exactly $1/2$ if $Z$ is binary, and $1$ if $Z$ is auxiliary). A prediction that differs from the bit is valid only if $Z\in W\setminus B$. Therefore, with $P$ the uniform distribution on $[N]$, \eqref{lim:auxiliary-tail} gives
\[
\E_{F,\sigma,\omega}\bigl[L_{P,h_{F,B}}(\widehat g)\bigr]
\geq\frac{\Pr[E]}2-\Pr[Z\in W\setminus B]
\geq\frac12\Bigl(1-\frac1N\Bigr)^m-\eta.
\]
Hence some $h_{F,B}\in\cK_N$ with $F\subseteq[N]$ has expected error at least this large. Letting $\eta\to0$ and taking the infimum over learners gives
\begin{equation}\label{lim:cofinite-risk-lower}
\RiskMin_m(\cK_N)\geq\frac12\Bigl(1-\frac1N\Bigr)^m.
\end{equation}
For $m\leq\lfloor N/2\rfloor$, Bernoulli's inequality bounds the right-hand side from below by $1/4$, whereas a $(1/16,1/16)$-PAC guarantee would give expected error at most $1/8$ by \eqref{eq:pac-expect}. This proves $m_{\cK_N}(1/16,1/16)>\lfloor N/2\rfloor$ and completes part~(ii).

Since $\cK_N\subseteq\cK_\infty$, letting $N\to\infty$ in \eqref{lim:cofinite-risk-lower} gives $\RiskMin_m(\cK_\infty)\geq1/2$. Conversely, the learner that ignores the sample and outputs a uniformly random constant predictor in $\{0,1\}$ has expected error exactly $1/2$ for every target, since exactly one of $0$ and $1$ is correct at each input. Hence $\RiskMin_m(\cK_\infty)=1/2$ for every $m$, and $\cK_\infty$ is not realizably PAC learnable, which proves part~(iii).

Finally, by part~(iv) the common profile does not distinguish the learnable class $\cG$ from the unlearnable class $\cK_\infty$. Nor does it bound the sample complexity: at $(\eps,\delta)=(1/16,1/16)$ such a bound would take the same value on every $\cK_N$, contradicting part~(ii) as $N\to\infty$.
\end{proof}

\begin{remark}[Full-set feedback for the cofinite construction]\label{rem:cofinite-full}
The scale-sensitive counterexample is specific to one-correct-label feedback. Every full correct-label set of $h_{F,B}$ contains an auxiliary label $a\in W\setminus B$, and such a label is correct at every input. Thus one observed set suffices: the learner that outputs its least auxiliary label at every input has error zero for every target in $\cK_\infty$. The scale-free obstruction of Theorem~\ref{thm:scale-free}, in contrast, holds under full-set feedback as well.
\end{remark}

\clearpage
\phantomsection
\addcontentsline{toc}{section}{References}
\bibliography{references}

@InProceedings{pmlr-v247-asilis24a,
  title = 	 {Regularization and Optimal Multiclass Learning},
  author =       {Asilis, Julian and Devic, Siddartha and Dughmi, Shaddin and Sharan, Vatsal and Teng, Shang-Hua},
  booktitle = 	 {Proceedings of Thirty Seventh Conference on Learning Theory},
  pages = 	 {260--310},
  year = 	 {2024},
  editor = 	 {Agrawal, Shipra and Roth, Aaron},
  volume = 	 {247},
  series = 	 {Proceedings of Machine Learning Research},
  month = 	 {30 Jun--03 Jul},
  publisher =    {PMLR},
  url = 	 {https://proceedings.mlr.press/v247/asilis24a.html}
}

@book{vapnik1974theory,
  author    = {Vapnik, Vladimir N. and Chervonenkis, Alexey Ya.},
  title     = {Theory of Pattern Recognition},
  publisher = {Nauka, Moscow},
  year      = {1974}
}

@article{vapnik1971uniform,
  author    = {Vapnik, Vladimir N. and Chervonenkis, Alexey Ya.},
  title     = {On the Uniform Convergence of Relative Frequencies of Events to Their Probabilities},
  journal   = {Theory of Probability \& Its Applications},
  volume    = {16},
  number    = {2},
  pages     = {264--280},
  year      = {1971},
  doi       = {10.1137/1116025}
}

@article{valiant1984theory,
  author    = {Valiant, Leslie G.},
  title     = {A Theory of the Learnable},
  journal   = {Communications of the ACM},
  volume    = {27},
  number    = {11},
  pages     = {1134--1142},
  year      = {1984}
}

@article{blumer1989learnability,
  author    = {Blumer, Anselm and Ehrenfeucht, Andrzej and Haussler, David and Warmuth, Manfred K.},
  title     = {Learnability and the {V}apnik-{C}hervonenkis Dimension},
  journal   = {Journal of the ACM},
  volume    = {36},
  number    = {4},
  pages     = {929--965},
  year      = {1989},
  doi       = {10.1145/76359.76371}
}

@article{natarajan1989learning,
  author    = {Natarajan, Balas K.},
  title     = {On Learning Sets and Functions},
  journal   = {Machine Learning},
  volume    = {4},
  number    = {1},
  pages     = {67--97},
  year      = {1989},
  doi       = {10.1007/BF00114804}
}

@article{ben1995characterizations,
  author    = {Ben-David, Shai and Cesa-Bianchi, Nicol{\`o} and Haussler, David and Long, Philip M.},
  title     = {Characterizations of Learnability for Classes of $\{0,\ldots,n\}$-Valued Functions},
  journal   = {Journal of Computer and System Sciences},
  volume    = {50},
  number    = {1},
  pages     = {74--86},
  year      = {1995},
  doi       = {10.1006/jcss.1995.1008}
}

@article{haussler1994predicting,
  author    = {Haussler, David and Littlestone, Nick and Warmuth, Manfred K.},
  title     = {Predicting $\{0,1\}$-Functions on Randomly Drawn Points},
  journal   = {Information and Computation},
  volume    = {115},
  number    = {2},
  pages     = {248--292},
  year      = {1994},
  doi       = {10.1006/inco.1994.1097}
}

@inproceedings{rubinstein2006shifting,
  author    = {Rubinstein, Benjamin I. P. and Bartlett, Peter L. and Rubinstein, J. Hyam},
  title     = {Shifting, One-Inclusion Mistake Bounds and Tight Multiclass Expected Risk Bounds},
  booktitle = {Advances in Neural Information Processing Systems},
  volume    = {19},
  pages     = {1193--1200},
  year      = {2006}
}

@inproceedings{daniely2011multiclass,
  author    = {Daniely, Amit and Sabato, Sivan and Ben-David, Shai and Shalev-Shwartz, Shai},
  title     = {Multiclass Learnability and the {ERM} Principle},
  booktitle = {Proceedings of the 24th Annual Conference on Learning Theory},
  series    = {Proceedings of Machine Learning Research},
  volume    = {19},
  pages     = {207--232},
  publisher = {PMLR},
  year      = {2011}
}

@inproceedings{daniely2014optimal,
  author    = {Daniely, Amit and Shalev-Shwartz, Shai},
  title     = {Optimal Learners for Multiclass Problems},
  booktitle = {Proceedings of The 27th Conference on Learning Theory},
  series    = {Proceedings of Machine Learning Research},
  volume    = {35},
  pages     = {287--316},
  publisher = {PMLR},
  year      = {2014}
}

@inproceedings{brukhim2022characterization,
  author    = {Brukhim, Nataly and Carmon, Daniel and Dinur, Irit and Moran, Shay and Yehudayoff, Amir},
  title     = {A Characterization of Multiclass Learnability},
  booktitle = {2022 IEEE 63rd Annual Symposium on Foundations of Computer Science (FOCS)},
  pages     = {943--955},
  publisher = {IEEE},
  year      = {2022},
  doi       = {10.1109/FOCS54457.2022.00093}
}

@inproceedings{charikar2023characterization,
  author    = {Charikar, Moses and Pabbaraju, Chirag},
  title     = {A Characterization of List Learnability},
  booktitle = {Proceedings of the 55th Annual ACM Symposium on Theory of Computing (STOC)},
  pages     = {1713--1726},
  publisher = {ACM},
  year      = {2023},
  doi       = {10.1145/3564246.3585190}
}

@inproceedings{alon2021theory,
  author    = {Alon, Noga and Hanneke, Steve and Holzman, Ron and Moran, Shay},
  title     = {A Theory of {PAC} Learnability of Partial Concept Classes},
  booktitle = {2021 IEEE 62nd Annual Symposium on Foundations of Computer Science (FOCS)},
  pages     = {658--671},
  publisher = {IEEE},
  year      = {2021},
  doi       = {10.1109/FOCS52979.2021.00070}
}

@inproceedings{cohen2025sample,
  author        = {Cohen, Alon and Erez, Liad and Hanneke, Steve and Koren, Tomer and Mansour, Yishay and Moran, Shay and Zhang, Qian},
  title         = {Sample Complexity of Agnostic Multiclass Classification: {N}atarajan Dimension Strikes Back},
  booktitle     = {Proceedings of the 58th Annual ACM Symposium on Theory of Computing (STOC)},
  pages         = {722--733},
  publisher     = {ACM},
  year          = {2026},
  doi           = {10.1145/3798129.3800787},
  eprint        = {2511.12659},
  archivePrefix = {arXiv}
}

@techreport{littlestone1986relating,
  author      = {Littlestone, Nick and Warmuth, Manfred K.},
  title       = {Relating Data Compression and Learnability},
  institution = {Department of Computer and Information Sciences, University of California at Santa Cruz},
  type        = {Technical Report},
  year        = {1986}
}

@article{moran2016sample,
  author    = {Moran, Shay and Yehudayoff, Amir},
  title     = {Sample Compression Schemes for {VC} Classes},
  journal   = {Journal of the ACM},
  volume    = {63},
  number    = {3},
  pages     = {21:1--21:10},
  year      = {2016},
  doi       = {10.1145/2890490}
}

@inproceedings{david2016supervised,
  author    = {David, Ofir and Moran, Shay and Yehudayoff, Amir},
  title     = {Supervised Learning through the Lens of Compression},
  booktitle = {Advances in Neural Information Processing Systems},
  volume    = {29},
  pages     = {2784--2792},
  year      = {2016}
}

@article{freund1997decision,
  author    = {Freund, Yoav and Schapire, Robert E.},
  title     = {A Decision-Theoretic Generalization of On-Line Learning and an Application to Boosting},
  journal   = {Journal of Computer and System Sciences},
  volume    = {55},
  number    = {1},
  pages     = {119--139},
  year      = {1997},
  doi       = {10.1006/jcss.1997.1504}
}

@inproceedings{brukhim2023improper,
  author    = {Brukhim, Nataly and Hanneke, Steve and Moran, Shay},
  title     = {Improper Multiclass Boosting},
  booktitle = {Proceedings of Thirty Sixth Conference on Learning Theory},
  series    = {Proceedings of Machine Learning Research},
  volume    = {195},
  pages     = {5433--5452},
  publisher = {PMLR},
  year      = {2023}
}

@inproceedings{bressan2025fine,
  author    = {Bressan, Marco and Brukhim, Nataly and Cesa-Bianchi, Nicol{\`o} and Esposito, Emmanuel and Mansour, Yishay and Moran, Shay and Thiessen, Maximilian},
  title     = {A Fine-grained Characterization of {PAC} Learnability},
  booktitle = {Proceedings of Thirty Eighth Conference on Learning Theory},
  series    = {Proceedings of Machine Learning Research},
  volume    = {291},
  pages     = {641--676},
  publisher = {PMLR},
  year      = {2025}
}

@inproceedings{aden2023optimal,
  author    = {Aden-Ali, Ishaq and Cherapanamjeri, Yeshwanth and Shetty, Abhishek and Zhivotovskiy, Nikita},
  title     = {Optimal {PAC} Bounds without Uniform Convergence},
  booktitle = {2023 IEEE 64th Annual Symposium on Foundations of Computer Science (FOCS)},
  pages     = {1203--1223},
  publisher = {IEEE},
  year      = {2023}
}

@article{auer2002nonstochastic,
  author    = {Auer, Peter and Cesa-Bianchi, Nicol{\`o} and Freund, Yoav and Schapire, Robert E.},
  title     = {The Nonstochastic Multiarmed Bandit Problem},
  journal   = {SIAM Journal on Computing},
  volume    = {32},
  number    = {1},
  pages     = {48--77},
  year      = {2002},
  doi       = {10.1137/S0097539701398375}
}

@inproceedings{kakade2008efficient,
  author    = {Kakade, Sham M. and Shalev-Shwartz, Shai and Tewari, Ambuj},
  title     = {Efficient Bandit Algorithms for Online Multiclass Prediction},
  booktitle = {Proceedings of the 25th International Conference on Machine Learning (ICML)},
  pages     = {440--447},
  publisher = {ACM},
  year      = {2008},
  doi       = {10.1145/1390156.1390212}
}

@inproceedings{daniely2013price,
  author    = {Daniely, Amit and Helbertal, Tom},
  title     = {The Price of Bandit Information in Multiclass Online Classification},
  booktitle = {Proceedings of the 26th Annual Conference on Learning Theory},
  series    = {Proceedings of Machine Learning Research},
  volume    = {30},
  pages     = {93--104},
  publisher = {PMLR},
  year      = {2013}
}

@article{long2020new,
  author    = {Long, Philip M.},
  title     = {New Bounds on the Price of Bandit Feedback for Mistake-Bounded Online Multiclass Learning},
  journal   = {Theoretical Computer Science},
  volume    = {808},
  pages     = {159--163},
  year      = {2020},
  doi       = {10.1016/j.tcs.2019.11.017}
}

@inproceedings{raman2024multiclass,
  author    = {Raman, Ananth and Raman, Vinod and Subedi, Unique and Mehalel, Idan and Tewari, Ambuj},
  title     = {Multiclass Online Learnability under Bandit Feedback},
  booktitle = {Proceedings of The 35th International Conference on Algorithmic Learning Theory},
  series    = {Proceedings of Machine Learning Research},
  volume    = {237},
  pages     = {997--1012},
  publisher = {PMLR},
  year      = {2024}
}

@inproceedings{erez2024real,
  author    = {Erez, Liad and Cohen, Alon and Koren, Tomer and Mansour, Yishay and Moran, Shay},
  title     = {The Real Price of Bandit Information in Multiclass Classification},
  booktitle = {Proceedings of Thirty Seventh Conference on Learning Theory},
  series    = {Proceedings of Machine Learning Research},
  volume    = {247},
  pages     = {1573--1598},
  publisher = {PMLR},
  year      = {2024}
}

@inproceedings{erez2024fast,
  author    = {Erez, Liad and Cohen, Alon and Koren, Tomer and Mansour, Yishay and Moran, Shay},
  title     = {Fast Rates for Bandit {PAC} Multiclass Classification},
  booktitle = {Advances in Neural Information Processing Systems},
  volume    = {37},
  pages     = {75152--75176},
  year      = {2024}
}

@inproceedings{filmus2024bandit,
  author    = {Filmus, Yuval and Hanneke, Steve and Mehalel, Idan and Moran, Shay},
  title     = {Bandit-Feedback Online Multiclass Classification: Variants and Tradeoffs},
  booktitle = {Advances in Neural Information Processing Systems},
  volume    = {37},
  pages     = {129605--129639},
  year      = {2024}
}

@inproceedings{hanneke2025universal,
  author    = {Hanneke, Steve and Shaeiri, Amirreza and Zhang, Qian},
  title     = {Universal Rates for Multiclass Learning with Bandit Feedback},
  booktitle = {Proceedings of Thirty Eighth Conference on Learning Theory},
  series    = {Proceedings of Machine Learning Research},
  volume    = {291},
  pages     = {2704--2756},
  publisher = {PMLR},
  year      = {2025}
}

@inproceedings{hanneke2026pac,
  author    = {Hanneke, Steve and Meng, Qinglin and Moran, Shay and Shaeiri, Amirreza},
  title     = {{PAC} Learning with Bandit Feedback: Sharp Sample Complexity in the Realizable Setting},
  booktitle = {Advances in Neural Information Processing Systems},
  year      = {2026},
  note      = {To appear. arXiv:2605.25678},
  eprint        = {2605.25678},
  archivePrefix = {arXiv}
}

@misc{zhang2026bandit,
  author        = {Zhang, Guangjian and Yang, Zhengyi and Wang, Xiaoyang},
  title         = {Bandit Multiclass {PAC} Learning: Structural Baselines and a Confidence Direct Sum},
  howpublished  = {arXiv preprint arXiv:2609.29694},
  year          = {2026},
  eprint        = {2609.29694},
  archivePrefix = {arXiv}
}

@inproceedings{raman2024online,
  author    = {Raman, Vinod and Subedi, Unique and Tewari, Ambuj},
  title     = {Online Learning with Set-Valued Feedback},
  booktitle = {Proceedings of Thirty Seventh Conference on Learning Theory},
  series    = {Proceedings of Machine Learning Research},
  volume    = {247},
  pages     = {4381--4412},
  publisher = {PMLR},
  year      = {2024}
}

@misc{pour2026learning,
  author        = {Pour, Alireza F. and Mansouri, Farnam and Ben-David, Shai},
  title         = {Learning with Multiple Correct Answers -- Regret Bounds under Different Feedback Models},
  howpublished  = {arXiv preprint arXiv:2602.09402},
  year          = {2026},
  eprint        = {2602.09402},
  archivePrefix = {arXiv}
}

@misc{shao2026partial,
  author        = {Shao, Shihao and Fang, Cong and Lin, Zhouchen and Tao, Dacheng},
  title         = {Partial Feedback Online Learning},
  howpublished  = {arXiv preprint arXiv:2601.21462},
  year          = {2026},
  eprint        = {2601.21462},
  archivePrefix = {arXiv}
}

@inproceedings{kosoy2026ambiguous,
  author    = {Kosoy, Vanessa},
  title     = {Ambiguous Online Learning},
  booktitle = {Proceedings of Thirty Ninth Conference on Learning Theory},
  series    = {Proceedings of Machine Learning Research},
  volume    = {336},
  pages     = {4229--4266},
  publisher = {PMLR},
  year      = {2026}
}

@inproceedings{joshi2026learning,
  author    = {Joshi, Nirmit and Li, Gene and Bhandari, Siddharth and Kasiviswanathan, Shiva Prasad and Ma, Cong and Srebro, Nathan},
  title     = {Learning to Answer from Correct Demonstrations},
  booktitle = {International Conference on Learning Representations},
  year      = {2026}
}

@inproceedings{kleinberg2024language,
  author    = {Kleinberg, Jon and Mullainathan, Sendhil},
  title     = {Language Generation in the Limit},
  booktitle = {Advances in Neural Information Processing Systems},
  volume    = {37},
  pages     = {66058--66079},
  year      = {2024}
}

@article{cour2011learning,
  author    = {Cour, Timothee and Sapp, Ben and Taskar, Ben},
  title     = {Learning from Partial Labels},
  journal   = {Journal of Machine Learning Research},
  volume    = {12},
  pages     = {1501--1536},
  year      = {2011}
}

@inproceedings{liu2014learnability,
  author    = {Liu, Li-Ping and Dietterich, Thomas G.},
  title     = {Learnability of the Superset Label Learning Problem},
  booktitle = {Proceedings of the 31st International Conference on Machine Learning},
  series    = {Proceedings of Machine Learning Research},
  volume    = {32},
  pages     = {1629--1637},
  publisher = {PMLR},
  year      = {2014}
}

@article{zhang2014review,
  author    = {Zhang, Min-Ling and Zhou, Zhi-Hua},
  title     = {A Review on Multi-Label Learning Algorithms},
  journal   = {IEEE Transactions on Knowledge and Data Engineering},
  volume    = {26},
  number    = {8},
  pages     = {1819--1837},
  year      = {2014}
}

@inproceedings{malach2024auto,
  author    = {Malach, Eran},
  title     = {Auto-Regressive Next-Token Predictors are Universal Learners},
  booktitle = {Proceedings of the 41st International Conference on Machine Learning},
  series    = {Proceedings of Machine Learning Research},
  volume    = {235},
  pages     = {34417--34431},
  publisher = {PMLR},
  year      = {2024}
}

@inproceedings{joshi2025theory,
  author    = {Joshi, Nirmit and Vardi, Gal and Block, Adam and Goel, Surbhi and Li, Zhiyuan and Misiakiewicz, Theodor and Srebro, Nathan},
  title     = {A Theory of Learning with Autoregressive Chain of Thought},
  booktitle = {Proceedings of Thirty Eighth Conference on Learning Theory},
  series    = {Proceedings of Machine Learning Research},
  volume    = {291},
  pages     = {3161--3212},
  publisher = {PMLR},
  year      = {2025}
}

@misc{hanneke2026sample,
  author        = {Hanneke, Steve and Mehalel, Idan and Moran, Shay},
  title         = {Sample Complexity of Autoregressive Reasoning: Chain-of-Thought vs.\ End-to-End},
  howpublished  = {arXiv preprint arXiv:2604.12013},
  year          = {2026},
  eprint        = {2604.12013},
  archivePrefix = {arXiv}
}

@inproceedings{lechner2024inherent,
  author    = {Lechner, Tosca and Ben-David, Shai},
  title     = {Inherent Limitations of Dimensions for Characterizing Learnability of Distribution Classes},
  booktitle = {Proceedings of Thirty Seventh Conference on Learning Theory},
  series    = {Proceedings of Machine Learning Research},
  volume    = {247},
  pages     = {3353--3374},
  publisher = {PMLR},
  year      = {2024}
}

@inproceedings{lechner2026no,
  author    = {Lechner, Tosca and Ben-David, Shai},
  title     = {No Scale Sensitive Dimension for Distribution Learning},
  booktitle = {Proceedings of the 37th International Conference on Algorithmic Learning Theory},
  series    = {Proceedings of Machine Learning Research},
  volume    = {313},
  pages     = {1--27},
  publisher = {PMLR},
  year      = {2026}
}

@article{bendavid2019learnability,
  author    = {Ben-David, Shai and Hrube{\v{s}}, Pavel and Moran, Shay and Shpilka, Amir and Yehudayoff, Amir},
  title     = {Learnability Can Be Undecidable},
  journal   = {Nature Machine Intelligence},
  volume    = {1},
  number    = {1},
  pages     = {44--48},
  year      = {2019}
}

@inproceedings{hanneke2023bandit,
  author    = {Hanneke, Steve and Yang, Liu},
  title     = {Bandit Learnability can be Undecidable},
  booktitle = {Proceedings of Thirty Sixth Conference on Learning Theory},
  series    = {Proceedings of Machine Learning Research},
  volume    = {195},
  pages     = {5813--5849},
  publisher = {PMLR},
  year      = {2023}
}

@article{hoeffding1963probability,
  author    = {Hoeffding, Wassily},
  title     = {Probability Inequalities for Sums of Bounded Random Variables},
  journal   = {Journal of the American Statistical Association},
  volume    = {58},
  number    = {301},
  pages     = {13--30},
  year      = {1963}
}

@article{vonneumann1928theorie,
  author    = {von Neumann, John},
  title     = {Zur {T}heorie der {G}esellschaftsspiele},
  journal   = {Mathematische Annalen},
  volume    = {100},
  number    = {1},
  pages     = {295--320},
  year      = {1928}
}

@book{odonnell2014analysis,
  author    = {O'Donnell, Ryan},
  title     = {Analysis of {B}oolean Functions},
  publisher = {Cambridge University Press},
  year      = {2014}
}

@article{mubayi2009bipartite,
  author    = {Mubayi, Dhruv and Vishwanathan, Sundar},
  title     = {Bipartite Coverings and the Chromatic Number},
  journal   = {The Electronic Journal of Combinatorics},
  volume    = {16},
  number    = {1},
  pages     = {N34},
  year      = {2009},
  doi       = {10.37236/272}
}

@article{huang2012counterexample,
  author    = {Huang, Hao and Sudakov, Benny},
  title     = {A Counterexample to the {A}lon-{S}aks-{S}eymour Conjecture and Related Problems},
  journal   = {Combinatorica},
  volume    = {32},
  number    = {2},
  pages     = {205--219},
  year      = {2012}
}

@inproceedings{balodis2021unambiguous,
  author    = {Balodis, Kaspars and Ben-David, Shalev and G{\"o}{\"o}s, Mika and Jain, Siddhartha and Kothari, Robin},
  title     = {Unambiguous {DNF}s and {A}lon-{S}aks-{S}eymour},
  booktitle = {2021 IEEE 62nd Annual Symposium on Foundations of Computer Science (FOCS)},
  pages     = {116--124},
  publisher = {IEEE},
  year      = {2021},
  doi       = {10.1109/FOCS52979.2021.00020}
}

@misc{fox2026note,
  author        = {Fox, Jacob},
  title         = {A Note on the {A}lon-{S}aks-{S}eymour Problem},
  howpublished  = {arXiv preprint arXiv:2605.28915},
  year          = {2026},
  eprint        = {2605.28915},
  archivePrefix = {arXiv}
}

@misc{pabbaraju2026optimal,
  author        = {Pabbaraju, Chirag},
  title         = {Optimal Unambiguous {DNF}s and {A}lon-{S}aks-{S}eymour},
  howpublished  = {arXiv preprint arXiv:2608.02533},
  year          = {2026},
  eprint        = {2608.02533},
  archivePrefix = {arXiv}
}

@book{kechris1995classical,
  author    = {Kechris, Alexander S.},
  title     = {Classical Descriptive Set Theory},
  series    = {Graduate Texts in Mathematics},
  volume    = {156},
  publisher = {Springer-Verlag},
  address   = {New York},
  year      = {1995},
  doi       = {10.1007/978-1-4612-4190-4}
}

@inproceedings{freund1997using,
  title     = {Using and Combining Predictors That Specialize},
  author    = {Freund, Yoav and Schapire, Robert E. and Singer, Yoram and Warmuth, Manfred K.},
  booktitle = {Proceedings of the 29th Annual ACM Symposium on Theory of Computing (STOC)},
  pages     = {334--343},
  year      = {1997}
}

@inproceedings{pmlr-v178-hopkins22a,
  author    = {Hopkins, Max and Kane, Daniel M. and Lovett, Shachar and Mahajan, Gaurav},
  title     = {Realizable Learning is All You Need},
  booktitle = {Proceedings of Thirty Fifth Conference on Learning Theory},
  series    = {Proceedings of Machine Learning Research},
  volume    = {178},
  pages     = {3015--3069},
  publisher = {PMLR},
  year      = {2022}
}

@inproceedings{hanneke2025representation,
  author    = {Hanneke, Steve and Meng, Qinglin and Shaeiri, Amirreza},
  title     = {Representation Preserving Multiclass Agnostic to Realizable Reduction},
  booktitle = {Proceedings of the 42nd International Conference on Machine Learning},
  series    = {Proceedings of Machine Learning Research},
  volume    = {267},
  pages     = {21995--22008},
  publisher = {PMLR},
  year      = {2025}
}

@inproceedings{hanneke2024multiclass,
  author    = {Hanneke, Steve and Raman, Vinod and Shaeiri, Amirreza and Subedi, Unique},
  title     = {Multiclass Transductive Online Learning},
  booktitle = {Advances in Neural Information Processing Systems},
  volume    = {37},
  pages     = {60403--60428},
  year      = {2024}
}

@inproceedings{hanneke2025list,
  author    = {Hanneke, Steve and Shaeiri, Amirreza},
  title     = {A Trichotomy for List Transductive Online Learning},
  booktitle = {Proceedings of the 42nd International Conference on Machine Learning},
  series    = {Proceedings of Machine Learning Research},
  volume    = {267},
  publisher = {PMLR},
  year      = {2025}
}

@inproceedings{hanneke2026adversary,
  author    = {Hanneke, Steve and Shaeiri, Amirreza},
  title     = {A Theory of Adversary-Directed Online Learning},
  booktitle = {Advances in Neural Information Processing Systems},
  year      = {2026}
}

@inproceedings{hanneke2025equivalent,
  author    = {Hanneke, Steve and Shaeiri, Amirreza and Wang, Hongao},
  title     = {For Universal Multiclass Online Learning, Bandit Feedback and Full Supervision are Equivalent},
  booktitle = {Proceedings of the 36th International Conference on Algorithmic Learning Theory},
  series    = {Proceedings of Machine Learning Research},
  volume    = {272},
  publisher = {PMLR},
  year      = {2025}
}

@inproceedings{hanneke2025nonuniform,
  author    = {Hanneke, Steve and Shaeiri, Amirreza and Wang, Hongao},
  title     = {Non-Uniform Multiclass Learning with Bandit Feedback},
  booktitle = {Advances in Neural Information Processing Systems},
  volume    = {38},
  year      = {2025}
}

@misc{hanneke2026optimalsauer,
  author        = {Hanneke, Steve and Meng, Qinglin and Moran, Shay and Shaeiri, Amirreza},
  title         = {An Optimal {S}auer Lemma Over $k$-ary Alphabets},
  howpublished  = {arXiv preprint arXiv:2604.12952},
  year          = {2026}
}

@misc{pabbaraju2026multiclass,
  author        = {Pabbaraju, Chirag},
  title         = {The Optimal Sample Complexity of Multiclass and List Learning},
  howpublished  = {arXiv preprint arXiv:2604.24749},
  year          = {2026}
}

\clearpage
\appendix
\section{Full-Set Learnability and the Joint DS Dimension Are Incomparable}\label{app:separations}

This appendix proves the last assertion of Theorem~\ref{thm:countable-full}: for full-set feedback and $\ell\geq2$, the joint DS dimension is neither necessary nor sufficient for learnability. We give two total classes with at most two correct labels per input, so both lie in the regime $\ell\geq2$ of that theorem. The first has joint DS dimension one but is not full-set PAC learnable (Theorem~\ref{thm:rank-example}); the second has infinite joint DS dimension but is full-set PAC learnable (Theorem~\ref{thm:large-ds-learnable}).

Both constructions use $\X=\N$ and a countable label space $\Y$ that contains $0$, $1$, $\bot$, and a fresh symbol $\langle t\rangle$ for each finite tuple $t$ used in the construction. We call these symbols \emph{tuple labels}; they are distinct from each other and from $0,1,\bot$, and a tuple label reveals its entries. In both classes a hypothesis accepts, at each input $i\leq n$, a bit $b_i$ and a tuple label that encodes $b_i$ together with some side information about the other inputs. The two constructions differ only in this side information. In the first, it couples the bits in pairs, which destroys DS witnesses but tells the learner almost nothing about the bit at an unseen input. In the second, it hands the learner a predictor that is consistent with the whole sample, but leaves each bit free to flip, which produces large witnesses.

\subsection{Joint DS Dimension One Does Not Imply Full-Set Learnability}\label{app:sep-rank}

Fix $n\geq2$, a vector $b\in\{0,1\}^n$, and integers $u_{ij}\in[n]$ for $1\leq i<j\leq n$. Define the offsets
\begin{equation}\label{app:eq:offset-identity}
r_{ij}=u_{ij},
\qquad
r_{ji}=u_{ij}+b_ib_j
\qquad(i<j),
\end{equation}
so that the two offsets of a pair differ exactly by $b_ib_j$, and write $r_i=(r_{ij})_{j\in[n]\setminus\{i\}}$, with the indices in increasing order. Define the total hypothesis
\begin{equation}\label{app:eq:offset-target}
h_{n,b,u}(i)=
\begin{cases}
\{b_i,\langle n,i,b_i,r_i\rangle\}, & i\leq n,\\
\{0\}, & i>n,
\end{cases}
\end{equation}
and let $\H^-$ consist of all these hypotheses, over every $n\geq2$ and all choices of $b$ and $u$.

Every correct label at an input $i\leq n$ determines $b_i$, since it is either $b_i$ itself or a tuple label whose third entry is $b_i$. Thus predicting a correct label at $i$ is as hard as guessing $b_i$. If the tuple labels recorded only $(n,i,b_i)$, the class would contain product witnesses of every size. The offsets destroy them: the tuple labels at two inputs $i<j$ reveal $r_{ij}$ and $r_{ji}$, hence the product $b_ib_j$, and this coupling of the two bits rules out DS witnesses on two inputs.

The coupling is of little use to a learner. It does not see the tuple label at the test input $i$, and the only trace of $b_i$ in its sample is the offset $r_{ji}=u_{ij}+b_i$ at observed inputs $j>i$ with $b_j=1$. When $u_{ij}$ is uniform on $[n]$, this offset reveals $b_i$ only if it equals $1$ or $n+1$, which happens with probability $1/n$. The lower bound therefore takes $n$ much larger than the sample size, so that the test input is unseen and its bit is almost uniform given the sample.

\begin{theorem}\label{thm:rank-example}
The class $\H^-$ satisfies
\[
\dDS(\H^-)=\dN(\H^-)=1,
\]
and the fixed list $\{0,1\}$ meets every correct-label set of $\H^-$. Nevertheless, for every sample size $m$, under full-set feedback,
\begin{equation}\label{app:eq:offset-minimax}
\inf_{\mathcal L}\ \sup_{h\in\H^-,\,P}\ \E\bigl[L_{P,h}(\mathcal L(\sigma))\bigr]=\frac12,
\end{equation}
where the infimum is over all learners $\mathcal L$ that receive $m$ examples, including randomized learners and randomized predictions. In particular, $\H^-$ is not full-set PAC learnable.
\end{theorem}

\begin{proof}
By Lemma~\ref{lem:orders}, the projection of a DS witness to two of its coordinates is again a DS witness, so for the dimensions it suffices to rule out DS witnesses on two inputs. We do this first for hypotheses with a common value of $n$, and then in general.

\paragraph{Two inputs and a fixed $n$.}
Fix $n$ and two inputs $i<j\leq n$, and suppose that the patterns on $(i,j)$ of the hypotheses $h_{n,b,u}$ with $(b,u)$ in a finite nonempty set $\Theta$ form a DS witness. Tildes carry over to offsets, so $\widetilde r$ is defined from $(\widetilde b,\widetilde u)$ by \eqref{app:eq:offset-identity}. Since $n$ is fixed, two members $(b,u),(\widetilde b,\widetilde u)$ of $\Theta$ have the same correct-label set at $i$ if and only if $(\widetilde b_i,\widetilde r_i)=(b_i,r_i)$. The \emph{DS condition at $j$} says that, for every $(b,u)\in\Theta$, the members that agree with $(b,u)$ at $i$ have no common correct label at $j$:
\begin{equation}\label{app:eq:offset-witness}
\bigcap_{\substack{(\widetilde b,\widetilde u)\in\Theta\\ (\widetilde b_i,\widetilde r_i)=(b_i,r_i)}}h_{n,\widetilde b,\widetilde u}(j)=\emptyset.
\end{equation}
The \emph{DS condition at $i$} says the same with the roles of $i$ and $j$ swapped: the members that agree with $(b,u)$ at $j$ have no common correct label at $i$.

Let $q=\max\{r_{ij}:(b,u)\in\Theta\}$. The idea is that, when $b_j=1$, changing $b_i$ from $1$ to $0$ while keeping the correct-label set at $j$, and hence the offset $r_{ji}=r_{ij}+b_i$, raises $r_{ij}$ by one. At the maximum $q$ this move is impossible, and the DS conditions propagate the obstruction to the other three bit pairs. We show in turn that no $(b,u)\in\Theta$ with $r_{ij}=q$ has $(b_i,b_j)$ equal to $(1,1)$, $(1,0)$, $(0,0)$, or $(0,1)$. Since $q$ is attained, this is a contradiction. In each case, fix such a $(b,u)$.

\emph{Case 1: $(b_i,b_j)=(1,1)$.} Then $r_{ji}=q+1$. Every member that agrees with $(b,u)$ at $j$ has $\widetilde b_j=1$ and $\widetilde r_{ij}+\widetilde b_i=\widetilde r_{ji}=q+1$, and since $\widetilde r_{ij}\leq q$, also $\widetilde b_i=1$. These members share the label $1$ at $i$, contradicting the DS condition at $i$.

\emph{Case 2: $(b_i,b_j)=(1,0)$.} Every member that agrees with $(b,u)$ at $i$ has $\widetilde b_i=1$ and $\widetilde r_{ij}=q$, so $\widetilde b_j=0$ by Case~1. These members share the label $0$ at $j$, contradicting the DS condition at $j$.

\emph{Case 3: $(b_i,b_j)=(0,0)$.} Then $r_{ji}=q$. Every member that agrees with $(b,u)$ at $j$ has $\widetilde b_j=0$, hence $\widetilde r_{ij}=\widetilde r_{ji}=q$, so $\widetilde b_i=0$ by Case~2. These members share the label $0$ at $i$, contradicting the DS condition at $i$.

\emph{Case 4: $(b_i,b_j)=(0,1)$.} Every member that agrees with $(b,u)$ at $i$ has $\widetilde b_i=0$ and $\widetilde r_{ij}=q$, so $\widetilde b_j=1$ by Case~3. These members share the label $1$ at $j$, contradicting the DS condition at $j$.

\paragraph{Two inputs in general.}
Now let $i<j$ be arbitrary inputs, and suppose that the patterns on $(i,j)$ of the hypotheses $h_{n,b,u}$ with $(n,b,u)$ in a finite nonempty set form a DS witness. We distinguish two cases.

If every triple has $n\geq j$, the tuple labels at both inputs reveal $n$, so patterns that agree at either input come from the same value of $n$. Hence the patterns with any one value of $n$ form a DS witness on $(i,j)$ by themselves, which is impossible by the fixed-$n$ case.

Otherwise some triple has $n<j$, and the DS condition at $j$ fails at it. Indeed, if $n<i$, its correct-label set at $i$ is $\{0\}$, so every member that agrees with it at $i$ also has $n<i$; if $i\leq n<j$, its tuple label at $i$ reveals $n$, so every member that agrees with it at $i$ has the same $n$. In both cases all these members have the correct-label set $\{0\}$ at $j$.

\paragraph{The dimensions.}
Thus $\H^-$ has no DS witness on two inputs, and $\dDS(\H^-)\leq1$. At a single input $i$, two hypotheses with $n\geq i$ and opposite bits $b_i$ have disjoint correct-label sets, since their tuple labels also differ. Hence $\dN(\H^-)\geq1$, and the inequality $\dN(\H^-)\leq\dDS(\H^-)$ of Lemma~\ref{lem:orders} gives the asserted equalities.

\paragraph{The upper bound in \eqref{app:eq:offset-minimax}.}
Every correct-label set of $\H^-$ contains exactly one of the labels $0$ and $1$, so a uniformly random constant predictor with value in $\{0,1\}$ has expected error exactly $1/2$ for every target and every distribution.

\paragraph{A random target.}
For the lower bound, fix $n$ and let $P_n$ be uniform on $[n]$. Choose
\[
B_i\sim\Unif(\{0,1\})\ \ (i\in[n]),
\qquad
U_{ij}\sim\Unif([n])\ \ (i<j),
\]
all independent, take the target $h_{n,B,U}$, and draw $m$ independent full-set examples from it. Write $R_{jk}$ for the offsets defined by \eqref{app:eq:offset-identity} with $B$ and $U$ in place of $b$ and $u$. Let $T$ be the set of distinct training inputs, and let $I$ be an independent test input with law $P_n$. On the event $\{I\notin T\}$, the sample reveals $B_j$ and the offsets $R_{jk}$, $k\neq j$, at each $j\in T$. Of these, only the offsets
\[
R_{jI}=U_{Ij}+B_I,\qquad j\in J=\{j\in T:\ j>I,\ B_j=1\},
\]
depend on $B_I$. Such an offset is uniform on $[n]$ if $B_I=0$ and uniform on $\{2,\ldots,n+1\}$ if $B_I=1$, so it is informative only outside $\{2,\ldots,n\}$. Define the event
\begin{equation}\label{app:eq:offset-interior}
E=\{I\notin T\}\cap\bigcap_{j\in J}\bigl\{R_{jI}\in\{2,\ldots,n\}\bigr\},
\end{equation}
which is determined by the sample and $I$.

Condition on the training inputs, on a test input $I\notin T$, and on the observed bits $B_j$, $j\in T$; this fixes $J$, and $B_I$ is still uniform. The block $(B_I,(U_{Ij})_{j\in J})$ consists of independent variables, and it is independent of all other observed offsets, which are functions of the bits $B_k$, $k\neq I$, and of the variables $U_{i'j'}$ outside this block. This gives two facts.
\begin{itemize}
\item[(a)] $\Pr[E]\geq(1-1/n)^{2m}$. Indeed, $\Pr[I\notin T]=(1-1/n)^m$. Given $I\notin T$ and either value of $B_I$, the at most $m$ offsets $R_{jI}$, $j\in J$, are conditionally independent, and each lies in $\{2,\ldots,n\}$ with probability $1-1/n$.
\item[(b)] On $E$, the bit $B_I$ is uniform on $\{0,1\}$ given the sample and $I$. Indeed, every observed value $R_{jI}\in\{2,\ldots,n\}$, $j\in J$, has likelihood $1/n$ under both values of $B_I$, and the rest of the sample does not depend on $B_I$. The same holds given also the independent randomness of the learner.
\end{itemize}

It remains to show that a correct prediction at $I$ reveals $B_I$. For a label $y$, let $\beta(y)\in\{0,1\}$ be the bit it points to: $\beta(1)=1$, $\beta(0)=\beta(\bot)=0$, and $\beta(\langle n',k,a,w\rangle)=a$. Set $\beta(y)=0$ for all remaining labels. Both correct labels at $I\in[n]$ point to $B_I$, so for every label $y$,
\[
\ind\{y\notin h_{n,B,U}(I)\}\geq\ind\{\beta(y)\neq B_I\}.
\]
Whatever the learner predicts at $I$ is a function of the sample, of $I$, and of randomness independent of $B_I$. By (b), on $E$ the prediction is incorrect with conditional probability at least $1/2$. Averaging over the target, the sample, and the randomness of the learner, and using (a),
\begin{equation}\label{app:eq:offset-risk-lower}
\E\bigl[L_{P_n,h_{n,B,U}}(\mathcal L(\sigma))\bigr]\geq\frac12\Pr[E]\geq\frac12(1-1/n)^{2m}.
\end{equation}
Some fixed target attains at least this average. Hence, for a fixed learner and a fixed $m$, the worst-case expected error is at least $\frac12(1-1/n)^{2m}$ for every $n$, and thus at least $1/2$. Together with the upper bound, this proves \eqref{app:eq:offset-minimax}. In particular, $\H^-$ is not full-set PAC learnable: by \eqref{eq:pac-expect}, a PAC guarantee with $\eps=\delta=1/8$ would bound the expected error by $1/4$ uniformly.
\end{proof}

\subsection{A Full-Set Learnable Class with Infinite Joint DS Dimension}\label{app:sep-large}

We identify a vector $a\in\{0,1\}^n$ with the function $x\mapsto a_x$ on $[n]$. For $n\geq1$, choose a vector $b\in\{0,1\}^n$ and functions $f_1,\ldots,f_n:[n]\to\{0,1\}$ for which there is a permutation $\pi$ of $[n]$ such that
\begin{equation}\label{app:eq:table-order}
f_{\pi(t)}(\pi(s))=b_{\pi(s)}
\qquad(1\leq s\leq t\leq n).
\end{equation}
Thus each function $f_i$ is correct at its own input and at all inputs that precede it in this order. Define the total hypothesis
\begin{equation}\label{app:eq:table-target}
h_{n,b,f}(i)=
\begin{cases}
\{b_i,\langle n,i,f_i\rangle\}, & i\leq n,\\
\{\bot\}, & i>n,
\end{cases}
\end{equation}
and let $\H^+$ consist of all these hypotheses.

The tuple label at $i$ reveals a function $f_i$ that predicts the bits at all inputs, but $f_i$ is reliable only at $i$ and at the inputs that precede it in a hidden order. The hidden order hurts the dimension but not the learner. On the one hand, an input $i$ may always be moved to the end of the order, where no other function needs to be correct at it, so its bit can be flipped without changing the correct-label sets elsewhere; this produces disjoint-neighbor witnesses of every size. On the other hand, the observed input that comes last in the order supplies a function that is correct at every observed input, so some predictor supplied by a single example is consistent with the whole sample. A learner that outputs such a predictor is a sample compression scheme of size one, and it generalizes at the usual rate.

\begin{theorem}\label{thm:large-ds-learnable}
The class $\H^+$ satisfies $\dDS(\H^+)=\infty$, and it has a deterministic full-set learner such that, for every target $h\in\H^+$, every distribution $P$ on $\N$, every $m\geq1$, and every $0<\eps<1$,
\begin{equation}\label{app:eq:table-tail}
\Pr\bigl[L_{P,h}(\widehat g_m)>\eps\bigr]\leq\min\bigl\{1,m(1-\eps)^{m-1}\bigr\},
\end{equation}
where $\widehat g_m$ is its output on $m$ examples. In particular, $\H^+$ is full-set PAC learnable.
\end{theorem}

\begin{proof}
We first construct witnesses of every size, and then analyze the learner described above.

\paragraph{Arbitrarily large witnesses.}
Fix $n$, and let $\H^+_n$ consist of the hypotheses in $\H^+$ with this value of $n$. This class is finite, and it is nonempty because $f_i=b$ for every $i$ satisfies \eqref{app:eq:table-order}. We show that the patterns of $\H^+_n$ on $(1,\ldots,n)$ form a disjoint-neighbor witness, so that $\dDS(\H^+)\geq\dpair(\H^+)\geq n$ by Lemma~\ref{lem:orders}; since $n$ is arbitrary, this gives $\dDS(\H^+)=\infty$.

Take a member $h_{n,b,f}$ of $\H^+_n$, a permutation $\pi$ as in \eqref{app:eq:table-order}, and an input $i\in[n]$. Let $b'$ be $b$ with coordinate $i$ flipped, and let $f'_i=b'$ and $f'_j=f_j$ for $j\neq i$. These parameters satisfy \eqref{app:eq:table-order} with the order $\pi'$ obtained from $\pi$ by moving $i$ to the last position. Indeed, for $j\neq i$, the inputs that precede $j$ under $\pi'$ also precede it under $\pi$ and differ from $i$, and $b'$ agrees with $b$ there and at $j$, so $f'_j=f_j$ is still correct at all of them. The input $i$ is last under $\pi'$, and $f'_i=b'$ is correct everywhere.

Therefore $h_{n,b',f'}\in\H^+_n$. It has the same correct-label set as $h_{n,b,f}$ at every $j\neq i$, since $b'_j=b_j$ and $f'_j=f_j$. At $i$, the correct-label sets $\{b_i,\langle n,i,f_i\rangle\}$ and $\{1-b_i,\langle n,i,b'\rangle\}$ are disjoint: their bits differ, and so do their tuple labels, because $f_i(i)=b_i\neq b'_i$. This proves the disjoint-neighbor property.

\paragraph{The learning rule.}
An example whose correct-label set contains a tuple label $c=\langle n,i,f\rangle$ supplies the predictor $g_c:\N\to\{0,1,\bot\}$ given by $g_c(x)=f(x)$ for $x\leq n$ and $g_c(x)=\bot$ for $x>n$; an example with correct-label set $\{\bot\}$ supplies the predictor $g_\bot$ that is identically $\bot$. Call a predictor $g$ \emph{valid on} a sample $\sigma=((x_t,h(x_t)))_{t=1}^m$ if $g(x_t)\in h(x_t)$ for all $t\in[m]$. The learner returns the first supplied predictor, in the order of the examples, that is valid on the sample, and an arbitrary predictor if there is none.

On every sample from a target $h=h_{n,b,f}\in\H^+$, some supplied predictor is valid, so the learner always returns one. If every observed correct-label set is $\{\bot\}$, then $g_\bot$ is valid. Otherwise, fix a permutation $\pi$ as in \eqref{app:eq:table-order} for the target, and take the observed input in $[n]$ that comes last under $\pi$. By \eqref{app:eq:table-order}, the function it supplies is correct at every observed input in $[n]$, and its extension by $\bot$ is correct at every observed input outside $[n]$.

\paragraph{Generalization.}
Fix the target and the distribution, and let $g_t$ be the predictor supplied by the $t$th example. Conditional on the $t$th example, $g_t$ is fixed and the other $m-1$ examples are still independent with law $\D_{P,h}$. Hence, if $L_{P,h}(g_t)>\eps$, then $g_t$ is correct at all of them with probability at most $(1-\eps)^{m-1}$, and
\[
\Pr\bigl[L_{P,h}(g_t)>\eps\text{ and }g_t\text{ is valid on the sample}\bigr]\leq(1-\eps)^{m-1}.
\]
The output is one of the $g_t$ and is valid on the sample, so a union bound over $t\in[m]$ proves \eqref{app:eq:table-tail}. Since $m(1-\eps)^{m-1}\leq me^{-\eps(m-1)}$, the right-hand side is at most $\delta$ once $m\geq C\eps^{-1}\bigl(\log(2/\eps)+\log(2/\delta)\bigr)$ for a universal constant $C$, so $\mfull_{\H^+}(\eps,\delta)=O\bigl(\eps^{-1}(\log(2/\eps)+\log(2/\delta))\bigr)$.
\end{proof}

\section{Comparing Packed and DS-Type Dimensions of Intersection Classes}\label{app:comparison}

This appendix proves Theorem~\ref{thm:comparison}. Throughout, $\ell\ge1$ and $0<\rho\le1$ are fixed, and we use the projections, partial intersection classes, one-inclusion hypergraphs and dimensions of Sections~\ref{sec:prelim-projections} and~\ref{sec:prelim-dimensions}. The class in question is $\cI_{\ell+1}(\H)$: its correct-label sets have at most $\ell$ labels, and the conflict blocks of its packed witnesses have at most $\ell+1$ members. The heart of the proof is the following extraction theorem.

\begin{theorem}[Witness extraction]\label{cmp:thm:extraction}
Fix $\ell\ge1$ and $0<\rho\le1$. There is a constant $c_{\ell,\rho}>0$ with the following property. For every $\rho$-packed witness of $\cI_{\ell+1}(\H)$ on a sequence of $n\ge1$ distinct inputs, and every choice of representing tuples for the patterns of its list, there are a set $U$ of inputs of this sequence and a disjoint-neighbor witness $W\subseteq\cI_{\ell+1}(\H)|_U$ such that
\[
 |U|\ge c_{\ell,\rho}\frac{n}{\log(n+2)},
\]
and every pattern of $W$ is represented by at most $\ell+1$ hypotheses drawn from the given representing tuples.
\end{theorem}

The proof is constructive: a finite procedure finds $U$ and $W$ from the given witness and tuples. With $d=\dpair(\cI_{\ell+1}(\H))$, the theorem gives $c_{\ell,\rho}\,n/\log(n+2)\le d$ for every $\rho$-packed witness on $n$ inputs, and \eqref{co:compare} follows by a short calculation; since $\dpair\le\dDS$ by Lemma~\ref{lem:orders}, so does the bound in terms of the DS dimension (Section~\ref{cmp:sec:proof}). The whole argument takes place inside finitely many hypotheses and labels, so $\X$ and $\Y$ may be arbitrary sets.

\paragraph{Outline.}
Fix a $\rho$-packed witness on $n$ inputs with list $v^1,\ldots,v^M$ (Definition~\ref{def:pack}). In each direction its conflict blocks have at most $\ell+1$ members and cover at least $\rho M$ entries, so there are at least $\rho Mn/(\ell+1)$ blocks in all. Section~\ref{cmp:sec:lemmas} states two lemmas that turn suitable configurations into disjoint-neighbor witnesses. The first applies to a product-like family of patterns indexed by $[r]^n$ in which, at every input $i$, the patterns with the same $i$th coordinate share a common label. The second applies to many conflict blocks whose sets at the varying input all meet one list of bounded width, and it yields a witness whose size is proportional to the number of blocks per pattern.

The main step, in Section~\ref{cmp:sec:reduction}, supplies such lists. Every pattern receives a list table of constant width that is determined by its \emph{history}: its correct-label sets on a common initial set of inputs and on a few further inputs that the procedure \emph{records}. The patterns of a conflict block differ only at the varying input, so unless this input is recorded, they have the same history and share one list. The reduction either finds a witness on $t+1$ inputs or records $O(t\log n)$ inputs per pattern, and the records must spoil only a small fraction of the blocks; so we take $t$ of order $n/\log n$, which is the source of the logarithm in Theorem~\ref{cmp:thm:extraction}. Section~\ref{cmp:sec:proof} combines the three lemmas, and Sections~\ref{cmp:sec:common-label} and~\ref{cmp:sec:list} prove the two extraction lemmas.

\subsection{Two Extraction Lemmas}\label{cmp:sec:lemmas}

We first need a few definitions.

\paragraph{Localization.}
Fix a $\rho$-packed witness of $\cI_{\ell+1}(\H)$ on distinct inputs $x_1,\ldots,x_n$, which we call the \emph{pool}, with list $v^1,\ldots,v^M$, collections $\cB_i$ of conflict blocks, and a representing tuple of at most $\ell+1$ hypotheses of $\H$ for each $v^a$. These hypotheses are defined on the whole pool, since their intersection $v^a$ is. Let $\H_0$ be the finite class of their restrictions to the pool, identify $x_i$ with $i$, and write $\Y_0$ for the finite set of labels used by $\H_0$. Every member of $\cI_R(\H_0)$ then corresponds to a member of $\cI_R(\H)$ supported on the pool, so every witness built from $\cI_R(\H_0)$ on $[n]$ is a witness of the same type for $\cI_R(\H)$ on the corresponding inputs, with representing tuples drawn from $\H_0$. The lemmas below hold for every finite class $\H_0$ of this kind, and all ties are broken by fixed orders on the inputs, the labels and the hypothesis tuples.

\paragraph{Partial patterns.}
The \emph{partial pattern} of $g\in\cI_R(\H_0)$ is the vector $(g(1),\ldots,g(n))\in(\cA\cup\{\star\})^n$, with $\star$ exactly at the inputs outside $\supp(g)$; a partial pattern without $\star$ is a pattern on the pool. For partial patterns $v,w$ and an input $i$, $v_{-i}=w_{-i}$ means that $v$ and $w$ agree at every input other than $i$; in particular, they are undefined at the same inputs other than $i$. A representing tuple of $g$, together with $\supp(g)$, determines its partial pattern.

\paragraph{Products with common labels.}
The first configuration has its patterns indexed by all vectors $b\in[r]^n$. At an input $i$, the correct-label sets of the patterns with $b_i=j$ form the $j$th \emph{group}. These sets may vary with the other coordinates of $b$, but they all contain a fixed label $a_i^j$, the \emph{common label} of the group, and they all lie in a set $Y_i^j$, where $Y_i^1,\ldots,Y_i^r$ have an empty intersection; thus no label is correct for all groups at $i$. If every set depended only on $b_i$, the patterns would form a product witness (Definition~\ref{def:nat}). The next lemma recovers such a product of fixed sets on a constant fraction of the inputs, even when the given patterns are intersections of many hypotheses. It is proved in Section~\ref{cmp:sec:common-label}.

\begin{lemma}[Extraction from a product with common labels]\label{cmp:lem:common-label}
Fix integers $\ell,R\ge1$ and $r\ge2$. There is a constant $a_{\ell,R,r}>0$ with the following property. Let $\H_0$ be a finite class of hypotheses defined on a finite pool, with correct-label sets of size at most $\ell$. Suppose that, on $n$ inputs of the pool, which we identify with $[n]$, sets $Y_i^j$, labels $a_i^j$, and patterns $A^b\in\cI_R(\H_0)|_{[n]}$, $b\in[r]^n$, satisfy
\[
 a_i^{b_i}\in A_i^b\subseteq Y_i^{b_i},
 \qquad
 \bigcap_{j=1}^rY_i^j=\emptyset
 \qquad(i\in[n]).
\]
Given representing tuples for these patterns, one can construct a set $U\subseteq[n]$ of at least $a_{\ell,R,r}n$ inputs and nonempty sets $T_i^j$, $i\in U$, $j\in[r]$, such that
\[
 \prod_{i\in U}\{T_i^j:j\in[r]\}\subseteq\cI_\ell(\H_0)|_U,
 \qquad
 \bigcap_{j=1}^rT_i^j=\emptyset\quad(i\in U).
\]
This product is a DS witness. For $r=2$, it is also a disjoint-neighbor witness.
\end{lemma}

For $r>2$, the product need not be a disjoint-neighbor witness. A second application of the lemma, to a binary product of intersections of its sets, turns it into one.

\begin{corollary}[Binary extraction]\label{cmp:cor:binary}
Under the hypotheses of Lemma~\ref{cmp:lem:common-label}, one can construct a disjoint-neighbor witness in $\cI_\ell(\H_0)$ on at least $a_{\ell,\ell^2,2}\,a_{\ell,R,r}\,n$ inputs.
\end{corollary}

\begin{proof}
Let $U$ and $T_i^j$ be given by Lemma~\ref{cmp:lem:common-label}, and fix $i\in U$. The sets $T_i^1,\ldots,T_i^r$ are nonempty, have at most $\ell$ labels, being values of members of $\cI_\ell(\H_0)$, and have an empty intersection. By Lemma~\ref{lem:small}, some minimal empty-intersection family among them has between two and $\ell+1$ members. Let $Z_i^2$ be its member with the largest index and $Z_i^1$ the intersection of the others; then $Z_i^1\ne\emptyset$ by minimality and $Z_i^1\cap Z_i^2=\emptyset$. Every pattern of the binary product $\prod_{i\in U}\{Z_i^1,Z_i^2\}$ is the intersection of $\ell$ patterns of the product of the lemma. Indeed, given a choice of $Z_i^1$ or $Z_i^2$ at every $i\in U$, write the chosen set at $i$ as $T_i^{j_{i,1}}\cap\cdots\cap T_i^{j_{i,\ell}}$, repeating indices if it is the intersection of fewer than $\ell$ of the sets. For $k\in[\ell]$, the pattern that takes the set $T_i^{j_{i,k}}$ at every $i\in U$ belongs to the product, because the product allows independent choices at different inputs, and the intersection of these $\ell$ patterns is the given pattern of the binary product. Each pattern of the product of the lemma is represented by at most $\ell$ hypotheses of $\H_0$, so
\[
 \prod_{i\in U}\{Z_i^1,Z_i^2\}\subseteq\cI_{\ell^2}(\H_0)|_U.
\]
Lemma~\ref{cmp:lem:common-label}, with $R=\ell^2$, $r=2$, $Y_i^j=Z_i^j$, and any label of $Z_i^j$ as $a_i^j$, now gives a binary product in $\cI_\ell(\H_0)$ with disjoint sets at every input, that is, a disjoint-neighbor witness, on at least $a_{\ell,\ell^2,2}|U|\ge a_{\ell,\ell^2,2}\,a_{\ell,R,r}\,n$ inputs. Both applications keep representing tuples drawn from $\H_0$.
\end{proof}

\paragraph{Lists.}
Lists of labels are standard in multiclass learning \cite{brukhim2022characterization,charikar2023characterization}; here a list succeeds when it hits a correct-label set. A \emph{list table} $Q$ on the pool assigns a nonempty set $Q(i)\subseteq\Y_0$ to every input $i\in[n]$, and its \emph{width} is $\max_i|Q(i)|$. It \emph{hits} a set $A$ at $i$ if $Q(i)\cap A\ne\emptyset$, and \emph{misses} it otherwise. For a pattern $v$ on the pool, we count the misses,
\[
 \miss(Q,v)=\bigl|\{i\in[n]:Q(i)\cap v_i=\emptyset\}\bigr|.
\]

The second configuration consists of many conflict blocks whose sets at the varying input all meet a list of bounded width. The list is required to hit the sets only at the varying input of each block; this local condition lets us, in Section~\ref{cmp:sec:proof}, discard different blocks in different directions. The lemma is proved in Section~\ref{cmp:sec:list}.

\begin{lemma}[Extraction under a bounded list]\label{cmp:lem:list}
Fix integers $\ell,R,L\ge1$. There is a constant $\beta_{\ell,R,L}>0$ with the following property. Let $\H_0$ be a finite class of hypotheses defined on a finite pool $[n]$, with correct-label sets of size at most $\ell$. Let $v^1,\ldots,v^M$ be partial patterns of $\cI_R(\H_0)$, with repetitions allowed and with given representing tuples, and let $Q$ be a list table of width at most $L$. For each input $i$, let $\cB_i$ be a family of pairwise disjoint subsets of $[M]$ such that every $B\in\cB_i$ satisfies
\[
 \begin{gathered}
 2\le|B|\le\ell+1,
 \qquad
 v^a_{-i}=v^{a'}_{-i}\quad(a,a'\in B),\\
 \bigcap_{a\in B}v_i^a=\emptyset,
 \qquad
 v_i^a\cap Q(i)\ne\emptyset\quad(a\in B).
 \end{gathered}
\]
If $N=\sum_i|\cB_i|>0$, one can construct a disjoint-neighbor witness in $\cI_\ell(\H_0)$ on at least $\beta_{\ell,R,L}N/M$ inputs.
\end{lemma}

\subsection{List Reduction}\label{cmp:sec:reduction}

Lemma~\ref{cmp:lem:list} needs one list table of bounded width for all the patterns of a block. The next lemma assigns to every pattern a list table that is determined by its history, in such a way that two patterns that differ only at an input recorded by neither of them have the same history; the procedure either does this or finds a disjoint-neighbor witness of the desired size.

\begin{lemma}[Conservative list reduction]\label{cmp:lem:reduction}
Fix $\ell\ge1$ and $0<\eta<1$. There exist an integer $J=J_\ell\ge1$ and constants $L_*=L_*(\ell)$ and $K=K_{\ell,\eta}$ with the following property. Let $v^1,\ldots,v^M$ be patterns of $\cI_{\ell+1}(\H_0)$ on $[n]$, with repetitions allowed, and let $t\ge1$ be an integer such that
\[
 u:=\lceil t/\eta\rceil<n,
 \qquad
 J(t+1)\le n.
\]
A finite procedure either constructs a disjoint-neighbor witness in $\cI_{\ell+1}(\H_0)$ on at least $t+1$ inputs, or assigns to every index $a\in[M]$ a history $\tau_a$, a record set $T_a\subseteq[n]$, and a list table $Q_a$ of width at most $L_*$ such that, with $\Lambda=K(t+1)\log(2en)$,
\begin{equation}\label{cmp:eq:reduction-cost}
 |T_a|\le u+\Lambda,
 \qquad
 \sum_{a=1}^M\miss(Q_a,v^a)\le M(\eta n+\Lambda).
\end{equation}
The history $\tau_a$ determines the list table $Q_a$, and for all $a,b\in[M]$ and $i\in[n]$,
\begin{equation}\label{cmp:eq:stability}
 v^a_{-i}=v^b_{-i},\quad i\notin T_a\cup T_b
 \quad\Longrightarrow\quad \tau_a=\tau_b.
\end{equation}
\end{lemma}

\paragraph{Proof idea.}
The proof has two phases. In the first, suppose that no $t+1$ inputs carry a disjoint-neighbor witness. Then the patterns on any $t+1$ inputs can be deleted one hyperedge at a time, and each deletion assigns a list of at most $\ell$ labels to an observation. Taking unions of these lists over the subsets of a random set of $u$ inputs gives lists of width $e^{O(t)}$, and a leave-one-out argument bounds their average number of misses by $\eta n$. In the second phase, we reduce the width to a constant in $O(\log(t+2))$ passes. A pass fixes the new list at an input before it looks at the correct-label set there, and it \emph{records} the input only when the new list misses a set that the old list hit. The records are controlled by a potential that counts, with weights, the binary products realized by the remaining candidate patterns: if one of these products uses $J(t+1)$ inputs, Lemma~\ref{cmp:lem:common-label} gives the witness, and otherwise every record divides the potential by a large factor. The history of a pattern consists of its sets at the initial and the recorded inputs. At an unrecorded input every pattern undergoes the same deletion step, and this gives the stability \eqref{cmp:eq:stability}.

\begin{proof}
Choose an integer $J\ge\max\{1,1/a_{\ell,\ell+1,2}\}$, where $a_{\ell,\ell+1,2}$ is the constant of Lemma~\ref{cmp:lem:common-label}, and an integer $L_*\ge\max\{16,\ell^4\}$. We fix in advance all orders and tie-breaking rules, the same for every index $a\in[M]$. Constants denoted by $C$ depend only on $\ell$, $\eta$ and $J$, constants denoted by $C_\ell$ depend only on $\ell$, and both may change from line to line.

\paragraph{Deletion and initialization.}
For every set $I\subseteq[n]$ of $t+1$ inputs, we first fix a rule that assigns lists to \emph{observations} on $I$, that is, to pairs $(i,w)$ of an input $i\in I$ and a pattern $w$ on $I\setminus\{i\}$. Start with the finite set of patterns of $\cI_{\ell+1}(\H_0)$ on $I$, and repeat the following while some patterns remain. Choose a remaining pattern $v$ and an input $i\in I$ such that $v_i$ meets $w_i$ for every remaining pattern $w$ with $w_{-i}=v_{-i}$. Assign the list $v_i$ to the observation $(i,v_{-i})$, and delete all remaining patterns with this observation, that is, the hyperedge $e_{i,v_{-i}}$ of the remaining set. If at some point no such choice exists while a nonempty set remains, then every remaining pattern has, in every direction $i\in I$, a remaining neighbor whose set at $i$ is disjoint from its own; so the remaining set is a disjoint-neighbor witness in $\cI_{\ell+1}(\H_0)$ on $t+1$ inputs, and we return it. Otherwise the process deletes every pattern on $I$. After a deletion no remaining pattern has the observation $(i,v_{-i})$, so each observation receives at most one list, and we give every unassigned observation a fixed list consisting of one label of $\Y_0$. All lists have at most $\ell$ labels. Moreover, the list $v_i$ meets the set at $i$ of every pattern deleted together with $v$. Hence every pattern $w$ on $I$ has an input $i\in I$, the one at which it was deleted, such that the list assigned to the observation $(i,w_{-i})$ meets $w_i$.

These rules give the initial lists. For a set $I_0\subseteq[n]$ of $u$ inputs and a pattern $v$ on $[n]$, let $Q_{v,I_0}(i)=v_i$ for $i\in I_0$, and for $i\notin I_0$ let $Q_{v,I_0}(i)$ be the union, over all sets $I'\subseteq I_0$ of $t$ inputs, of the list assigned by the rule on $I'\cup\{i\}$ to the observation $(i,v|_{I'})$. Since $\binom ut\le(eu/t)^t$ and $u\le2t/\eta$, every such list table has width at most
\begin{equation}\label{cmp:eq:initial-width}
 L_0:=\max\Bigl\{\ell,\ell\binom ut\Bigr\}\le\exp\bigl(C(t+1)\bigr).
\end{equation}

We now bound the misses for a random $I_0$. Fix a pattern $v$ on $[n]$ and a set $I_1\subseteq[n]$ of $u+1$ inputs, which exists since $u<n$. We claim that at most $t$ inputs $i\in I_1$ have $Q_{v,I_1\setminus\{i\}}(i)\cap v_i=\emptyset$. Indeed, otherwise let $I\subseteq I_1$ be a set of $t+1$ such inputs. The pattern $v|_I$ was deleted by the rule on $I$ at some input $i\in I$, so the list assigned to the observation $(i,v|_{I\setminus\{i\}})$ meets $v_i$. Since $I\setminus\{i\}$ is a set of $t$ inputs of $I_1\setminus\{i\}$, this list is contained in $Q_{v,I_1\setminus\{i\}}(i)$, a contradiction. Now let $I_0$ be a uniformly random set of $u$ inputs. Equivalently, we may first draw a uniform set $I_1$ of $u+1$ inputs, then choose $i$ uniformly from $I_1$ and set $I_0=I_1\setminus\{i\}$. This gives the same joint law for $I_0$ and a uniform input $i\notin I_0$. Since $Q_{v,I_0}$ never misses on $I_0$, the claim gives
\[
 \E\bigl[\miss(Q_{v,I_0},v)\bigr]\le\frac{t(n-u)}{u+1}\le\eta n,
\]
where the last inequality uses $u\ge t/\eta$. Choose one set $I_0$ that minimizes $\sum_a\miss(Q_{v^a,I_0},v^a)$; the same set is used for every index. Write $Q_{a,0}=Q_{v^a,I_0}$. By averaging,
\begin{equation}\label{cmp:eq:initial-misses}
 \sum_{a=1}^M\miss(Q_{a,0},v^a)\le\eta Mn.
\end{equation}
The set $I_0$ is the first part of every record set $T_a$, and the sets $v^a_i$, $i\in I_0$, are the first part of every history $\tau_a$.

\paragraph{Encoded profiles.}
The initial lists have few misses in total, but their width may be exponential in $t$, and we reduce it in passes. To decide which labels a shorter list must keep, we measure how many binary products the remaining candidate patterns can still realize, since a large binary product is exactly what Lemma~\ref{cmp:lem:common-label} needs. Products are counted after coding every set by its trace on the old list, and a random coloring of the labels outside the list makes this coding faithful for disjointness: retained sets with disjoint codes are disjoint.

Formally, a pass starts from a list table $Q$ of width at most $L\ge2$. The \emph{reset class} of $Q$ is obtained by restricting every partial pattern $A$ of $\cI_{\ell+1}(\H_0)$ on $[n]$ to its hit domain,
\[
 A\longmapsto A\big|_{\{i\in\supp(A):\,A_i\cap Q(i)\ne\emptyset\}};
\]
it consists of partial patterns of $\cI_{\ell+1}(\H_0)$, the one with empty support included. Thus every set attained at an input $i$ by a member of the reset class meets $Q(i)$. Independently at every input $i$, color each label $y\in\Y_0\setminus Q(i)$ by a uniform label $z_i(y)\in Q(i)$. A set $E$ at $i$ is \emph{retained} if
\[
 E_Q:=E\cap Q(i)\ne\emptyset
 \qquad\text{and}\qquad
 z_i(E\setminus Q(i))\subseteq E_Q,
\]
and it is then encoded by $E_Q$; a set that is not retained becomes $\star$. For a set $V$ of partial patterns, let $V_z$ be the set of their encodings. A \emph{profile} $\pi$ of $V_z$ consists of a set $\supp(\pi)$ of inputs and, at every $i\in\supp(\pi)$, an unordered pair of disjoint nonempty encoded sets, such that every choice of one set from each pair is realized by a member of $V_z$ defined on all of $\supp(\pi)$. The profile with empty support is a profile of $V_z$ whenever $V\ne\emptyset$. Let $\Pi(V_z)$ be the set of profiles of $V_z$.

\paragraph{Extraction from a large profile.}
Before every pass, we enumerate the profiles of $V_z$ for the reset class $V$ of the current list and for all of the finitely many colorings $z$. Suppose that some profile $\pi$ has $|\supp(\pi)|\ge J(t+1)$. Restrict it to $J(t+1)$ inputs of its support, identified with $[J(t+1)]$, order the two sets of each pair, and for every $b\in[2]^{J(t+1)}$ choose a member $A^b$ of the reset class whose encoding realizes the choice $b$. These are partial patterns of $\cI_{\ell+1}(\H_0)$ defined on all $J(t+1)$ inputs, and we keep representing tuples for them. At an input $i$ of the restricted support, let $a_i^j$ be any label of the $j$th encoded set and $Y_i^j=\bigcup_{b:b_i=j}A_i^b$, for $j\in[2]$. An encoded set is contained in every set it encodes, so $a_i^{b_i}\in A_i^b\subseteq Y_i^{b_i}$. Moreover $Y_i^1\cap Y_i^2=\emptyset$: a common label in $Q(i)$ would lie in both encoded sets, and a common label $y\notin Q(i)$ would have its color $z_i(y)$ in both, whereas the two encoded sets are disjoint. Lemma~\ref{cmp:lem:common-label}, with $R=\ell+1$ and $r=2$, therefore constructs a disjoint-neighbor witness in $\cI_\ell(\H_0)\subseteq\cI_{\ell+1}(\H_0)$ on at least $a_{\ell,\ell+1,2}J(t+1)\ge t+1$ inputs, which we return.

Otherwise, for every coloring, every profile of the reset class has support of size less than $J(t+1)$. The same bound holds for every class obtained from the reset class by conditioning on correct-label sets and deleting inputs. Indeed, the realizers of a profile of such a class are restrictions of members of the reset class, and their encodings at the inputs of the support do not depend on the colors at the deleted inputs; so the profile is also a profile of the reset class under any extension of the coloring. This bound on the profiles controls the cost of the updates below.

\paragraph{The profile potential.}
During a pass with list table $Q$ of width at most $L$, the \emph{states} are nonempty sets $V$ of partial patterns on the unprocessed inputs obtained from the reset class in this way. For such $V$ put
\[
 \Psi_Q(V)=\E_z\Bigl[\sum_{\pi\in\Pi(V_z)}\lambda^{|\supp(\pi)|}\Bigr],
 \qquad \lambda=L^{2(\ell-1)},
\]
where $z$ is the random coloring of the unprocessed inputs. The profile with empty support gives $\Psi_Q(V)\ge1$. At each input, an encoded set is a nonempty subset of $Q(i)$ with at most $\ell$ labels, so there are at most $C_\ell(L+1)^{2\ell}$ pairs of encoded sets. Supports have fewer than $J(t+1)\le n$ inputs, so bounding the number of supports of size $j$ by $\binom nj$ and using $\sum_{j\le J(t+1)}\binom nj\le\bigl(en/(J(t+1))\bigr)^{J(t+1)}$, valid since $J(t+1)\le n$, gives
\begin{equation}\label{cmp:eq:potential-size}
 1\le\Psi_Q(V),
 \qquad
 \log\Psi_Q(V)\le C_\ell J(t+1)\Bigl[\log\frac{en}{J(t+1)}+\log(L+1)\Bigr].
\end{equation}
Throughout the pass, $L$, $\lambda$, and the coloring law stay fixed; the coloring law is restricted to the unprocessed inputs as inputs are deleted. Deleting an input does not increase the potential, since every profile of the smaller class is a profile of the larger one with the same support.

For a set $A$ attained at an unprocessed input $i$ by a member of $V$, the \emph{exact child} $V_A$ is the set of members of $V$ whose set at $i$ is exactly $A$, with $i$ deleted. The next inequality is the core of the proof: conditioning on disjoint sets splits the potential.

\begin{claim}
If $A_1,\ldots,A_k$ are pairwise disjoint sets attained at an unprocessed input $i$ by members of a state $V$, then
\begin{equation}\label{cmp:eq:child-potential}
 \sum_{j=1}^k\Psi_Q(V_{A_j})\le\Psi_Q(V).
\end{equation}
\end{claim}

\begin{proof}
Fix the colors at all unprocessed inputs other than $i$; the encoded children $(V_{A_j})_z$ depend only on them. Let $\pi$ be a profile of exactly $k_\pi\ge1$ of these children, so that $\pi$ contributes $k_\pi\lambda^{|\supp(\pi)|}$ to the left-hand side. We show that, in expectation over the colors at $i$, the profile $\pi$ and its extensions to $i$ contribute at least as much to $\Psi_Q(V)$.

First, $\pi$ is a profile of $V_z$ whatever the colors at $i$, since its realizers come from members of $V$. Next, consider two children that realize $\pi$, corresponding to $A_j$ and $A_{j'}$. If both sets are retained at $i$, their encodings are nonempty and disjoint, and adding this pair at $i$ extends $\pi$ to a profile of $V_z$ on $\supp(\pi)\cup\{i\}$: either choice at $i$, together with any choice in $\pi$, is realized by the corresponding child. Different pairs give different extensions.

It remains to bound the probability that both sets are retained. Every $A_j$ meets $Q(i)$ and has at most $\ell$ labels, so it has at most $\ell-1$ labels outside $Q(i)$, each of which receives a color in $A_j\cap Q(i)$ with probability at least $1/L$; hence $A_j$ is retained with probability at least $L^{-(\ell-1)}$. The retention events of $A_j$ and $A_{j'}$ depend on the colors of disjoint sets of labels, so they are independent, and both sets are retained with probability at least $L^{-2(\ell-1)}=1/\lambda$. Each extension therefore has expected weight at least $\lambda^{|\supp(\pi)|+1}/\lambda=\lambda^{|\supp(\pi)|}$, and the contribution of $\pi$ and its extensions is at least
\[
 \Bigl(1+\binom{k_\pi}2\Bigr)\lambda^{|\supp(\pi)|}\ge k_\pi\lambda^{|\supp(\pi)|}.
\]
Deleting $i$ from an extension of $\pi$ gives back $\pi$, so different profiles have different extensions. Summing over $\pi$ and averaging over the remaining colors proves \eqref{cmp:eq:child-potential}.
\end{proof}

\paragraph{Conservative updates.}
Inequality \eqref{cmp:eq:child-potential} allows a short list to hit every set whose exact child keeps a large fraction of the potential. We now describe one pass for a target pattern $v$ (one of the $v^a$), a list table $Q$ of width at most $L$, and an integer $m\ge1$. Let $V$ be the reset class of $Q$, and process the inputs in a fixed order. At input $i$, call a set $A$ attained at $i$ by a member of $V$ \emph{heavy} if $\Psi_Q(V_A)>\Psi_Q(V)/(m+1)$. By \eqref{cmp:eq:child-potential}, a family of pairwise disjoint heavy sets has at most $m$ members. Let $Q'(i)$ be the union of a maximal such family, chosen by the fixed rule as a function of the state, or a fixed singleton if no set is heavy. Then $|Q'(i)|\le\ell m$, and by maximality $Q'(i)$ meets every heavy set. The list $Q'(i)$ is computed and stored before the target set $v_i$ is examined.

Now $v_i$ is examined. If $Q(i)$ meets $v_i$ and $Q'(i)$ misses it, record the pair $(i,v_i)$ and replace $V$ by $V_{v_i}$. Otherwise record nothing and delete input $i$ from all members of $V$, without conditioning. In both cases the restriction of $v$ to its hit domain, with the processed inputs deleted, stays in $V$; so $V$ is never empty, and $\Psi_Q(V)\ge1$ throughout the pass. In particular, at a recorded input the set $v_i$ is attained. It is not heavy, since $Q'(i)$ misses it, so conditioning on it divides the potential by at least $m+1$, while every other step is a deletion and does not increase the potential. By \eqref{cmp:eq:potential-size}, the number of inputs recorded in one pass is at most
\begin{equation}\label{cmp:eq:pass-records}
 \frac{C_\ell J(t+1)}{\log(m+1)}\Bigl[\log\frac{en}{J(t+1)}+\log(L+1)\Bigr].
\end{equation}
A new miss, that is, an input at which $Q$ hits $v_i$ and $Q'$ misses it, is always recorded. Hence $\miss(Q',v)$ exceeds $\miss(Q,v)$ by at most the number of inputs recorded in the pass.

\paragraph{Reduction to constant width.}
We now choose the parameters so that the width decreases quickly and the total number of records remains bounded. Start from the width bound $L=L_0$ of \eqref{cmp:eq:initial-width}. While $L>L_*$, run a pass with
\[
 m=\Bigl\lceil\frac{\sqrt L}{\ell}\Bigr\rceil,
 \qquad
 L_{\mathrm{next}}=2\sqrt L,
\]
and replace $L$ by $L_{\mathrm{next}}$. Since $L>L_*\ge\ell^4$, we have $\ell\le L^{1/4}$, hence $\ell m\le\sqrt L+\ell\le2\sqrt L$, so the new list has width at most $L_{\mathrm{next}}$, and $\log(m+1)\ge\log(\sqrt L/\ell)\ge\frac14\log L$. The width bounds $L_0,L_1,\ldots$ are the same for every index, and they satisfy $\log L_{j+1}-\log4=\frac12(\log L_j-\log4)$. A pass runs only while $L_j>L_*\ge16$, that is, while $\log L_j-\log4>\log4$, so over the passes the numbers $\log L_j-\log4$ halve from one pass to the next, start below $\log L_0\le C(t+1)$ by \eqref{cmp:eq:initial-width}, and stay above $\log4$. Hence, with $j$ running over the passes,
\[
 \#\{\text{passes}\}\le C\log(2t+2),
 \qquad
 \sum_j\frac1{\log L_j}\le\sum_j\frac1{\log L_j-\log4}\le\frac2{\log4},
\]
since the terms of the second sum double from one pass to the next and the last is less than $1/\log4$.

If, for some index, the scan before some pass finds a profile with support of size at least $J(t+1)$, we return the witness extracted from it. Otherwise, by \eqref{cmp:eq:pass-records}, $\log(L+1)\le2\log L$, $\log(m+1)\ge\frac14\log L$, and $J(t+1)\ge t+1$, a pass with width bound $L$ records at most
\[
 C(t+1)\Bigl[\frac{\log(en/(t+1))}{\log L}+1\Bigr]
\]
inputs. Summing over the passes with the two bounds above, the passes record at most
\[
 C(t+1)\Bigl[\log\frac{en}{t+1}+\log(2t+2)\Bigr]
 =C(t+1)\log(2en)\le\Lambda
\]
inputs per index, for a suitable $K=K_{\ell,\eta}$. Let $T_a$ consist of $I_0$ and the inputs recorded in the passes for the index $a$, and let $Q_a$ be the final list table, of width at most $L_*$. Then $|T_a|\le u+\Lambda$. Every new miss is recorded, so $\miss(Q_a,v^a)\le\miss(Q_{a,0},v^a)+\Lambda$, and \eqref{cmp:eq:initial-misses} gives the second bound in \eqref{cmp:eq:reduction-cost}.

\paragraph{Stability.}
The history $\tau_a$ consists of the sets $v^a_i$, $i\in I_0$, and of the pairs recorded in all passes, each with its pass index. It determines every list and every state of the run for the index $a$. Indeed, the sets on $I_0$ determine $Q_{a,0}$, and the width bounds are fixed in advance. Within a pass, the state before an input $i$ determines $Q'(i)$; then the history tells whether $i$ is recorded, and if so, the set to condition on; otherwise $i$ is deleted. In particular, equal histories give equal final lists.

Now let $v^a_{-i}=v^b_{-i}$ and $i\notin T_a\cup T_b$. Since $I_0\subseteq T_a$, the set $I_0$ avoids $i$, so the two runs see the same sets on $I_0$ and start from the same list. Suppose that a pass starts from the same list in both runs. Then the two reset classes coincide, and we show by induction over the inputs that the two states coincide. Equal states give the same list $Q'(j)$ at the next input $j$. If $j\ne i$, then $v^a_j=v^b_j$, so the two runs make the same decision, record the same pair if any, and reach the same state. If $j=i$, neither run records $i$, so both delete it. Hence the pass records the same pairs and produces the same list in both runs. Induction over the passes gives $\tau_a=\tau_b$, which proves \eqref{cmp:eq:stability}.
\end{proof}

\subsection{Proof of the Comparison Theorem}\label{cmp:sec:proof}

We now prove Theorem~\ref{cmp:thm:extraction} and then deduce Theorem~\ref{thm:comparison}. The main point is the choice of the target size $t$ in Lemma~\ref{cmp:lem:reduction}: records and misses cost $\eta n+O(t\log n)$ inputs per pattern, while the packed witness has $\rho n/(\ell+1)$ blocks per pattern on average, so with $t$ a small multiple of $n/\log n$ most blocks survive, and Lemma~\ref{cmp:lem:list} gives a witness on order $n$ inputs, more than the $t+1$ we need.

\begin{proof}[Proof of Theorem~\ref{cmp:thm:extraction}]
Fix a $\rho$-packed witness of $\cI_{\ell+1}(\H)$ on $n\ge1$ inputs, with list $v^1,\ldots,v^M$ and collections $\cB_1,\ldots,\cB_n$ of conflict blocks, and fix the representing tuples of its patterns. As in Section~\ref{cmp:sec:lemmas}, we localize to the finite class $\H_0$ of the hypotheses in these tuples and identify the inputs with $[n]$; thus $v^1,\ldots,v^M$ are patterns of $\cI_{\ell+1}(\H_0)$ defined on all of $[n]$.

\paragraph{Constants.}
Put $\alpha=\rho/(\ell+1)$ and $\eta=\alpha/4$, let $J$, $L_*$ and $K$ be the constants of Lemma~\ref{cmp:lem:reduction} for $\ell$ and $\eta$, and write $\beta=\beta_{\ell,\ell+1,L_*}$ for the constant of Lemma~\ref{cmp:lem:list}. All of them depend only on $\ell$ and $\rho$. Choose $c>0$ with
\[
 c\le\min\Bigl\{\frac\eta4,\ \frac1{4J},\ \frac{\alpha}{4(2/\eta+4K)},\ \frac{\beta\alpha}{4}\Bigr\},
\]
and set
\[
 t=\Bigl\lfloor\frac{cn}{\log(2en)}\Bigr\rfloor.
\]
We construct a disjoint-neighbor witness on more than $cn/\log(2en)$ inputs whose patterns are intersections of at most $\ell+1$ hypotheses of $\H_0$. This proves the theorem with $c_{\ell,\rho}=c/2$, since $\log(2en)\le2\log(n+2)$ for $n\ge1$.

\paragraph{Target size zero.}
If $t=0$, then $cn/\log(2en)<1$, and a witness on one input suffices. The blocks of direction $1$ cover at least $\rho M>0$ indices, so there is a block $B\in\cB_1$. Its sets $v^a_1$ are intersections of up to $\ell+1$ hypotheses, while the witness must lie in $\cI_\ell(\H_0)$, so we pass to single hypotheses. The sets $g(1)$, over the hypotheses $g$ in the tuples of the patterns $v^a$, $a\in B$, are nonempty sets of at most $\ell$ labels, and they have an empty intersection because $\bigcap_{a\in B}v^a_1=\emptyset$. By Lemma~\ref{lem:small} they contain a minimal empty-intersection family $A_1,\ldots,A_r$ with $2\le r\le\ell+1$. The sets $\bigcap_{j<r}A_j$ and $A_r$ are then disjoint and nonempty, and they are the values at input $1$ of intersections of at most $\ell$ hypotheses of $\H_0$; as one-coordinate patterns, they form a disjoint-neighbor witness in $\cI_\ell(\H_0)$ on one input.

\paragraph{Applying the list reduction.}
Suppose now that $t\ge1$. Then $t+1\le2t\le2cn/\log(2en)$. As $t\ge1$ forces $n\ge2$, the bound $c\le\eta/4$ gives $u=\lceil t/\eta\rceil\le n/4+1<n$, and the bound $c\le1/(4J)$ gives $J(t+1)\le n/2$. So Lemma~\ref{cmp:lem:reduction} applies to the patterns $v^1,\ldots,v^M$ with target size $t$. If it constructs a disjoint-neighbor witness in $\cI_{\ell+1}(\H_0)$ on at least $t+1>cn/\log(2en)$ inputs, we return this witness. Otherwise it assigns to every index $a\in[M]$ a history $\tau_a$, a record set $T_a$, and a list $Q_a$ of width at most $L_*$ that satisfy \eqref{cmp:eq:reduction-cost} and \eqref{cmp:eq:stability}, with $\Lambda=K(t+1)\log(2en)$. It remains to find, in this case, a family of blocks to which Lemma~\ref{cmp:lem:list} applies.

\paragraph{Discarding blocks hit by records or misses.}
The blocks of direction $i$ have at most $\ell+1$ members and cover at least $\rho M$ indices, so there are at least $\alpha M$ of them, and at least $\alpha Mn$ blocks in all. Discard a block $B\in\cB_i$ if some $a\in B$ records $i$, that is, $i\in T_a$, or has a miss at $i$, that is, $Q_a(i)\cap v^a_i=\emptyset$. Since the blocks of direction $i$ are disjoint, an index $a$ with a record or a miss at $i$ lies in at most one block of direction $i$. By \eqref{cmp:eq:reduction-cost}, the number of discarded blocks is therefore at most
\[
 \sum_{a=1}^M\bigl(|T_a|+\miss(Q_a,v^a)\bigr)\le M(u+\eta n+2\Lambda).
\]
Since $u\le t/\eta+1\le2t/\eta\le2cn/\eta$ and $2\Lambda=2K(t+1)\log(2en)\le4Kcn$, the choice of $c$ gives
\[
 u+\eta n+2\Lambda\le\Bigl[\eta+c\Bigl(\frac2\eta+4K\Bigr)\Bigr]n\le\frac\alpha2\,n.
\]
Thus at least $\alpha Mn/2$ blocks remain. For every remaining block $B\in\cB_i$ and every $a\in B$, we have $i\notin T_a$ and $Q_a(i)\cap v^a_i\ne\emptyset$.

\paragraph{Grouping the patterns by their histories.}
Partition the indices $[M]$ into groups of indices with the same history. Let $B\in\cB_i$ be a remaining block and $a,b\in B$. Then $v^a_{-i}=v^b_{-i}$ and $i\notin T_a\cup T_b$, so $\tau_a=\tau_b$ by \eqref{cmp:eq:stability}. Hence every remaining block lies in a single group. Since the history determines the list, all indices of a group $G$ have a common list $Q_G$ of width at most $L_*$. Let $M_G=|G|$, and let $N_G$ be the number of remaining blocks contained in $G$. Since $\sum_GM_G=M$ and $\sum_GN_G\ge\alpha Mn/2$, weighted averaging gives a group $G$ satisfying
\[
 \frac{N_G}{M_G}\ge\frac\alpha2\,n>0.
\]

\paragraph{Applying the bounded-list lemma.}
Fix this group $G$, and for each $i\in[n]$ let $\cB_i^G$ be the set of remaining blocks of direction $i$ contained in $G$. Consider the patterns $v^a$, $a\in G$, which are patterns of $\cI_{\ell+1}(\H_0)$ defined on all of $[n]$, with their representing tuples, and the list $Q_G$. Every $B\in\cB_i^G$ is a conflict block of the packed witness, so $2\le|B|\le\ell+1$, its patterns agree outside $i$, and their sets at $i$ have an empty intersection; moreover, $Q_G(i)=Q_a(i)$ hits $v^a_i$ for every $a\in B$. The blocks of $\cB_i^G$ are disjoint, and $\sum_i|\cB_i^G|=N_G>0$. Thus Lemma~\ref{cmp:lem:list}, with $R=\ell+1$ and $L=L_*$, applied to the $M_G$ patterns of $G$, constructs a disjoint-neighbor witness in $\cI_\ell(\H_0)$ on at least
\[
 \beta\,\frac{N_G}{M_G}\ge\frac{\beta\alpha}2\,n\ge2cn\ge t+1
\]
inputs, where the second inequality uses $c\le\beta\alpha/4$ and the last uses $t+1\le2cn/\log(2en)$. This finishes the proof.
\end{proof}

We now deduce the comparison theorem.

\begin{proof}[Proof of Theorem~\ref{thm:comparison}]
We first take $d=\dpair(\cI_{\ell+1}(\H))<\infty$. Let a $\rho$-packed witness of $\cI_{\ell+1}(\H)$ on $n\ge1$ distinct inputs be given. Theorem~\ref{cmp:thm:extraction}, applied with any choice of representing tuples, gives a disjoint-neighbor witness of $\cI_{\ell+1}(\H)$ on a subsequence $U$ of these inputs with $|U|\ge c_{\ell,\rho}n/\log(n+2)>0$. Hence
\[
 d\ge|U|\ge c_{\ell,\rho}\frac{n}{\log(n+2)}.
\]
If $d=0$, this rules out every $\rho$-packed witness of positive size, so $\dpackr{\rho}(\cI_{\ell+1}(\H))=0$ and \eqref{co:compare} holds. Let $d\ge1$. Since $\log(n+2)\le\log(2en)$, we have
\[
 n\le Cd\log(2en),\qquad C=\max\{1,1/c_{\ell,\rho}\}.
\]
Put $x=n/d$, so that $\log(2en)=\log(2ed)+\log x$. The tangent-line inequality $\log x\le x/(2C)+\log(2C)-1$ gives
\[
 x\le C\bigl[\log(2ed)+\log x\bigr]\le\frac x2+C\log(4Cd),
\]
so $x\le2C\log(4Cd)$. Since $\log(d+2)\ge\log3>1$, we have $\log(4Cd)\le\log(4C)+\log(d+2)\le(1+\log(4C))\log(d+2)$, and therefore
\[
 n\le2C(1+\log(4C))\,d\log(d+2).
\]
The constant depends only on $\ell$ and $\rho$, and $n$ was the size of an arbitrary $\rho$-packed witness. Taking the supremum over all witnesses proves \eqref{co:compare} for the disjoint-neighbor dimension.

Now let $d=\dDS(\cI_{\ell+1}(\H))<\infty$. By Lemma~\ref{lem:orders}, $\dpair(\cI_{\ell+1}(\H))\le d$, so the disjoint-neighbor case applies, and the bound in terms of $d$ follows because $z\mapsto z\log(z+2)$ is increasing for $z\ge0$.
\end{proof}

\subsection{Proof of the Common-Label Lemma}\label{cmp:sec:common-label}

In this subsection we prove Lemma~\ref{cmp:lem:common-label}. For a hypothesis $h\in\H_0$ and an input $i$ of the pool, we write $h(i)$ for the correct-label set of $h$ at $i$. We first prove two auxiliary results: a coverage lemma for partial classes over $[r]$, and a noise inequality on $[r]^n$. We then prove the common-label lemma by induction on the sizes of prescribed sets, called cores.

\paragraph{Partial finite-alphabet coverage.}
Fix $r\ge2$. A \emph{partial class} on $[n]$ over the alphabet $[r]$ is a set $\cF\subseteq([r]\cup\{\star\})^{[n]}$, where $f_i=\star$ means that $f$ is undefined at $i$, and $\supp(f)=\{i:f_i\ne\star\}$. A set $U\subseteq[n]$ is \emph{shattered} by $\cF$ if every vector $b\in[r]^U$ is realized on $U$, that is, some $f\in\cF$ with $U\subseteq\supp(f)$ has $f_i=b_i$ for all $i\in U$. The empty set is shattered exactly when $\cF$ is nonempty. For $r=2$, the largest size of a shattered set is the VC dimension of a partial concept class \cite{alon2021theory}.

\begin{lemma}[Partial coverage]\label{cmp:lem:coverage}
Let $n\ge1$, $r\ge2$, and $0<\kappa\le1/r$, and let $\cF\subseteq([r]\cup\{\star\})^{[n]}$ be nonempty. Let $D$ be the largest size of a set shattered by $\cF$. Suppose that for every $b\in[r]^n$ there is $f^b\in\cF$ with support $E_b$ such that $f^b_i=b_i$ for all $i\in E_b$, and that
\[
 \frac1{r^n}\sum_{b\in[r]^n}|E_b|\ge\Bigl(1-\frac1r+\kappa\Bigr)n.
\]
Then $D\ge1$ and
\[
 n\le C_{r,\kappa}D,
 \qquad C_{r,\kappa}=4a_r^2\kappa^{-4},
 \qquad a_r=\frac1{\log(r/(r-1))}.
\]
A shattered set of size at least $n/C_{r,\kappa}$, together with a member of $\cF$ realizing each vector on it, can be found by finite enumeration.
\end{lemma}

For fixed lists $g_1,\ldots,g_n\subseteq[r]$ of size $r-1$, a uniform vector $b\in[r]^n$ has $b_i\in g_i$ at a $1-1/r$ fraction of the coordinates on average; the lemma says that a class agreeing with random vectors on noticeably more coordinates shatters a set of linear size. The main idea is to compare two bounds on a family $\cG$ of sequences of lists of $r-1$ symbols such that every member of $\cF$ is \emph{covered} by some sequence $g\in\cG$, that is, $f_i\in g_i$ for all $i\in\supp(f)$. On the one hand, for a uniformly random vector $b$, a concentration bound shows that the best sequence in $\cG$ satisfies $b_i\in g_i$ at no more than $(1-1/r)n+\sqrt{n\log|\cG|/2}$ coordinates on average, so the hypothesis forces $\log|\cG|\ge2\kappa^2n$. On the other hand, we construct $\cG$ by reading each member of $\cF$ coordinate by coordinate. At each step we propose a list excluding the symbol whose conditional class has the fewest shattered sets. If the member takes that symbol, we \emph{record} the coordinate and change the list to cover it. Each record divides the number of shattered sets by at least $r/(r-1)$. Since $\cF$ shatters at most $(en/D)^D$ sets, there are $O(D\log(n/D))$ records, and the records determine the sequence of lists; this gives $\log|\cG|=O(D\log^2(n/D))$, and the two bounds together give $n=O(D)$.

\begin{proof}
\emph{The potential.} For a partial class $V$ on a set of coordinates, let $\Phi(V)$ be the number of sets shattered by $V$, the empty set included; thus $\Phi(V)\ge1$ whenever $V\ne\emptyset$. For a coordinate $i$ and a symbol $c\in[r]$, let $V_c$ consist of the members of $V$ with value $c$ at $i$, with coordinate $i$ deleted, and let $V_{-i}$ consist of all members of $V$ with coordinate $i$ deleted. Then
\begin{equation}\label{cmp:eq:coverage-potential}
 \sum_{c=1}^r\Phi(V_c)\le(r-1)\Phi(V),
 \qquad
 \Phi(V_{-i})\le\Phi(V).
\end{equation}
The second inequality holds because every set shattered by $V_{-i}$ is shattered by $V$. For the first, let $U$ be a set of coordinates other than $i$ that is shattered by exactly $\nu\ge1$ of the classes $V_c$. Then $U$ is shattered by $V$, and if $\nu=r$, so is $U\cup\{i\}$. Thus $U$ contributes $\nu$ to the left-hand side, and it accounts for one set counted by $\Phi(V)$ if $\nu<r$ and for two if $\nu=r$. Since $\nu\le r-1$ in the first case and $r\le2(r-1)$ in the second, the contribution of $U$ is at most $r-1$ times the number of sets it accounts for. Distinct sets $U$ account for distinct sets counted by $\Phi(V)$, which proves the claim.

\emph{Lists from few records.} Fix an order on the coordinates and a rule for breaking ties. For each $f\in\cF$ we read the coordinates in order and build a sequence $g=(g_1,\ldots,g_n)$ of lists of $r-1$ symbols that covers $f$. The state is a partial class $V$ on the unread coordinates, initially $\cF$. At coordinate $i$, let $c$ be a symbol minimizing $\Phi(V_c)$. If $f_i\ne c$, put $g_i=[r]\setminus\{c\}$ and replace $V$ by $V_{-i}$. If $f_i=c$, \emph{record} $i$, put $g_i=[r]\setminus\{c'\}$ for the least $c'\ne c$, and replace $V$ by $V_c$. In both cases $f_i\in g_i$ if $f$ is defined at $i$, and the restriction of $f$ to the unread coordinates stays in $V$, so that $\Phi(V)\ge1$ throughout. By \eqref{cmp:eq:coverage-potential}, the minimizing symbol satisfies $\Phi(V_c)\le(1-1/r)\Phi(V)$, so every record multiplies $\Phi(V)$ by at most $1-1/r$, while the other steps do not increase it. Hence every $f$ has at most $\mu:=\lfloor a_r\log\Phi(\cF)\rfloor$ records.

The recorded coordinates determine the whole run: given them, we can replay the procedure, conditioning on the current minimizing symbol at a recorded coordinate and deleting the coordinate otherwise. Thus the sequences of lists form a family $\cG$ that covers every member of $\cF$. Since every shattered set has at most $D$ elements,
\begin{equation}\label{cmp:eq:coverage-count}
 |\cG|\le\sum_{\nu=0}^{\min\{\mu,n\}}\binom n\nu,
 \qquad
 \Phi(\cF)\le\sum_{\nu=0}^{D}\binom n\nu.
\end{equation}

\emph{The lower bound on $|\cG|$.} Let $\mathbf b$ be uniform on $[r]^n$. For a fixed $g\in\cG$, the number of $i$ with $\mathbf b_i\in g_i$ is a sum of $n$ independent Bernoulli variables with mean $1-1/r$. By Hoeffding's lemma and the standard bound on the expected maximum of finitely many subgaussian variables,
\[
 \E\max_{g\in\cG}\bigl|\{i:\mathbf b_i\in g_i\}\bigr|
 \le\Bigl(1-\frac1r\Bigr)n+\sqrt{\frac{n\log|\cG|}2}.
\]
On the other hand, $f^{\mathbf b}$ is covered by some $g\in\cG$, and $\mathbf b_i=f^{\mathbf b}_i\in g_i$ for every $i\in E_{\mathbf b}$; thus $|E_{\mathbf b}|$ is at most the maximum on the left. By the hypothesis, $\kappa n\le\sqrt{n\log|\cG|/2}$, that is,
\begin{equation}\label{cmp:eq:coverage-entropy}
 2\kappa^2n\le\log|\cG|.
\end{equation}
In particular $|\cG|>1$, so $\mu\ge1$ by \eqref{cmp:eq:coverage-count}. Hence $\Phi(\cF)>1$, and $\cF$ shatters a nonempty set, that is, $D\ge1$.

\emph{The upper bound on $|\cG|$.} Put $\xi=n/D\ge1$. If $\xi\le16a_r^2$, the claim holds, because $\kappa\le1/2$ gives $C_{r,\kappa}\ge64a_r^2$. Otherwise, \eqref{cmp:eq:coverage-count} and the bound $\sum_{\nu\le D}\binom n\nu\le(en/D)^D$ give
\[
 \mu\le\mu_0:=a_rD\log(e\xi)<\frac n2,
\]
where the last inequality follows from $\log(e\xi)\le2\sqrt\xi$ and $\xi>16a_r^2$. The function $\nu\mapsto\nu\log(en/\nu)$ is increasing on $(0,n]$, and $a_r\log(e\xi)\ge1$; hence, by \eqref{cmp:eq:coverage-count},
\[
 \log|\cG|
 \le\mu\log\frac{en}{\mu}
 \le\mu_0\log\frac{en}{\mu_0}
 =a_rD\log(e\xi)\log\frac{e\xi}{a_r\log(e\xi)}
 \le a_rD\log^2(e\xi).
\]
Together with \eqref{cmp:eq:coverage-entropy}, this gives $\xi\le a_r\log^2(e\xi)/(2\kappa^2)$. Since $(1+\tau)e^{-\tau/4}\le2$ for all $\tau\ge0$ (its maximum, attained at $\tau=3$, is $4e^{-3/4}$), taking $\tau=\log\xi$ gives $\log^2(e\xi)\le4\sqrt\xi$. Therefore $\sqrt\xi\le2a_r\kappa^{-2}$, that is, $n\le4a_r^2\kappa^{-4}D$.

Finally, $\cF$ is finite, so a largest shattered set and a member realizing each vector on it can be found by enumeration.
\end{proof}

\paragraph{A finite-alphabet noise inequality.}
For the light inputs in the proof below, we need a form of the hypercontractive inequality \cite{odonnell2014analysis} on the product space $[r]^n$. We include the short proof.

\begin{lemma}[Noise inequality]\label{cmp:lem:noise}
Fix $r\ge2$ and $0<\theta<1$. There is $p\in(1,2)$, depending only on $r$ and $\theta$, such that the following holds for every $n\ge0$. Let $\mathbf u,\mathbf v\in[r]^n$ be random vectors whose coordinate pairs $(\mathbf u_i,\mathbf v_i)$ are independent, and each pair is a common uniform symbol with probability $\theta$ and two independent uniform symbols otherwise. Then for all $f,g:[r]^n\to\{0,1\}$,
\[
 \E\bigl[f(\mathbf u)g(\mathbf v)\bigr]\le\bigl(\E[f(\mathbf u)]\,\E[g(\mathbf v)]\bigr)^{1/p}.
\]
\end{lemma}

\begin{proof}
We use the uniform probability measure on $[r]$ and on $[r]^n$, and write $\|f\|_p=(\E|f|^p)^{1/p}$. The proof has two steps: we first prove a one-coordinate bound for the noise operator, and then extend it to $[r]^n$ by the standard tensorization argument.

\emph{One coordinate.} Let $f\ge0$ be a function on $[r]$, and write $f=a+b$ with $a=\E f$ and $\E b=0$. We claim that
\begin{equation}\label{cmp:eq:noise-convexity}
 \|f\|_p^2\ge a^2+(p-1)\|b\|_p^2
 \ge a^2+(p-1)r^{1-2/p}\|b\|_2^2,
 \qquad 1<p\le2.
\end{equation}
If $a=0$, then $f=0$ and there is nothing to prove. Otherwise, for $0\le\tau<1$ put $z_\tau=a+\tau b$, which equals $(1-\tau)a+\tau f$ and is therefore positive, and let $\phi(\tau)=\E z_\tau^p$ and $\psi(\tau)=\phi(\tau)^{2/p}$. Differentiating twice, and dropping a term that is nonnegative because $p\le2$, we obtain
\[
 \psi''(\tau)\ge2(p-1)\,\phi(\tau)^{(2-p)/p}\,\E\bigl[b^2z_\tau^{p-2}\bigr].
\]
H\"older's inequality with exponents $2/p$ and $2/(2-p)$, applied to $|b|^p=(b^2z_\tau^{p-2})^{p/2}z_\tau^{p(2-p)/2}$, shows that $\phi(\tau)^{(2-p)/p}\E[b^2z_\tau^{p-2}]\ge\|b\|_p^2$. Since $\psi(0)=a^2$ and $\psi'(0)=2a^{2-p}\cdot a^{p-1}\E b=0$, integrating twice gives $\psi(\tau)\ge a^2+(p-1)\tau^2\|b\|_p^2$, and letting $\tau\to1$ proves the first inequality in \eqref{cmp:eq:noise-convexity}. The second follows from $\|b\|_p\ge r^{1/2-1/p}\|b\|_2$, which is the comparison between the $\ell_p$ and $\ell_2$ norms of the vector $(b(1),\ldots,b(r))$.

For $0\le\lambda\le1$, let $T_\lambda f=\lambda f+(1-\lambda)\E f$. Choose $p<2$ so close to $2$ that $(p-1)r^{1-2/p}\ge\theta$; this is possible because the left-hand side tends to $1$ as $p\to2$. Since $T_{\sqrt\theta}f=a+\sqrt\theta\,b$ and $\E b=0$, \eqref{cmp:eq:noise-convexity} gives
\[
 \|T_{\sqrt\theta}f\|_2^2=a^2+\theta\|b\|_2^2\le\|f\|_p^2.
\]

\emph{Tensorization.} On $[r]^n$, let $T_\lambda$ act as the one-coordinate operator in every coordinate; it maps nonnegative functions to nonnegative functions. We show $\|T_{\sqrt\theta}f\|_2\le\|f\|_p$ for $f\ge0$ by induction on $n$. Split the coordinates into a block $x$ of $n-1$ coordinates and one further coordinate $y$, and write $T_x$ and $T_y$ for the operators $T_{\sqrt\theta}$ acting on these blocks, so that $T_{\sqrt\theta}=T_xT_y$. The induction hypothesis, applied for each fixed $y$, Minkowski's integral inequality (which applies because $p\le2$), and the one-coordinate bound, applied for each fixed $x$, give
\[
 \|T_xT_yf\|_{L_y^2(L_x^2)}
 \le\|T_yf\|_{L_y^2(L_x^p)}
 \le\|T_yf\|_{L_x^p(L_y^2)}
 \le\|f\|_{L_x^p(L_y^p)}=\|f\|_p.
\]

\emph{The correlated pair.} Given $\mathbf u$, each $\mathbf v_i$ equals $\mathbf u_i$ with probability $\theta$ and is uniform otherwise, so $\E[g(\mathbf v)\mid\mathbf u]=T_\theta g(\mathbf u)$. The operators $T_\lambda$ are self-adjoint and satisfy $T_\lambda T_{\lambda'}=T_{\lambda\lambda'}$. Hence, by the Cauchy--Schwarz inequality and the bound just proved,
\[
 \E\bigl[f(\mathbf u)g(\mathbf v)\bigr]
 =\langle f,T_\theta g\rangle
 =\langle T_{\sqrt\theta}f,T_{\sqrt\theta}g\rangle
 \le\|f\|_p\|g\|_p
 =\bigl(\E f\,\E g\bigr)^{1/p},
\]
where the last equality holds because $f$ and $g$ are indicator functions.
\end{proof}

\paragraph{Extraction by core refinement.}
We now prove Lemma~\ref{cmp:lem:common-label}, using the notation of its statement. The main idea is to replace each common label $a_i^j$ by \emph{cores}, one for each of the $R$ positions of a representing tuple, and to induct on the \emph{deficiency}, the total number of labels by which the cores fall short of size $\ell$. Fix, for every vector $b$, a representing tuple whose hypotheses contain the cores of the groups chosen by $b$, and draw $b$ uniformly at random. Call an input heavy if, given its group, the hypothesis at some position of the random tuple contains a fixed label outside its core with probability at least a small constant $\delta$. If at least half of the inputs are heavy, we add such a label to a core at each heavy input. This lowers the deficiency. The chosen tuples remain compatible with the enlarged cores on more than a $1-1/r$ fraction of the heavy inputs on average. Lemma~\ref{cmp:lem:coverage} then gives a large set of inputs on which we can apply the induction hypothesis. If at least half of the inputs are light, every label outside a core rarely lies in the corresponding hypothesis. We then intersect hypotheses taken from a few correlated copies of the random vector. When the copies agree on group $j$ at input $i$, the intersection contains a fixed set $T_i^j$, defined by the cores of that group. The noise inequality shows that, conditional on this agreement, the intersection equals $T_i^j$ with probability close to one. Lemma~\ref{cmp:lem:coverage}, applied to these intersections, then yields the product of the sets $T_i^j$ on a constant fraction of the inputs.

\begin{proof}[Proof of Lemma~\ref{cmp:lem:common-label}]
Represent each pattern $A^b$ by a tuple of exactly $R$ hypotheses of $\H_0$, repeating hypotheses if needed. If $n=0$, the unique pattern on the empty set is the required witness. If $\ell=1$, all correct-label sets are singletons, so a nonempty intersection of hypotheses equals each of them on its support. Then $A_i^b=\{a_i^{b_i}\}$, and the patterns $A^b$ themselves form the product $\prod_{i\in[n]}\bigl\{\{a_i^j\}:j\in[r]\bigr\}$, which lies in $\cI_1(\H_0)|_{[n]}$; its sets at each input $i$ have empty intersection, since $a_i^j\in Y_i^j$. Thus we may take $a_{1,R,r}=1$, and we assume from now on that $\ell\ge2$ and $n\ge1$.

\paragraph{Constants.}
Set $\theta=1-1/(8r)$, let $p\in(1,2)$ be given by Lemma~\ref{cmp:lem:noise} for $r$ and $\theta$, and put $\epsilon=2/p-1>0$. Fix $0<\delta<1$ such that
\begin{equation}\label{cmp:eq:cl-delta}
 2\ell\delta+\ell\delta^\epsilon\le\frac1{8r}.
\end{equation}
With the constants $C_{r,\kappa}$ of Lemma~\ref{cmp:lem:coverage}, write
\[
 c_{\ell,R}=\sum_{\nu=1}^{\min\{\ell,R\}}\binom R\nu,
 \qquad C_H=C_{r,\delta/r},
 \qquad C_L=C_{r,1/(2r)},
\]
and $\zeta_\Delta=1/\bigl(2c_{\ell,R}C_L(2C_H)^\Delta\bigr)$ for every integer $\Delta\ge0$.

\paragraph{Cores and the induction statement.}
Let $Y_i^j$ be sets with $\bigcap_{j=1}^rY_i^j=\emptyset$ at every input $i\in[n]$. A \emph{core system} consists of nonempty sets $C_s^j(i)$, for $i\in[n]$, positions $s\in[R]$, and groups $j\in[r]$, such that
\begin{equation}\label{cmp:eq:cl-cores}
 |C_s^j(i)|\le\ell,
 \qquad
 \emptyset\ne\bigcap_{s=1}^RC_s^j(i)\subseteq Y_i^j.
\end{equation}
A tuple $(h_1,\ldots,h_R)$ of hypotheses of $\H_0$ is \emph{compatible with group $j$ at $i$} if
\[
 C_s^j(i)\subseteq h_s(i)\quad(s\in[R]),
 \qquad
 \emptyset\ne\bigcap_{s=1}^Rh_s(i)\subseteq Y_i^j.
\]
Since the sets $Y_i^j$ need only have an empty total intersection, a tuple may be compatible with several groups at the same input. The \emph{deficiency} at $i$ is
\[
 \Delta(i)=\sum_{j=1}^r\sum_{s=1}^R\bigl(\ell-|C_s^j(i)|\bigr).
\]
We prove the following statement by induction on the integer $\Delta\ge0$, for every $n\ge1$ and all such sets $Y_i^j$ and cores.
\begin{quote}
Suppose that $\Delta(i)\le\Delta$ for every $i\in[n]$, and that for every $b\in[r]^n$ there is a tuple $(h_{b,1},\ldots,h_{b,R})$ of hypotheses of $\H_0$ that is compatible with group $b_i$ at every $i\in[n]$. Then one can construct a set $U\subseteq[n]$ with $|U|\ge \zeta_\Delta n$ and nonempty sets $T_i^j$ such that
\[
 \prod_{i\in U}\{T_i^j:j\in[r]\}\subseteq\cI_\ell(\H_0)|_U,
 \qquad
 \bigcap_{j=1}^rT_i^j=\emptyset\quad(i\in U).
\]
\end{quote}
The lemma follows from the case $C_s^j(i)=\{a_i^j\}$. Indeed, \eqref{cmp:eq:cl-cores} holds, since $a_i^j\in A_i^b\subseteq Y_i^j$ for every $b$ with $b_i=j$. The representing tuple of $A^b$ is compatible with group $b_i$ at every $i$, since each of its hypotheses contains $A_i^b\ni a_i^{b_i}$ and their intersection is $A_i^b\subseteq Y_i^{b_i}$. All deficiencies equal $rR(\ell-1)$, so the statement gives the lemma with
\[
 a_{\ell,R,r}=\zeta_{rR(\ell-1)}=\frac1{2c_{\ell,R}C_L(2C_H)^{rR(\ell-1)}}.
\]

\paragraph{Heavy and light inputs.}
Assume the hypotheses of the induction statement, and let $\mathbf b$ be uniform on $[r]^n$. Call an input $i$ \emph{heavy} if some group $j$, position $s$, and label $y\notin C_s^j(i)$ satisfy
\begin{equation}\label{cmp:eq:cl-heavy}
 \Pr\bigl[y\in h_{\mathbf b,s}(i)\bigm|\mathbf b_i=j\bigr]\ge\delta,
\end{equation}
and \emph{light} otherwise. If $\Delta=0$, every core has $\ell$ labels, so a core contained in a correct-label set equals it, and no input is heavy. Either at least $n/2$ inputs are heavy, or at least $n/2$ inputs are light; we treat the two cases in turn.

\paragraph{Heavy inputs.}
Suppose that the set $I_{\mathrm h}$ of heavy inputs has at least $n/2$ elements; then $\Delta\ge1$. At each $i\in I_{\mathrm h}$, choose a triple $(j,s,y)$ satisfying \eqref{cmp:eq:cl-heavy} and add $y$ to $C_s^j(i)$. Since the probability in \eqref{cmp:eq:cl-heavy} is positive, some $b$ with $b_i=j$ has $y\in h_{b,s}(i)$. The tuple of this $b$ is compatible with group $j$ at $i$, so the enlarged core is contained in $h_{b,s}(i)$ and still has at most $\ell$ labels. The intersection of the enlarged cores remains nonempty because it contains the previous intersection. Each enlarged core is contained in the corresponding hypothesis of this tuple, so their intersection is still contained in $\bigcap_sh_{b,s}(i)\subseteq Y_i^j$. The other cores are unchanged. Thus the enlarged cores on $I_{\mathrm h}$ satisfy \eqref{cmp:eq:cl-cores}, and their deficiency is at most $\Delta-1$ at every input of $I_{\mathrm h}$.

It remains to find a large set on which every vector has a tuple compatible with the enlarged cores. Let $\cF$ be the partial class on $I_{\mathrm h}$ consisting of all $f$ for which some tuple of hypotheses of $\H_0$ is compatible, for the enlarged cores, with group $f_i$ at every $i\in\supp(f)$. Fix $i\in I_{\mathrm h}$ with chosen triple $(j,s,y)$. If $\mathbf b_i\ne j$, the corresponding cores are unchanged, so the tuple of $\mathbf b$ remains compatible at $i$. If $\mathbf b_i=j$, it remains compatible exactly when $y\in h_{\mathbf b,s}(i)$, an event of conditional probability at least $\delta$. Hence the expected number of inputs of $I_{\mathrm h}$ at which the tuple of $\mathbf b$ remains compatible is at least $(1-1/r+\delta/r)|I_{\mathrm h}|$. For each vector $b'\in[r]^{I_{\mathrm h}}$, choose an extension $b\in[r]^n$ that maximizes the number of such inputs, and let $f^{b'}$ agree with $b'$ at these inputs and be undefined at the other inputs of $I_{\mathrm h}$. Then $f^{b'}\in\cF$, and since a maximum is at least an average, the supports of the functions $f^{b'}$ have average size at least $(1-1/r+\delta/r)|I_{\mathrm h}|$. Lemma~\ref{cmp:lem:coverage}, with $\kappa=\delta/r$, gives a set $U\subseteq I_{\mathrm h}$ shattered by $\cF$ with
\[
 |U|\ge\frac{|I_{\mathrm h}|}{C_H}\ge\frac n{2C_H}.
\]
Shattering means that every vector of $[r]^U$ has a tuple compatible with the enlarged cores at every input of $U$. The induction hypothesis for $\Delta-1$, applied on $U$ (relabeled as $[|U|]$), constructs the required product on at least
\[
 \zeta_{\Delta-1}|U|\ge\frac{\zeta_{\Delta-1}}{2C_H}\,n=\zeta_\Delta n
\]
inputs.

\paragraph{Light inputs: a certificate.}
Suppose now that at least $n/2$ inputs are light. For a light input $i$ and a position $s\in[R]$, put $E_s(i)=\bigcap_{j=1}^rC_s^j(i)$. These sets have empty intersection, since by \eqref{cmp:eq:cl-cores}
\[
 \bigcap_{s=1}^RE_s(i)=\bigcap_{j=1}^r\bigcap_{s=1}^RC_s^j(i)\subseteq\bigcap_{j=1}^rY_i^j=\emptyset.
\]
We find a small subfamily with empty intersection. Some $E_s(i)$ has at most $\ell-1$ labels. Indeed, if all of them had $\ell$ labels, then each core $C_s^j(i)$, which contains $E_s(i)$ and has at most $\ell$ labels, would equal $E_s(i)$ and so would not depend on $j$; the nonempty set $\bigcap_sC_s^1(i)$ would then lie in every $Y_i^j$, a contradiction. Take such an $s$, and for each label of $E_s(i)$ choose a position $s'$ such that $E_{s'}(i)$ omits it. This gives a nonempty \emph{certificate} $S_i\subseteq[R]$ with $|S_i|\le\ell$ and $\bigcap_{s\in S_i}E_s(i)=\emptyset$.

There are at most $c_{\ell,R}$ possible certificates, so some set $I_{\mathrm l}$ of light inputs with a common certificate $S$ satisfies $|I_{\mathrm l}|\ge n/(2c_{\ell,R})$. List $S$ as $s_1,\ldots,s_\nu$; if $S$ is a singleton, repeat its member and set $\nu=2$. Thus $2\le\nu\le\ell$. For $i\in I_{\mathrm l}$ and $j\in[r]$, the sets
\[
 T_i^j=\bigcap_{a=1}^\nu C_{s_a}^j(i)
\]
are nonempty, since they contain $\bigcap_sC_s^j(i)$, and $\bigcap_jT_i^j=\bigcap_{s\in S}E_s(i)=\emptyset$. We now show that intersections of $\nu$ hypotheses realize the product of these sets on a large subset of $I_{\mathrm l}$.

\paragraph{Light inputs: correlated copies.}
Draw a uniform \emph{hub} $\mathbf b\in[r]^n$ and copies $\mathbf b^{(1)},\ldots,\mathbf b^{(\nu)}\in[r]^n$, independently across inputs, as follows: at each input, with probability $\theta$ all copies equal the symbol of the hub, and otherwise they are independent uniform symbols, independent also of the hub. Each copy is uniform on $[r]^n$, and any two copies form a pair as in Lemma~\ref{cmp:lem:noise}. Put
\[
 P=\bigcap_{a=1}^\nu h_{\mathbf b^{(a)},s_a},
\]
restricted to the inputs where this intersection is nonempty. The same $\nu\le\ell$ hypotheses of $\H_0$ are intersected at every input, so every outcome of $P$ belongs to $\cI_\ell(\H_0)$.

Fix $i\in I_{\mathrm l}$ and $j\in[r]$. Condition on $\mathbf b_i=j$ and on the event that all copies have symbol $j$ at $i$; given $\mathbf b_i=j$, this event has probability at least $\theta$. The other inputs keep their joint law, so each copy is distributed as a uniform vector conditioned on having symbol $j$ at $i$, and the bounds that define lightness apply to it. Moreover, the tuple of each copy is compatible with group $j$ at $i$, so $h_{\mathbf b^{(a)},s_a}(i)\supseteq C_{s_a}^j(i)$ for every $a$, and hence $T_i^j\subseteq P_i$. All probabilities in the rest of this paragraph are conditional on these events, and we bound the probability that $P_i$ contains a label outside $T_i^j$. Write $C'=C_{s_1}^j(i)$ and $C''=C_{s_2}^j(i)$, and note that $P_i\subseteq h_{\mathbf b^{(1)},s_1}(i)\cap h_{\mathbf b^{(2)},s_2}(i)$. A label of $P_i$ outside $T_i^j$ is of one of three kinds.
\begin{enumerate}
\item[(i)] A label of $C'\setminus T_i^j$. It lies outside $C_{s_a}^j(i)$ for some $a$, so it lies in $P_i\subseteq h_{\mathbf b^{(a)},s_a}(i)$ with probability less than $\delta$.
\item[(ii)] A label of $C''\setminus C'$. It lies in $P_i$ only if it lies in $h_{\mathbf b^{(1)},s_1}(i)$ outside $C'$, which again has probability less than $\delta$.
\item[(iii)] A label $y\notin C'\cup C''$. It lies in $P_i$ only if it lies in both $h_{\mathbf b^{(1)},s_1}(i)$ and $h_{\mathbf b^{(2)},s_2}(i)$.
\end{enumerate}
As $|C'|,|C''|\le\ell$, labels of kinds (i) and (ii) contribute at most $2\ell\delta$ to the probability that $P_i\ne T_i^j$. For a label $y$ of kind (iii) and $a\in\{1,2\}$, let $p_{a,y}$ be the probability that $y\in h_{\mathbf b^{(a)},s_a}(i)$. Then $p_{a,y}<\delta$ by lightness, and $\sum_yp_{a,y}\le\ell$, because correct-label sets have at most $\ell$ labels. The two events $y\in h_{\mathbf b^{(a)},s_a}(i)$, $a=1,2$, are determined by the coordinates of $\mathbf b^{(1)}$ and $\mathbf b^{(2)}$ at the other $n-1$ inputs, which form a pair as in Lemma~\ref{cmp:lem:noise}. That lemma, applied to the two indicator functions, gives
\[
 \sum_{y\notin C'\cup C''}\Pr\bigl[y\in h_{\mathbf b^{(1)},s_1}(i)\cap h_{\mathbf b^{(2)},s_2}(i)\bigr]
 \le\sum_{y\notin C'\cup C''}(p_{1,y}p_{2,y})^{1/p}
 \le\delta^\epsilon\sum_{y\notin C'\cup C''}\sqrt{p_{1,y}p_{2,y}}
 \le\ell\delta^\epsilon,
\]
where the second inequality uses $(p_{1,y}p_{2,y})^{1/p-1/2}\le\delta^{2/p-1}=\delta^\epsilon$, and the last one is the Cauchy--Schwarz inequality. This also applies when $s_1=s_2$, since the two hypotheses come from different copies. By \eqref{cmp:eq:cl-delta}, the conditional probability that $P_i\ne T_i^j$ is at most $1/(8r)$. Removing the conditioning gives
\begin{equation}\label{cmp:eq:cl-good}
 \Pr\bigl[\text{all copies agree with the hub at $i$, and }P_i=T_i^{\mathbf b_i}\bigr]
 \ge\theta\Bigl(1-\frac1{8r}\Bigr)
 =\theta^2
 \ge1-\frac1{4r},
\end{equation}
and the right-hand side exceeds $1-1/r+1/(2r)$.

\paragraph{Light inputs: coverage.}
Call an input $i\in I_{\mathrm l}$ \emph{good} for an outcome of the hub and the copies if the event in \eqref{cmp:eq:cl-good} occurs at $i$. Let $\cF'$ be the partial class on $I_{\mathrm l}$ consisting of all $f$ for which some outcome of $P$ satisfies $P_i=T_i^{f_i}$ for every $i\in\supp(f)$; when some sets $T_i^j$ coincide, several symbols are allowed at $i$. For each vector $b'\in[r]^{I_{\mathrm l}}$, condition the hub to agree with $b'$ on $I_{\mathrm l}$. Choose the remaining values of the hub and the copies so that the number of good inputs is at least its conditional expectation. Let $f^{b'}$ agree with $b'$ at these good inputs and be undefined elsewhere on $I_{\mathrm l}$. Then $f^{b'}\in\cF'$, and by \eqref{cmp:eq:cl-good} the supports of these functions have average size at least $(1-1/r+1/(2r))|I_{\mathrm l}|$. Lemma~\ref{cmp:lem:coverage}, with $\kappa=1/(2r)$, gives a set $U\subseteq I_{\mathrm l}$ shattered by $\cF'$ with
\[
 |U|\ge\frac{|I_{\mathrm l}|}{C_L}\ge\frac n{2c_{\ell,R}C_L}\ge \zeta_\Delta n.
\]
Shattering means that for every vector $b''\in[r]^U$ some outcome of $P$ satisfies $P_i=T_i^{b''_i}$ for all $i\in U$. Restricted to $U$, these outcomes therefore form exactly the product $\prod_{i\in U}\{T_i^j:j\in[r]\}$, and each of them is the restriction to $U$ of a member of $\cI_\ell(\H_0)$ defined on $U$. Since $\bigcap_jT_i^j=\emptyset$ for every $i\in U$, this proves the induction statement in the light case.

\paragraph{Closing the induction.}
If $\Delta=0$, all inputs are light, and the light case applies. If $\Delta\ge1$, either the heavy case reduces the statement to $\Delta-1$ on a set of at least $n/(2C_H)$ inputs, or the light case constructs the product directly. This proves the induction statement for every $\Delta$, and hence the lemma with $a_{\ell,R,r}=\zeta_{rR(\ell-1)}$. All choices range over finite sets, so the construction is a finite procedure, and every pattern of the product is represented by a tuple of at most $\ell$ hypotheses of $\H_0$.

The product is a DS witness: in direction $i$, fixing the sets at all other inputs of $U$ leaves exactly the distinct sets $T_i^j$, $j\in[r]$, at input $i$, and their intersection is empty; this holds also when sets of different groups coincide. For $r=2$, the sets $T_i^1$ and $T_i^2$ are disjoint, so every pattern of the product has a disjoint neighbor in every direction, and the product is also a disjoint-neighbor witness.
\end{proof}

\subsection{Proof of the Bounded-List Lemma}\label{cmp:sec:list}

We now prove Lemma~\ref{cmp:lem:list}. The blocks may lie in all directions and conflict through arbitrary labels, so we first hash the correct-label sets into codes over a finite alphabet; then most blocks conflict through one fixed family $\mathcal J$ of $r$ codes. Averaging over the multiplicities of the coded patterns gives a family of words over $[r]$ with many complete hyperedges per word, and compression turns it into a down-set, which projects onto a full cube $[r]^E$ with $|E|$ of order $N/M$; undoing the compressions preserves this projection. Reserved symbols of the codes then give common labels on $E$, and Corollary~\ref{cmp:cor:binary} extracts the witness.

\begin{proof}[Proof of Lemma~\ref{cmp:lem:list}]
Identify the pool with $[n]$; as before, every correct-label set below is a subset of the finite set $\Y_0$ of labels used by $\H_0$. All tie-breaking rules are fixed in advance. Every pattern selected below is the restriction of some $v^a$ to a set of coordinates at which it is defined, and it is represented by the given representing tuple of $v^a$; since $\cI_R(\H_0)$ is closed under domain restrictions, it is a pattern of $\cI_R(\H_0)$.

\paragraph{Encoding the correct-label sets.}
Order each list, and let $\iota_i:Q(i)\to[|Q(i)|]$ be the resulting bijection; we call the symbols of $[L]$ \emph{reserved}. Put
\[
 N_0=\ell(\ell+1),\qquad m_0=N_0(N_0-1),
\]
so that $\binom{N_0}2/m_0=1/2$.
Independently for every coordinate $i$ and every label $y\in\Y_0\setminus Q(i)$, choose a uniform value $\chi_i(y)\in[m_0]$, and extend $\iota_i$ by $\iota_i(y)=L+\chi_i(y)$. Thus every label receives a symbol in $[L+m_0]$, the labels of $Q(i)$ receive distinct reserved symbols, and no other label receives a reserved symbol. The \emph{code} of a correct-label set $A$ at $i$ is
\[
 \phi_i(A)=\iota_i(A)=\{\iota_i(y):y\in A\}\qquad\text{if }A\cap Q(i)\ne\emptyset,
\]
and $\phi_i(A)=\star$ if $A$ misses $Q(i)$; we also put $\phi_i(\star)=\star$. Every code other than $\star$ has between one and $\ell$ symbols and contains a reserved symbol, so it belongs to the alphabet
\[
 \mathcal Z=\bigl\{Z\subseteq[L+m_0]:1\le|Z|\le\ell,\ Z\cap[L]\ne\emptyset\bigr\},
\]
which depends only on $\ell$ and $L$. For a partial pattern $v$ we call $\phi(v)=(\phi_1(v_1),\ldots,\phi_n(v_n))$ its \emph{coded pattern}.

A common label of sets that meet the list gives a common symbol of their codes, but hash collisions can create further common symbols. Call a block \emph{surviving} if the codes of its sets have an empty intersection; we show that at least half of the blocks survive in expectation. Let $B\in\cB_i$. Its sets $v_i^a$, $a\in B$, all meet $Q(i)$, so their codes are not $\star$, and together they contain at most $|B|\ell\le N_0$ labels outside $Q(i)$. By the union bound over pairs, these labels receive distinct hash values with probability at least $1-\binom{N_0}{2}/m_0=1/2$, and then $B$ survives. Indeed, a common reserved symbol would be the index of a label of $Q(i)$ lying in every $v_i^a$, $a\in B$. A common symbol $L+w$ would come, in every $v_i^a$, from a label outside $Q(i)$ with hash value $w$, and since these labels have distinct hash values, it would come from one label lying in every $v_i^a$. Either case contradicts $\bigcap_{a\in B}v_i^a=\emptyset$. Hence we may fix hash values for which at least $N/2$ blocks survive; from now on the codes are fixed.

The codes are nonempty sets of at most $\ell$ symbols. By Lemma~\ref{lem:small}, the distinct codes of a surviving block therefore contain a minimal empty-intersection family, which has between two and $\ell+1$ members. The number of such families of symbols of $\mathcal Z$ is at most
\[
 \Gamma=\sum_{j=2}^{\ell+1}\binom{|\mathcal Z|}{j},
\]
which depends only on $\ell$ and $L$. Assign to every surviving block one such family contained in its codes, and fix a family $\mathcal J=\{Z_1,\ldots,Z_r\}$, with $2\le r\le\ell+1$, that is assigned to at least $N/(2\Gamma)$ blocks. The selected blocks now have the same family $\mathcal J$ of conflicting codes, even when they lie in different directions.

\paragraph{Selecting complete collections.}
We now pass from the blocks to the coded patterns. Let $V$ be the set of distinct coded patterns $\phi(v^a)$, $a\in[M]$, and let $\omega(w)$ be the number of indices $a\in[M]$ with $\phi(v^a)=w$, so that $\sum_{w\in V}\omega(w)=M$. A \emph{complete collection} in direction $i$ is a set of $r$ members of $V$ that agree at every coordinate other than $i$ and take the values $Z_1,\ldots,Z_r$ at $i$. Its members differ only at $i$, so a complete collection determines its direction. For $V'\subseteq V$ we write $\cE(V')$ for the set of complete collections contained in $V'$.

Every block $B\in\cB_i$ assigned to $\mathcal J$ determines a complete collection $e_B$ in direction $i$. Indeed, the coded patterns $\phi(v^a)$, $a\in B$, agree off $i$, because the partial patterns $v^a$ agree off $i$, including their domains; and their codes at $i$ include $Z_1,\ldots,Z_r$, so those among them with these codes at $i$ form a complete collection. Now fix $e\in\cE(V)$ with direction $i$. The blocks $B$ with $e_B=e$ belong to $\cB_i$, so they are pairwise disjoint, and each of them contains, for every $w\in e$, an index $a$ with $\phi(v^a)=w$. Hence at most $\min_{w\in e}\omega(w)$ blocks determine $e$, and summing over $e$ gives
\begin{equation}\label{cmp:eq:list-weight}
 \sum_{e\in\cE(V)}\min_{w\in e}\omega(w)\ge\frac{N}{2\Gamma}.
\end{equation}

We remove the multiplicities by splitting $V$ into layers. For an integer $s\ge1$, let $V_s=\{w\in V:\omega(w)\ge s\}$. A complete collection $e$ lies in $V_s$ exactly when $\min_{w\in e}\omega(w)\ge s$, so
\[
 \sum_{s\ge1}|\cE(V_s)|=\sum_{e\in\cE(V)}\min_{w\in e}\omega(w),
 \qquad
 \sum_{s\ge1}|V_s|=\sum_{w\in V}\omega(w)=M.
\]
By weighted averaging and \eqref{cmp:eq:list-weight}, some nonempty layer satisfies
\begin{equation}\label{cmp:eq:list-density}
 \frac{|\cE(V_s)|}{|V_s|}\ge\frac{N}{2\Gamma M}=:\xi.
\end{equation}

Next we fix the values outside $\mathcal J$. Partition $V_s$ according to the set of coordinates at which the value lies outside $\mathcal J$ (the value $\star$ included) and the values at these coordinates. The members of a complete collection agree off its direction and take values in $\mathcal J$ at it, so each complete collection lies in one part, and by the same averaging some part $V'$ satisfies $|\cE(V')|\ge\xi|V'|$. Let $I$ be the set of coordinates at which the members of $V'$ take values in $\mathcal J$. Deleting the other coordinates, at which all members of $V'$ agree, and identifying $Z_j$ with $j$ for $j\in[r]$, we obtain a family $F\subseteq[r]^I$ of words with $|F|=|V'|$. The complete collections of $V'$ become the hyperedges of the one-inclusion hypergraph of $F$ with $r$ vertices, that is, the sets of $r$ words of $F$ that agree off some $i\in I$ and take all values of $[r]$ at $i$. We still call these hyperedges complete collections and write $\cE(F)$ for their set, so that
\[
 |\cE(F)|\ge\xi|F|.
\]

\paragraph{Compression and recovery of a product.}
It remains to find a large set of coordinates on which $F$ projects onto a full cube, and we do this by compression, or shifting, a standard technique for one-inclusion graphs \cite{haussler1994predicting}. For a family $G\subseteq[r]^I$ and $i\in I$, the \emph{compression} of $G$ in direction $i$ replaces, in every hyperedge of $G$ in direction $i$ with $\nu$ words, the $\nu$ values at $i$ by $1,\ldots,\nu$. It keeps the number of words, the set of restrictions of the words to $I\setminus\{i\}$, and the sizes of the hyperedges in direction $i$.

\begin{claim}
Compression keeps $|G|$ and does not decrease $|\cE(G)|$.
\end{claim}

\begin{proof}
Compression in direction $i$ keeps the sizes of the hyperedges in direction $i$, hence the number of complete collections in direction $i$. Let $i'\ne i$, fix the values at all coordinates other than $i$ and $i'$, and for $j\in[r]$ let $\Sigma_j\subseteq[r]$ be the set of values at $i$ of the words of $G$ with these values and with value $j$ at $i'$. The complete collections in direction $i'$ among these words correspond to the values at $i$ that occur with every $j$. Compression replaces each $\Sigma_j$ by $\{1,\ldots,|\Sigma_j|\}$, so their number changes from
\[
 \Bigl|\bigcap_{j=1}^r\Sigma_j\Bigr|
 \qquad\text{to}\qquad
 \min_j|\Sigma_j|\ge\Bigl|\bigcap_{j=1}^r\Sigma_j\Bigr|.
\]
Summing over the fixed values proves the claim.
\end{proof}

Every compression that changes the family strictly decreases the sum of the coordinate values of its words. Repeated compression therefore stops after finitely many steps at a family $F_*$ fixed by every compression, and $|\cE(F_*)|\ge\xi|F_*|$ by the claim. Every hyperedge of $F_*$ in a direction $i$ takes the values $1,\ldots,\nu$ at $i$ for some $\nu$. So $F_*$ is a down-set: lowering a value greater than one of a word of $F_*$ by one gives a word of $F_*$, and repeating this shows that $F_*$ contains every word that is coordinatewise at most a word of $F_*$. In particular, a hyperedge of $F_*$ in direction $i$ is a complete collection exactly when it contains a word with value $r$ at $i$, and this word is unique. Hence
\[
 \sum_{w\in F_*}|\{i\in I:w_i=r\}|=|\cE(F_*)|\ge\xi|F_*|,
\]
and some word $w^*\in F_*$ takes the value $r$ at every coordinate of a set $E\subseteq I$ with $|E|\ge\xi$. As $F_*$ is a down-set, it contains, for every $b\in[r]^E$, the word that equals $b$ on $E$ and $w^*$ off $E$. Thus $F_*|_E=[r]^E$.

We now undo the compressions one at a time, in reverse order, and show that each family $G$ in the sequence from $F$ to $F_*$ satisfies $G|_E=[r]^E$. Suppose that $G'$ is the compression of $G$ in direction $i$ and $G'|_E=[r]^E$. If $i\notin E$, then $G|_E=G'|_E$, since compression keeps the restrictions to $I\setminus\{i\}$. If $i\in E$, fix $b'\in[r]^{E\setminus\{i\}}$, and let $w\in G'$ restrict to $b'$ on $E\setminus\{i\}$ and take the value $r$ at $i$. The hyperedge of $w$ in direction $i$ in $G'$ takes the values $1,\ldots,\nu$ at $i$ and contains $r$, so it has $r$ words. The words of $G$ that agree with $w$ off $i$ form a hyperedge of the same size, so they take all $r$ values at $i$, and they restrict to all $r$ extensions of $b'$ to $E$. As $b'$ was arbitrary, $G|_E=[r]^E$. In particular, $F|_E=[r]^E$.

\paragraph{Recovering groups with a common label.}
Each word of $F$ comes from a coded pattern of $V'$ by deleting coordinates at which all members of $V'$ agree. Hence, by the last paragraph, for every $b\in[r]^E$ there is an index $a(b)\in[M]$ with $\phi_i(v_i^{a(b)})=Z_{b_i}$ for all $i\in E$. In particular, $v^{a(b)}$ is defined on $E$, and we let $A^b$ be its restriction to $E$, a pattern of $\cI_R(\H_0)$ represented by the representing tuple of $v^{a(b)}$. For every $j\in[r]$, fix a reserved symbol $z_j\in Z_j\cap[L]$, and for $i\in E$ put
\[
 a_i^j=\iota_i^{-1}(z_j),
 \qquad
 Y_i^j=\bigcup_{b:\,b_i=j}A_i^b.
\]
The label $a_i^j$ is well defined and lies in $A_i^b$ whenever $b_i=j$. Indeed, for such $b$ the symbol $z_j$ belongs to $Z_j=\iota_i(A_i^b)$, and the only label with the reserved symbol $z_j$ is the label of $Q(i)$ with index $z_j$; so this label exists and lies in $A_i^b$. Hence
\[
 a_i^{b_i}\in A_i^b\subseteq Y_i^{b_i}\qquad(b\in[r]^E,\ i\in E).
\]
Moreover, $\bigcap_{j=1}^rY_i^j=\emptyset$ for every $i\in E$. Indeed, a label $y$ in this intersection lies in some $A_i^b$ with $b_i=j$, for every $j$, so its symbol $\iota_i(y)$ belongs to every $Z_j$; this contradicts $\bigcap_{j=1}^rZ_j=\emptyset$.

Thus the hypotheses of Lemma~\ref{cmp:lem:common-label} hold on the $|E|$ inputs of $E$, with the representing tuples of the partial patterns $v^{a(b)}$, and Corollary~\ref{cmp:cor:binary} constructs a disjoint-neighbor witness in $\cI_\ell(\H_0)$ on at least $a_{\ell,\ell^2,2}\,a_{\ell,R,r}|E|$ inputs, whose patterns are represented by hypotheses of $\H_0$. Since $|E|\ge\xi=N/(2\Gamma M)$ and $2\le r\le\ell+1$, this is at least $\beta_{\ell,R,L}N/M$ with
\[
 \beta_{\ell,R,L}=\frac{a_{\ell,\ell^2,2}}{2\Gamma}\min_{2\le r\le\ell+1}a_{\ell,R,r},
\]
which depends only on $\ell$, $R$, and $L$. This finishes the proof.
\end{proof}


\section{Connection with Cost-Sensitive PAC Learning}\label{app:cost-sensitive}

For finite label spaces, the realizable full-set characterization can also be derived from the cost-sensitive framework of \cite{bressan2025fine}. We first encode partial set-valued hypotheses as total hypotheses over a finite alphabet, and then relate the cube dimensions of \cite{bressan2025fine} to $\dN$. Combined with their characterization, this gives a second proof that realizable full-set learnability is equivalent to $\dN(\H)<\infty$ (Theorem~\ref{thm:A}). Their eliminators also give a sample-complexity bound, which we compare with our direct construction; neither bound dominates the other.

Throughout this appendix, $k=|\Y|<\infty$, $1\le\ell\le k$, and $\H\ne\emptyset$. As in \eqref{eq:allowed}--\eqref{eq:q}, $\cA$ is the collection of nonempty sets of at most $\ell$ labels, and $q=|\cA|$.

\subsection{Encoding the Learning Problem}\label{jc:sec:encoding}

Recall that $h(x)=\star$ means that $h$ is undefined at $x$. Put $\Sigma=\cA\cup\{\star\}$, and let $\H_\Sigma\subseteq\Sigma^{\X}$ consist of the same functions as $\H$, now with $\star$ treated as an ordinary target symbol. The actions also range over $\Sigma$. Fix $s(A)\in A$ for each $A\in\cA$ and any $s(\star)\in\Y$, and call $s(B)$ the label decoded from the action $B$. The cost of the action $B$ against the target symbol $A$ is
\begin{equation}\label{jc:eq:cost}
 w(B,A)=
 \begin{cases}
  \ind\{s(B)\notin A\},&A\in\cA,\\
  0,&A=\star,
 \end{cases}
 \qquad B\in\Sigma.
\end{equation}
This cost has zero diagonal, as required in \cite{bressan2025fine}. For every $x\in\supp(h)$, the cost $w(B,h(x))=\ind\{s(B)\notin h(x)\}$ is exactly the validity loss of the decoded label $s(B)$, and every label $y\in\Y$ is decoded from the action $\{y\}$, since $s(\{y\})=y$.

In the cost-sensitive problem, the input distribution may assign mass to points where the original target is undefined. We first show that this does not change qualitative learnability.

\begin{lemma}\label{jc:lem:encoding}
The class $\H$ is realizably PAC learnable from full sets if and only if $\H_\Sigma$ is $(w,0)$-learnable in the sense of \cite{bressan2025fine}.
\end{lemma}

\begin{proof}
Suppose first that $\H_\Sigma$ is $(w,0)$-learnable. A full-set sample labeled by $h\in\H$ is a sample for $\H_\Sigma$ whose inputs lie in $\supp(h)$. Run the cost-sensitive learner on it at accuracy $\eps/(\ell+1)$ and decode its predictions by $s$. Since the cost is the validity loss on $\supp(h)$, this gives a full-set learner. If the learner outputs distributions over actions, we predict a most probable decoded label, which by Lemma~\ref{lem:mode} multiplies the error by at most $\ell+1$.

Conversely, let $m=\mfull_\H(\eps,\delta/2)$. Fix a target $h\in\H$ and an arbitrary input distribution $P$, and write $p=P(\supp(h))$. The idea is to discard the examples where the target is undefined. Draw
\[
 N=\left\lceil\frac{2}{\eps}\left(m+4\log\frac2\delta\right)\right\rceil
\]
examples and discard those labeled $\star$. If at least $m$ examples remain, run the full-set learner on the first $m$ of them and encode the returned predictor $g$ by the singleton actions $\{g(x)\}$; otherwise return a fixed prediction.

If $p\le\eps$, every predictor has $w$-risk at most $\eps$, because the cost vanishes outside $\supp(h)$. Now suppose that $p>\eps$. Then $Np>2m+8\log(2/\delta)$, and since the number $Z$ of retained examples has mean $Np$, a Chernoff bound gives
\[
 \Pr[Z<m]\le\exp(-Np/8)\le\delta/2.
\]
Conditional on the retained positions, the retained inputs are independent draws from $P$ conditioned on $\supp(h)$. Thus, conditional on $Z\ge m$, with probability at least $1-\delta/2$ the predictor $g$ has validity error at most $\eps$ under this conditional distribution. Its $w$-risk is $p$ times this error, hence at most $p\eps\le\eps$. The two failure probabilities sum to at most $\delta$.
\end{proof}

\subsection{Conflicts and Cube Dimensions}\label{jc:sec:dimensions}

In the terminology of \cite{bressan2025fine}, a family $K\subseteq\Sigma$ is \emph{dice-attainable at threshold zero} if
\[
 \forall\,\mu\in\Delta(K)\ \ \exists\,\nu\in\Delta(\Sigma):\quad \E_{A\sim\mu,\,B\sim\nu}\bigl[w(B,A)\bigr]=0,
\]
and it is \emph{obstructed} otherwise. Since $\star$ costs nothing, only the defined symbols of $K$ matter: put $I(K)=\bigcap_{A\in K\setminus\{\star\}}A$, with the intersection of the empty family interpreted as $\Y$. We claim that $K$ is dice-attainable at threshold zero exactly when $I(K)\ne\emptyset$.

Indeed, a label $y\in I(K)$ gives the action $\{y\}$, which has zero cost against every symbol of $K$. Conversely, if $I(K)=\emptyset$, take the target symbol uniformly from $K\setminus\{\star\}$. Every decoded label is excluded by at least one of these sets, so every action distribution incurs cost at least $1/|K\setminus\{\star\}|>0$. Thus the obstructed families are precisely
\begin{equation}\label{jc:eq:obstructions}
 \mathfrak J_0=\{K\subseteq\Sigma:I(K)=\emptyset\}.
\end{equation}

Recall the collection $\mathfrak C$ of minimal empty-intersection families in $\cA$ and its size $\Gamma$ from \eqref{eq:finite-Gamma}. By Lemma~\ref{lem:small}, every $J\in\mathfrak C$ has between two and $\ell+1$ members. Every member of $\mathfrak C$ is obstructed, and every $K\in\mathfrak J_0$ contains some $J\in\mathfrak C$, namely a minimal subfamily of $K\setminus\{\star\}$ with empty intersection. Hence the minimal obstructed families are exactly the members of $\mathfrak C$, and none of them contains $\star$.

For $J\subseteq\Sigma$ with $|J|\ge2$, the $J$-cube dimension $d_J$ of $\H_\Sigma$ is the supremum of $n$ such that $J^n\subseteq\H_\Sigma|_S=\{(h(x_1),\ldots,h(x_n)):h\in\H_\Sigma\}$ for some sequence $S=(x_1,\ldots,x_n)$ of distinct inputs, with entries equal to $\star$ allowed \cite[Definition~2]{bressan2025fine}. If $J\subseteq\cA$, the cube uses only defined values, so $J^n\subseteq\H_\Sigma|_S$ is equivalent to $J^n\subseteq\H|_S$. The next lemma shows that $\dN$ is sandwiched between the cube dimensions of the conflict families.

\begin{lemma}\label{jc:lem:dimensions}
With the maximum and sum over an empty family both interpreted as zero,
\begin{equation}\label{jc:eq:dimensions}
 \max_{J\in\mathfrak C}d_J
 \ \le\ \dN(\H)
 \ \le\ \sum_{J\in\mathfrak C}d_J
 \ \le\ \Gamma\max_{J\in\mathfrak C}d_J.
\end{equation}
Moreover, all the dimensions $d_K$, $K\in\mathfrak J_0$, are finite if and only if all the dimensions $d_J$, $J\in\mathfrak C$, are finite.
\end{lemma}

\begin{proof}
A $J$-cube with $J\in\mathfrak C$ is a product witness, with the family $J$ at every input (Definition~\ref{def:nat}); this proves the first inequality. For the second, take a product witness $\cA_1\times\cdots\times\cA_n$ and choose a minimal empty-intersection family $J_i\subseteq\cA_i$ at every coordinate. Group the coordinates according to the chosen family: for $J\in\mathfrak C$, let $T_J=\{i:J_i=J\}$. The projection onto $T_J$ contains a $J$-cube, so $|T_J|\le d_J$. Summing gives $n\le\sum_Jd_J$, and the second inequality follows because the witness was arbitrary. The last inequality follows by bounding each summand by the maximum.

For the final assertion, every obstructed family $K$ contains a $J\in\mathfrak C$. A $K$-cube contains a $J$-cube on the same inputs, so $d_K\le d_J$. Conversely, every $J\in\mathfrak C$ is itself obstructed.
\end{proof}

Theorem~4 of \cite{bressan2025fine} characterizes $(w,0)$-learnability by finiteness of the cube dimensions of all obstructed families. Combined with the two lemmas, it gives the characterization, for partial hypotheses as well.

\begin{corollary}\label{jc:eq:characterization}
The following are equivalent:
\begin{enumerate}[(i)]
 \item $\H$ is realizably PAC learnable from full sets;
 \item $d_J<\infty$ for every $J\in\mathfrak C$;
 \item $\dN(\H)<\infty$.
\end{enumerate}
\end{corollary}

\begin{proof}
By Lemma~\ref{jc:lem:encoding} and Theorem~4 of \cite{bressan2025fine}, (i) holds if and only if $d_K<\infty$ for every $K\in\mathfrak J_0$, which by Lemma~\ref{jc:lem:dimensions} is equivalent to (ii). Since the alphabet is finite, $\Gamma<\infty$, so the sum in \eqref{jc:eq:dimensions} has finitely many terms and \eqref{jc:eq:dimensions} shows that (ii) is equivalent to (iii).
\end{proof}

\subsection{Quantitative Bounds}\label{jc:sec:quantitative}

The reduction also yields a sample-complexity bound. Here the eliminators of \cite{bressan2025fine} are trained directly on full-set samples, whose inputs lie in the support of the target, so the filtering step of Lemma~\ref{jc:lem:encoding} is not needed. Assume $d=\dN(\H)<\infty$, and put
\[
 \mathfrak C_+=\{J\in\mathfrak C:d_J>0\},\qquad
 \Gamma_+=|\mathfrak C_+|,\qquad
 D_{\mathrm{sum}}=\sum_{J\in\mathfrak C}d_J.
\]

\begin{proposition}
If $\Gamma_+\ge1$, then
\begin{equation}\label{jc:eq:elimination-rate}
 \mfull_\H(\eps,\delta)
 \le O_\ell\!\left(
 \frac{D_{\mathrm{sum}}+\Gamma_+\log(2\Gamma_+/\delta)}{\eps}
 \right).
\end{equation}
If $\Gamma_+=0$, a single predictor, fixed in advance, is valid on $\supp(h)$ for every $h\in\H$.
\end{proposition}

The learner trains one eliminator for each $J\in\mathfrak C$ and predicts a common label of the sets that survive. Only the families in $\mathfrak C$ are needed, since an eliminator for $J$ excludes a member of every obstructed family containing $J$.

\begin{proof}
Fix a target $h\in\H$ and a distribution $P$ supported on $\supp(h)$. A $J$-eliminator is a function $e_J:\X\to J$, and its error is $\Pr_{x\sim P}[e_J(x)=h(x)]$. Theorem~10 of \cite{bressan2025fine} gives error at most $\eta$ with confidence $1-\xi$ from $O((d_J+\log(1/\xi))/\eta)$ examples. If an eliminator is randomized, taking a most probable symbol multiplies its error by at most $|J|\le\ell+1$, a factor absorbed in \eqref{jc:eq:elimination-rate}.

If $d_J=0$, then at every input some member of $J$ is absent from $\{h'(x):h'\in\H\}$. Fix an order on $J$ and choose the first such member as $e_J(x)$; this eliminator has zero error and needs no examples. For $J\in\mathfrak C_+$, put
\[
 a_J=d_J+\log\frac{2\Gamma_+}{\delta},\qquad
 T=\sum_{J\in\mathfrak C_+}a_J,\qquad
 \eta_J=\frac{\eps a_J}{T},\qquad
 \xi_J=\frac{\delta}{\Gamma_+}.
\]
Since $(d_J+\log(1/\xi_J))/\eta_J\le a_J/\eta_J=T/\eps$, each eliminator requires $O_\ell(T/\eps)$ examples. Train them all on the same full-set sample, viewed as labeled examples for $\H_\Sigma$. A union bound shows that, with probability at least $1-\delta$, every trained eliminator has error at most $\eta_J$; independence between the eliminators is not required.

For a test input $x$, remove all the excluded symbols and let
\[
 \mathcal R(x)=\cA\setminus\{e_J(x):J\in\mathfrak C\}.
\]
No $J\in\mathfrak C$ is contained in $\mathcal R(x)$, since its chosen symbol $e_J(x)$ was removed. Thus, if $\mathcal R(x)\ne\emptyset$, its members have a common label, for otherwise $\mathcal R(x)$ would contain a minimal empty-intersection family, which lies in $\mathfrak C$. Predict the first such label, or a fixed label if $\mathcal R(x)=\emptyset$. Whenever no eliminator excludes the true set $h(x)$, that set remains in $\mathcal R(x)$ and the prediction is valid. On the simultaneous success event, the risk is therefore at most $\sum_{J\in\mathfrak C_+}\eta_J=\eps$. If $\Gamma_+=0$, all eliminators are exact and the same construction needs no examples.
\end{proof}

\paragraph{Comparison with Corollary~\ref{cor:finite-full-rate}.}
Our direct bound is
\begin{equation}\label{jc:eq:direct-rate}
 \mfull_\H(\eps,\delta)
 \le O_\ell\!\left(
 \frac{d\log((d+1)(k+1)/\eps)+\log(2/\delta)}{\eps}
 \right).
\end{equation}
By \eqref{jc:eq:dimensions}, $D_{\mathrm{sum}}\ge d$, and since only the $\Gamma_+$ families in $\mathfrak C_+$ contribute to $D_{\mathrm{sum}}$, each with $d_J\le d$, also $D_{\mathrm{sum}}\le\Gamma_+d$. Neither bound dominates the other. When few families have positive dimension, \eqref{jc:eq:elimination-rate} can be smaller, and it has no logarithmic factor in $d$ or $1/\eps$; the inverse-linear dependence on $\eps$ is already present in \cite[Corollary~13]{bressan2025fine}. In exchange, it sums contributions from the fixed conflict families.

Our bound avoids this sum because the potential of Lemma~\ref{lem:potential-inequality} counts products in which the conflict family may vary from one coordinate to another. Its mistake bound on $n$ inputs is $O_\ell(d\log(1+n\Gamma))$ for $d\ge1$. Since $q\le(k+1)^\ell$ and $\Gamma\le(q+1)^{\ell+1}$, we have $\log\Gamma=O_\ell(\log(k+1))$, which gives the logarithmic dependence on $k$ in \eqref{jc:eq:direct-rate} for fixed $\ell$.

The sum over conflict families can be much larger than $d$. For $k\ge2$ and $d\ge1$, take all functions from a $d$-point domain to $\cA$. Then $\dN=d_J=d$ for every $J\in\mathfrak C$, so $D_{\mathrm{sum}}=\Gamma d$. Moreover, $\Gamma\ge\binom{k}{2}$ because each pair of distinct singleton sets is a minimal conflict family. This compares the two upper bounds; it does not show that the family factor is necessary.

The constructions also differ in what they provide. A PAC $J$-eliminator excludes one target symbol for a fixed family $J$. Lemmas~\ref{lem:potential-inequality} and~\ref{lem:predict} give a learner that predicts labels in $\Y$ and changes its state only after a mistake. It satisfies the transductive guarantee of Theorem~\ref{thm:online}, and its possible runs form the finite cover of Lemma~\ref{lem:finite-cover}, used in the agnostic and bandit arguments.

The reduction above concerns the realizable full-set equivalence. In general, demonstrations and bandit answers do not reveal the encoded target symbol, while countable label spaces need not admit the finite collection of conflict families used here. The main text treats these settings with additional constructions.

\section{Hypergraph Coloring with Forbidden Crossings}\label{app:coloring}

The counting argument behind Theorem~\ref{thm:online} has a purely combinatorial counterpart, which we present here. It bounds the chromatic number of a uniform hypergraph whose hyperedges are covered by a few complete multipartite hypergraphs, no two of which cross. For graphs whose edges are partitioned into complete bipartite graphs, this is the setting of the Alon--Saks--Seymour problem, and Remark~\ref{rem:ass} compares our bound with the known ones.

Throughout, $\chi(G)$ denotes the \emph{weak chromatic number} of a hypergraph $G$, the minimum number of colors in a \emph{proper} vertex coloring, that is, one with no monochromatic hyperedge. For $W\subseteq V(G)$ we write $G[W]$ for the induced hypergraph on $W$, and we call $W$ \emph{independent} if it contains no hyperedge. For pairwise disjoint nonempty sets $A^1,\ldots,A^r$, we write $K^{(r)}(A^1,\ldots,A^r)$ for the complete $r$-partite $r$-uniform hypergraph with these vertex classes; its hyperedges are the sets that contain exactly one vertex of each $A^a$. Two such hypergraphs \emph{cross} if every vertex class of one meets every vertex class of the other.

\begin{theorem}\label{thm:coloring}
Let $r\geq2$ and $m\geq1$ be integers, let $G$ be an $r$-uniform hypergraph, and let $K_i=K^{(r)}(A_i^1,\ldots,A_i^r)$, $i\in[m]$, be complete $r$-partite $r$-uniform hypergraphs such that every hyperedge of $G$ is a hyperedge of some $K_i$. Suppose that no two of the $K_i$ cross, that is, for every pair $i\neq j$ there are $a,b\in[r]$ such that
\begin{equation}\label{app:eq:fc}
A_i^a\cap A_j^b=\emptyset.
\end{equation}
Then, with $\alpha=(r-1)/r$ and $L=\lceil\log_{1/\alpha}m\rceil+1$,
\[
\chi(G)\leq\sum_{t=0}^{L}m^t\alpha^{\binom t2}.
\]
In particular,
\begin{equation}\label{app:eq:coloring-exp}
\chi(G)\leq\exp\biggl(\frac{(\log m)^2}{2\log\bigl(\frac{r}{r-1}\bigr)}+O_r\bigl(\log(m+1)\bigr)\biggr).
\end{equation}
\end{theorem}

The proof follows the potential argument of Section~\ref{ff:sec:products}. For $W\subseteq V(G)$, call an index $i\in[m]$ \emph{active} on $W$ if $W$ meets every class of $K_i$, and let $\nu(W)$ be the number of indices active on $W$; this count replaces the potential $\Phi$. Inside a class of an active $K_i$, the index $i$ is no longer active, and since no two of the $K_j$ cross, every other active index remains active in at most $r-1$ of the $r$ classes. This is the analogue of Lemma~\ref{lem:potential-inequality}: some class of $K_i$ keeps at most an $\alpha$-fraction of the other active indices. Removing one such class for each active index leaves an independent set, which gets a single color, and recursing on the removed classes gives depth $O_r(\log m)$ and the stated bound.

\begin{proof}
If $W\subseteq V(G)$ contains a hyperedge, this hyperedge belongs to some $K_i$ and has a vertex in each $A_i^a$, so $i$ is active on $W$. Hence $\nu(W)=0$ implies that $W$ is independent. Moreover, $\nu$ is monotone: $W'\subseteq W$ implies $\nu(W')\leq\nu(W)$.

\paragraph{A contraction inequality.}
Fix $W$ with $s=\nu(W)\geq1$, and let $i$ be active on $W$. We claim that
\begin{equation}\label{app:eq:contraction}
\sum_{a=1}^{r}\nu(W\cap A_i^a)\leq(r-1)(s-1).
\end{equation}
Indeed, the classes $A_i^1,\ldots,A_i^r$ are pairwise disjoint, so each set $W\cap A_i^a$ misses the other classes of $K_i$, and $i$ is inactive on it. An index that is inactive on $W$ is inactive on all these subsets as well. Finally, an index $j\neq i$ that is active on $W$ is active on at most $r-1$ of the sets $W\cap A_i^1,\ldots,W\cap A_i^r$, since otherwise $W\cap A_i^a\cap A_j^b\neq\emptyset$ for all $a,b\in[r]$, contrary to \eqref{app:eq:fc}. Since there are $s-1$ such indices $j$, summing over all indices gives \eqref{app:eq:contraction}.

\paragraph{A recursive coloring.}
The $r$ integers on the left of \eqref{app:eq:contraction} sum to at most $(r-1)(s-1)$, so the smallest of them is at most $\alpha(s-1)$, and we may choose $a_i\in[r]$ such that
\begin{equation}\label{app:eq:small}
\nu(W\cap A_i^{a_i})\leq\bigl\lfloor\alpha(s-1)\bigr\rfloor.
\end{equation}
We make this choice for every index active on $W$. Let $I(W)$ be the set of these $s$ indices, and put
\[
W_0=W\setminus\bigcup_{i\in I(W)}A_i^{a_i}.
\]
The set $W_0$ is independent, because every hyperedge contained in $W$ lies in some $K_i$ with $i\in I(W)$ and therefore meets $A_i^{a_i}$. Partition $W\setminus W_0$ into at most $s$ disjoint sets $W_i\subseteq W\cap A_i^{a_i}$, $i\in I(W)$; by \eqref{app:eq:small} and monotonicity, $\nu(W_i)\leq\lfloor\alpha(s-1)\rfloor$. Give $W_0$ one color, and color each $G[W_i]$ recursively with its own palette, the palettes being disjoint. The resulting coloring of $G[W]$ is proper, since each color class lies either in $W_0$ or in one of the sets $W_i$, where it is a color class of a proper coloring of $G[W_i]$.

To count the colors, let $F(0)=1$ and
\[
F(s)=1+sF\bigl(\lfloor\alpha(s-1)\rfloor\bigr)\qquad(s\geq1).
\]
Since $\lfloor\alpha(s-1)\rfloor$ is smaller than $s$ and nondecreasing in $s$, an induction on $s$ shows that $F$ is nondecreasing. We now show by induction on $\nu(W)$ that $\chi(G[W])\leq F(\nu(W))$ for every $W\subseteq V(G)$. If $\nu(W)=0$, then $W$ is independent and one color suffices. If $\nu(W)=s\geq1$, the coloring above uses one color on $W_0$ and, by the induction hypothesis, at most $F(\nu(W_i))\leq F(\lfloor\alpha(s-1)\rfloor)$ colors on each of the at most $s$ sets $W_i$, hence at most $1+sF(\lfloor\alpha(s-1)\rfloor)=F(s)$ colors in total. Monotonicity of $F$ is needed here because we only know an upper bound on $\nu(W_i)$. Since $\nu(V(G))\leq m$, we obtain $\chi(G)\leq F(m)$.

\paragraph{Solving the recurrence.}
Let $s_0=m$, let $s_{j+1}=\lfloor\alpha(s_j-1)\rfloor$ while $s_j\geq1$, and keep $s_{j+1}=0$ once $s_j=0$. Then $s_j\leq m\alpha^j$ for every $j$, and since $L-1\geq\log_{1/\alpha}m$, we get $s_{L-1}\leq1$ and hence $s_L=0$. Unrolling $F(s_j)=1+s_jF(s_{j+1})$ until the sequence reaches zero, and then using $s_j\leq m\alpha^j$, gives
\[
F(m)\leq\sum_{t=0}^{L}\prod_{j=0}^{t-1}s_j\leq\sum_{t=0}^{L}m^t\alpha^{\binom t2}.
\]
This is the first bound of the theorem. For the second, put $\beta=\log(1/\alpha)=\log(r/(r-1))$. The logarithm of $m^t\alpha^{\binom t2}$ is $t\log m-\frac\beta2t(t-1)$, a quadratic in $t$ whose maximum over the reals is $\frac{(\log m)^2}{2\beta}+\frac12\log m+\frac\beta8$. Since there are $L+1$ terms,
\[
\log F(m)\leq\frac{(\log m)^2}{2\beta}+\frac12\log m+\frac\beta8+\log(L+1)=\frac{(\log m)^2}{2\beta}+O_r\bigl(\log(m+1)\bigr).
\]
This finishes the proof.
\end{proof}

When the hyperedge sets of the $K_i$ are disjoint, no two of them cross, which gives the following corollary.

\begin{corollary}[Multipartite edge partitions]\label{cor:partitions}
Fix $r\geq2$. If the hyperedge set of an $r$-uniform hypergraph $G$ is partitioned into the hyperedge sets of $m\geq1$ complete $r$-partite $r$-uniform hypergraphs, then $\chi(G)$ satisfies \eqref{app:eq:coloring-exp}.
\end{corollary}

\begin{proof}
Let the complete $r$-partite hypergraphs be $K_i=K^{(r)}(A_i^1,\ldots,A_i^r)$, $i\in[m]$, and fix $i\neq j$. If all $r^2$ intersections $A_i^a\cap A_j^b$ were nonempty, we could choose $v_a\in A_i^a\cap A_j^a$ for each $a\in[r]$. These vertices are distinct, since the classes of $K_i$ are disjoint, and $\{v_1,\ldots,v_r\}$ is a hyperedge of both $K_i$ and $K_j$, contrary to the disjointness of their hyperedge sets. Thus $K_i$ and $K_j$ do not cross, and Theorem~\ref{thm:coloring} applies.
\end{proof}

\begin{remark}\label{rem:ass}
For $r=2$, Corollary~\ref{cor:partitions} concerns graphs whose edge set is partitioned into $m$ complete bipartite graphs, the setting of the Alon--Saks--Seymour problem. For such graphs, \cite{mubayi2009bipartite} proved a quasipolynomial upper bound with leading exponent $(\log m)^2/(2\log 2)$, which is also what Theorem~\ref{thm:coloring} gives for $r=2$, and \cite{fox2026note} improved the exponent to $(\log m)^2/(4\log 2)+O(\log(m+1))$. For graph edge partitions, therefore, Theorem~\ref{thm:coloring} is weaker than the best known bound; it extends the quasipolynomial order to $r\geq3$ and to covers whose parts may overlap but do not cross. In the other direction, Alon, Saks and Seymour conjectured that $\chi\leq m+1$ for such graphs, and \cite{huang2012counterexample} disproved this conjecture. Strengthening the lower bound of \cite{balodis2021unambiguous}, \cite{pabbaraju2026optimal} constructed such graphs with chromatic number $\exp(\Omega((\log m)^2))$. Thus the exponent $(\log m)^2$ is necessary up to a constant factor already when $r=2$.
\end{remark}

\section{Standard Borel Instance Spaces}\label{app:measurability}

The main proofs assume that $\X$ is countable, so that the rules a learner builds on a finite pool, such as the transductive rules of Lemma~\ref{lem:packed-weak} and the minimax predictions of Theorem~\ref{thm:family-conversion}, need no measurability conditions. This appendix extends Theorems~\ref{thm:A}--\ref{thm:F} to standard Borel instance spaces under two mild assumptions on $\H$. The label space stays countable, and the sample-complexity bounds of Theorems~\ref{thm:A}--\ref{thm:E} are unchanged. We choose the pool rules measurably and, for the agnostic demonstration learner, replace independent rounding at individual inputs by rounding on finitely many Borel sets.

\subsection{Assumptions and Statement}\label{app:meas-assumptions}

Let $\X$ be a nonempty standard Borel space, let $\Y$ be a nonempty countable discrete space, and fix $\ell\ge1$. Equip $\cA\cup\{\star\}$ with its discrete $\sigma$-algebra. We may assume $\H\ne\emptyset$, since otherwise the learning requirements are vacuous. We impose the following assumptions on $\H$:
\begin{enumerate}[(i)]
\item Every $h\in\H$ is Borel measurable.
\item For every $n\ge1$ and $v\in(\cA\cup\{\star\})^n$, the set
\[
 E_{n,v}=\bigl\{(x_1,\ldots,x_n)\in\X^n:
       (h(x_1),\ldots,h(x_n))=v\text{ for some }h\in\H\bigr\}
\]
is Borel.
\end{enumerate}
Assumption~(ii), unlike~(i), is a condition on the class as a whole: whether $\H$ realizes a given finite pattern must depend measurably on the inputs. Both assumptions hold when $\X$ is countable or $\H$ is a countable class of Borel functions, but we do not require countability.

Two facts about standard Borel spaces are used throughout: equality of inputs is a Borel condition, and $\X$ admits a Borel injection into $[0,1]$, hence a Borel linear order. We use this order wherever the proofs sort a finite pool by an enumeration of $\X$; sorting finitely many inputs is Borel.

\begin{theorem}\label{thm:standard-borel-extension}
Let $\X$ be a standard Borel space and let $\H$ satisfy assumptions \textup{(i)}--\textup{(ii)}. Take all distributions to be Borel probability measures and all domain restrictions, including those in the intersection classes, to be restrictions to Borel sets. Require the prediction map of a learner, or each of its label probabilities for randomized predictions, to be jointly Borel in the sample, the random seed and the test input. Then the conclusions of Theorems~\ref{thm:A}--\ref{thm:F} remain valid, with the same sample-complexity bounds in Theorems~\ref{thm:A}--\ref{thm:E}.
\end{theorem}

By Fubini's theorem, these conventions make the error of a learner's output a Borel function of the sample and the seed, so the risks and success events are measurable. The necessity directions need nothing new, since they use finite witnesses and finitely supported distributions; the explicit lower-bound and separating classes, including those of Theorem~\ref{thm:G}, have countable domains; and the comparison in Appendix~\ref{app:comparison} is combinatorial. It remains to make the upper-bound constructions measurable. On a general $\X$, Theorem~\ref{thm:family-conversion} also needs its prediction families to depend measurably on the pool; Section~\ref{app:meas-families} states these conditions and checks them for the families of Sections~\ref{sec:finite-one} and~\ref{sec:countable-one}.

\paragraph{Intersection classes.}
Restricting the domains in~\eqref{eq:intersections} to Borel sets changes no finite trace, and hence no dimension, because a restriction on a finite pool can be realized by restricting to the finitely many inputs where the desired pattern is defined. Assumption~(i) passes to $\cI_R(\H)$, since the constituent hypotheses and the restricting domain are Borel.

For assumption~(ii), fix a pool size $n$ and a partial pattern $v\in(\cA\cup\{\star\})^n$. Call a tuple $(w^1,\ldots,w^r)$ of patterns in $(\cA\cup\{\star\})^n$ \emph{compatible with $v$} if
\[
 w^a_i\ne\star\ \text{for all }a\in[r]\quad\text{and}\quad\bigcap_{a=1}^r w^a_i=v_i
 \qquad\text{whenever }v_i\ne\star,
\]
with no condition at the positions where $v_i=\star$. Then $v$ is realized by $\cI_R(\H)$ on $(x_1,\ldots,x_n)$ if and only if $x_i=x_j$ implies $v_i=v_j$ and
\[
 (x_1,\ldots,x_n)\in\bigcup_{r=1}^R\ \bigcup_{(w^1,\ldots,w^r)\text{ compatible with }v}\ \bigcap_{a=1}^rE_{n,w^a}.
\]
Indeed, if $h_1,\ldots,h_r$ realize a compatible tuple, restricting their intersection to the finite set $\{x_i:v_i\ne\star\}$ realizes $v$, where the first condition ensures that no input carries both a defined and an undefined entry of $v$; conversely, a representing tuple of an intersection realizing $v$ realizes a compatible tuple. The first condition is Borel, and the union is countable because $\cA\cup\{\star\}$ is countable, so this event is Borel by~(ii).

\subsection{Choosing the Pool Rules Measurably}\label{app:meas-selection}

The compactness arguments in Lemma~\ref{lem:packed-weak} and Theorem~\ref{thm:family-conversion} choose, for each pool, a point satisfying countably many closed constraints, and the pool determines which constraints are in force. The following lemma makes this choice Borel whenever the pool determines the constraints in a Borel way.

\begin{lemma}[Borel choice under closed constraints]\label{lem:meas-selection}
Let $K$ be a compact metrizable space and $C_1,C_2,\ldots$ closed subsets of $K$, and for $b\in\{0,1\}^{\N}$ let $K_b=\bigcap_{j:b_j=1}C_j$. Let $T$ be a standard Borel space and $S\mapsto b(S)$ a Borel map from $T$ to $\{0,1\}^{\N}$ such that $K_{b(S)}\ne\emptyset$ for every $S\in T$. Then there is a Borel map $\zeta:T\to K$ with $\zeta(S)\in K_{b(S)}$ for every $S\in T$.
\end{lemma}

\begin{proof}
For each $j$, the set of pairs $(b,z)$ with $b_j=0$ or $z\in C_j$ is closed, so the relation $\{(b,z):z\in K_b\}$ is closed in $\{0,1\}^{\N}\times K$. Its projection $\{b:K_b\ne\emptyset\}$ is closed, being the projection of a closed set along a compact factor, and each section $K_b$ is compact. The Arsenin--Kunugui theorem \cite[Theorem~18.18]{kechris1995classical} therefore gives a Borel map $b\mapsto z_b\in K_b$ on this projection, and $\zeta(S)=z_{b(S)}$ is a composition of Borel maps.
\end{proof}

\paragraph{The packed weak learner.}
Let $d=\dpack(\H)<\infty$ and fix the pool size $n$. The set $\cP_{\le1}(\Y)$ of sub-probability vectors from the proof of Lemma~\ref{lem:packed-weak} is closed in $[0,1]^{\Y}$ with the product topology, so the product $K$ of its copies indexed by the countably many pairs $(i,u)$ with $i\in[n]$ and $u\in\cA^{n-1}$ is compact metrizable. Each pattern $v\in\cA^n$ gives the closed constraint
\[
 \sum_{i=1}^n\left(1-\sum_{y\in v_i}z^{i,v_{-i}}_y\right)
 \le d+n\gamma,
\]
which is in force on a pool $S=(x_1,\ldots,x_n)$ exactly when $v\in\H|_S$, that is, when $S\in E_{n,v}$. Enumerating $\cA^n$ as $v_1,v_2,\ldots$, the map $b_j(S)=\ind\{S\in E_{n,v_j}\}$ is Borel by~(ii), and the argument of Lemma~\ref{lem:packed-weak} shows that the constraints in force are satisfiable, so Lemma~\ref{lem:meas-selection} gives a Borel choice of a rule for each pool. The rest of that proof stays Borel: completing the missing mass $1-\sum_yz^{i,u}_y$, a countable sum of coordinates, and averaging over the finitely many permutations of the pool, which gives the symmetric kernel of the leave-one-out argument.

\subsection{Finite Rules, Compression, and Prediction Families}\label{app:meas-families}

Apart from the reduction, which uses Lemma~\ref{lem:meas-selection} again, the remaining constructions use on each pool only countably many tests of the form $S\in E_{n,v}$ and finitely many comparisons of inputs or of Borel quantities, so their measurability can be read off from their definitions.

\paragraph{Zero-example predictors.}
If the relevant Natarajan or packed dimension is zero, all the correct-label sets $h(x)$ at a fixed input $x$ share a label, since otherwise Lemma~\ref{lem:small} gives a witness on one input. For $y\in\Y$, the set of inputs at which $y$ belongs to all these sets is
\[
 \X\setminus\bigcup_{\substack{A\in\cA\\y\notin A}}E_{1,A},
\]
where $E_{1,A}$ is the set of inputs $x$ with $h(x)=A$ for some $h\in\H$. This set is Borel, so predicting the first such label in the enumeration of $\Y$ is Borel. At inputs where every hypothesis is undefined, all labels qualify, and any prediction is allowed there.

\paragraph{Finite label spaces.}
For finite $\Y$, the conservative rule of Section~\ref{sec:finite-full} uses only which partial patterns are realized on its pool. Each such test is Borel by~(ii), so the potential, the updates and the fixed tie-breaking rules are Borel, and enumerating the possible runs gives Borel covers. The exponential-weights and bandit learners built from these covers are therefore measurable.

\paragraph{Compression.}
The compression constructions are measurable once the full-set weak kernel is jointly Borel. On a finite sample, each boosting round chooses among finitely many multisets of examples, and we take the first one that satisfies the required loss inequality. Reconstruction takes finite averages, chooses the first most probable label, and corrects the predictions at finitely many sample inputs, which is Borel because equality of inputs is. Empirical risk minimization and validation compare finitely many Borel quantities.

\paragraph{Prediction families.}
On an ordered pool $S=(x_1,\ldots,x_N)$, identify a function on $\supp(S)$ with its vector of values in $\Y^N$. For a general $\X$ we add the following conditions to the hypotheses of Theorem~\ref{thm:family-conversion}:
\begin{enumerate}[(a)]
\item the family $F_{h,S}$ depends only on $S$ and the pattern $v=(h(x_1),\ldots,h(x_N))\in\cA^N$;
\item for each such $v$, this family is a Borel function of $S$, with values in the countable space of finite subsets of $\Y^N$;
\item its members agree at repeated inputs of $S$;
\item permuting the pool and the pattern permutes the coordinates of every vector in the family.
\end{enumerate}
The finite-label families of Section~\ref{sec:finite-one} and the message families of Section~\ref{co:sec-families} satisfy them. For both, sort the pool and its pattern by the Borel order, construct the family there, and return its vectors to the original order; this gives~(d). The finite-label families enumerate the runs of the conservative rule, reading a cover vector at the first occurrence of each input, which gives~(c); these runs depend only on the pattern and on realization tests, which gives~(a) and~(b). The message families decode the admissible messages of Section~\ref{co:sec-families}; for a fixed pattern there are finitely many of them, their admissibility is determined by the pattern, and the decoder is a compression reconstruction, hence jointly Borel by the previous paragraph. The size bounds and intersection properties are exactly those already proved.

\paragraph{The reduction.}
It remains to make the minimax choice in the proof of Theorem~\ref{thm:family-conversion} measurable. Recall that the learner observes the labels at the positions $J$ of a pool $S=(x_1,\ldots,x_N)$, that $U=[N]\setminus J$, and that it needs distributions $(p_z)_{z\in\supp(U)}$ satisfying~\eqref{eq:red-goal} for every retained hypothesis, where $\beta$ is the constant defined at the start of that proof. By condition~(a), membership of $h$ in $\H_J$ depends only on its pattern $v\in\cA^N$ on $S$, and $v$ is retained if $S\in E_{N,v}$, $y_i\in v_i$ for every $i\in J$, and~\eqref{eq:red-retained} holds for every member of its family; each of these tests is Borel. Since the set $\supp(U)$ varies with the pool, we index the distributions by positions instead and work in the compact metrizable space $\Delta_{\le1}(\Y)^N$. We require $p_i=p_j$ whenever $x_i=x_j$, and for each retained pattern $v$ we impose the constraint~\eqref{eq:red-compactness},
\[
 \frac1{|U|}\sum_{i\in U}\sum_{y\in v_i}p_i(y)\ge1-\beta.
\]
These constraints are closed, the set of constraints in force is a Borel function of the data, and the proof of Theorem~\ref{thm:family-conversion} shows that they are satisfiable when some pattern is retained. Lemma~\ref{lem:meas-selection} therefore gives a Borel choice, and completing the missing mass violates no constraint. If no pattern is retained, a Borel event, we use the point mass on the fixed label. No measurable selection of a hypothesis from $\H$ is needed.

The proof also needs this choice to depend only on the multiset $S$ and the multiset $\{(x_i,y_i):i\in J\}$ of observed examples. We therefore apply the selection to a canonical arrangement of the data, a Borel function of these two multisets: the observed examples first, sorted by the Borel order on inputs and then by the enumeration of $\Y$, followed by the unobserved inputs in sorted order. By condition~(d), the retained patterns are rearranged accordingly.

\subsection{Randomization and the Remaining Arguments}\label{app:meas-final}

A jointly Borel kernel $p_\sigma(\cdot\mid x)$ can be realized as a measurable predictor with a single random seed. Write $\Y=\{y_1,y_2,\ldots\}$, take $\xi\sim\Unif([0,1))$ independent of everything else, and let $g_{\sigma,\xi}(x)$ be the first label $y_j$ such that
\[
 \xi<\sum_{a=1}^j p_\sigma(y_a\mid x).
\]
The partial sums increase to $1$, so such a label exists. The map $(\sigma,\xi,x)\mapsto g_{\sigma,\xi}(x)$ is jointly Borel with label distribution $p_\sigma(\cdot\mid x)$ at every input, so it has the same expected risk as the kernel; the error is evaluated at a single test input, so predictions at distinct inputs need not be independent. Conversely, averaging a learner over its seed preserves joint measurability by Fubini's theorem.

The probability arguments for Theorems~\ref{thm:A}--\ref{thm:E} now concern Borel random variables on standard Borel spaces. In particular, the conditional law of the demonstrated label given the input, used in the reduction, exists by the disintegration theorem \cite{kechris1995classical}, so their exchangeability, concentration, compression and confidence-amplification arguments apply as written. It remains to extend the agnostic demonstration learner of Theorem~\ref{thm:F}.

\subsection{Agnostic Learning from Demonstrations}\label{app:meas-agnostic}

Assume that $\H$ is nondegenerate and locally compatible, with $\dDS(\widetilde\H)<\infty$. We first check that the learner of Lemma~\ref{agn:lem:expected} is measurable. The only probability argument that needs a different construction is the rounding step of Lemma~\ref{agn:lem:confidence}. We replace its independent choices at individual inputs by independent choices on the gaps between finitely many sampled inputs.

\paragraph{The regions and the learner.}
The two regions are Borel, and membership in each local alphabet is Borel in the input, since
\[
 \X_{\mathrm{mix}}=\bigcup_{\substack{A\in\cA\\|A|\ge2}}E_{1,A},
 \qquad
 \{x\in\X_{\mathrm{mix}}:a\in\Y_x\}=\X_{\mathrm{mix}}\cap E_{1,\{a\}}.
\]
The collapsed class inherits assumptions~(i)--(ii): its hypotheses are Borel, and each finite-pattern realization event is a countable union of original realization events. Restriction to either Borel region preserves these properties.

On the mixed region, enumerate $\Y_x$ in the fixed order on $\Y$, encode its labels by their ranks in $[\ell]$, and encode $\diamond$ by $\ell+1$. This encoding is Borel in $x$. A realization event for the encoded class is a countable union, over original patterns, of the corresponding events $E_{n,v}$ intersected with the Borel conditions specifying the encoding. Thus the encoded class also satisfies~(i)--(ii). Its finite covers are Borel by Section~\ref{app:meas-families}, and the specialist updates and finite permutation averages give a jointly Borel kernel in Lemma~\ref{agn:lem:expected}. The learner on the pure region is already covered by the full-set extension with $\ell=1$.

\paragraph{Gaps of a sample.}
Fix a Borel injection $t:\X_{\mathrm{mix}}\to[0,1]$. Given inputs $X_1,\ldots,X_u$, let $U=\{X_1,\ldots,X_u\}$ and list $t(U)$ as $z_1<\cdots<z_s$. Put $z_0=-\infty$, $z_{s+1}=+\infty$, and
\[
 C_j=\{x\in\X_{\mathrm{mix}}:z_j<t(x)<z_{j+1}\},\qquad 0\le j\le s.
\]
These gaps partition $\X_{\mathrm{mix}}\setminus U$. Their membership relations are Borel in the sample and the test input. The next lemma shows that a fresh sample makes all their masses small, even when the input distribution has atoms.

\begin{lemma}\label{lem:meas-gaps}
Let $X_1,\ldots,X_u$ be independent with law $\mu$ on $\X_{\mathrm{mix}}$, where $u\ge1$. For every $0<\tau<1$,
\[
 \Pr\bigl[\max_{0\le j\le s}\mu(C_j)>\tau\bigr]
 \le(u+1)(1-\tau)^u\le(u+1)e^{-u\tau}.
\]
\end{lemma}

\begin{proof}
Let $F$ be the distribution function of $t(X)$ for $X\sim\mu$. Couple the sample as $t(X_i)=F^{-1}(V_i)$, where the $V_i$ are independent uniform variables on $(0,1)$ and $F^{-1}$ is the generalized inverse. The gap between consecutive observed values $a<b$ has mass $F(b-)-F(a)$. The interval $(F(a),F(b-))$ contains no $V_i$, so its length is at most the largest spacing between consecutive ordered $V_i$'s, including the two exterior spacings. The same holds for an exterior gap, with $F(-\infty)=0$ and $F(+\infty-)=1$.

The $u+1$ uniform spacings have a joint distribution uniform on the simplex of nonnegative vectors summing to one. Each spacing therefore exceeds $\tau$ with probability $(1-\tau)^u$, by translating that coordinate by $\tau$ and rescaling the simplex. A union bound proves the claim.
\end{proof}

\paragraph{Completing the confidence argument.}
We now extend Lemma~\ref{agn:lem:confidence} to standard Borel instance spaces. Let $\mu$ be the input marginal on the mixed region and choose a Borel version $\pi_x$ of the conditional law of the demonstrated label. With $c_h,w_h$ and $\pi_x(p)$ as in Section~\ref{agn:sec:mixed}, the conditional comparison becomes
\begin{equation}\label{meas:eq:comparison}
 \Delta_h(p)=\int\bigl(c_h(x)-w_h(x)\pi_x(p)\bigr)\,d\mu(x).
\end{equation}
All these functions are Borel. Keep $\alpha=\eps/4$, $m_0$, $K$ and $\tau$ from the proof of Lemma~\ref{agn:lem:confidence}, and replace its second block size by
\[
 u=\left\lceil\frac2\tau\log\frac8{\delta\tau}\right\rceil.
\]
Use the same formula for the third block size $v$, with this new value of $u$.

Step~1 is unchanged: averaging the $K$ kernels gives $\Delta_h(\bar p)\le2\alpha$ except with probability $\delta/4$, for each fixed $h$. Use the second block to form $U$ and the gaps $C_j$. Lemma~\ref{lem:meas-gaps} gives $\max_j\mu(C_j)\le\tau$ except with probability $(u+1)e^{-u\tau}\le\delta/4$.

Step~3 is also unchanged, since $U$ is finite: use the third block to choose a most frequently demonstrated label $\widehat a_x\in\Y_x$ at each $x\in U$. Let $p^U$ equal $\bar p$ off $U$ and put mass one on $\widehat a_x$ at $x\in U$. The same union bound over at most $\ell^u$ maps on $U$, followed by~\eqref{meas:eq:comparison}, gives
\[
 \Delta_h(p^U)\le\Delta_h(\bar p)+\alpha
\]
except with probability $\delta/4$.

For the rounding step, draw one independent uniform seed $\xi_j$ for each gap $C_j$. At every $x\in C_j$, sample from $\bar p(\cdot\mid x)$ by the inverse-CDF rule of Section~\ref{app:meas-final}, using the same seed $\xi_j$ throughout this gap. On $U$, predict $\widehat a_x$. This defines an admissible predictor $\widehat g$ that is jointly Borel in the examples, the seeds and the test input. There are at most $u+1$ seeds, which can be encoded in a single uniform seed.

Fix $h$ and condition on all examples. The contribution of gap $C_j$ to the change in comparison is
\[
 Z_j=\int_{C_j}w_h(x)\bigl(\pi_x(\bar p)-\pi_x(\widehat g)\bigr)\,d\mu(x).
\]
At each $x\in C_j$, the rounded label has law $\bar p(\cdot\mid x)$, so Fubini's theorem gives $\E[Z_j\mid\text{examples}]=0$. The $Z_j$ are independent, since they use different seeds. Their range lengths are at most $\mu(C_j)$: the first term in each integral is fixed, while $\int_{C_j}w_h(x)\pi_x(\widehat g)\,d\mu(x)$ lies in $[0,\mu(C_j)]$. On the event that all gaps have mass at most $\tau$, we have $\sum_j\mu(C_j)^2\le\tau$. Hence Hoeffding's inequality gives
\[
 \Pr\bigl[\Delta_h(\widehat g)-\Delta_h(p^U)>\alpha\mid\text{examples}\bigr]
 =\Pr\bigl[\textstyle\sum_jZ_j>\alpha\mid\text{examples}\bigr]
 \le e^{-2\alpha^2/\tau}\le\delta/4.
\]
The four failure events have total probability at most $\delta$. Outside their union, $\Delta_h(\widehat g)\le2\alpha+\alpha+\alpha=\eps$. This proves the confidence lemma on a standard Borel instance space.

The observable comparison now gives the mixed-region guarantee by choosing a fixed near-optimal hypothesis, as in Proposition~\ref{agn:prop:mixed}. Since the regions are Borel, the gluing argument of Lemma~\ref{agn:lem:glue} combines this learner with the pure-region learner measurably. Necessity and the degenerate obstruction use finite witnesses and finitely supported distributions, so they need no change. This proves the extension of Theorem~\ref{thm:F} and completes the proof of Theorem~\ref{thm:standard-borel-extension}.
\end{document}